\documentclass[preprints,article,accept,pdftex,moreauthors]{Definitions/mdpi} 
\usepackage{easyReview,rotating,multirow,pdflscape,adjustbox,multirow,wasysym}
\usepackage{framed} 
\colorlet{shadecolor}{red!10}
\usepackage{tcolorbox}
\tcbuselibrary{breakable}
\usepackage{amssymb}
\usepackage{pifont}
\usepackage[normalem]{ulem}
\usepackage[hyphens]{url}

\makeatletter

\newcommand{\lsc}[1]{\textcolor{red}{#1}}
\DeclareMathOperator*{\dom}{dom}
\DeclareMathOperator*{\cod}{cod}
\newcommand{\eq}{\textsf{Eq}}
\newcommand{\nq}{\textsf{NEq}}
\newcommand{\impl}{\ensuremath{\twoheadrightarrow}}
\newcommand{\mig}{\ensuremath{\impl^\textsf{nspec}}}
\newcommand{\miss}{\ensuremath{\impl^\textbf{None}}}
\newcommand{\minst}{\ensuremath{\impl^{\downarrow}}}

\newcommand{\mstar}{\ensuremath{\impl^\star}}
\newcommand{\ndiff}{\ensuremath{\omega}}
\usepackage{enumitem,booktabs,cancel}
\newlist{competitors}{enumerate}{1}
\setlist[competitors]{label=T\arabic*. , leftmargin=*, font=\bfseries}
\newlist{questions}{enumerate}{2}
\setlist[questions,1]{label=RQ \textnumero\arabic* , leftmargin=*, font=\bfseries}
\setlist[questions,2]{label=(\alph*),ref=\thequestionsi(\alph*)}
\newlist{requirements}{enumerate}{2}
\setlist[requirements,1]{label=Req \textnumero\arabic* , leftmargin=*, font=\bfseries}
\setlist[requirements,2]{label=(\alph*),ref=\therequirementsi(\alph*)}
\usepackage{xcolor}
\usepackage{graphicx}
\usepackage{easyReview,subcaption}
\usepackage{graphicx}
\usepackage[frozencache,cachedir=_minted]{minted} 
\usepackage{braket}
\usepackage{caption}
\usepackage{amsthm}
\usepackage{amsmath}
\usepackage[mathscr]{euscript}
\usepackage{amssymb,arydshln}
\newtheorem{lemma}[theorem]{Lemma}
\newtheorem{corollary}[theorem]{Corollary}
\newtheorem{example}[theorem]{Example}
\newtheorem{definition}[theorem]{Definition}
\usepackage{colortbl}

\SetupFloatingEnvironment{listing}{name=Source Code}
\PassOptionsToPackage{hypertexnames=false}{hyperref} 

\usepackage[super]{nth}

\usepackage{algorithm}
\usepackage[noEnd=true]{algpseudocodex}

\usepackage{soul}
\soulregister\cite7
\soulregister\ref7
\soulregister\pageref7
\soulregister\gls7

\if@todonotes@disabled

\else

\fi

\newcommand{\sem}[1]{\ensuremath{[\![#1]\!]}}
\usepackage[acronym,nonumberlist]{glossaries}
\makeglossaries
\newacronym{qa}{QA}{Question Answering}
\newacronym{rag}{RAG}{Retrieval-Augmented Generation}
\newacronym{fol}{FOL}{First-Order Logic}
\newacronym{nlp}{NLP}{Natural Language Processing}
\newacronym{ai}{AI}{Artificial Intelligence}
\newacronym{mg}{MG}{Montague Grammar}
\newacronym{ggg}{GGG}{Generalised Graph Grammar}
\newacronym{gql}{GQL}{Graph Query Language}
\newacronym{ml}{ML}{Machine Learning}
\newacronym{nn}{NN}{Neural Network}
\newacronym{meudb}{meuDB}{Multi-Word Entity Unit DataBase}
\newacronym{meu}{MEU}{Multi-Word Entity Unit}
\newacronym{gpe}{GPE}{GeoPolitical Entity}
\newacronym{ari}{ARI}{Adjusted Random Index}
\newacronym{amr}{AMR}{Abstract Meaning Representation}
\newacronym{ri}{RI}{Random Index}
\newacronym{gsm}{GSM}{Generalised Semistructured Model}
\newacronym{mgr}{MGR}{Montague Grammar Representation}
\newacronym{dg}{DG}{Dependency Graph}
\newacronym{lr}{LR}{Logical Representation}
\newacronym{ir}{IR}{Information Retrieval}
\newacronym[\glslongpluralkey={Universal Dependencies}, \glsshortpluralkey={UDs}]{ud}{UD}{Universal Dependency}
\newacronym{gwl}{GwL}{Graphs with Logic}
\newacronym{dfs}{DFS}{Depth-First Search}
\newacronym{bfs}{BFS}{Breadth-First Search}
\newacronym{pos}{POS}{Part of Speech}
\newacronym{ahc}{HAC}{Hierarchical Agglomerative Clustering}
\newacronym{mc}{MCL}{Markov Clustering}
\newacronym{lassi}{LaSSI}{Logical, Structural and Semantic text Interpretation}
\newacronym{cwa}{CWA}{Closed-World Assumption}
\newacronym{llm}{LLM}{Large Language Model}
\newacronym{kb}{KB}{Knowledge Base}
\newacronym{sg}{SG}{Simple Graph}
\newacronym{lg}{LG}{Logical Graph}
\newacronym{ovr}{OvR}{One-versus-Rest}
\newacronym{dt}{DT}{Decision Tree}
\newacronym{rdbms}{RDBMS}{Relational Database Management System}
\newacronym{gevai}{GEVAI}{General, Explainable, and Verified Artificial Intelligence}
\makeglossaries

\usepackage{tabularray}
\definecolor{Silver}{rgb}{0.825,0.825,0.825}
\definecolor{Alto}{rgb}{0.85,0.85,0.85}
\definecolor{Mercury}{rgb}{0.875,0.875,0.875}
\definecolor{Gallery}{rgb}{0.9,0.9,0.9}
\definecolor{WildSand}{rgb}{0.925,0.925,0.925}

\firstpage{1} 
\pubvolume{1}
\issuenum{1}
\articlenumber{0}
\pubyear{2025}
\copyrightyear{2024}
\datereceived{ } 
\daterevised{ } 
\dateaccepted{ } 
\datepublished{ } 
\hreflink{https://doi.org/} 

\Title{{Verified Language Processing with Hybrid Explainability: A Technical Report}}

\TitleCitation{Verified Language Processing with Hybrid Explainability: A Technical Report}

\Author{Oliver Robert Fox\orcidA{}, Giacomo Bergami\orcidB{} and Graham Morgan\orcidC{}}

\AuthorNames{Oliver Robert Fox, Giacomo Bergami and Graham Morgan}

\isAPAStyle{%
       \AuthorCitation{Fox, O.R., Bergami, G., \& Morgan, G.}
         }{%
        \isChicagoStyle{%
        \AuthorCitation{Fox, Oliver Robert, Giacomo Bergami, and Graham Morgan.}
        }{
        \AuthorCitation{Fox, O.R.; Bergami, G.; Morgan, G.}
        }
}

\address{School of Computing, Faculty of Science, Agriculture and Engineering, Newcastle University, \linebreak  Newcastle Upon Tyne NE4 5TG, UK;  {o.fox3@newcastle.ac.uk (O.R.F.); graham.morgan@newcastle.ac.uk (G.M.)}}

\corres{\hangafter=1 \hangindent=1.05em \hspace{-0.82em}Correspondence: {giacomo.bergami@newcastle.ac.uk} }

\conference{LaSSI: Logical, Structural, and Semantic text Interpretation. Fox, O.R.; Bergami, G; Morgan, G. Database Engineered Applications. 28th International Symposium, IDEAS 2024, Bayonne, France, August 26--29, 2024, Springer} 

\abstract{The volume and diversity of digital information have led to a growing reliance on \acrfull{ml} techniques, such as \acrfull{nlp}, for interpreting and accessing appropriate data. While vector and graph embeddings represent data for similarity tasks, current state-of-the-art pipelines lack guaranteed explainability, failing to determine similarity for given full texts accurately. These considerations can also be applied to classifiers exploiting generative language models with logical prompts, which fail to correctly distinguish between logical implication, indifference, and inconsistency, despite being explicitly trained to recognise the first two classes. We present a novel pipeline designed for \textit{hybrid explainability} to address this. Our methodology combines graphs and logic to produce \acrfull{fol} representations, creating machine- and human-readable representations through \acrfull{mg}. Preliminary results indicate the effectiveness of this approach in accurately capturing full text similarity. To the best of our knowledge, this is the first approach to differentiate between implication, inconsistency, and indifference for text classification tasks. To address the limitations of existing approaches, we use three self-contained datasets annotated for the former classification task to determine the suitability of these approaches in capturing sentence structure equivalence, logical connectives, and spatiotemporal reasoning. We also use these data to compare the proposed method with language models pre-trained for detecting sentence entailment. The results show that the proposed method outperforms state-of-the-art models, indicating that natural language understanding cannot be easily generalised by training over extensive document corpora. 
	 This work offers a step toward more transparent and reliable \acrfull{ir} from extensive textual data.}


\DeclareMathOperator*{\argmax}{arg\,max}
\DeclareMathOperator*{\argmin}{arg\,min}
\definecolor{lightGray}{gray}{0.9}
\definecolor{darkGray}{gray}{0.8}

\usepackage[toc,page,header]{appendix}
\usepackage{minitoc}

\makeatletter
\newcommand{\supplementaryname}{Supplement}
\newcommand{\supplementarytitles}[1]{\gdef\@appendixtitles{#1}}

\newcommand{\supplementarysections}[1]{\gdef\@appendixsections{#1}}

\newcommand{\supplementarystart}[1]{\gdef\@appendixstart{#1}}

\renewcommand{\appendixstart}{%
	\setcounter{section}{0}%
	\setcounter{subsection}{0}%
	\setcounter{subsubsection}{0}%
	\setcounter{figure}{0}
	\setcounter{table}{0}
	\setcounter{scheme}{0}
	\setcounter{chart}{0}
	\setcounter{boxenv}{0}
	\setcounter{equation}{0}
	\setcounter{theorem}{0}
	\setcounter{lemma}{0}
	\setcounter{corollary}{0}
	\setcounter{proposition}{0} 
	\setcounter{characterization}{0} 
	\setcounter{property}{0} 
	\setcounter{problem}{0} 
	\setcounter{example}{0} 
	\setcounter{examplesanddefinitions}{0} 
	\setcounter{remark}{0} 
	\setcounter{definition}{0} 
	\setcounter{hypothesis}{0}
	\setcounter{notation}{0}
	\setcounter{algorithm}{0}
}

\newcommand{\supplement}{%
	\pagebreak
	\appendix\appendixstart
	{\centering \Huge Supplementary Material}
	\addcontentsline{toc}{section}{Appendix} 
	\parttoc 
	
	\renewcommand{\thesection}{\Roman{section}} 
	\renewcommand{\thesubsection}{\thesection.\arabic{subsection}}
	\titleformat {\section} [block] {\raggedright\bfseries} {%
		\ifthenelse{\equal{\@appendixtitles}{yes}}{%
			\supplementaryname~\thesection.%
		}{%
			\supplementaryname~\thesection~%
		}
	} {0pt} {}
	\titlespacing {\section} {0pt} {12pt} {3pt}
	\titleformat {\subsection} [block] {\raggedright\itshape} {%
		\ifthenelse{\equal{\@appendixtitles}{yes}}{%
			\supplementaryname~\thesubsection.%
		}{%
			\supplementaryname~\thesubsection%
		}
	} {0pt} {}
	\titlespacing {\section} {0pt} {12pt} {3pt}
	\titleformat {\subsubsection} [block] {\raggedright\selectfont} {%
		\ifthenelse{\equal{\@appendixtitles}{yes}}{%
			\supplementaryname~\thesubsubsection.%
		}{%
			\supplementaryname~\thesubsubsection%
		}
	} {0pt} {}
	\titlespacing {\section} {0pt} {12pt} {3pt}
	\gdef\theHsection{\@Roman\c@section.}
	\gdef\theHsubsection{\@Roman\c@section.\@arabic\c@subsection}
	\csname appendixmore\endcsname
	\renewcommand{\thefigure}{S\arabic{figure}}
	\renewcommand{\thetable}{S\arabic{table}}
	\renewcommand{\thescheme}{S\arabic{scheme}}
	\renewcommand{\thechart}{S\arabic{chart}}
	\renewcommand{\theboxenv}{S\arabic{boxenv}}
	\renewcommand{\theequation}{S\arabic{equation}}
	\renewcommand{\thetheorem}{S\arabic{theorem}}
	\renewcommand{\thelemma}{S\arabic{lemma}}
	\renewcommand{\thecorollary}{S\arabic{corollary}}
	\renewcommand{\theproposition}{S\arabic{proposition}} 
	\renewcommand{\thecharacterization}{S\arabic{characterization}}
	\renewcommand{\theproperty}{S\arabic{property}}
	\renewcommand{\theproblem}{S\arabic{problem}}
	\renewcommand{\theexample}{S\arabic{example}}
	\renewcommand{\theexamplesanddefinitions}{S\arabic{examplesanddefinitions}}
	\renewcommand{\theremark}{S\arabic{remark}}
	\renewcommand{\thedefinition}{S\arabic{definition}}
	\renewcommand{\thehypothesis}{S\arabic{hypothesis}}
	\renewcommand{\thenotation}{S\arabic{notation}}
	\renewcommand{\thealgorithm}{S\arabic{algorithm}}
}
\makeatother

\begin{document}
%

\section{Introduction}
Factoid sentences are commonly used to characterise news \cite{DBLP:conf/tac/ZhangSSHLPLZWWJ18} and word definitions from dictionaries \cite{DBLP:conf/coling/YoshidaTH82,ayse-etal-2011-extraction}, as they can be easily used to recognise conflicting opinions, as well as to represent the majority of a sentence type contained in \glspl{kb} such as ConceptNet 5.5 \cite{DBLP:conf/aaai/SpeerCH17} or DBpedia \cite{10.5555/1785162.1785216,mendes-etal-2012-dbpedia}. Automated extraction of \gls{kb} nodes and edges from full text is limited, as it results in some sentences not being correctly parsed and rendered as full-text nodes \cite{bergami2019frameworksupportingimprecisequeries}. This is a major limitation when addressing the possibility of answering common-sense questions, as a machine cannot easily interpret the latter information, thus leading to low-accuracy results (55.9\% \cite{talmor-etal-2019-commonsenseqa}). To improve these results soon, we need a technique that provides a machine-readable representation of spurious text within the graph while also ensuring the correctness of the representation, going beyond the strict boundaries of a graph \gls{kb} representation. We can consider this the dual problem of querying bibliographical metadata using a query language as close as possible to natural language. Notably, given the untrustworthiness of existing \gls{nlp} approaches to \acrfull{ir}, librarians still rely on domain-oriented query languages  \cite{Kreutz2022}. By providing a more trustworthy and verifiable representation of a full text in natural language, we can then generate a more reliable intermediate representation of the text that can be used to query the bibliographical data \cite{10.1145/2588555.2594519,10.1007/978-3-031-38499-8_29}. 

An adequate semantic representation of the sentences should capture both the semantic nature of the data as well as recognise implication, inconsistency, or indifference as classification outcomes over pairs of sentences. To date, this has not been considered in literature, as inconsistency is merged with indifference \cite{bao-etal-2024-abstract}; we have either similarity or entailment classification, but not the classification of conflicting information. 
Thus, none of the available datasets for \gls{nlp} support training abilities to address the spread of misinformation through logical reasoning. Pre-trained language models (Section \ref{sec:rw_soa}) are often characterised as ``stochastic parrots'' \cite{10.1145/3442188.3445922}, they exhibit significant deficiencies in formal reasoning due to their reliance on learned statistical patterns from vector representations rather than proper logical inference \cite{mirzadeh2024gsmsymbolicunderstandinglimitationsmathematical, badyal2023intentionalbiasesllmresponses}. On the other hand, sound inference is fundamentally grounded in logic, for which well-known reasoning algorithms already exist without the need for hardcoding truth derivation rules \cite{Harrison_2009}. Pre-trained language models merely mimic logical processes without formally incorporating them or, if they do through prompting, they provide limited reasoning ability restricted to a subset of all possible natural deduction rules (Section \ref{sec:litrev_llm}, T5: DeBERTaV2+AMR-LDA \cite{bao-etal-2024-abstract}). On the other hand, we could ultimately reason through text by first deriving a logic-based representation of the text through \gls{mg} \cite{Montague+1975+94+121}, for then subsequently applying logical inference steps.  While doing so, we also want to avoid any contradiction of information as \textit{ex falso [sequitur] quodlibet} (``from falsehood, anything follows''). This begs the question: how do we reason with negation?

Classical logic approaches face the ``principle of explosion'', where a single contradiction can render an entire system nonsensical by deriving arbitrary facts from false premises. Paraconsistent reasoning, by contrast, acknowledges and isolates a contradiction to avoid the above  \cite{bb, hawking1988}. This can be used in realistic misinformation detection when considering data coming from disparate sources: the work by Zhang et al. \cite{DBLP:conf/tac/ZhangSSHLPLZWWJ18} already motivated the effectiveness of doing so for ``repairing'' the inference steps of AI algorithms over real-world contradictory data when handling misinformation online: thus, this paper puts this reasoning to the extreme by incorporating paraconsistency reasoning at the heart of the inference step rather than postponing at the end to address potential reasoning flaws. At the time of the writing, no logical inference algorithm is known to reason paraconsistently despite their formalisation in the aforementioned theoretical literature.

We continue our previous work on \gls{lassi} \cite{ideas2024b} on addressing the former limitations while addressing the following questions:

\begin{questions}


\item \textit{Is it possible to create an plug-and-play and explainable \gls{nlp} pipeline for sentence representation?} By exploiting white-box reasoning, we can ultimately visualise the outcome of each inference and sentence generation step (Section \ref{explainstudy}). Furthermore, by designing the pipeline in a modular way, it is easy to replace each single component to adapt to different linguistic needs. Through declarative context-free rewritings for NLP representation (Section \ref{sec:conclusion}), we ensure pipeline versatility by changing the inner rules rather than requiring code changes, as well as future extensibility. Therefore, we can see how the full text is transformed at every stage, allowing any errors to be identified and corrected using a human-in-the-loop approach. 

\item \label{rq2}\textit{Can pre-trained language models correctly capture the notion of sentence similarity?} The previous result should imply the impossibility of accurately deriving the notion of equivalence, as entailment implies equivalence through if-and-only-if relationships but not vice versa. 
 Meanwhile, the notion of sentence indifference should be kept distinct from the notion of conflict. 
  We designed empirical experiments with certain datasets to address the following sub-questions:

\begin{questions}
\item\label{rqn2a} \textit{Can pre-trained language models capture propositional calculus?} These experiments work on the following considerations: given that \gls{fol} is more expressive than propositional calculus, and given that pre-trained models are assumed to reason on arbitrary sentences, thus representable in FOL as per Montague's assumption \cite{Montague+1975+94+121}, and given that propositional calculus is less expressive but included in  \gls{fol}, any inability to capture propositional calculus will also invalidate the possibility of extending the approach to soundly making inferences in FOL. Current experiments (Section \ref{sub:logical_connectives}) show that pre-trained language models cannot adequately capture the information contained in logical connectives from propositional calculus.

\item\label{rqn2b} \textit{Can pre-trained language models distinguish between active and passive sentences?} 
Experiments (Section \ref{sub:active-passive}) show that their intermediate representation (AMR, tokens, embeddings)  is insufficient for distinguishing them faithfully. 

\item\label{rqn2c} \textit{Can pre-trained language models correctly capture minimal FOL extensions for  spatiotemporal reasoning?} Spatiotemporal reasoning requires specific part-of and is-a reasoning requiring minimal FOL extensions (eFOL). The paper's results validate the observations from \ref{rqn2a}, thus remarking the impossibility of these approaches to reason on eFOL (Section \ref{sub:spatio}). 
Additional experiments clearly remark the impossibility of such models to derive correct solutions even after re-training (Section \ref{explainstudy}).
\end{questions}


\end{questions}


To address the former limitations, we propose \gls{lassi}, an instance of the \gls{gevai} framework (Section \ref{sec:gevai}) providing modular phases: after addressing \textit{discourse integration} in the \textit{a priori} phase concerning entity recognition and semantic ambiguity through entity type association, the \textit{ad hoc} phase applies derivational and declarative rules (when possible) to transform full-text representations into final logical formulae. While doing so, we further address \textit{discourse integration} through pronoun resolution (\supplementaryname~\ref{sup:pronres}) and addressing ambiguities from the generated \gls{ud} parsing by exploiting syntactic and morphological information through our proposed Upper Ontology, Parmenides. Finally, the \textit{ex post} phase explains sentences' logical entailment by providing a confidence score as a result of reasoning paraconsistently over Parmenides for semantic information, while plotting the reasoning outcomes graphically (Section \ref{explainstudy}). We also offer a pipeline ablation study for testing the different stages of \textit{discourse integration} (Section \ref{subsec:pip_abla}) and compare our explainers to other textual explainers (SHAP and LIME) using state-of-the-art methodologies to tokenise text and correlate it with the predicted class (Section \ref{explainstudy}). Last, we draw our conclusions and indicate some future research directions (Section \ref{sec:conclusion}). The datasets for this paper are available online (\url{https://osf.io/g5k9q/}, Accessed on 14 May 2025).

\section{Related Works}

\subsection{General Explainable and Verified Artificial Intelligence (GEVAI)}\label{sec:gevai}

A recent survey by Seshia et al. introduced the notion of verified \gls{ai} \cite{VAI}, through which we seek greater control by exploiting reliable and safe approaches to derive a specification $\Phi$ describing abstract properties of the data $\mathfrak{S}$. Through verifiability, the specification itself $\Phi$ can be used as a witness of the correctness of the proposed approach by determining whether the data satisfy such specification, i.e., $\mathfrak{S}\vDash\Phi$ (\textit{formal verification}).  This survey also revealed that, at the time of its writing, providing a truly verifiable approach of \gls{nlp} is still an open challenge, remaining unresolved with current techniques. In fact, specification $\Phi$ is not simply considered the outcome of a classification model or the result of applying a traditional explainer, but rather a compact representation of the data in terms of the properties it satisfies in a machine- and human-understandable form. Furthermore, as remarked in our recent survey \cite{gevai}, the possibility of explaining the decision-making process in detail, even in the learning phase, goes hand in hand with using an abstract and logical representation of the data. However, if one wants to use a numerical approach to represent the data, such as when using \glspl{nn} and producing sentence embeddings from transformers (Section \ref{sec:rw_soa}), then one is forced to reduce the explanation of the entire learning process to the choice of weights useful for minimising the loss function and to the loss function itself \cite{Paper2}. A possible way to partially overcome this limitation is to jointly train a classifier with an explainer  \cite{Paper1}, which might then pinpoint the specific data features leading to the classification outcome \cite{Paper3}: as current explainers mainly state how a single feature affects the classification outcome, they mainly lose information on the correlations between these features, which are extremely relevant in \gls{nlp} (semantic) classification tasks. More recent approaches \cite{10.1007/978-3-031-38499-8_29} have attempted to revive previous non-training-based approaches, showing the possibility of representing a full sentence with a query via semantic parsing  \cite{10.1145/2588555.2594519}. 
 A more recent approach also enables sentence representation in logical format rather than ensuring an SQL representation of the text. As a result, the latter can also be easily rewritten and used for inference purposes. Notwithstanding the former, researchers have not covered all the rewriting steps required to capture different linguistic functions and categorise their role within the sentence, unlike in this study. 
  Furthermore, while the authors of \cite{10.1007/978-3-031-38499-8_29} attempted to answer questions, our study takes a preliminary step back. We first test the suitability of our proposed approach to derive correct sentence similarity from the logical representation. Then, we then tackle the possibility of using logic-based representations to answer questions and ensure the correct capturing of multi-word entities within the text while differentiating between the main entities based on the properties specifying them.

Our latest work also highlights the potential for achieving verification when combined with explainability, making the data understandable to both humans and machines \cite{gevai}. This identifies three distinct phases that should be considered prerequisites for achieving good explanations: \textit{First}, within the first \textit{a priori} explanation, unstructured data should achieve a higher structural representation level by deriving additional contextual information from the data and its environment. \textit{Second}, the \textit{ad hoc} explanation should provide an explainable way through which a specification is extracted from the data, where provenance mechanisms help trace all the data processing steps. If represented as a logical program, the specification can also ensure both human and machine understandability by being expressed in an unambiguous format. \textit{Lastly}, the \textit{ex post} phase (\textit{post hoc} in \cite{Paper1}) should further refine the previously generated specifications by achieving better and more fine-grained explainability. Therefore, we can derive even more accessible results and facilitate comparisons between models, while also enabling them to be compared with other data. Our Section \ref{sec:methodology} reflects these phases.

\begin{landscape}
	\begin{table}[H]
		\caption{A comparative table between the competing pre-trained language model approaches and LaSSI.\label{tab3}}
		\begin{adjustbox}{max width=1.6\textwidth}
			\begin{tabular}{l|clll|l|l}
				& \multicolumn{3}{c|}{Sentence Transformers (Section \ref{litrev:st})}                                                                                                                                                                 & Neural \gls{ir} (Section \ref{litrev:nir})                         & Generative \acrfull{llm} (Section \ref{sec:litrev_llm})                                                                                                                                                                      & GEVAI (Section \ref{sec:gevai})                                                                                                                                                                                                                                    \\
				& \multicolumn{1}{l}{MPNet \cite{song2020mpnet}}                               & RoBERTa \cite{liu2019roberta} & \multicolumn{1}{l|}{MiniLMv2 \cite{wang2021minilmv2multiheadselfattentionrelation}} & ColBERTv2 \cite{santhanam-etal-2022-colbertv2}               & DeBERTaV2+AMR-LDA \cite{bao-etal-2024-abstract}                                                                                                                                                                     & LaSSI (\textbf{This Paper})                                                                                                                                                                                                                                        \\ \hline
				Task                                                                                                  & \multicolumn{3}{c|}{Document Similarity}                                                                                                                                                                                                              & Query Answering                                                               & Entailment Classification                                                                                                                                                                                                            & \multicolumn{1}{l|}{Paraconsistent Reasoning}                                                                                                                                                                                                                                                            \\ \cline{2-7} 
				Sentence Pre-Processing                                                     & \multicolumn{4}{c|}{Word Tokenisation + Position Encoding}                                                                                                                                                                                                                                                                            & \begin{tabular}[c]{@{}l@{}}\textbullet AMR with Multi-Word Entity Recognition\\ \textbullet AMR Rewriting\end{tabular}                                                                                 & \multicolumn{1}{l|}{\begin{tabular}[c]{@{}l@{}}\textbullet Dependency Parsing\\ \textbullet Generalised Graph Grammars\\ \textbullet Multi-Word Entity Recognition\\ \textbullet Logic Function Rewriting\end{tabular}} \\ \cline{2-7} 
				\begin{tabular}[c]{@{}l@{}} Similarity/Relationship\\ inference\end{tabular} & \multicolumn{1}{l|}{\begin{tabular}[c]{@{}l@{}}Permutated Language \\ Modelling\end{tabular}} & \multicolumn{1}{l|}{--}                        & \multicolumn{1}{l|}{\begin{tabular}[c]{@{}l@{}}Annotated Training\\ Dataset\end{tabular}}            & \begin{tabular}[c]{@{}l@{}}Factored by\\ Tokenisation\end{tabular}            & \begin{tabular}[c]{@{}l@{}}\textbullet Logical Prompts\\ \textbullet Contrastive Learning\end{tabular}                                                                                                 & \multicolumn{1}{l|}{\multirow{2}{*}{\begin{tabular}[c]{@{}l@{}}\textbullet Knowledge Base-driven Similarity\\ \textbullet TBox Reasoning\end{tabular}}}                                                                                               \\ \cline{2-6}
				Learning Strategy                                                    & \multicolumn{2}{c|}{Static Masking}                                                                                                            & \multicolumn{1}{l|}{Dynamic Masking}                                                                 & \begin{tabular}[c]{@{}l@{}}Annotated Training\\ Dataset\end{tabular}          & \begin{tabular}[c]{@{}l@{}}\textbullet Autoregression\\ \textbullet Sentence Distance Minimisation\end{tabular}                                                                                        & \multicolumn{1}{l|}{}                                                                                                                                                                                                                                                               \\ \cline{2-7} 
				Final Representation                                                        & \multicolumn{3}{c|}{One vector per sentence}                                                                                                                                                                                                          & Many vectors per sentence                                                     & Classification outcome                                                                                                                                                                                                               & \multicolumn{1}{l|}{Extended \acrfull{fol}}                                                                                                                                                                                                                        \\ \hline
				Pros                                                                                                  & \multicolumn{3}{c|}{Deriving Semantic Similarity through Learning}                                                                                                                                                                                    & \begin{tabular}[c]{@{}l@{}}Generalisation of document\\ matching\end{tabular} & Deriving Logical Entailment through Learning                                                                                                                                                                                         & \multicolumn{1}{l|}{\begin{tabular}[c]{@{}l@{}}\textbullet Reasoning Traceability\\ \textbullet Paraconsistent Reasoning\\ \textbullet Non biased by documents\end{tabular}}                                                           \\ \cline{2-7} 
				Cons                                                                        & \multicolumn{4}{c|}{\begin{tabular}[c]{@{}c@{}}\textbullet Cannot express propositional calculus\\ \textbullet Semantic similarity does not entail implication capturing\end{tabular}}                                                                                                                  & \begin{tabular}[c]{@{}l@{}}\textbullet Inadequacy of AMR\\ \textbullet Reasoning limited by Logical Prompts\\ \textbullet Biased by probabilistic reasoning\end{tabular} & \multicolumn{1}{l|}{Heavily Relies on Upper Ontology}                                                                                                                                                                                                                                                    \\ \hline
			\end{tabular}
		\end{adjustbox}
	\end{table}
\end{landscape}

\subsection{Pre-Trained Language Models}\label{sec:rw_soa}
We now introduce our competing approaches, which all work by assuming that information can be distilled from a large set of annotated documents and is suitable training tasks, leading to a model representation minimising the loss function over an additional training dataset. The characterisation of \glspl{llm} as ``stochastic parrots'' \cite{10.1145/3442188.3445922} posits that their proficiency lies in mimicking statistical patterns from vast training data rather than genuine comprehension, leading to significant challenges such as hallucination and the amplification of societal biases. Research has explored these issues, even noting how biases can be intentionally manipulated \cite{badyal2023intentionalbiasesllmresponses}, private information from prompt data can be revealed \cite{duan2023flocksstochasticparrotsdifferentially}, and issues with semantic leakage \cite{gonen-etal-2025-liking}. To mitigate these problems, researchers propose a combination of methodological and technical solutions. Existing human-in-the-loop approaches look at creating expert-annotated datasets in specialised fields like mental health and nutrition counselling, advocating improved data quality and reduced bias \cite{balloccu2024askexpertssourcinghighquality}. The rules declared in Parmenides, \gls{ggg} rewriting, and the Logical Functions in Table \ref{tab:logicalFunction} are all expressed declaratively, meaning these can be changed to alter results for the most correct outcome. By using logic, we are minimising bias, which is known to affect pre-trained language models \cite{10298519}, and alleviating the causes of semantic leakage \cite{gonen-etal-2025-liking}. On the other hand, symbolic \glspl{nn} such as \cite{Tsamoura_Hospedales_Michael_2021} are never considering negation: if a system does not use logic combined with negation, then there is no possibility for the system to reason over contradictory and real world data sentences, which also contain negations. We focus on pre-trained language models for sentence similarity and logical prediction tasks. \tablename~\ref{tab3} summarises our findings.

\subsubsection{Sentence Transformers}\label{litrev:st}
Google introduced transformers \cite{NIPS2017_3f5ee243} as a compact way to encode semantic representations of data into numerical vectors, usually within a Euclidean space, through a preliminary tokenisation process. After converting tokens and their positions into vector representations, a final transformation layer provides the final vector representation for the entire sentence. The overall architecture seeks to learn a vector representation for an entire sentence, maximising the probability distribution over the extracted tokens. This is ultimately achieved through a loss minimisation task that depends on the transformer's architecture of choice; while \textit{masking}  considers predicting the masked out tokens by learning a conditional probability distribution over the non-masked one, \textit{autoregression} learns a stationary distribution for the first token and a conditional probability distribution aiming to predict the subsequent tokens, which are gradually unmasked. 
  While sentence transformers adopt the former approach, generative \glspl{llm} (Section \ref{sec:litrev_llm}) use the latter.

Pre-trained sentence transformer models are extensively employed to turn text into vectors known as embeddings and are fine-tuned on many datasets for general-purpose tasks such as semantic search, grouping, and retrieval. Nanjing University of Science and Technology and Microsoft Research jointly created MPNet \cite{song2020mpnet}, which aims to consider the dependency among predicted tokens through permuted language modelling while considering their position within the input sentence. RoBERTa \cite{liu2019roberta}, a collaborative effort between the University of Washington and Facebook AI, is an improvement over traditional BERT models, where masking only occurs at data pre-processing, by performing \textit{dynamic masking}, thus generating a masking pattern every time a training token sequence is fed to the model. The authors also recognised the positive effect of hyperparameter tuning over the resulting model, hence systematising the training phase while considering additional documents.  Lastly, Microsoft Research \cite{wang2021minilmv2multiheadselfattentionrelation} took an opposite direction on the hyperparameter tuning challenge: rather than consider hundreds of millions of parameters, MiniLMv2 considers a simpler approach compressing large transformers via pre-trained models, where a small student model is trained to mimic the pre-trained one. Furthermore, the authors exploited a contrastive learning objective for maximising the sentence semantics' similarity mapping: given a training dataset composed of pairs of full text sentences, the prediction task is to match one sentence from the pair, and then the other is given.


Recent surveys on the expressive power of transformer-based approaches, mainly for capturing text semantics, reveal some limitations in their reasoning capabilities. \textit{First}, when two sentences are unrelated, the attention mechanisms are dominated by the last output vector \cite{Paper1}, which might easily lead to hallucination and untrustworthy results such as the ones due by semantic leakage \cite{gonen-etal-2025-liking}. \textit{Second}, theoretical results have suggested that these approaches are unable to reason on propositional calculus \cite{10.1162/tacl_a_00663}. 
 If the impossibility of simple logical reasoning during the learning phase is confirmed, this would strongly undermine the possibility of relying on the resulting vector representation for determining complex sentence similarity. \textit{Lastly}, while these approaches' ability to represent synonymy relations and carry out multi-word name recognition is recognised, their ability to discard parts of the text deemed irrelevant is well known to result in some difficulty with capturing higher-level knowledge structures \cite{Paper1}. 
  That said, if a word is then considered a stop word, it will not be used in the similarity learning mechanism, and the semantic information will be permanently lost. On the contrary, a learning approach exploiting either \gls{amr} and \gls{ud} graphs (\figurename~\ref{fig:dependency_sentences}, see \supplementaryname(\supplementaryname~\ref{nlpsec}) can potentially limit this information loss. Section \ref{sec:litrev_llm} discusses more powerful generative-based approaches that attempt to overcome the limitations above.

\subsubsection{Neural \acrfull{ir}}\label{litrev:nir}

\gls{ir} concerns retrieving full text documents given a full text query. 
Classical approaches tokenise the query into words of interest and retrieve documents within a corpus, ranking them based on the presence of tokens \cite{Manning_Raghavan}. Neural \gls{ir} improved over classical \gls{ir}, which was text-bound without considering semantic information. After representing queries and documents as vectors, relevance is computed through the dot product. Early versions exploited transformers, representing documents and queries as single vectors. Late interaction approaches like ColBERTv2 by Keshav et al. \cite{santhanam-etal-2022-colbertv2} provide a finer granularity representation by encoding the former into multi-vectors. After finding each document token maximising the dot product with a given query token, the final document ranking score is defined by summing all the maximising dot products. Training is then performed to maximise the matches of the given queries with human-annotated documents, marked as positive or negative matches for each query. Please observe that although this approach might help maximise the recall of the documents based on their semantic similarity to the query, the query tokenisation phase might lose information concerning the correlation between the different tokens occurring within the document, thus potentially disrupting any structural information occurring across query tokens. On the other hand, retaining semantic information concerning the relationships between entities leads us to a better logical and semantic representation of the text, as our proposed approach proves. This paper considers benchmarks against ColBERTv2 through the pre-trained RAGatouille v0.0.9 (\url{https://github.com/AnswerDotAI/RAGatouille}, Accessed on 22 April 2025) library.


\subsubsection{Generative \acrfull{llm}}\label{sec:litrev_llm}
As a result of the autoregressive tasks generally adopted by generative \gls{llm} models, when the system is asked about concepts on which it was not trained initially, it tends to invent misleading information \cite{Hicks2024}. This is inherently due to the probabilistic reasoning embedded within the model \cite{10.1145/3442188.3445922},  not accounting for inherent semantic contradiction implicitly resulting from the data through explicit rule-based approaches \cite{Grounding,bergami2019frameworksupportingimprecisequeries}. These models do not account for probabilistic reasoning by contradiction, with facts given as conjunctions of elements, leading to the inference of unlikely facts \cite{Kyburg1961-KYBPAT-2,bb}. 
 All these consequences lead to \textit{hallucinations}, which cannot be trusted to verify the inference outcome \cite{NASA}.
 
 \begin{figure}[p]
	\centering
		\subfloat[\gls{amr} for ``\textit{Newcastle and Brighton have traffic}'': ``\textit{and}'' is conjoining the subject with the direct object, losing grammatical distinction between them (\texttt{op1}, \texttt{op2}, \texttt{op3}).]{
			{\makebox[1\textwidth][c]{
				\includegraphics[width=0.6\textwidth]{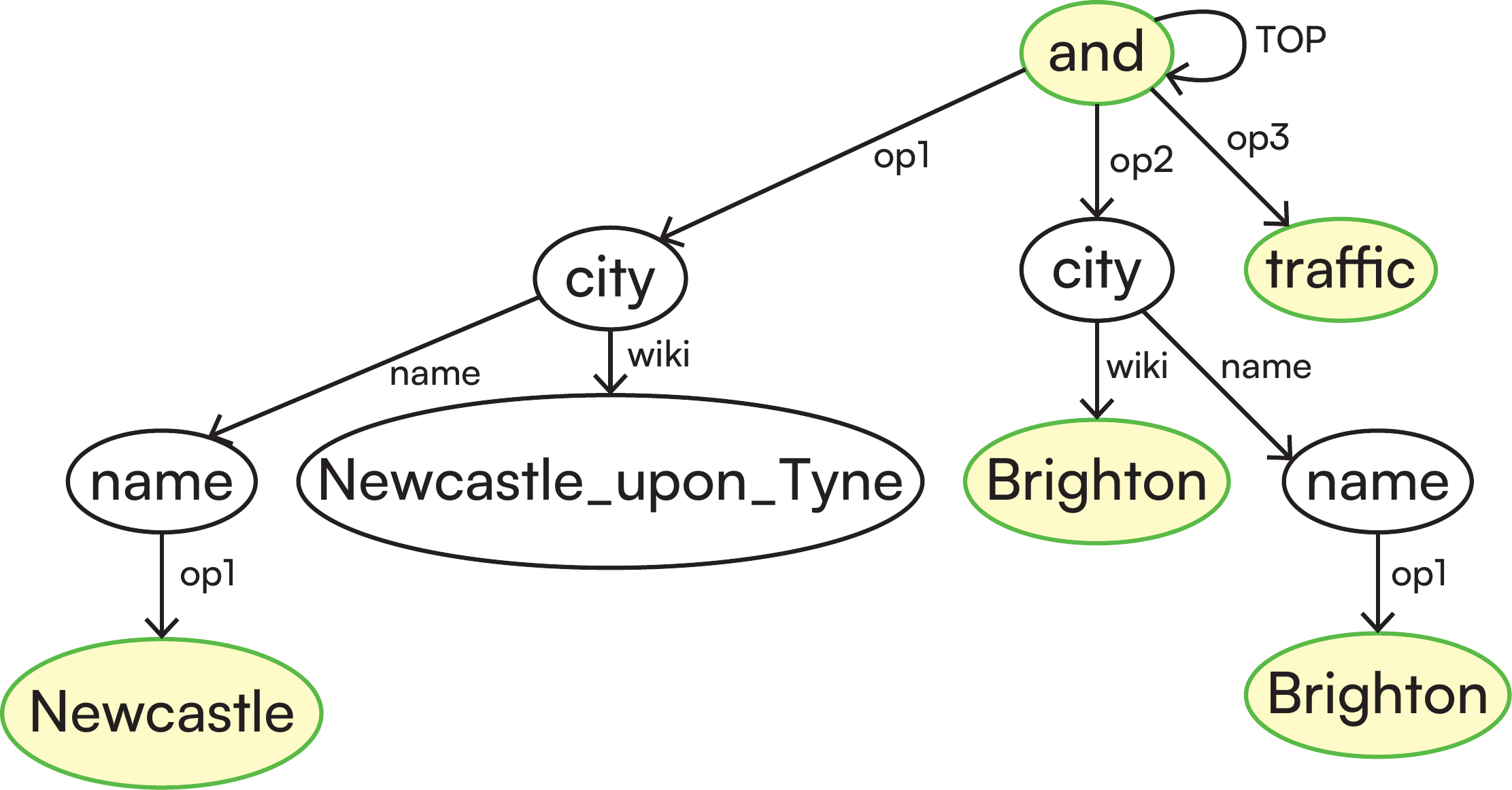}
				}
			}
		\label{fig:AMRbn}

	}
	\vspace{10pt}
\subfloat[POS tagging (below) and \gls{ud} (above) for ``\textit{Newcastle and Brighton have traffic}'': there is a clear distinction between subject and direct object, ``\textit{Newcastle}'' and ``\textit{Brighton}'' are joined by a \texttt{conj}, and the verb ``\textit{have}'' relates to ``\textit{traffic}'' through a \texttt{dobj}.]{
		{	
			\makebox[1\textwidth][c]{
				\includegraphics[width=0.6\textwidth]{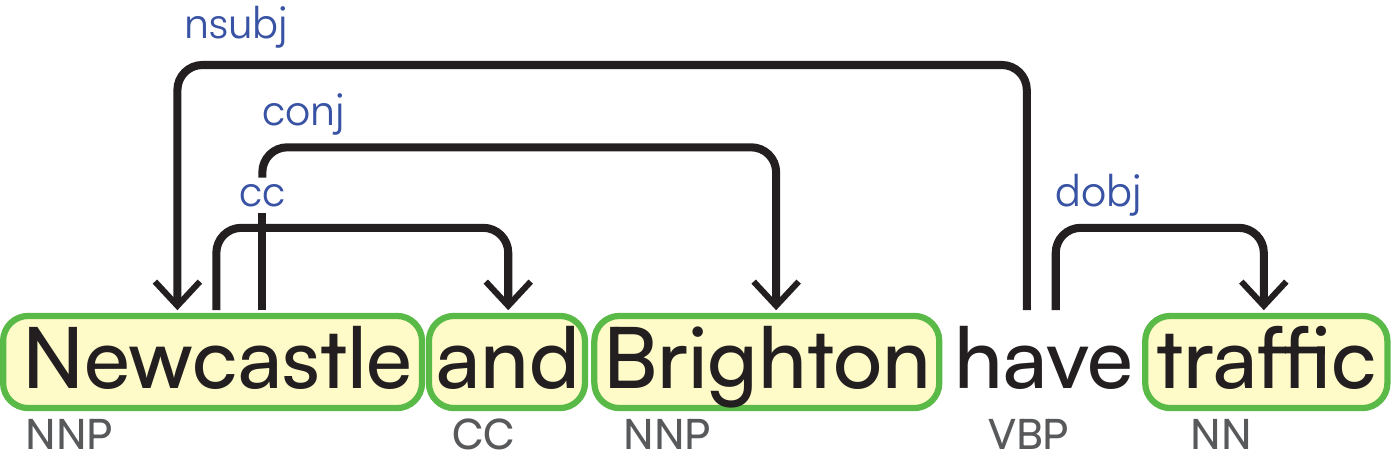}
			}
		}
		\label{fig:classic_sentence}
	}
	\vspace{10pt}
        	\subfloat[\gls{amr} for ``\textit{There is traffic in Newcastle but not in the city centre}'': despite \texttt{contrast-01} highlighting the opposite sentiment between ``\textit{traffic in Newcastle}'' and ``\textit{city centre}'', verb information is lost.]{
        	{
        		\makebox[1\textwidth][c]{
				\includegraphics[width=0.6\textwidth]{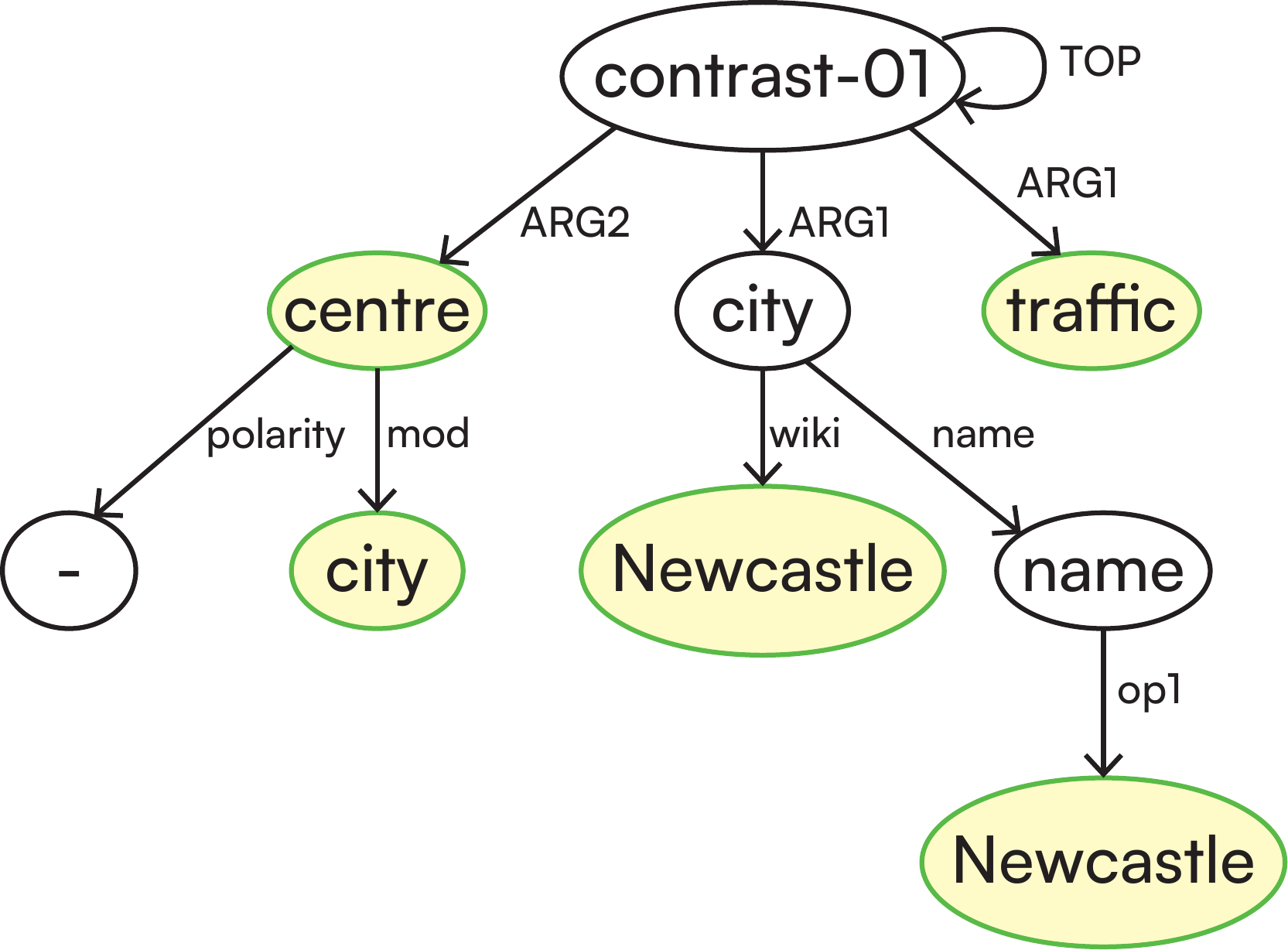}}}
		\label{fig:amrNCC}
	}
	\vspace{10pt}
\subfloat[POS tagging (below) and \gls{ud}(above) for ``\textit{There is traffic in Newcastle but not in the city centre}'': all the desired information is retained, besides a negation dependency that should be extracted from ``\textit{not}''.]{
	{\makebox[1\textwidth][c]{
		\includegraphics[width=0.815\textwidth]{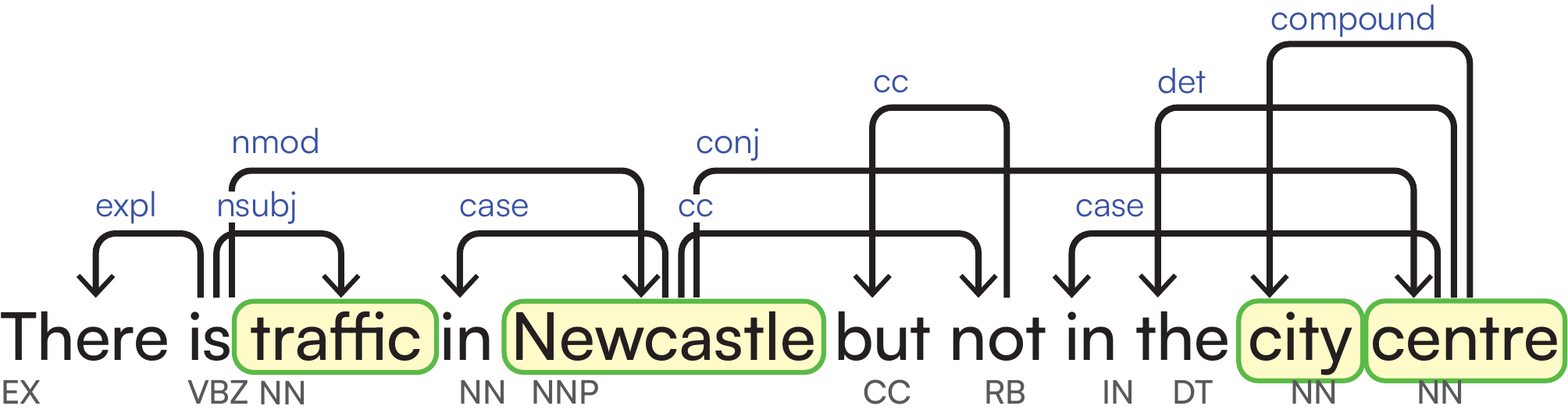}}}

		\label{fig:negation_sentences}
	}
	\caption{Visualisation of the differences between \gls{pos} tagging, \gls{amr} graphs, and \glspl{ud}, providing more explicit relationships between words. \glspl{amr} were generated through AMREager (\url{https://bollin.inf.ed.ac.uk}, Accessed on 24 April 2025), while \glspl{ud} were generated using StanfordNLP \cite{DBLP:conf/emnlp/ChenM14}. Graphs are highlighted to word correspondences, represented as graph nodes.}
	\label{fig:dependency_sentences}
\end{figure}

DeBERTaV2+AMR-LDA, proposed by Qiming Bao et al. \cite{bao-etal-2024-abstract}, is a state-of-the-art model supporting sentence classification through logical reasoning using a generative \gls{llm}. The model can conclude whether the first given sentence entails the second or not, thus attempting to overcome the above limitations of \glspl{llm}. After deriving an \gls{amr} of a full text sentence, the graphs are rewritten to obtain logically equivalent sentence representations for equivalent sentences. AMR-LDA is used to augment the input prompt before feeding it into the model, where prompts are given for logical rules of interest to classify the notion of logical entailment throughout the text. Contrastive learning is then used to identify logical implications by learning a distance measure between different sentence representations, aiming to minimise the distance between logically entailing sentences while maximising the distance between the given sentence and the negative example. This approach has several limitations: \textit{First}, the authors only considered equivalence rules that frequently occur in the text and not all of the possible equivalence rules, thus heavily limiting the reasoning capabilities of the model. \textit{Second}, in doing so, the model does not exploit contextual information from the knowledge graphs to consider part-of and is-a relationships relevant for deriving correct entailment implications within spatiotemporal reasoning. \textit{Third}, due to the lack of paraconsistent reasoning, the model cannot clearly distinguish whether the missing entailment is due to inconsistency or whether the given facts are not correlated. \textit{Lastly}, the choice of using \gls{amr} heavily impacts the ability of the model to correctly distinguish different logical functions of the sentence within the text. 

The present study overcomes the limitations above in the following manner: \textit{First}, we avoid hardcoding all possible logical equivalence rules by interpreting each formula using classical Boolean-valued semantics for each atom within the sentences. After generating a truth table with all the atoms, we then evaluate the Boolean-valued semantics for each atom combination (\appendixname~\ref{app:formulartabular}). In doing so, we avoid the explosion problem by reasoning paraconsistently, thus removing the conflicting worlds (also \appendixname~\ref{app:formulartabular}). \textit{Second}, we introduce a new compact logical representation, where entities within the text are represented as functions (Section \ref{sec:logrewr}); the logical entailment of the atoms within the logical representation is then supported by a \gls{kb} expressing complex part-of and is-a relationships (\appendixname~\ref{propSem}). \textit{Third}, we consider a three-fold classification score through the \texttt{confidence} score (Definition \ref{def:confidence}): while $1.0$ and $0.0$ can be used to differentiate between implication and inconsistency, intermediate values will capture indifference. \textit{Lastly}, we use \gls{ud} graphs rather than \gls{amr} graphs (\supplementaryname~\ref{nlpsec} and \ref{sec:igc}), similarly to recent attempts at providing reliable rule-based \gls{qa} \cite{10.1007/978-3-031-38499-8_29}. 

This study considered benchmarking against the pre-trained LLM classifier, which was made available through HuggingFace by the original paper's authors (\texttt{AMR-LE-DeBERTa-V2-\\XXLarge-Contraposition-Double-Negation-Implication-Commutative-Pos-Neg-1-3}).


\startlandscape
\begin{figure}[H]
	\centering
	\includegraphics[width=\textwidth]{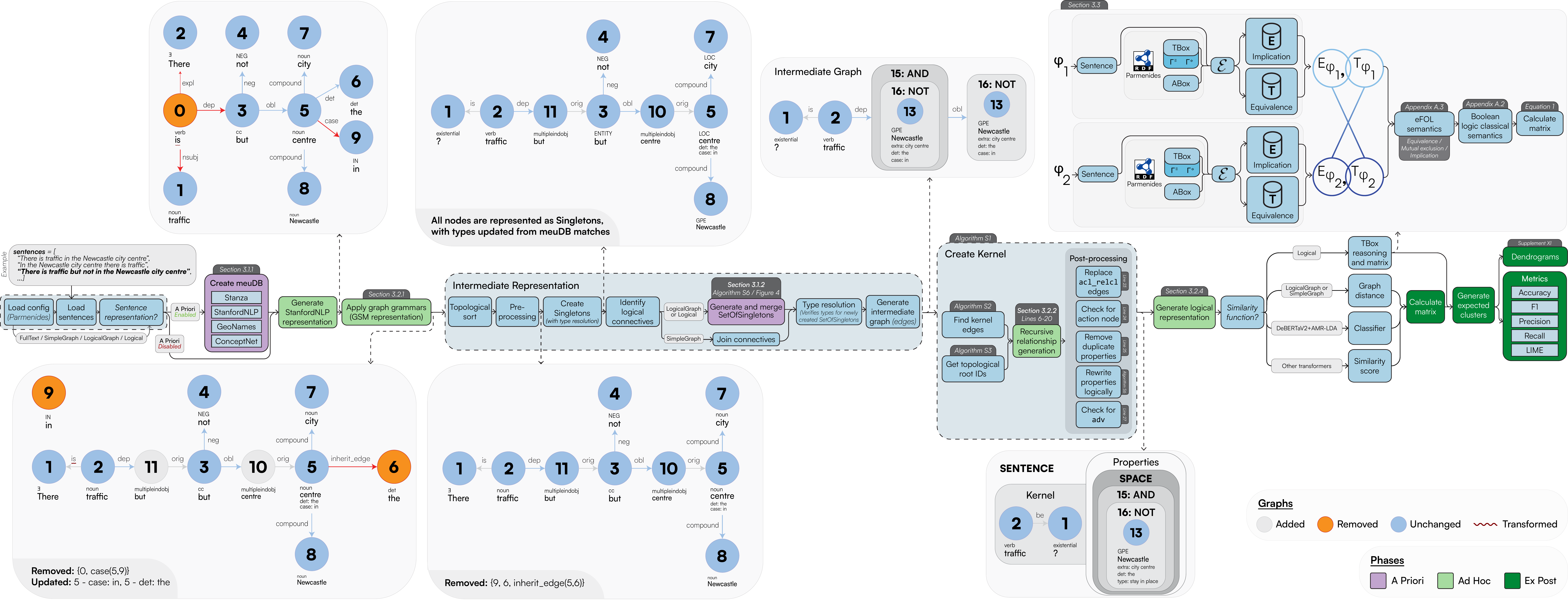}
	\caption{Detailed view of \figurename~	\ref{fig:pipeline}: The pipeline shows a running example of the sentence \#2 from \ref{rqn2c}, ``\textit{There is traffic but not in the Newcastle city centre}'': graphs provide the representation returned by specific pipeline tasks, where colours highlight the performed changes. We retain colours from \figurename~\ref{fig:pipeline} for linking across the same tasks. Due to page limitations, we refer to \figurename~\ref{fig:setofsingletonsgeneration} for a detailed description of the transformation needed to generate an Intermediate Graph after identifying the logical connectives. We also refer to Sentence 2 occurring in \figurename~\ref{fig:lassi_explanations} for a graphical representation of both the final logical representation of the sentence (Section \ref{sec:logrewr}) as well as a high-level representation of the reasoning process (Section \ref{sec:expostexmpl})} \label{fig:full-example}
\end{figure}
\finishlandscape

\section{Materials and Methods}\label{sec:methodology}
\textit{Let $\alpha$ and $\beta$ be full text sentences. In this paper, we consider only factoid sentences that can at most represent existentials, expressing the omission of knowledge to, at some point, be injected with new, relevant information. $\tau$ represents a transformation function, in which the vector and logical representations are denoted as $\tau(\alpha)=A$ and $\tau(\beta)=B$ for $\alpha$ and $\beta$, respectively.}

\begin{figure}[H]
	\includegraphics[width=1\linewidth]{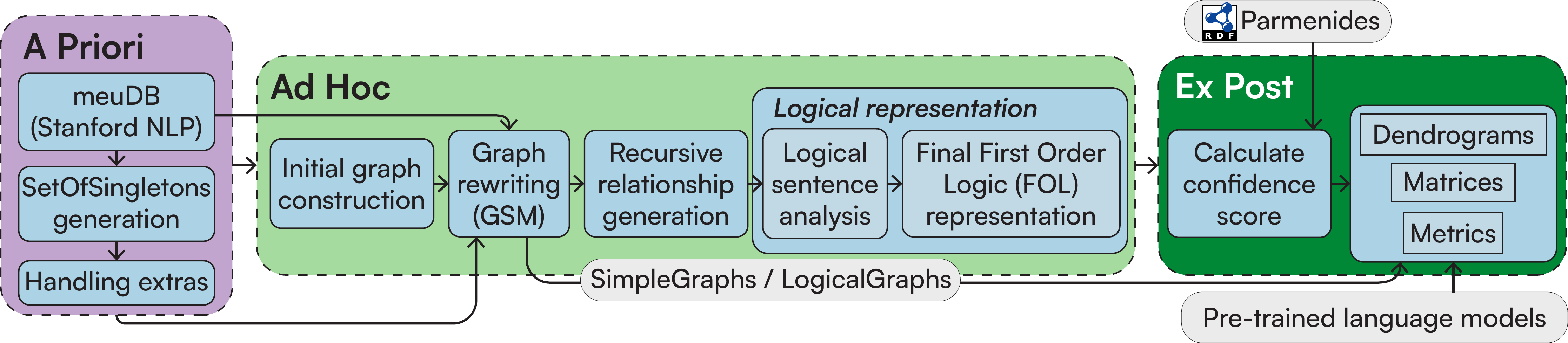}
	\caption{LaSSI Pipeline: Operational description of the pipeline, reflecting the outline of this section.}
	\label{fig:pipeline}
\end{figure}

\textit{From $\tau$, we want to derive a logical interpretation through $\varphi$ while capturing the common-sense notions from the text. 
We then need a binary function $\varphi_\tau$ that expresses this for each transformation $\tau$ (Section \ref{theorimposs}). \figurename~\ref{fig:pipeline} offers a birds-eye view of the entire pipeline as narrated in the present paper. \figurename~\ref{fig:full-example} details the former by adding references to specific parts of the paper while providing a running example.}

\subsection{A Priori}\label{sub:apriori}
\textit{In the \textit{a priori} explanation phase, we aim to enrich the semantic information for each word (Section \ref{sec:san}) to subsequently recognise multi-word entities (Section \ref{subsub:setofsingletons}) with extra information (i.e., specifications, \supplementaryname~\ref{typehier}) by leveraging the former. This information will be used to pair the intermediate syntactic and morphological sentence representation achieved through subsequent graph rewritings (Section \ref{sec:adhoc}) with the semantic interpretation derived from the phase narrated within the forthcoming subsections.}

The main data structure used to nest dependent clauses represented as relationships occurring at any recursive sentence level is the \texttt{Singleton}. This is a curated class used throughout the pipeline to portray entities within the graph from the given full text. This also represents the atomic representation of an entity (Section \ref{sec:san}); it includes each word of a multi-word entity (Section \ref{subsub:setofsingletons}) and is defined with the following attributes: \texttt{id}, \texttt{named\_entity}, \texttt{properties}, \texttt{min}, \texttt{max}, \texttt{type}, \texttt{confidence}, and \texttt{kernel}. When \texttt{kernel} is none, properties mainly refer to the entities, thus including the aforementioned specifications (\supplementaryname~\ref{typehier}); otherwise, they refer to additional entities leading to logical functions associated with the sentence. \texttt{Kernel} is used when we want to represent an entire sentence as a coarser-grained component of our pipeline: this is defined as a \texttt{relationship} between a source and target mediated by an edge label (representing a verb), while an extra Boolean attribute reflects its \textit{negation} (Section \ref{sec:recursive}). 
The source and target are also \texttt{Singletons} as we want to be able to use a kernel as a source or target of another kernel (e.g., to express causality relationships) so that we have consistent data structures across all entities at all stages of the rewriting. The \texttt{properties} of the kernel could include spatiotemporal or other additional information, represented as a dictionary, which is used later to derive logical functions through logical sentence analysis (Section \ref{sec:lsa}).

\subsubsection{Syntactic Analysis using Stanford CoreNLP}\label{sec:san} This step aims to extract syntactic information from the input sentences $\alpha$ and $\beta$ using Stanford CoreNLP. A Java service within our LaSSI pipeline utilises Stanford CoreNLP to process the full text, generating annotations for each word. These annotations include base forms (lemmas), \gls{pos} tags, and morphological features, providing a foundational understanding of the sentence structure while considering entity recognition. 

The \textit{\gls{meudb}} contains information about all variations of each word in a given full text. This could refer to American and British spellings of a word like ``\textit{centre}'' and ``\textit{center}'', or typos in a word like ``\textit{interne}'' instead of ``\textit{internet}''. Each entry in the \gls{meudb} represents an entity match appearing within the full text, with some collected from specific \texttt{sources}, including \textbf{GeoNames} \cite{10.1145/2533888.2533938} for geographical places, \textbf{SUTime} \cite{sutime} for recognising temporal entities, \textbf{Stanza} \cite{qi2020stanza} and our curated \textbf{Parmenides ontology} for detecting entity types, and \textbf{ConceptNet} \cite{DBLP:conf/aaai/SpeerCH17} for generic real-world entities. 
Depending on the trustworthiness of each source, we also associate a confidence weight: for example, as the GeoNames gazetteer contains noisy entity information \cite{10.1145/2533888.2533938}, we multiply the entity match uncertainty by 0.8 as determined in our previous research  \cite{ideas2024b}. Each match also carries the following additional information:

\begin{itemize}
	\item \texttt{\textbf{start}} and \texttt{\textbf{end}} characters respective to their character position within the sentence: these constitute provenance information that is also pertained in the \textit{ad hoc} explanation phase (Section \ref{sec:adhoc}), thus allowing the enrichment of purely syntactic sentence information with a more semantic one.
	\item \texttt{\textbf{text}} value referring to the original matched text.
	\item \texttt{\textbf{monad}} for the possible replacement value:
	\begin{itemize}
		\item \supplementaryname~\ref{supp:handleincspell}  details that this might eventually replace words in the logical rewriting stage.
	\end{itemize}
\end{itemize}

Changes were made to the \gls{meu} matching to improve its efficiency in recognising all possibilities of a given entity. In our previous solution, only the original text was used. Now, we perform a fuzzy match through PostgreSQL for lemmatised versions of given words \cite{postgres} rather than through Python code directly to boost the recognition of multi-word entities by assembling other single-word entities. Furthermore, when generating the resolution for \glspl{meu}, a \textit{typed match} is also performed when no match is initially found from Stanford NLP, so the type from the \gls{meudb} is returned for the given \gls{meu}.

This categorisation subsequently allows the representation of each single named entity occurring in the text to be represented as a \texttt{Singleton} as discussed before.

\subsubsection{Generation of \texttt{SetOfSingletons}}\label{subsub:setofsingletons}

\begin{figure}[H]
	\centering
	\includegraphics[width=.8\linewidth]{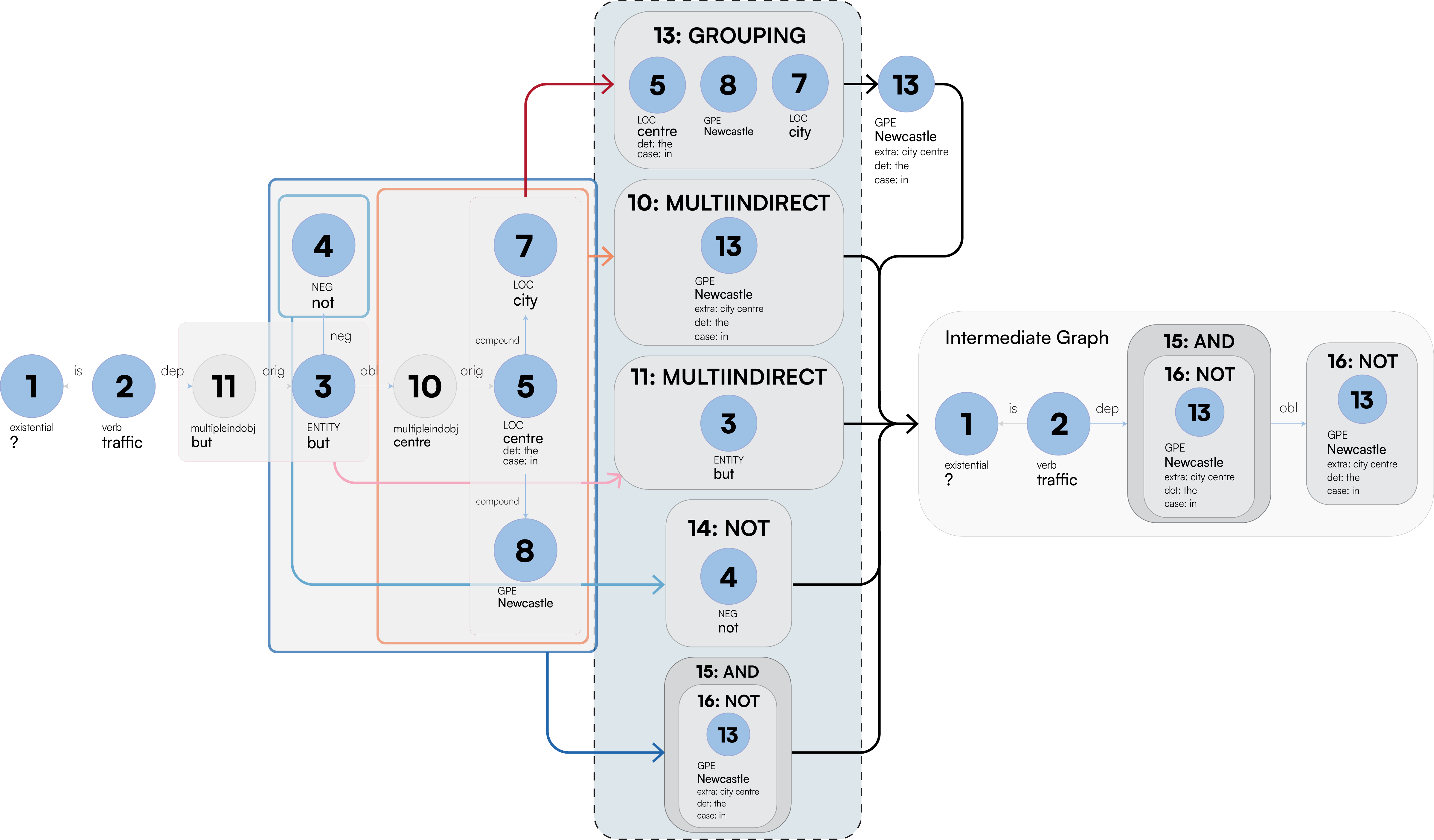}
	\caption{Continuing the example from \figurename~\ref{fig:full-example}, we show how different types of \texttt{SetOfSingletons} generated from distinct \gls{ud} relationships lead to Intermediate Graphs. We showcase coordination (e.g., \texttt{AND} and \texttt{NOT}), multi-word entities (e.g., \texttt{GROUPING}), and multiple logical functions (e.g. \texttt{MULTIINDIRECT}). The sequence of changes highlighted in the central column are applied by visiting the graph in lexicographical order \cite{math12172677} so to start changing the graph from the n+odes with fewer edge dependencies.}
	\label{fig:setofsingletonsgeneration}
\end{figure}

A \texttt{SetOfSingletons} is a specific type of \texttt{Singleton} containing multiple \texttt{entities}, an array of \texttt{Singleton}s. 
As showcased by \figurename~\ref{fig:setofsingletonsgeneration}, a group of items is generated by coalescing distinct entities grouped into clusters as indicated by \glspl{ud} relationships, such as the coordination of several other entities or sentences (\texttt{conj}), the identification of multi-word entities (\texttt{compound}), or the identification of multiple logical functions attributed to the same sentence (\texttt{multipleindobj}, derived after the \gls{ggg} rewriting of the original \glspl{ud} graph). Each \texttt{SetOfSingletons} can be associated with types.

We now illustrate the proposed \texttt{SetOfSingleton} type according to the application order from the example given in 	\figurename~\ref{fig:setofsingletonsgeneration}:

\begin{description}
	\item[Multi-Word Entities:] Algorithm \ref{alg:merge_singletons} performs node grouping \cite{DBLP:conf/btw/JunghannsPR17} over the the nodes connected by \texttt{compound} edge labels while efficiently visiting the graph using a \gls{dfs} search. After this, we identify whether a subset of these nodes acts as a specification (\texttt{extra}) to the primary entity of interest or whether it should be treated as a single entity. This is computed as follows: after generating all the possible ordered grouping of words, we associate each group to a type as derived by their corresponding \gls{meudb} match. Through the typing information, we then decide to keep the most specific type as the main entity, while leaving the most general one as a specification (\texttt{extra}). While doing so, we also consider the confidence of the fuzzy string matching through the \gls{meudb}. \supplementaryname~\ref{mweandextras} provides further algorithmic details on how LaSSI performs this computation.
\end{description}

\begin{figure}[H]
		\centering
		\includegraphics[width=\textwidth]{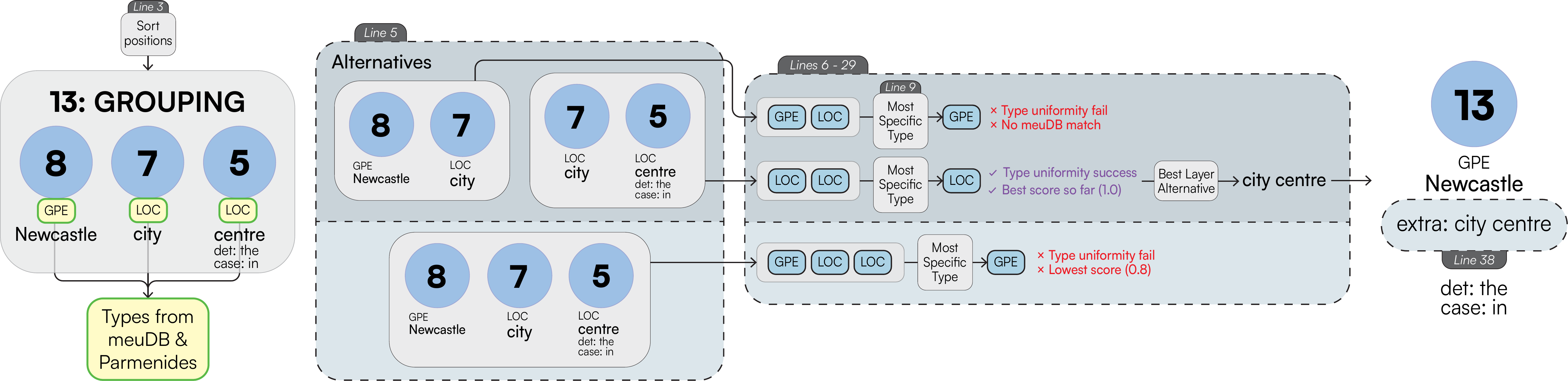}
		\caption{Continuing the example from \figurename~\ref{fig:setofsingletonsgeneration}, we display how \texttt{SetOfSingletons} with type \texttt{GROUPING} are rewritten into \texttt{Singletons} with an \texttt{extra} property, modifying the main entity with additional spatial information. Line numbers refer to distinct subsequent phases from  Algorithm \ref{alg:merge_singletons}.}
		\label{fig:algo1explain}
	\end{figure}

	\begin{example}
		After coalescing the \texttt{compound} relationships from \figurename~\ref{fig:setofsingletonsgeneration}, we would like to represent the grouping ``\textit{Newcastle city centre}'' as a \texttt{Singleton} with a {core} entity ``\textit{Newcastle}'' and an \texttt{extra} ``\textit{city centre}''. \figurename~\ref{fig:algo1explain} sketches the main phases of Algorithm \ref{alg:merge_singletons} leading to this expected result. For our example, the possible ordered permutations of the entities within \texttt{GROUPING} are: ``\textit{Newcastle city}'', ``\textit{city centre}'', and ``\textit{Newcastle city centre}''. Given these alternatives, ``\textit{Newcastle city centre}'' returns a confidence of 0.8 and ``\textit{city centre}'' returns the greatest confidence of \texttt{1.0}, so our chosen alternative is [\texttt{city}, \texttt{centre}]. As ``\textit{Newcastle}'' is the entity having the most specific type, this is selected as our \texttt{chosen\_entity}, and subsequently, ``\textit{city centre}'' becomes the \texttt{extra} property to be added to ``\textit{Newcastle}'', resulting in our final \texttt{Singleton: Newcastle[extra:city centre]}.
		
		For Simplistic Graphs, ``\textit{Newcastle upon Tyne}'' would be represented as one \texttt{Singleton} with no \texttt{extra} property. 
	\end{example}

\begin{description}
	\item[Multiple Logical Functions:] Due to the impossibility of graphs to represent n-ary relationships, we group multiple adverbial phrases into one \texttt{SetOfSingleton}. These will be then semantically disambiguated by their function during the Logical Sentence Analysis (Section \ref{sec:lsa}). \figurename~\ref{fig:setofsingletonsgeneration} provides a simple example, where each \texttt{MULTIINDIRECT} contains either one adverbial phrase or a conjunction. \supplementaryname~\ref{mweandextras} provides a more compelling example, where such \texttt{SetOfSingleton} actually contains more \texttt{Singletons}.
	
	\item [Coordination:] For coordination induced by \texttt{conj} relationships, we can derive a coordination type to be \texttt{AND}, \texttt{NEITHER}, or \texttt{OR}. This is derived through an additional \texttt{cc} relationship on any given node through a \gls{bfs} that will determine the type.
\end{description}

Last, LaSSI also handles \texttt{compound\_prt} relationships; unlike the above, these are coalesced into one \texttt{Singleton} as they represent a compound word:  $(shut)\xrightarrow{compound\_prt}(down)$ becomes $(shut\ down)$, and are not therefore represented as a \texttt{SetOfSingleton}.  

%

\subsection{Ad Hoc}\label{sec:adhoc}

\begin{figure}[H]
	\includegraphics[width=\linewidth]{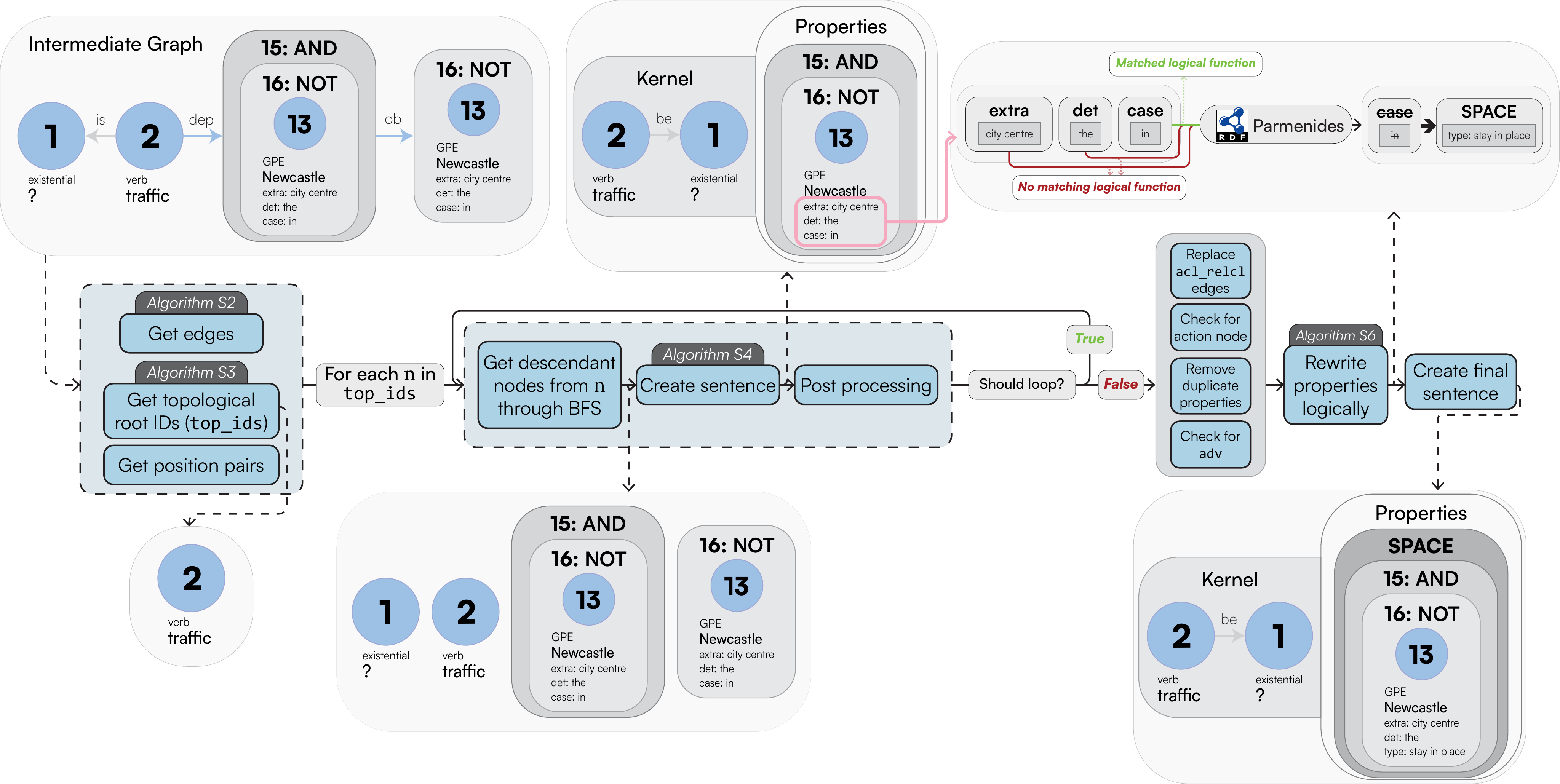}
	\caption{A broad view of the \textit{ad hoc} phase of the pipeline: with the intermediate graph generated, we iterate over each topologically sorted root node ID and create the final sentence representation.}
	\label{fig:rewrite_props2}
\end{figure}

This phase provides a gradual ascent of the data representation ladder through which raw full text data are represented as logical programs via intermediate rewriting steps (\figurename~\ref{fig:rewrite_props2}), thus achieving the most expressive representation of the text. As this provides an algorithm to extract a specification from each sentence, providing both a human- and machine-interpretable representation, we refer to this phase as an \textit{ad hoc}  explanation phase, where information is ``mined'' structurally and semantically from the text.


The transformation function, $\tau$, takes our full text enriched with semantic information from the previous phase and rewrites it into a final suitable format whereby a semantic similarity metric can be used: either a vector-based cosine similarity, which is a traditional graph-based similarity metric where both node and edge similarity are given through vector-based similarity, potentially capturing the logical connectives represented within each node; or our proposed support-based metric requiring a logical representation for the sentences. These are then used to account for a different transformation function $\tau$: when considering classical vector-based transformers, we consider those available through the HuggingFace library. For our proposed logical approach, the full text must be transformed as we need a representation that the system can understand to calculate an accurate similarity value produced from only relevant information. 

\begin{figure}[h]
	\centering
	
	\begin{minipage}{\textwidth}
		\centering
		\begin{minipage}{\textwidth}
			\centering
			\includegraphics[width=\linewidth,height=5cm,keepaspectratio]{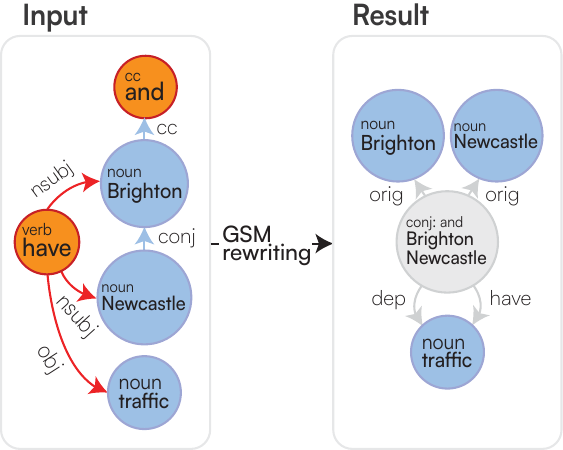}
			\vspace{20pt}
		\end{minipage}
		\begin{minipage}{0.55\textwidth}
			\centering
			\includegraphics[width=\linewidth,height=7cm,keepaspectratio]{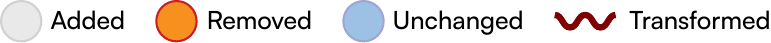}
		\end{minipage}
	\end{minipage}
	
			

	\caption{Transforming the \gls{ggg} rewriting of a \gls{ud} graph into a graph Intermediate Representation for ``\textit{Newcastle and Brighton have traffic''}.}
	\label{fig:transformations}\label{fig:input}
\end{figure}

To obtain this, we have distinct subsequent rewriting phases, where more contextual information is gradually added on top of the original raw information: after generating a semistructured representation of the full text by enriching the text with \glspl{ud} as per Stanford NLP (\textbf{Input} in \figurename~\ref{fig:input}, \supplementaryname~\ref{sec:igc}), we apply a preliminary graph rewriting phase that aims to generate similar graphs for sentences, where one is the permutation of the other or simply differs from the active/passive form (\textbf{Result} in \figurename~\ref{fig:input}, Section \ref{gsmrewr}). 
At this stage, we also derive a cluster of nodes (referred to as the \texttt{SetOfSingletons}) that can be used later on to differentiate the main entity to the concept that the kernel entity is referring to (\supplementaryname~\ref{typehier}). After this, we acknowledge the recursive nature of complex sentences by visiting the resulting graph in topological order, thus generating minimal sentences first (\textit{kernels}) to then merge them into a complex and nested sentence structure (Section \ref{sec:recursive}). After this phase, we extract each linguistic \textit{logical function} occurring within each minimal sentence using a rule-based approach exploiting the semantic information associated with each entity as derived from the \textit{a priori} phase (Section \ref{sec:lsa}). This then leads to the final logical form of a sentence (Section \ref{sec:logrewr}), generating the following logical representation for \figurename~\ref{fig:input}:
\[\mathtt{has(Newcastle, traffic)\wedge\mathtt{has(Brighton, traffic)}}\]

\subsubsection{Graph Rewriting with the \gls{gsm}}\label{gsmrewr}
This step employs the proposed \gls{gsm} \cite{10.1145/3609429.3609433} to refine the initial graph and capture shallow semantic relationships, merely acknowledging the syntactic nature of the sentence without accounting for the inner node semantic information. Traditional graph rewriting methods, such as those for property graphs \cite{bonifati2024transformingpropertygraphs}, are  insufficient for expressing rewriting rules such as the ones required for Montague Grammar (Supplement \ref{sub:grammatical_structure}). They struggle with creating entirely new graph structures or require restructuring existing ones. To overcome these limitations, we leverage graph grammars \cite{math12172677} within the DatagramDB framework. The DatagramDB database rewrites the initial graph represented using a \gls{gsm} data structure, incorporating logical connectives and representing verbs as edges between nodes as shown in \figurename~\ref{fig:input}; among the other operations, this phase normalises active and passive verbs by consistently generating one single edge per verb. Here, the source identifies the entity performing the action described by the edge label. For transitive verbs, the targets might provide information regarding the entity receiving the action from the agent. This restructuring better reflects the syntactic structure and prepares the graph for the final logical rewriting step. If this does not occur, we either flatten out each \texttt{SetOfSingleton} node into one \texttt{Singleton} node (\textbf{\glspl{sg}}) or we only retain the logical connectors and flatten out the rest (\textbf{\glspl{lg}}). Thus, all forthcoming substeps are considered relevant to obtaining only the final logical representation of a sentence (Section \ref{sec:recursive}, Section 3.2.3, and Section \ref{sec:logrewr}). Given that the scope of our work is on the main semantic pipeline and not on actual graph rewriting queries, which were already analysed in our previous work \cite{math12172677}, we refer to the online query for more information on the rewriting covered by our current solution (\url{https://github.com/LogDS/LaSSI/blob/32ff1df2df7d824619f9a84e7ae7d7f6e4842cb0/LaSSI/resources/gsm\_query.txt}, Accessed on 29 March 2025). 

\subsubsection{Recursive Relationship Generation}\label{sec:recursive}
{In this phase, we carry out some additional graph rewriting operations that generate binary relationships representing unary and binary predicates by considering semantic information for both edge labels and the topological orders of the sentences. While the former are clearly represented as binary relationships with a \textbf{none} target argument and usually refer to intransitive verbs, the latter are usually associated with transitive verbs. Either subjects or targets explicitly missing from the text and not expressed as pronouns are resolved as fresh variables, which will then be bound in the logical rewriting phase into an existential quantifier. Given that this phase also needs to consider semantic information, this rewriting cannot be completely covered by any graph grammar or general \gls{gql} and, therefore, cannot be entirely addressed in the former phase. This motivates us to hardcode this phase directly in a programming language rather than flexibly represent this through rewriting rules like any other phase within the pipeline. }

Unlike in our previous contribution \cite{ideas2024b}, we now cover the recursive nature of subordinate clauses \cite{facultyOfLanguage} by employing a \gls{dfs} topological sort \cite{math12172677}, whereby the deepest nodes in our always acyclic graph are accounted for first. Previously, no pre-processing occurred and the graph was read in directly from the rewritten \gls{ggg} output. 

\begin{figure}[H]
	\includegraphics[width=\linewidth]{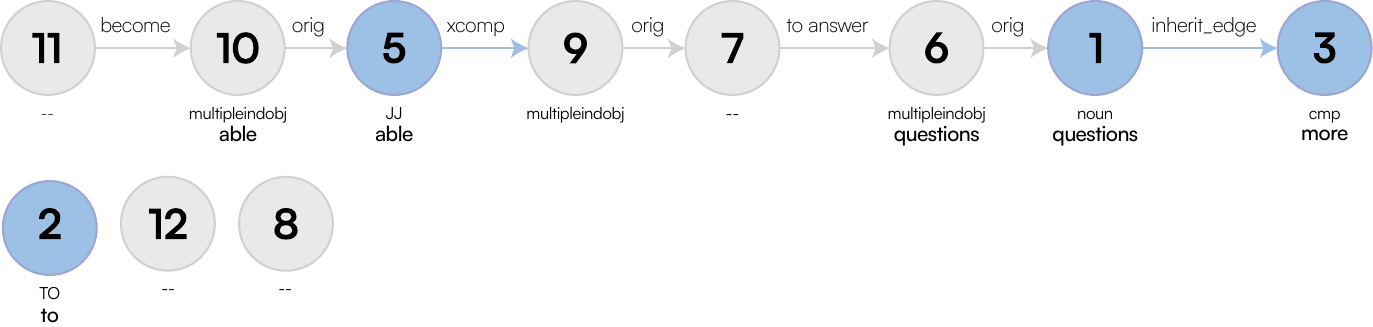}
	\caption{\gls{ggg} rewritten graph for the sentence ``\textit{become able to answer more questions}'' from Example \ref{ex2}: node IDs are overlayed in the middle of each node. {Each node represents a word in the sentence, and each edge represents a \gls{ud}.}}
	\label{fig:top_sort_ex}
\end{figure}

\begin{example}\label{ex2}
	Figure \ref{fig:top_sort_ex} shows an example output from DatagramDB. The generated JSON file lists the IDs in the following order: 1, 6, 7, 8, 9, 10, 11, 12, 5, 2, 3. However, once our topological sort is performed, this becomes 3, 1, 2, 6, 7, 9, 5, 8, 10, 11, 12, where our `deepest' nodes are at the start of the list. Subsequent filtering culls nodes from the list that are no longer needed: the edge label between nodes 1 and 3 is \texttt{inherit\_edge}, which means all properties of node 3 are added to node 1, and thus, node 3 is removed. Nodes 12 and 8 contain no information, so they can also be removed. Finally, node 2 (``to'') has already been inherited into the edge label ``\textit{\textbf{to} answer}'', so it is also removed (because it does not have any parents or children). This results in the final sorted list: 1, 6, 7, 9, 5, 10, 11. Our list of \texttt{nodes} within the pipeline is kept in this topological order throughout. Therefore, we can retrieve all {roots} from the graph to create our {kernels}. 
\end{example}

Unlike the previous simplistic example, most real-world sentences often have a hierarchical structure, where components within the sentence depend on prior elements \cite{math12172677}. Topological sorts then take care of these more complex situations.

Algorithm \ref{alg:overall_kernel_construction} discussed in \supplementaryname~\ref{app:recursiveSentenceRewr} considers each clause root in inverse topological order, thus rewriting each root-induced subgraph into a nested relationship that can be ultimately represented as a nested \texttt{Singleton}. While doing so, we recognise \textit{semi-modal verbs} and appropriately recognise their prepositions, while differentiating them from \textit{phrasal verbs}. In this phase, we also provide pronoun resolution by replacing them with the entity they refer to as indicated by \gls{ud} semantic parsing. 


\subsubsection{Logical Sentence Analysis}\label{sec:lsa} 
Given the properties extracted from the previous phase\footnote{The previous phase provided a preliminary rewriting, where a new relationship is derived from each verb occurring within the pipeline and connecting the agents performing and receiving the action; information concerning additional entities and pronouns occurring within the sentence is collected among the properties associated with the relationship.}, we now want to rewrite such properties by associating each occurring entity with its logical function within the sentence and recognising any associated adverb or preposition while considering the type of verb and entity of interest. This rewriting mechanism exploits simple grammar rules found in any Italian linguistic book (\supplementaryname~\ref{sec:itling}), and is therefore easily accessible. To avoid hardcoding these rules in the pipeline, we declaratively represent them as instances of \texttt{LogicalRewriteRule} concepts within our Parmenides Upper Ontology (\figurename~\ref{code:logical_rule_space}). These rules can be easily interpreted as Horn clauses (\figurename~\ref{subfig:asLogic}). These are used in LaSSI to support further logical functions by extending the rules within the ontology rather than changing the codebase.

\begin{figure}[!h]
	\begin{subfigure}[t]{0.45\textwidth}
		\begin{minted}[fontsize=\footnotesize, framesep=2mm, autogobble, breaklines]{turtle}
			parmenides:logrule%2F1 a parmenides:LogicalRewritingRule ;
			rdfs:label "logrule/1"^^xsd:string ;
			parmenides:SingletonHasBeenMatchedBy "SUTime"^^xsd:string ;
			parmenides:logicalConstructName "time"^^xsd:string ;
			parmenides:logicalConstructProperty "defined"^^xsd:string ;
			parmenides:preposition "as soon as"^^xsd:string,
			"at"^^xsd:string,
			"in"^^xsd:string,
			"on"^^xsd:string ;
			parmenides:rule_order 1 .
			parmenides:logrule%2F12 a parmenides:LogicalRewritingRule ;
			rdfs:label "logrule/12"^^xsd:string ;
			parmenides:abstract_entity false ;
			parmenides:logicalConstructName "space"^^xsd:string ;
			parmenides:logicalConstructProperty "stay in place"^^xsd:string ;
			parmenides:preposition "in"^^xsd:string,
			"into"^^xsd:string ;
			parmenides:rule_order 12 .
		\end{minted}
		\caption{Rewriting rules \#~1 and \#~12 for generating logical functions from kernel properties.}\label{code:logical_rule_space}\label{code:logical_rule_time}
	\end{subfigure}%
	~ 
	\begin{subfigure}[t]{0.45\textwidth}
		\begin{minted}[fontsize=\footnotesize, framesep=2mm, autogobble, breaklines]{turtle}
			<https://logds.github.io/parmenides#log%2Fspace%2Fstay+in+place> a parmenides:LogicalFunction ;
			rdfs:label "log/space/stay in place"^^xsd:string ;
			parmenides:argument "property"^^xsd:string ;
			parmenides:attachTo "Kernel"^^xsd:string ;
			parmenides:logicalConstructName "space"^^xsd:string ;
			parmenides:logicalConstructProperty "stay in place"^^xsd:string .
			parmenides:log%2Ftime%2Fdefined a parmenides:LogicalFunction ;
			rdfs:label "log/time/defined"^^xsd:string ;
			parmenides:argument "property"^^xsd:string ;
			parmenides:attachTo "Kernel"^^xsd:string ;
			parmenides:logicalConstructName "time"^^xsd:string ;
			parmenides:logicalConstructProperty "defined"^^xsd:string .
		\end{minted}
		\caption{Characterisation of where the logical functions for \texttt{SPACE} of type \texttt{stay in place} and \texttt{TIME} of type \texttt{defined} should reside, that is, as properties of the relationship.}\label{code:logical_function_space}
	\end{subfigure}
	\vspace{.5cm}
	
	
	\begin{subfigure}[t]{\textwidth}
		{
			\[\begin{split}
				(\{\textrm{``in''}, \textrm{ ``into''}\} \cap \texttt{singleton}) \ne \emptyset \wedge \texttt{hasAbstractEntity(singleton)} \Rightarrow \\
				\qquad \texttt{SPACE: singleton[type:stay in place]}
			\end{split}\]
			\[\begin{split}
				(\{\textrm{``as soon as''}, \textrm{``at''}, \textrm{ ``in''}, \textrm{``on''}\} \cap \texttt{singleton}) \ne \emptyset \wedge \texttt{singleton}\colon\texttt{SUTime} \Rightarrow\\ \qquad \texttt{TIME: singleton[type:defined]}
			\end{split}\]
		}
		
		\caption{Logical representation of the rules in \figurename~\ref{code:logical_rule_space}.}\label{subfig:asLogic}
	\end{subfigure}
	\caption{Fragments of the Parmenides Upper Ontology encoding rules for capturing logical functions. } 
\end{figure}

\begin{figure}[H]
	\includegraphics[width=\linewidth]{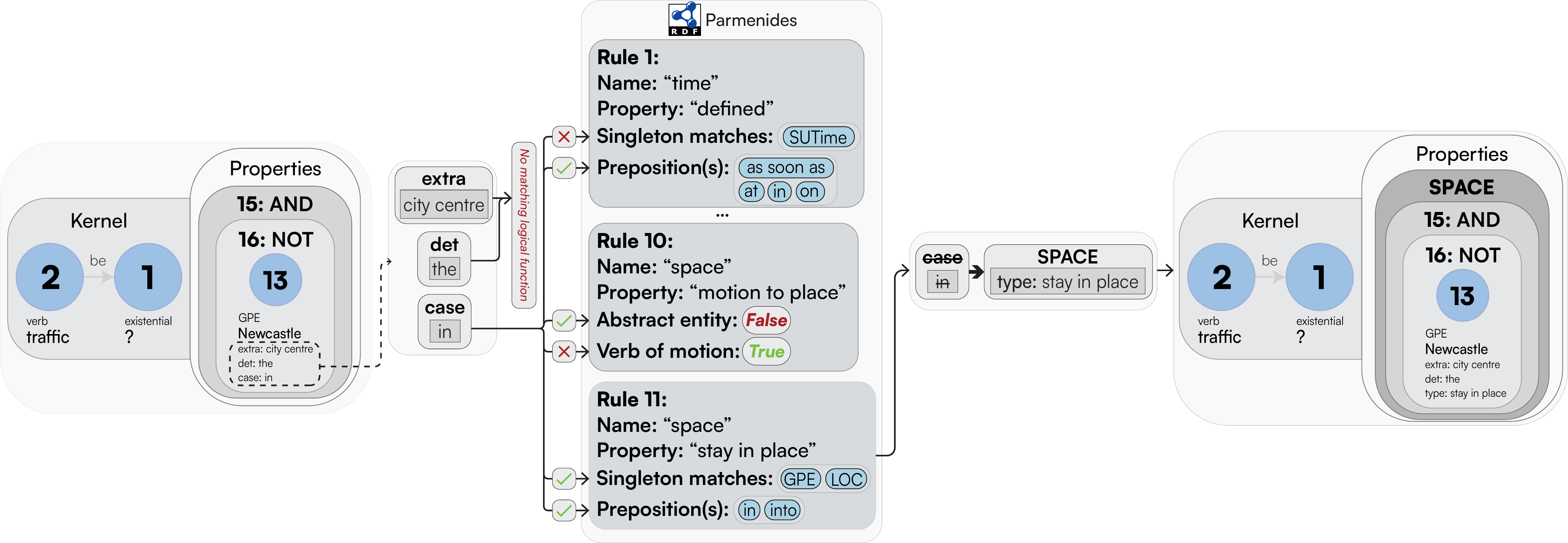}
	\caption{An abstraction of Algorithm \ref{alg:rewrite_props}, continuing the example from \figurename~\ref{fig:algo1explain}: we show how properties are matched against the Parmenides \gls{kb} to find a logical function that match each given property, to thus be rewritten and replaced within the final logical representation.}
	\label{fig:rewrite_props}
\end{figure}

\figurename~\ref{fig:rewrite_props} summarises the contribution of Algorithm \ref{alg:rewrite_props} (Supplement \ref{sec:fullSemantics}): for each sentence node (either \texttt{SetOfSingletons} or \texttt{Singleton}), we obtain its properties and test all the rules as stored in Parmenides in declaration order. Upon the satisfaction of the premises, the rule determines how to rewrite the matched content as either properties of the relationship or of the aforementioned node.

\begin{example}
	Concerning \figurename~\ref{code:logical_rule_space}, we are looking for a property that contains a preposition of either ``\textit{in}'' or ``\textit{into}'', and is not an \textit{abstract concept} entity (i.e., not concrete objects/people from the real world). An example sentence that would match this rule is ``\textit{characters in movies}''. Before rewriting with Algorithm \ref{alg:rewrite_props}, we obtain: \texttt{be(characters, ?)[nmod(characters, movies[2:in])]}. The \texttt{nmod} edge is matched to the rewriting rule, and is thus rewritten based on the properties of the matched logical function, presented in \figurename~\ref{code:logical_function_space}, whereby it should be attached to the kernel, resulting in: \texttt{be(characters, ?1)[(SPACE:movies[(type:stay in place)])]}. 
\end{example}

\begin{example}
	`\textit{Group of reindeer}' is initially rewritten as:
	\begin{center}
		\texttt{be(group, ?)[(nmod(group, reindeer[(2:of)]))]}
	\end{center}
	
	After determining the specification as per Line \ref{line:rewrite_nmod} in Algorithm \ref{alg:rewrite_props}, we obtain some redundancies:
	\begin{center}
		\texttt{be(group, ?)[SPECIFICATION:group[(extra:reindeer)[2:of]]]}
	\end{center}
	
	We have the source containing \texttt{group}, with properties that are also of the same entity, but with the additional information of \texttt{reindeer}; therefore, on Line \ref{line:property_replace} in Algorithm \ref{alg:rewrite_props}, we replace the source with the property and subsequently obtain: 
	
	\begin{center}
		\texttt{be(group[(extra:reindeer)[2:of]], ?)}
	\end{center}
\end{example}

Rule premises may include prepositions from \texttt{case} \cite{case_ud} properties like `\textit{of}', `\textit{by}', and `\textit{in}', or predicates based on verbs from \texttt{nmod} \cite{nmod_ud} relationships, and whether they are causative or movement verbs. 
There are many different types, like `space' and `time', and the \texttt{property} further clarifies the \texttt{type}. For `space', you might have `motion to place', implying the property has a motion from one place to another, or `stay in place', indicating that the sentence's location is static. For \texttt{time}, we might have `defined' for `on Saturdays' or `continuous' for `during', implying the time for the given sentence has yet to occur (\tablename~\ref{tab:logicalFunction}).

\begin{example}
	The sentence ``\textit{Traffic is flowing in Newcastle city centre, on Saturdays}'' is initially rewritten as: \texttt{flow(Traffic, None)[(GPE:Newcastle[(extra:city centre), (4:in)]), (DATE:Saturdays[(9:on)])]}. We have both the location of ``\textit{Newcastle}'' and time of ``\textit{Saturdays}''. Given the rules from \figurename~\ref{code:logical_rule_time}, the sentence would match the \texttt{DATE} property  and \texttt{GPE}. After the application of the rules, the relationship is rewritten as: 
	\begin{center}
		\texttt{flow(Traffic, None)[(SPACE:Newcastle[(type:stay in place), (extra:city centre)]), (TIME:Saturdays[(type:defined)])]}
	\end{center}
	
	Due to lemmatisation, the edge label becomes \texttt{flow} from ``\textit{is flowing}''.
\end{example}

For conciseness, additional details for how such a matching mechanism works are presented in \supplementaryname~\ref{sec:fullSemantics}.

\subsubsection{Final First-Order Logic (FOL) Representation}\label{sec:logrewr}
Finally, we derive a logical representation in \gls{fol}. Each entity is represented as one single function, where the arguments provide the name of the entity, its potential specification value, and any adjectives associated with it (\texttt{cop}), as well as any explicit quantification. These are pivotal for spatial information from which we can determine if all the parts of the area ({$\Box t$}) or just some of these ({$\lozenge t$}) are considered. This characterisation is not represented as \gls{fol} universal or existential quantifiers, as they are only used to refine the intended interpretation of the function representing the spatial entity. Transitive verbs are then always represented with binary propositions, while intransitive verbs are always represented as unary ones; for both, their names refer to the associated verb name. If any ellipsis from the text makes an implicit reference to either of the arguments, these are replaced with a fresh variable, which is then bound with an existential quantifier. For both functions and propositions, we provide a minor syntax extension that does not substantially affect its semantics, rather than using shorthand to avoid expressing additional function and proposition arguments referring to further logical functions and entities associated with them. We then introduce explicit properties $p$ as key--value multimaps. Among these, we also consider a special constant (or $0$-ary function) \textbf{None}, identifying that one argument is missing relevant information. We then derive the following syntax, which can adequately represent factoid sentences like those addressed in the present paper: 

\[\begin{cases}\label{folsyntax}
	t\colon\textsc{Term} ::= &x\in\textsc{Var}\;|\;{\Box}\textsc{func}_p(\texttt{name},\texttt{specification},\texttt{cop})\;|\\
	& {\lozenge}\textsc{func}_p(\texttt{name},\texttt{specification},\texttt{cop})\\
	p_1\colon \textsc{Proposition}::=&u_p(t)\;|\;b_p(t,t')\\
	A\colon\textsc{Formula}::=&p_1\;|\;\neg A\;|\;A\wedge A'\;|\;A\vee A'\;|\;\exists x. A
\end{cases}\]

Throughout this paper, when an entity ``foo'' is associated with only a name and has no explicit all/some representation, this will be rendered as $\lozenge\textrm{foo}$. When ``foo'' comes with a specification ``bar'' and has no explicit all/some representation, this is represented as $\lozenge\textrm{foo [of] }\textit{bar}$.

Given the intermediate representation resulting from Section \ref{sec:recursive}, we  rewrite any logical connective occurring within either the relationships' properties or within the remaining \texttt{SetOfSingletons} as logical connectives within the \gls{fol} representation, and represent each \texttt{Singleton} as a single function. Each free variable in the formula is bound to a single existential quantifier. When possible, negations potentially associated with specifications of a specific function are then expanded and associated with the proposition containing such function as a term.

\subsection{Ex Post}\label{sec:expostexmpl}

	The \textit{ex post} explanation phase details the similarity of two full text sentences through a similarity score over a representation derived from the previous phase. When considering traditional transformer approaches representing sentences as semantic vectors, we consider traditional cosine similarity metrics (\supplementaryname~\ref{sec:fulltexcos} and \ref{neural-ir}). For \glspl{llm}, we consider normalised confidence score (Supplement \ref{debertaprocessing}). When considering graphs representable as collections of edges, we consider alignment-driven similarities, for which node and edge similarity is defined via the cosine similarity over their full text representation (\supplementaryname~\ref{sec:galignment}). 

We decompose the two full text sentences into atoms, which we expand over through a common approach, Upper Ontologies \cite{DBLP:conf/fois/NilesP01,DBLP:conf/isko-uk/Winstanley23}, where TBox reasoning rules are hardcoded to ensure the correctness of the inference approach. We derive the equivalence through atoms expanded as propositions using the Parmenides \gls{kb}, providing relationships occurring between two propositions, while remarking entailment, equivalence, or indifference (Appendix \ref{app:tabular-semantics}). This culminates to calculating the confidence value of $\varphi_1 \Rightarrow \varphi_2$ and $\varphi_2 \Rightarrow \varphi_1$. Thus, we derive two truth tables for each sentence, each extracted from  possible worlds that hold for all atoms within each sentence, which we then equi-join. By naturally joining all the derived tables together into $\mathcal{T}$, including the tabular semantics associated with each formula $A$ and $B$ (\appendixname~\ref{app:formulartabular}), we trivially reason paraconsistently by only considering the worlds that do not contain contradicting assumptions \cite{hawking1988}. We refer to Section \ref{sec:areval} for some high-level graphical representation of this phase. We express confidence from Eq. \ref{eq:conf} as follows:
\begin{equation}
	\label{ourDef}
	\textsf{support}(s,t)=\textup{avg}\,\pi_{t}(\sigma_{s=1}(\mathcal{T}))
\end{equation} 

While the metric summarises the logic-based sentence relatedness, $\sigma_{s=1}(\mathcal{T})$ provides the full possible world combination table for the number derived. {By following the interpretation of the support score, we then derive a score of 1 when the premise entails the consequence, 0 when the implication derives a contradiction, and any value in between otherwise, noting the estimated ratio of possible worlds as stated above. Thus, the aforementioned score induces a three-way classification for the sentences of choice.} 

At this stage, we must derive a logic-driven similarity score to overcome the limitations of current symmetrical measures that cannot capture logical entailment. We can then re-formulate the classical notion of confidence from association rule mining \cite{10.5555/857189.857670}, which implicitly follows the steps of entailment and provides an estimate for conditional probability. 
From each sentence $\alpha$ and its logical representation $A$, we need to derive the set of circumstances $W(A)$ or worlds in which we trust the sentence will hold. As confidence values are always normalised between 0 and 1, these give us the best metric to represent the degree of trustworthiness of information accurately. We can then rephrase the definition of confidence for logical formulae as follows:
\begin{definition}[Confidence]\label{def:confidence}
	Given two logically represented sentences $A$ and $B$, let $W(A)$ and $W(B)$ represent the set of possible worlds where $A$ and $B$ hold, respectively. Then, the \textit{confidence} metric, denoted as $\mathtt{confidence}(A, B)$, is defined based on its bag semantics as:
	
	\begin{equation}\label{eq:conf}
		\mathtt{confidence}(A, B) = \begin{cases}
			\frac{|W(A) \cap W(B)|}{|W(A)|} & W(A)\neq\emptyset\\
			0 & \textup{oth.}\\
		\end{cases}
	\end{equation}
	
	Please observe that the only formula with an empty set of possible worlds is logically equivalent to the universal falsehood $\bot$; thus, $W(\bot)=\emptyset$.
\end{definition}

\appendixname~\ref{app:formulartabular} formalises the definition of the tabular semantics in terms of relational algebra, thus showcasing the possibility of enumerating all the worlds for which one formula holds while circumscribing them to the propositions that define the formula.

\begin{example}
	\label{ex:alicebob}
	Consider the sentences $\alpha'$: ``Alice plays football'' and $\beta'$: ``Bob plays football''.  We can represent these logically as binary propositions $p_1:=\mathtt{play(Alice, football)}$ and $p_2:=\mathtt{play(Bob, football)}$, respectively. Given the sentences $\alpha$: ``Alice and Bob play football'' and $\beta$: ``Either Alice or Bob play football'', we can see that the former propositions can be considered atoms defining the current sentences. We then represent them as $A=p_1\wedge p_2$ and $B=p_1\vee p_2$, where both $\mathcal{B}(A)=\mathcal{B}(B)=\{p_1,p_2\}$. Thus, we can easily derive the set of the possible worlds from the ones arising from all the possible combinations of the truth and falsehood of each proposition, as shown in \tablename~\ref{table:exPostTruthTable}:
	Thus, given the corresponding truth table from \tablename~\ref{table:exPostTruthTable}, we derive the following values for the bag semantics of $A$ and $B$: $W(A)=\{\#3,\#4\}$ and $W(B)=\{\#2,\#4\}$.
	
	We can use these values to determine the confidence value when $s \Rightarrow t$ and $t \Rightarrow s$. In our scenario, the resulting equi-join $\mathcal{T}$ matches the one from \tablename~\ref{table:exPostTruthTable}, as all the propositions are indifferent.
	
	To first find $s \Rightarrow t$, we find $W(s) \cap W(t)$, meaning the number of times $s$ and $t$ are true, which is 1 when both $a$ and $b$ are true. 
	Then, we find $W(s)$, the number of times $s$ is true, which is also 1 when both $a$ and $b$ are true. Substituting into the confidence metric, we obtain $\frac{1}{1}=100\%$, meaning that when $a$ is true, it is \textbf{certain} that $b$ is also true; like when Alice and Bob are playing football, it must therefore hold that either are playing.
	
	Alternatively, to find $t \Rightarrow s$; we find $W(t) \cap W(s)$, which is 1, the same when both $a$ and $b$ are true. Then, we find $W(t)$, the number of times $t$ is true, which is 3 when either $a$ or $b$ are true. Substituting into the confidence metric, we obtain $\frac{1}{3}=33.\dot{3}\%$, meaning that when $b$ is true, there is only a $33.\dot{3}\%$ chance that $a$ is true; aW when either Alice or Bob play football, one cannot be certain whether \textbf{both} are playing.
	
\end{example}

\begin{table}[H]
	\centering
	\caption{Truth table of values for sentences $\alpha'$ and $\beta'$.}
	\label{table:exPostTruthTable}
	\begin{tabular}{|l|l|l|l|l|}
		\hline
		\rowcolor[HTML]{EFEFEF} 
		\#W & \textbf{$p_1$} & \textbf{$p_2$} & \textbf{$A: p_1 \wedge p_2$} & \textbf{$B: p_1 \vee p_2$} \\ \hline
		\#1 & 0 & 0 & 0 & 0  \\ \hline
		\#2 &0 & 1 & 0 & 1  \\ \hline
		\#3 &1 & 0 & 0 & 1  \\ \hline
		\#4 &1 & 1 & 1 & 1  \\ \hline
	\end{tabular}
\end{table}


\section{Results}\label{sec:results}

{The Results section} is structured as follows: \textit{First}, we demonstrate  the impossibility of deriving the notion of logical entailment via any symmetric similarity function (Section \ref{theorimposs}). We base our argument on the following observation: given that current symmetric metrics are better suited to capture sentence similarity through clustering due to their predisposition of representing equivalence rather than entailment, we show that assuming symmetrical metrics also leads to incorrect clustering outcomes, based on which dissimilar or non-equivalent sentences are grouped. \textit{Second}, we provide empirical benchmarks showing the impossibility of achieving this through classical transformer-based approaches as outlined in the Introduction (Section \ref{sub:clustering}).  
All of these components provide a pipeline ablation study concerning the different rewriting stages occurring within the \textit{ad hoc} phase of the pipeline, while more thorough considerations addressing disabling the \textit{a priori} phase are presented in the Discussion section (Section \ref{subsec:pip_abla}). For improving paper's readibility, we moved our scalability results to \appendixname~\ref{sub:benchmark}, where we address the scalability (also referred to as AI latency) of our proposed approach by considering sub-sents of full text sentences appearing as nodes for the ConceptNet common-sense knowledge graph (Appendix \ref{sub:benchmark}). 

The experiments were run on a Linux Desktop machine with the following specifications: \textbf{CPU:} 12th Gen Intel i9-12900 (24) @ 5GHz; \textbf{memory:} 32 GB DDR5. The raw data for the results, including confusion matrices and generated final logical representations, can be found on OSF.io (\url{https://osf.io/g5k9q/}, Accessed on 28 March 2025).

\subsection{Theoretical Results}\label{theorimposs}

As the notion of verified \gls{ai} incentivises inheriting logic-driven notions for ensuring the correctness of the algorithms \cite{VAI}, we leverage the logical notion of \textbf{soundness} \cite{Hinman_2005} to map the common-sense interpretation of a full text into a machine-readable representation (as a logical rule or a vector embedding); a rewriting process is sound if the rewritten representation logically follows from the original sentence, which then follows the notion of correctness. For the sake of the current paper, we limit our interest to capturing logical entailment as generally intended from two sentences. Hence, we are interested in the following definition of soundness.
\begin{definition}
	\textbf{Weak Soundness}, in the context of sentence rewriting, refers to the preservation of the original semantic meaning of the logical implication of two sentences $\alpha$ and $\beta$. Formally:
	\[
	\text{Weak Soundness} \colon \text{if } \vdash_S \alpha\sqsubseteq\beta, \text{ then also } \vDash {\varphi_\tau}(\alpha,\beta)
	\]
	where $S$ is the common-sense interpretation of the sentence, $\sqsubseteq$ is the notion of logical entailment between textual concepts, and $\varphi_\tau$ is a predicate deriving the notion of entailment from the $\tau$ transformation of a sentence through the choice of a preferred similarity metric $\mathcal{S}$.  
\end{definition}

In the context of this paper, we are then interested in capturing sentence dissimilarities in a non-symmetrical way, thus capturing the notion of logical entailment:

\begin{equation}
	\text{Correctness} \colon \alpha \sqsubseteq \beta \wedge \beta \not\sqsubseteq \alpha \Rightarrow \varphi_\tau(\alpha,\beta) \wedge \neg\varphi_\tau(\beta,\alpha).
\end{equation}

Any symmetric similarity metrics  (including  cosine similarity and edge-based graph alignment) cannot be used to express logical entailment (Section \ref{subsub:cosine}), while the notion of confidence adequately captures the notion of logical implication by design (Section \ref{proofs:confidence}). All proofs for the forthcoming lemmas are presented in \appendixname~\ref{proofs}.

\subsubsection{Cosine Similarity}\label{subsub:cosine}

The  above entails that we can always derive a threshold value $\theta$ above which we can deem as one sentence implying the other, thus enabling the following definition.

\begin{definition}\label{phitau}
	Given $\alpha$ and $\beta$ are full texts and $\tau$ is the vector embedding of the full text, we derive entailment from any similarity metric $\mathcal{S}$ as follows:
	\[
	\varphi_\tau(\alpha,\beta) \iff \exists\theta.\mathcal{S}(\tau(\alpha),\tau(\beta)) > \theta,
	\]
	where $\theta$ is a constant threshold. This definition allows us to express implications as exceeding a similarity threshold.
\end{definition}

As cosine similarity captures the notion of similarity, and henceforth an approximation of a notion of equivalence, we can clearly see that this metric is symmetric.

\begin{lemma}\label{def:cosine}
	Cosine similarity is symmetric
\end{lemma}

Symmetry breaks the capturing of directionality for logical implication. 
Symmetric similarity metrics can lead to situations where soundness is violated. For instance, if $A \Rightarrow B$ holds based on a symmetric metric, then $B \Rightarrow A$ would also hold, even if it is not logically valid. We derive implication from similarity when the similarity metric for $A \Rightarrow B$ is different to $B \Rightarrow A$, given that $A \not\equiv B$, and this shows that one thing might imply the other but not vice versa. To enable the identification of implication across different similarity functions, we entail the notion of implication via the \textbf{similarity value} as follows.

\begin{lemma}\label{lemma:symm}
	All symmetric metrics $\mathcal{S}$ trivialise logical implication:
	\[\mathcal{S} \text{ symm} \wedge \varphi_\tau(\alpha,\beta)\Rightarrow \varphi_\tau(\beta,\alpha)\]
\end{lemma}

Since symmetric metrics such as cosine similarity cannot capture the directionality of implication, they cannot fully represent logical entailment. This limitation highlights the need for alternative approaches to model implication accurately, thus violating our intended notion of correctness.

\subsubsection{Confidence Metrics}\label{proofs:confidence}
In contrast to the former, we show that the confidence metric presented in Section \ref{sec:expostexmpl} produces a value that aims to express logical entailment under the assumption that the $\varphi$ transformation from the full text to a logical representation is correct.

\begin{lemma}\label{lem:eqsameconf}
	When two sentences are equivalent, they always have the same confidence.
	\[A \equiv B \Leftrightarrow \mathtt{confidence}(A,B) = \mathtt{confidence}(B,A) = 1 \]
	
\end{lemma}

As a corollary of the above, this shows that our confidence metric is non-symmetric.

\begin{corollary}\label{coroll:one}
	Confidence provides an adequate characterisation of logical implication.
	\[\neg(A\Rightarrow B) \Rightarrow \mathtt{confidence}(A,B)<1\]
\end{corollary}

This observation leads to the definition of the following notion of $\varphi_\tau$ given $\tau$, the processing pipeline.

\begin{definition}\label{phitau2}
	Given that $\alpha$ and $\beta$ are the full text and $\tau$ is the logical representation of the full text derived in the \textit{ad hoc}  phase (Section \ref{sec:adhoc}), we derive entailment from any similarity metric $\mathcal{S}$ as follows:
	\[
	\varphi_\tau(\alpha,\beta) \iff \mathtt{confidence}(\tau(\alpha),\tau(\beta)) =1
	\]
\end{definition}

\subsection{Classification}\label{sub:clustering}

While the previous section provided the basis for demonstrating the theoretical impossibility of achieving a perfect notion of implication, the following experiments aimed to test this argument from a different perspective. For making the paper more concise, we refer to \appendixname~\ref{app:clustering} for further information concerning the preliminary clustering steps, through which we derive the upper threshold $\theta$ separating the indifferent and conflicting sentences from the equivalent and entailing ones.


We want to determine the ability to not only identify which sentences are equivalent but also ascertain their ability to differentiate between logical entailment, indifference, and inconsistent representation. To do so, we need to first annotate each sentence within our proposed dataset to indicate such a difference. Given that none of the proposed approaches were explicitly trained to recognise conflicting arguments, we then derive an upper threshold value $\vartheta$ separating the conflicting sentences from the rest by taking the maximal similarity score between the pair of sentences expected to be contradictory from the manual annotation. If $\vartheta>\theta$, we consider only one threshold value separating entailing and contradictory data ($\vartheta=\theta$). Thus, we consider all similarity values above $\theta$ prediction values for a logical entailment and all values lower than $\vartheta$ as predictive of a conflict between the two sentences. Otherwise, we determine an indifference relationship. Please observe that different clustering algorithms might lead to different $\theta$ scores: given this, we will also expect to see variations within the classification scores discussed in \supplementaryname~\ref{supp:classmetric}. Thus, our classification results will explicitly report the name of the paired clustering algorithm leading to the specific classification outcome.

As pre-trained language models are assumed to generalise their semantic understanding of the text through extensive training on a large corpus of data, these models should also be able to capture semantic and structural nuances provided in completely new datasets. To remove any semantic ambiguity derivable from analysing complex sentences with multiple entities, we considered smaller ones for controlled experiments under the different scenarios discussed in the Introduction  (\figurename~\ref{subf:sentences_ab}, \ref{subf:sentences_cm}, and  \figurename~\ref{subf:sentences_newc}) so provide empirical validation to preliminary theoretical results \cite{10.1162/tacl_a_00663}. The set of sentences, alongside their expected clusters and classification outcomes, are freely available for reproducibility purposes (\url{https://osf.io/g5k9q/}, Accessed on 25 April).

We used the following transformers $\tau$ available on HuggingFace \cite{HuggingFaceTransformers} from the current state-of-the-art research papers discussed in Section \ref{sec:rw_soa}:
\begin{competitors}
	\item all-MiniLM-L6-v2 \cite{wang2021minilmv2multiheadselfattentionrelation}
	\item all-MiniLM-L12-v2 \cite{wang2021minilmv2multiheadselfattentionrelation}
	\item all-mpnet-base-v2 \cite{song2020mpnet}
	\item all-roberta-large-v1 \cite{liu2019roberta}
	\item DeBERTaV2+AMR-LDA \cite{bao-etal-2024-abstract} 
	\item ColBERTv2+RAGatouille  \cite{santhanam-etal-2022-colbertv2} 
\end{competitors}

For all the approaches, we consider all similarity metrics $s(A,B)$ as already normalised between 0 and 1, as discussed in Section \ref{sec:expostexmpl}. We can derive a distance function from this by subtracting the value from one; hence, $d(A,B):=1-s(A,B)$. Since not all of these approaches lead to symmetric distance functions and given that the clustering algorithms work under symmetric distance functions, we obtain a symmetric definition by averaging the distance values as follows:

\[\overline{d}(a,b):=\frac{1}{2}(d(a,b)+d(b,a))\]

\subsubsection{Capturing Logical Connectives and Reasoning}\label{sub:logical_connectives}

We check if the logical connectives are treated as stop words and then ignored by transformers. This is demonstrated through sentences in \figurename~\ref{subf:sentences_ab}, whereby one sentence might imply another, but not vice versa. However, with our clustering approaches, we only want to cluster sentences that are similar in \textit{both} directions; therefore, our clusters have only one element, as shown in \figurename~\ref{subf:exp_clusters_ab}.

\begin{figure}[!h] 
	\subfloat[Sentences]{\label{subf:sentences_ab}
		\begin{minipage}[t]{0.45\textwidth}
			\vspace{0pt}
			\begin{enumerate}[]
				\setcounter{enumi}{-1}
				\item Alice plays football
				\item Bob plays football
				\item Alice and Bob play football
				\item Alice or Bob play football
				\item Alice doesn't play football
				\item Bob doesn't play football
				\item Dan plays football
				\item Neither Alice nor Bob play football
			\end{enumerate}
		\end{minipage}
	}
	\hfill
	\subfloat[Expected clusters]{\label{subf:exp_clusters_ab}
		\begin{minipage}[t]{0.45\textwidth}
			\vspace{0pt}
			\begin{itemize}[leftmargin=*]
				\item[] \hspace*{\dimexpr-\labelindent-\labelwidth-\labelsep+\itemindent} \{0\}: \hspace{\itemindent}  Alice plays football
				\item[] \hspace*{\dimexpr-\labelindent-\labelwidth-\labelsep+\itemindent} \{1\}: \hspace{\itemindent}  Bob plays football
				\item[] \hspace*{\dimexpr-\labelindent-\labelwidth-\labelsep+\itemindent} \{2\}: \hspace{\itemindent}  Alice and Bob play football
				\item[] \hspace*{\dimexpr-\labelindent-\labelwidth-\labelsep+\itemindent} \{3\}: \hspace{\itemindent}  Alice or Bob play football
				\item[] \hspace*{\dimexpr-\labelindent-\labelwidth-\labelsep+\itemindent} \{4\}: \hspace{\itemindent}  Alice doesn't play football
				\item[] \hspace*{\dimexpr-\labelindent-\labelwidth-\labelsep+\itemindent} \{5\}: \hspace{\itemindent}  Bob doesn't play football
				\item[] \hspace*{\dimexpr-\labelindent-\labelwidth-\labelsep+\itemindent} \{6\}: \hspace{\itemindent}  Dan plays football
				\item[] \hspace*{\dimexpr-\labelindent-\labelwidth-\labelsep+\itemindent} \{7\}: \hspace{\itemindent}  Neither Alice nor Bob play football
			\end{itemize}
		\end{minipage}
	}
	\centering
	\caption{Sentences and expected clusters for \ref{rqn2a}{, where no sentences are clustered together, as no sentence is perfectly similar to another in both directions}.}
	\label{sentences:ab}
\end{figure}

In \tablename~\ref{table:metric_results_ab}, we use both macro and weighted average scores, as all our datasets are heavily unbalanced. Furthermore, as different clustering algorithms might derive different upper threshold values $\theta$, we kept the distinction between the different classification outcomes leading to a difference in classification. Our logical approach excels, achieving perfect scores across all classification scores. The preliminary stage of our pipeline, \gls{sg}, achieves the worse scores across the board, thus clearly indicating that just targeting node similarity through node and edge embedding is insufficient for fully capturing sentence structure, even after \gls{ggg} rewriting. \glspl{lg} slightly improve over \glspl{sg} due to the presence of a logical operation; this approach is shown to improve over classical sentence transformer approaches, with a few exceptions.

\begin{table}[!h]
	\caption{Classification scores for \ref{rqn2a} sentences; the best value for each row is highlighted in bold, blue text and the worst values are highlighted in red. The classes are distributed as follows: implication: 15; inconsistency: 16; indifference: 33.}
	\label{table:metric_results_ab}\centering
	\begin{tblr}{
			width = 0.5\textwidth,
			column{4-12} = {colsep=2pt,font=\footnotesize},
			row{1} = {font=\normalsize},
			cells = {c},
			row{1} = {Alto},
			cell{2}{1} = {r=2}{Alto},
			cell{2}{2} = {r=2}{Mercury},
			cell{2}{3} = {Gallery},
			cell{3}{3} = {Gallery},
			cell{4}{1} = {r=4}{Alto},
			cell{4}{2} = {r=2}{Mercury},
			cell{4}{3} = {Gallery},
			cell{5}{3} = {Gallery},
			cell{6}{2} = {r=2}{Mercury},
			cell{6}{3} = {Gallery},
			cell{7}{3} = {Gallery},
			cell{8}{1} = {r=4}{Alto},
			cell{8}{2} = {r=2}{Mercury},
			cell{8}{3} = {Gallery},
			cell{9}{3} = {Gallery},
			cell{10}{2} = {r=2}{Mercury},
			cell{10}{3} = {Gallery},
			cell{11}{3} = {Gallery},
			cell{12}{1} = {r=4}{Alto},
			cell{12}{2} = {r=2}{Mercury},
			cell{12}{3} = {Gallery},
			cell{13}{3} = {Gallery},
			cell{14}{2} = {r=2}{Mercury},
			cell{14}{3} = {Gallery},
			cell{15}{3} = {Gallery},
			vlines,
			hline{1-2,4,8,12,16} = {-}{},
			hline{3,5,7,9,11,13,15} = {3-12}{},
			hline{6,10,14} = {2-12}{},
		}
		\textbf{Metric}     & \textbf{Average}  & \textbf{Clustering}         & \textbf{\glspl{sg}} & \textbf{\glspl{lg}} & \textbf{Logical} & \textbf{T1}   & \textbf{T2}   & \textbf{T3}   & \textbf{T4}   & \textbf{T5}   & \textbf{T6} \\
		Accuracy & -- & \textbf{HAC} & \lsc{0.28} & 0.38 & \textbf{\textcolor{blue}{1.00}} & 0.34 & 0.34 & 0.36 & 0.36 & 0.36 & 0.39 \\
		& & \textbf{$k$-Medoids}& \lsc{0.28} & 0.38 & \textbf{\textcolor{blue}{1.00}} & 0.34 & 0.34 & 0.36 & 0.36 & 0.36 & 0.39 \\
		F1 & Macro & \textbf{HAC} & \lsc{0.25} & 0.41 & \textbf{\textcolor{blue}{1.00}} & 0.37 & 0.37 & 0.39 & 0.39 & 0.34 & 0.42 \\
		& & \textbf{$k$-Medoids}& \lsc{0.25} & 0.41 & \textbf{\textcolor{blue}{1.00}} & 0.37 & 0.37 & 0.39 & 0.39 & 0.34 & 0.42 \\
		& Weighted & \textbf{HAC} & \lsc{0.18} & 0.31 & \textbf{\textcolor{blue}{1.00}} & 0.26 & 0.26 & 0.29 & 0.29 & 0.24 & 0.33 \\
		& & \textbf{$k$-Medoids}& \lsc{0.18} & 0.31 & \textbf{\textcolor{blue}{1.00}} & 0.26 & 0.26 & 0.29 & 0.29 & 0.24 & 0.33 \\
		Precision & Macro & \textbf{HAC} & \lsc{0.27} & 0.48 & \textbf{\textcolor{blue}{1.00}} & 0.42 & 0.42 & 0.51 & 0.51 & 0.28 & 0.56 \\
		& & \textbf{$k$-Medoids}& \lsc{0.27} & 0.48 & \textbf{\textcolor{blue}{1.00}} & 0.42 & 0.42 & 0.51 & 0.51 & 0.28 & 0.56 \\
		& Weighted & \textbf{HAC} & \lsc{0.20} & 0.38 & \textbf{\textcolor{blue}{1.00}} & 0.30 & 0.30 & 0.43 & 0.43 & \lsc{0.20} & 0.51 \\ 
		& & \textbf{$k$-Medoids}& \lsc{0.20} & 0.38 & \textbf{\textcolor{blue}{1.00}} & 0.30 & 0.30 & 0.43 & 0.43 & \lsc{0.20} & 0.51 \\
		Recall & Macro & \textbf{HAC} & \lsc{0.38} & 0.50 & \textbf{\textcolor{blue}{1.00}} & 0.47 & 0.47 & 0.48 & 0.48 & 0.49 & 0.51 \\
		& & \textbf{$k$-Medoids}& \lsc{0.38} & 0.50 & \textbf{\textcolor{blue}{1.00}} & 0.47 & 0.47 & 0.48 & 0.48 & 0.49 & 0.51 \\
		& Weighted & \textbf{HAC} & \lsc{0.28} & 0.38 & \textbf{\textcolor{blue}{1.00}} & 0.34 & 0.34 & 0.36 & 0.36 & 0.36 & 0.39 \\
		& & \textbf{$k$-Medoids}& \lsc{0.28} & 0.38 & \textbf{\textcolor{blue}{1.00}} & 0.34 & 0.34 & 0.36 & 0.36 & 0.36 & 0.39 \\   
	\end{tblr}
	
\end{table}

 In fact, ColBERTv2+RAGatouille provides better performances than other sentence transformers, and  DeBERTaV2+AMR-LDA shows poor performance, thus indicating the unsuitability of this approach to provide reasoning over other logical operators not captured in the training phase. 
Moreover, our proposed logical representation improves over \glspl{lg} through the tabular reasoning phase, due to the classical Boolean interpretation of the formulae: this indicates that graph similarity alone cannot be used to fully capture the essence of logical reasoning. In competing approaches, the weighted average is consistently lower compared to the macro average, thus suggesting that the misclassification task is not necessarily ascribable due to the imbalance nature of the dataset, 
while also suggesting that the majority class (indifference) was mainly misrepresented. 

\subsubsection{Capturing Simple Semantics and Sentence Structure}\label{sub:active-passive}


The sentences in \figurename~\ref{subf:sentences_cm} are all variations of the same subjects (a cat and mouse), with different actions and active/passive relationships. The dataset is set up to produce sentences with similar words but in a different order, allowing for determination of whether the sentence embedding understands context from structure rather than edit distance. \figurename~\ref{subf:exp_clusters_cm} shows the expected clusters for the \ref{rqn2b} dataset. Here, 0 and 1 are clustered together as the subject's action on the direct object is the same in both: ``\textit{the cat eats the mouse}'' is equivalent to ``\textit{the mouse is eaten by the cat}''. Similarly, sentences 2 and 3 are the same, but with the action reversed (the mouse eats the cat in both).

\begin{figure}[h] 
	\subfloat[Sentences]{\label{subf:sentences_cm}
		\begin{minipage}[t]{0.475\textwidth}
			\vspace{0pt}
			\begin{enumerate}[]
				\setcounter{enumi}{-1}
				\item The cat eats the mouse 
				\item The mouse is eaten by the cat 
				\item The mouse eats the cat 
				\item The cat is eaten by the mouse 
				\item The cat doesn't eat the mouse 
				\item The mouse doesn't eat the cat 
			\end{enumerate}
		\end{minipage}
	}
	\hfill
	\subfloat[Expected clusters]{\label{subf:exp_clusters_cm}
		\begin{minipage}[t]{0.475\textwidth}
			\vspace{0pt}
			\begin{itemize}[leftmargin=*]
				\item[] \hspace*{\dimexpr-\labelindent-\labelwidth-\labelsep+\itemindent} \{0, 1\}: \hspace{\itemindent}
				\begin{itemize}[leftmargin=1.5pt, labelsep=2pt]
					\item The cat eats the mouse
					\item The mouse is eaten by the cat
				\end{itemize}
				\item[] \hspace*{\dimexpr-\labelindent-\labelwidth-\labelsep+\itemindent} \{2, 3\}: \hspace{\itemindent}
				\begin{itemize}[leftmargin=1.5pt, labelsep=2pt]
					\item The mouse eats the cat
					\item The cat is eaten by the mouse
				\end{itemize}
				\item[] \hspace*{\dimexpr-\labelindent-\labelwidth-\labelsep+\itemindent} \{4\}: \hspace{\itemindent} The cat doesn't eat the mouse
				\item[] \hspace*{\dimexpr-\labelindent-\labelwidth-\labelsep+\itemindent} \{5\}: \hspace{\itemindent} The mouse doesn't eat the cat
			\end{itemize}
		\end{minipage}
	}
	\centering
	\caption{Sentences and expected clusters for \ref{rqn2b}.}
	\label{sentences:cm}
\end{figure}

\tablename~\ref{table:metric_results_cm} provides a more in-depth analysis of the situation where, instead of being satisfied with the possibility that the techniques mentioned above can capture the notion of equivalence between sentences, we also require that they can make finer-grained semantic distinctions. The inability of DeBERTaV2+AMR-LDA to fully capture sentence semantics might be reflected in the choice of the AMR as, in this dataset, the main differences across the data are based on the presence of negation and of sentences in both active and passive form. This supports the evidence that transformer-based approaches providing one single vector generally provide better results through masking and tokenisation. Differently from the previous set of experiments, subdividing the text encoding into multiple different sentences proved to be ineffective for the ColBERTv2+RAGatouille approach, as subdividing short sentences into different tokens for the derivation of a vector results in complete loss of the semantic information captured by the sentence structure. In this scenario, the proposed logical approach is mainly supported by tabular semantics, where most of the sentences are simply represented as two variants of atoms, potentially being negated. This indicates the impossibility of fully capturing logical inference through graph structure alone. In this scenario, precision scores appear almost the same, independent of the clustering and averaging technique. The Accuracy and F1 scores show that the similarity score is very near to a random choice.

\begin{table}[!h]
	\caption{Classification scores for \ref{rqn2b} sentences, with the best value for each row highlighted in bold, blue text and the worst values highlighted in red. The classes are distributed as such: Implication: 10, Inconsistency: 8, Indifference: 18.}
	\label{table:metric_results_cm}\centering
	\begin{tblr}{
			width = 0.5\textwidth,
			column{4-12} = {colsep=2pt,font=\footnotesize},
			row{1} = {font=\normalsize},
			cells = {c},
			row{1} = {Alto},
			cell{2}{1} = {r=2}{Alto},
			cell{2}{2} = {r=2}{Mercury},
			cell{2}{3} = {Gallery},
			cell{3}{3} = {Gallery},
			cell{4}{1} = {r=4}{Alto},
			cell{4}{2} = {r=2}{Mercury},
			cell{4}{3} = {Gallery},
			cell{5}{3} = {Gallery},
			cell{6}{2} = {r=2}{Mercury},
			cell{6}{3} = {Gallery},
			cell{7}{3} = {Gallery},
			cell{8}{1} = {r=4}{Alto},
			cell{8}{2} = {r=2}{Mercury},
			cell{8}{3} = {Gallery},
			cell{9}{3} = {Gallery},
			cell{10}{2} = {r=2}{Mercury},
			cell{10}{3} = {Gallery},
			cell{11}{3} = {Gallery},
			cell{12}{1} = {r=4}{Alto},
			cell{12}{2} = {r=2}{Mercury},
			cell{12}{3} = {Gallery},
			cell{13}{3} = {Gallery},
			cell{14}{2} = {r=2}{Mercury},
			cell{14}{3} = {Gallery},
			cell{15}{3} = {Gallery},
			vlines,
			hline{1-2,4,8,12,16} = {-}{},
			hline{3,5,7,9,11,13,15} = {3-12}{},
			hline{6,10,14} = {2-12}{},
		}
		\textbf{Metric}     & \textbf{Average}  & \textbf{Clustering}         & \textbf{\glspl{sg}} & \textbf{\glspl{lg}} & \textbf{Logical} & \textbf{T1} & \textbf{T2} & \textbf{T3} & \textbf{T4} & \textbf{T5} & \textbf{T6} \\
		
		Accuracy & -- & \textbf{HAC} & 0.44 & 0.39 & \textbf{\textcolor{blue}{1.00}} & 0.50 & 0.50 & 0.44 & 0.50 & \lsc{0.19} & 0.39  \\
		& & \textbf{$k$-Medoids}& 0.44 & 0.39 & \textbf{\textcolor{blue}{1.00}} & 0.50 & 0.50 & 0.44 & 0.50 & \lsc{0.19} & 0.39  \\
		F1 & Macro  & \textbf{HAC} & 0.46 & 0.38 & \textbf{\textcolor{blue}{1.00}} & 0.52 & 0.52 & 0.45 & 0.52 & \lsc{0.13} & 0.36  \\
		&   & \textbf{$k$-Medoids}& 0.46 & 0.38 & \textbf{\textcolor{blue}{1.00}} & 0.52 & 0.52 & 0.45 & 0.52 & \lsc{0.13} & 0.36  \\
		& Weighted  & \textbf{HAC} & 0.36 & 0.30 & \textbf{\textcolor{blue}{1.00}} & 0.49 & 0.49 & 0.43 & 0.49 & \lsc{0.09} & 0.27  \\
		&  & \textbf{$k$-Medoids}& 0.36 & 0.30 & \textbf{\textcolor{blue}{1.00}} & 0.49 & 0.49 & 0.43 & 0.49 & \lsc{0.09} & 0.27  \\
		Precision & Macro & \textbf{HAC} & 0.42 & 0.34 & \textbf{\textcolor{blue}{1.00}} & 0.51 & 0.51 & 0.52 & 0.51 & \lsc{0.09} & 0.29 \\
		&  & \textbf{$k$-Medoids}& 0.42 & 0.34 & \textbf{\textcolor{blue}{1.00}} & 0.51 & 0.51 & 0.52 & 0.51 & \lsc{0.09} & 0.29 \\
		& Weighted  & \textbf{HAC} & 0.33 & 0.27 & \textbf{\textcolor{blue}{1.00}} & 0.51 & 0.51 & 0.57 & 0.51 & \lsc{0.06} & 0.23  \\
		&  & \textbf{$k$-Medoids}& 0.33 & 0.27 & \textbf{\textcolor{blue}{1.00}} & 0.51 & 0.51 & 0.57 & 0.51 & \lsc{0.06} & 0.23 \\
		Recall & Macro & \textbf{HAC} & 0.58 & 0.52 & \textbf{\textcolor{blue}{1.00}} & 0.56 & 0.56 & 0.52 & 0.56 & \textcolor{red}{0.29} & 0.53  \\
		&  & \textbf{$k$-Medoids}& 0.58 & 0.52 & \textbf{\textcolor{blue}{1.00}} & 0.56 & 0.56 & 0.52 & 0.56 & \lsc{0.29} & 0.53 \\
		& Weighted  & \textbf{HAC} & 0.44 & 0.39 & \textbf{\textcolor{blue}{1.00}} & 0.50 & 0.50 & 0.44 & 0.50 & \lsc{0.19} & 0.39 \\
		&  & \textbf{$k$-Medoids}& 0.44 & 0.39 & \textbf{\textcolor{blue}{1.00}} & 0.50 & 0.50 & 0.44 & 0.50 & \lsc{0.19} & 0.39 \\
		
	\end{tblr}
	
\end{table}

\subsubsection{Capturing Simple Spatiotemporal Reasoning}\label{sub:spatio}
We considered multiple scenarios involving traffic in Newcastle, presented in \figurename~\ref{subf:sentences_newc}, which have been extended from our previous paper to include more permutations of the sentences. This was done to obtain multiple versions of the same sentence that should be treated equally by ensuring that the rewriting of each permutation is the same, therefore resulting in 100\% similarity. We consider this dataset as a benchmark over a part/existential semantics $\lozenge t$, thus assuming that all the potential quantifiers being omitted refer to an existential (e.g., \textit{somewhere} in Newcastle, in \textit{some} city centers, on \textit{some} Saturdays). In addition, we consider negation as a flattening out process.

\begin{figure}[!h]
	\subfloat[Sentences]{\label{subf:sentences_newc}
		\begin{minipage}[t]{0.475\textwidth}
			\vspace{0pt}
			\begin{enumerate}[]
				\footnotesize
				\setcounter{enumi}{-1}
				\item There is traffic in the Newcastle city centre
				\item In the Newcastle city centre there is traffic
				\item There is traffic but not in the Newcastle city centre
				\item Newcastle city centre is trafficked
				\item It is busy in Newcastle
				\item Saturdays have usually busy city centers
				\item In Newcastle city center on Saturdays, traffic is flowing
				\item Traffic is flowing in Newcastle city centre, on Saturdays
				\item On Saturdays, traffic is flowing in Newcastle city centre
				\item Newcastle city centre has traffic
				\item Newcastle city center does not have traffic
				\item Newcastle has traffic but not in the city centre
				\item The busy Newcastle city centre is closed for traffic
			\end{enumerate}
		\end{minipage}
	}
	\hfill
	\subfloat[Expected clusters]{\label{subf:exp_clusters_newc}
		\begin{minipage}[t]{0.475\textwidth}
			\vspace{0pt}
			\footnotesize
			\begin{itemize}[leftmargin=*]
				\item[] \hspace*{\dimexpr-\labelindent-\labelwidth-\labelsep+\itemindent} \{0, 1, 9\}: \hspace{\itemindent}
				\begin{itemize}[leftmargin=1.5pt, labelsep=2pt]
					\item There is traffic in the Newcastle city centre
					\item In the Newcastle city centre there is traffic
					\item Newcastle city centre has traffic
				\end{itemize}
				
				\item[] \hspace*{\dimexpr-\labelindent-\labelwidth-\labelsep+\itemindent} \{2\}: \hspace{\itemindent}  There is traffic but not in the Newcastle city centre
				\item[] \hspace*{\dimexpr-\labelindent-\labelwidth-\labelsep+\itemindent} \{3\}: \hspace{\itemindent}  Newcastle city centre is trafficked
				
				\item[] \hspace*{\dimexpr-\labelindent-\labelwidth-\labelsep+\itemindent} \{4\}: \hspace{\itemindent} It is busy in Newcastle
				\item[] \hspace*{\dimexpr-\labelindent-\labelwidth-\labelsep+\itemindent} \{5\}: \hspace{\itemindent} Saturdays have usually busy city centers
				\item[] \hspace*{\dimexpr-\labelindent-\labelwidth-\labelsep+\itemindent} \{6, 7, 8\}: \hspace{\itemindent}
				\begin{itemize}[leftmargin=1.5pt, labelsep=2pt]
					\item In Newcastle city center on Saturdays, traffic is flowing
					\item Traffic is flowing in Newcastle city centre, on Saturdays
					\item On Saturdays, traffic is flowing in Newcastle city centre
				\end{itemize}
				\item[] \hspace*{\dimexpr-\labelindent-\labelwidth-\labelsep+\itemindent} \{10\}: \hspace{\itemindent} Newcastle city center does not have traffic
				\item[] \hspace*{\dimexpr-\labelindent-\labelwidth-\labelsep+\itemindent} \{11\}: \hspace{\itemindent}  Newcastle has traffic but not in the city centre
				
				\item[] \hspace*{\dimexpr-\labelindent-\labelwidth-\labelsep+\itemindent} \{12\}: \hspace{\itemindent} The busy Newcastle city centre is closed for traffic
			\end{itemize}
		\end{minipage}
	}
	\centering
	\caption{Sentences and expected clusters for \ref{rqn2c}.}
	\label{sentences:newc}
\end{figure}
%

The clustering results now seem in line with the classification scores from \tablename~\ref{table:metric_results_newc}: DeBERTaV2+AMR-LDA consistently provided low scores, being slightly improved by our preliminary simplistic \gls{sg} representation. Despite the good ability of \glspl{lg} to cluster the sentences, they less favourably capture correct sentence classifications, as (consistently with the previous results) they are outperformed by competing transformer-based approaches which, still, provided less than random scores in terms of accuracy. Even in this scenario, the competing approaches exhibited a lower weighted average compared to the macro one, thus indicating that the results were not biased by the unbalanced dataset. 

\begin{table}[!h]
	\caption{Classification scores for \ref{rqn2c} sentences, with the best value for each row highlighted in bold, blue text and the worst values highlighted in red. The classes are distributed as such: Implication: 32, Inconsistency: 27, Indifference: 110.}\label{table:metric_results_newc}
	\centering
	\begin{tblr}{
			width = 0.5\textwidth,
			column{4-12} = {colsep=2pt,font=\footnotesize},
			row{1} = {font=\normalsize},
			cells = {c},
			row{1} = {Alto},
			cell{2}{1} = {r=2}{Alto},
			cell{2}{2} = {r=2}{Mercury},
			cell{2}{3} = {Gallery},
			cell{3}{3} = {Gallery},
			cell{4}{1} = {r=4}{Alto},
			cell{4}{2} = {r=2}{Mercury},
			cell{4}{3} = {Gallery},
			cell{5}{3} = {Gallery},
			cell{6}{2} = {r=2}{Mercury},
			cell{6}{3} = {Gallery},
			cell{7}{3} = {Gallery},
			cell{8}{1} = {r=4}{Alto},
			cell{8}{2} = {r=2}{Mercury},
			cell{8}{3} = {Gallery},
			cell{9}{3} = {Gallery},
			cell{10}{2} = {r=2}{Mercury},
			cell{10}{3} = {Gallery},
			cell{11}{3} = {Gallery},
			cell{12}{1} = {r=4}{Alto},
			cell{12}{2} = {r=2}{Mercury},
			cell{12}{3} = {Gallery},
			cell{13}{3} = {Gallery},
			cell{14}{2} = {r=2}{Mercury},
			cell{14}{3} = {Gallery},
			cell{15}{3} = {Gallery},
			vlines,
			hline{1-2,4,8,12,16} = {-}{},
			hline{3,5,7,9,11,13,15} = {3-12}{},
			hline{6,10,14} = {2-12}{},
		}
		\textbf{Metric}     & \textbf{Average}  & \textbf{Clustering}         & \textbf{\glspl{sg}} & \textbf{\glspl{lg}} & \textbf{Logical} & \textbf{T1} & \textbf{T2} & \textbf{T3} & \textbf{T4} & \textbf{T5} & \textbf{T6} \\
		Accuracy & -- & \textbf{HAC} & \lsc{0.21}& 0.23& \textbf{\textcolor{blue}{1.00}} & 0.28& 0.29& 0.27& 0.29& \lsc{0.21}& 0.29\\
		& & \textbf{$k$-Medoids} & \lsc{0.21}& 0.23& \textbf{\textcolor{blue}{1.00}} & 0.28& 0.29& 0.27& 0.29& \lsc{0.21}& 0.29\\
		F1 & Macro & \textbf{HAC} & 0.24& 0.28& \textbf{\textcolor{blue}{1.00}} & 0.37& 0.38& 0.34& 0.38& \lsc{0.21}& 0.37\\
		&  & \textbf{$k$-Medoids} & 0.24& 0.28& \textbf{\textcolor{blue}{1.00}} & 0.37& 0.37& 0.34& 0.38& \lsc{0.21}& 0.37\\
		& Weighted & \textbf{HAC} & 0.13& 0.15& \textbf{\textcolor{blue}{1.00}} & 0.21& 0.22& 0.18& 0.22& \lsc{0.11}& 0.20\\
		&  & \textbf{$k$-Medoids} & 0.13& 0.15& \textbf{\textcolor{blue}{1.00}} & 0.21& 0.20& 0.18& 0.22& \lsc{0.11}& 0.20\\
		Precision & Macro & \textbf{HAC} & 0.35& 0.37& \textbf{\textcolor{blue}{1.00}} & 0.46& 0.49& 0.35& 0.54& \lsc{0.20}& 0.37\\
		& & \textbf{$k$-Medoids} & 0.35& 0.37& \textbf{\textcolor{blue}{1.00}} & 0.46& 0.36& 0.35& 0.54& \lsc{0.20}& 0.37\\
		& Weighted & \textbf{HAC} & 0.20& 0.20& \textbf{\textcolor{blue}{1.00}} & 0.37& 0.43& 0.19& 0.53& \lsc{0.11}& 0.20\\
		&  & \textbf{$k$-Medoids} & 0.20& 0.20& \textbf{\textcolor{blue}{1.00}} & 0.37& 0.20& 0.19& 0.53& \lsc{0.11}& 0.20\\
		Recall & Macro & \textbf{HAC} & \lsc{0.41}& 0.46& \textbf{\textcolor{blue}{1.00}} & 0.54& 0.54& 0.52& 0.54& \lsc{0.41}& 0.56\\
		&  & \textbf{$k$-Medoids} & \lsc{0.41}& 0.46& \textbf{\textcolor{blue}{1.00}} & 0.54& 0.56& 0.52& 0.54& \lsc{0.41}& 0.56\\
		& Weighted & \textbf{HAC} & \lsc{0.21}& 0.23& \textbf{\textcolor{blue}{1.00}} & 0.28& 0.29& 0.27& 0.29& \lsc{0.21}& 0.29\\
		&  & \textbf{$k$-Medoids} & \lsc{0.21}& 0.23& \textbf{\textcolor{blue}{1.00}} & 0.28& 0.29& 0.27& 0.29& \lsc{0.21}& 0.29\\      
	\end{tblr}
	
\end{table}

\section{Discussion}\label{sec:discussion}
This section begins with a brief motivation on our design choice for considering shorter sentences before discussing more complex ones (Section \ref{explanaStudy}). It then continues with a more detailed ablation study regarding our pipeline (Section \ref{subsec:pip_abla}), as well as comparing the different types of explainability achievable by current explainers if compared to our proposed approach (Section \ref{explainstudy}). \supplementaryname~\ref{supp:consider} provides further some preliminary considerations carried out when analysing the logical representation output as returned by our pipeline.

\subsection{Using Short Sentences}\label{explanaStudy}
We restricted our current analysis to full texts with no given structure, as in ConceptNet \cite{DBLP:conf/aaai/SpeerCH17}, instead of being parsed as semantic graphs. If there are no major ellipses in a sentence, it can be fully represented using \textit{propositional logic}; otherwise, we need to exploit existential quantifiers (which are now supported by the present pipeline).  In propositional logic, the truth value of a complex sentence depends on the truth values of its simpler components (i.e., propositions) and the connectives that join them. Therefore, using short sentences for logical rewriting is essential, as their validity directly influences that of larger, more complex sentences constructed from them. If the short sentences are logically sound, the resulting rewritten sentences will also be logically sound and we can ensure that each component of the rewritten sentence is a well-formed formula, thereby maintaining logical consistency throughout the process. For example, consider the sentence ``\textit{It is busy in Newcastle city centre because there is traffic}'' This sentence can be broken down into two short sentences: ``\textit{It is busy in Newcastle city centre}'' and ``\textit{There is traffic}''. These short sentences can be represented as propositions, such as $P$ and $Q$. The original sentence can then be expressed as $Q \wedge P$. Thus, through the use of short sentences, we ensure that the overall sentence adheres to the rules of propositional logic \cite{Harrison_2009}.

\subsection{LaSSI Ablation Study}\label{subsec:pip_abla}
We implicitly performed an initial ablation study in Section \ref{sub:clustering}: our three \textit{ad hoc} representation steps (\glspl{sg}, \glspl{lg}, and Logical) demonstrate how the introduction (or removal) of stages in the pipeline can enforce proper semantic and structural understanding of sentences given different scenarios (\ref{rq2}). Here, we extend this study by disabling the \textit{a priori} phase of each pipeline stage and comparing the results, thus implicitly limiting the possibility of reasoning through entities being expressed within our Parmenides KB. As the \textit{ex post} phase mainly reflects the computation of the confidence score required for the logical phase to derive a sensible score, the generation of the scores (both for the clustering and the classification tasks), and the generation of the dendrograms, it was not considered in this analysis. Thus, we focused on the different stages of the \textit{ad hoc} phase while considering the dis-/enabling of the \textit{a priori} phase.

Concerning the datasets for \ref{rqn2a} and \ref{rqn2b}, we obtained the same dendrograms and clustering results: this should be ascribed to any missing semantic entities of interest, as mainly common entities were involved. As a consequence, we obtain the same clustering and classification results.  Due to the redundant nature of these plots, the dendrogram plots have been moved to \supplementaryname~\ref{sub:incorrect_sv_rel}.


We now discuss the dendrograms and classification outcomes for the \ref{rqn2c} dataset. \figurename~\ref{fig:clustering_disabled_dend_newcastle} shows different dendrograms. When disabling the \textit{a priori} phase and the consequent \gls{meudb} match, we can see that fewer entities are matched, thus resulting in a slight decrease in the similarity values. Here, the difference lies in the logical implication of the sentences, and more marked results regarding the logical representation of sentences can be noticed. As sentences are finally reduced to the number of possible worlds that their atoms can generate, we can better appreciate the differences in similarity with a more marked gradient. Making a further comparison with the clusters, we observe that the further semantic rewriting step undertaken in the final part of the \textit{ad hoc}  phase is the one that fully guarantees uniformity of the representation of sentences, which is maintained despite failure to recognise the entities in a correct way. To better discriminate the loss of precision due to the lack of recognition of the main entities, we provide the obtained results in  \tablename~\ref{table:metric_results_newc_disabled_ad_hoc}. It can be clearly seen that \glspl{sg} are not affected by the \textit{a priori} phase as, in this section, no further semantic rewriting is performed and all the nodes are flattened out as merely nodes containing only textual information. On the other hand, the \textit{a priori} phase seems to negatively affect \glspl{lg}, as further distinction between the main entity and specification does not improve the transformer-based node and edge similarity, given that part of the information is lost.  As the  \textit{ex post}   phase requires the preliminary recognition of entities for computation of the confidence score, and given that disabling the \textit{a priori} phase leads to the missed recognition of entities, the improvement in scores can be merely ascribed to the correctness of the reasoning abilities implemented through Boolean-based classical semantics. Comparing the results for the logical representation with those of the competing approaches in \tablename~\ref{table:metric_results_newc}, we see that---notwithstanding the lack of multi-entity recognition, which is instrumental in connecting the entities within the intermediate representation leading to the final graph one---the usage of a proper logical reasoning mechanism allowed our solution to still achieve globally better classification scores, when compared to those used for comparison.

\begin{table}[h]
	\caption{Classification scores for \ref{rqn2c} sentences, comparing transformation stages with \textit{a priori} phase disabled in each case.}
	\label{table:metric_results_newc_disabled_ad_hoc}\centering
	\begin{tblr}{
			width = 0.5\textwidth,
			column{4-9} = {colsep=2pt,font=\footnotesize},
			row{1} = {font=\normalsize},
			cells = {c},
			row{1} = {Alto},
			row{2} = {Mercury},
			row{3} = {WildSand},
			cell{1}{1} = {r=3}{},
			cell{1}{2} = {r=3}{},
			cell{1}{3} = {r=3}{},
			cell{1}{4} = {c=2}{},
			cell{1}{6} = {c=2}{},
			cell{1}{8} = {c=2}{},
			cell{2}{4} = {c=2}{},
			cell{2}{6} = {c=2}{},
			cell{2}{8} = {c=2}{},
			cell{4}{1} = {r=2}{Alto},
			cell{4}{2} = {r=2}{Mercury},
			cell{4}{3} = {Gallery},
			cell{5}{3} = {Gallery},
			cell{6}{1} = {r=4}{Alto},
			cell{6}{2} = {r=2}{Mercury},
			cell{6}{3} = {Gallery},
			cell{7}{3} = {Gallery},
			cell{8}{2} = {r=2}{Mercury},
			cell{8}{3} = {Gallery},
			cell{9}{3} = {Gallery},
			cell{10}{1} = {r=4}{Alto},
			cell{10}{2} = {r=2}{Mercury},
			cell{10}{3} = {Gallery},
			cell{11}{3} = {Gallery},
			cell{12}{2} = {r=2}{Mercury},
			cell{12}{3} = {Gallery},
			cell{13}{3} = {Gallery},
			cell{14}{1} = {r=4}{Alto},
			cell{14}{2} = {r=2}{Mercury},
			cell{14}{3} = {Gallery},
			cell{15}{3} = {Gallery},
			cell{16}{2} = {r=2}{Mercury},
			cell{16}{3} = {Gallery},
			cell{17}{3} = {Gallery},
			vlines,
			hline{1,4,6,10,14,18} = {-}{},
			hline{2-3} = {4-9}{},
			hline{5,7,9,11,13,15,17} = {3-9}{},
			hline{8,12,16} = {2-9}{},
		}
		\textbf{Metric}     & \textbf{Average}  & \textbf{Clustering}         & \textbf{\glspl{sg}} &            & \textbf{\textbf{\glspl{lg}}} &            & \textbf{Logical} &            \\
		&                   &                             & \textbf{Ad Hoc}     &            & \textbf{Ad Hoc}              &            & \textbf{Ad Hoc}  &            \\
		&                   &                             & \textbf{Y}          & \textbf{N} & \textbf{Y}                   & \textbf{N} & \textbf{Y}       & \textbf{N} \\
		Accuracy & -- & \textbf{HAC} & \lsc{0.21}& \lsc{0.21}& 0.23& 0.24& \textbf{\textcolor{blue}{1.00}} & 0.33\\
		&  & \textbf{$k$-Medoids} & \lsc{0.21}& \lsc{0.21}& 0.23& 0.24& \textbf{\textcolor{blue}{1.00}} & 0.33\\
		F1 &Macro   & \textbf{HAC} & \lsc{0.24}& \lsc{0.24}& 0.28& 0.30& \textbf{\textcolor{blue}{1.00}} & 0.42\\
		&  & \textbf{$k$-Medoids} & \lsc{0.24}& \lsc{0.24}& 0.28& 0.30& \textbf{\textcolor{blue}{1.00}} & 0.42\\
		& Weighted   & \textbf{HAC} & \lsc{0.13}& \lsc{0.13}& 0.15& 0.16& \textbf{\textcolor{blue}{1.00}} & 0.23\\
		&  & \textbf{$k$-Medoids} & \lsc{0.13}& \lsc{0.13}& 0.15& 0.16& \textbf{\textcolor{blue}{1.00}} & 0.23\\
		Precision & Macro  & \textbf{HAC} & \lsc{0.35}& \lsc{0.35}& 0.37& 0.37& \textbf{\textcolor{blue}{1.00}} & 0.36\\
		&   & \textbf{$k$-Medoids} & \lsc{0.35}& \lsc{0.35}& 0.37& 0.37& \textbf{\textcolor{blue}{1.00}} & 0.36\\
		& Weighted   & \textbf{HAC} & \lsc{0.20}& \lsc{0.20}& \lsc{0.20}& \lsc{0.20}& \textbf{\textcolor{blue}{1.00}} & \lsc{0.20}\\
		&   & \textbf{$k$-Medoids} & \lsc{0.20}& \lsc{0.20}& \lsc{0.20}& \lsc{0.20}& \textbf{\textcolor{blue}{1.00}} & \lsc{0.20}\\
		Recall &Macro   & \textbf{HAC} & \lsc{0.41}& \lsc{0.41}& 0.46& 0.48& \textbf{\textcolor{blue}{1.00}} & 0.63\\
		&  & \textbf{$k$-Medoids} & \lsc{0.41}& \lsc{0.41}& 0.46& 0.48& \textbf{\textcolor{blue}{1.00}} & 0.63\\
		& Weighted   & \textbf{HAC} & \lsc{0.21}& \lsc{0.21}& 0.23& 0.24& \textbf{\textcolor{blue}{1.00}} & 0.33\\
		&  & \textbf{$k$-Medoids} & \lsc{0.21}& \lsc{0.21}& 0.23& 0.24& \textbf{\textcolor{blue}{1.00}} & 0.33\\
	\end{tblr}
	
\end{table}

\subsection{Explainability Study}\label{explainstudy}

{The research paradigm known as Design Science Research  focusses on the creation and verification of information science prescriptive knowledge while assessing how well it fits with research objectives \cite{0f2317e9-c1a1-3424-acbd-8bf00e2160a0}. While the previous set of experiments show that the current methodology attempts to overcome limitations of current state-of-the-art approaches, this section will focus on determining the suitability of LaSSI at explaining the reason leading to the final confidence score, to be used both as a similarity score and as a classification outcome. The remainder of the current subsection will be then structured according to the rigid framework of Design Science Research as identified by Johannesson and Perjons \cite{DBLP:books/sp/JohannessonP21}. This will ensure the objectiveness of our outlined considerations.}

\subsubsection{Explicate Problem}

While considering explanation classification for textual content, we seek an explainable methodology motivating why the classifier returned the expected class for a specific text. At the time of the writing, given that the pre-trained language models act as black boxes, the only possible way to derive the explanation for the classification outcome from the text is to train another white-box classifier, often referred to as an Explainer. This acts as an additional classifier correlating single features to the classification outcome, thus potentially introducing further classification errors \cite{Lundberg2020}. Currently, explainers for textual classification tasks weight each specific word or passage of the text: despite such characterisation being sufficient for sentiment analysis \cite{10.1007/978-981-19-8234-7_6} or misinformation detection \cite{Paper3}, these cannot adequately represent the notion of semantic entailment requiring the definition of a correlation between premise and consequence as occurring within the text, also requiring to target deeper semantic correlations across two distinct parts of the given implication.

\subsubsection{Define Requirements}

Given that current explainers cannot explicitly derive any trained model reasoning using explanations similar to the chain of thought prompting \cite{DBLP:conf/nips/Wei0SBIXCLZ22} as they merely correlate the features occurring within the text with the classification label, requiring this assumption will bias the evaluation against real-world explainers. 
The above considerations limit the correctness and desiderata to the basic characteristics that an explainer must possess:

\begin{requirements}
	\item \label{perfdegr} The trained model used by the explainer should minimise the degradation of classification performances.

	\item \label{intuitive} The explainer should provide an intuitive explanation of the motivations why the text correlates with the classification outcome.
	
	\item \label{conn} The explainer should derive connections between semantically entailing words towards the classification task.
	\begin{requirements}
		\item \label{simplereq} The existence of one single feature should not be sufficient to derive the classification: when this occurs, the model will overfit a specific dataset rather than learning to understand the general context of the passage.
	\end{requirements}
\end{requirements}

\subsubsection{Design And Develop}\label{subsub:design-and-develop}

At the time of the writing, both LIME \cite{LIME,LIME2,VisaniBC20} and SHAP \cite{SHAP,SHAP2} values require an extra off-the-shelf classifier to support explaining words or passages within the text in relation to the classification label. We then pre-process our annotated dataset to create pairs of strings as in \supplementaryname~\ref{debertaprocessing}, for then associating the expected classification outcome. The resulting corpus  is  used to fit the following models:

\begin{description}
	\item[TF-IDFVec+DT:] TF-IDF Vectorisation \cite{10.5555/3285754} is a straightforward approach to represent each document within a corpus as a vector, where each dimension describes the TF-IDF value \cite{Manning_Raghavan} for each word in the document. After vectorising the corpus, we fit a \gls{dt} for learning the correlation between word frequency and classification outcome. Stopwords such as ``the'' typically have high IDF scores, as they might frequently occur within the text. We retain all the occurring words to minimise our bias when training the classifier. As this violates \ref{simplereq}, we decide to pair this mechanism with the following, being attention-based.
	
	\item[DistilBERT+Train:] DistilBERT \cite{DBLP:journals/corr/abs-1910-01108} is a transformer model designed to be fine-tuned on tasks that use entire sentences (potentially masked) to make decisions \cite{DBLP:conf/taai/BaiCMS20}. It uses the WordPiece subword segmentation to extract features from the full text. We use this transformer to go beyond straightforward word tokenisation as the former. Thus, this approach will not violate \ref{simplereq} if the attention mechanism will not focus on one single word to draw conclusions, thus remarking their impossibility to draw correlations across the two sentences.
\end{description}

The resulting trained model is then fed to a LIME and SHAP explainer explaining how single word frequencies (TF-IDFVec+DT) or sentence parts (DistilBERT+Train) correlate with the expected classification label.

\subsubsection{Artifact Evaluation}\label{sec:areval}
We decide to train the previous models over the \ref{rqn2c} dataset, as this is more semantically rich:  correlations across entailing sentences are quintessential, while both term similarity and logical connectives should be considered. So, to avoid any potential bias a classifier introduces when providing the classification labels, we train the models from the former section directly on the annotated dataset.

\paragraph*{Performance Degradation}

We discuss \ref{perfdegr}: \tablename~\ref{tab:perefdeg} showcases a straightforward \gls{dt} and frequency-based classification task outperform a re-trained language model. While the former model clearly over-fits over the term frequency distribution, thus potentially leading to deceitful explanations, the latter might still derive wrong explanations due to low model precision. Higher values on the weighted averages entail that the classifiers are biased towards the majority class, indifference.

\begin{table}[!h]
	\caption{Performance degradation when training a preliminary model used by the explainer to correlate parts of text to the classification label. Refer to Section \ref{sub:spatio} for classification results from LaSSI. }\label{tab:perefdeg}
	\begin{adjustbox}{max width=1\textwidth}
		\begin{tabular}{lccccccc}
			\toprule
			& \multirow{2}{*}{Accuracy} & \multicolumn{2}{c}{F1} & \multicolumn{2}{c}{Precision} & \multicolumn{2}{c}{Recall} \\ \cline{3-8}
			&          & Macro & Weighted & Macro & Weigthed & Macro & Weighted \\
			\midrule
			TF-IDFVec+DT & 0.95 & 0.93 & 0.94 & 0.95 & 0.94 & 0.92 & 0.94 \\ 
			DistilBERT+Train & \textcolor{red}{0.76} & \textcolor{red}{0.51} & \textcolor{red}{0.69} & \textcolor{red}{0.45} & \textcolor{red}{0.64} & \textcolor{red}{0.61} & \textcolor{red}{0.76} \\
			\bottomrule
		\end{tabular}
	\end{adjustbox}
	
\end{table}

\paragraph*{Intuitiveness}

We discuss \ref{intuitive}: an intuitive explanation should clearly show \textit{why} the model made a specific classification based on the input text, ideally in a way that aligns with human understanding or, at a minimum, reveals the model's internal logic. 

\begin{figure}[!hp]
	\centering
	\subfloat[LIME explanation using TF-IDFVec+DT, highlighting the word ``\textit{the}'' in both \uline{NOT Indifference} and \uline{Implication}, leading to a 100\% confidence for an \uline{Implication} classification.]{
		\includegraphics[width=1\textwidth]{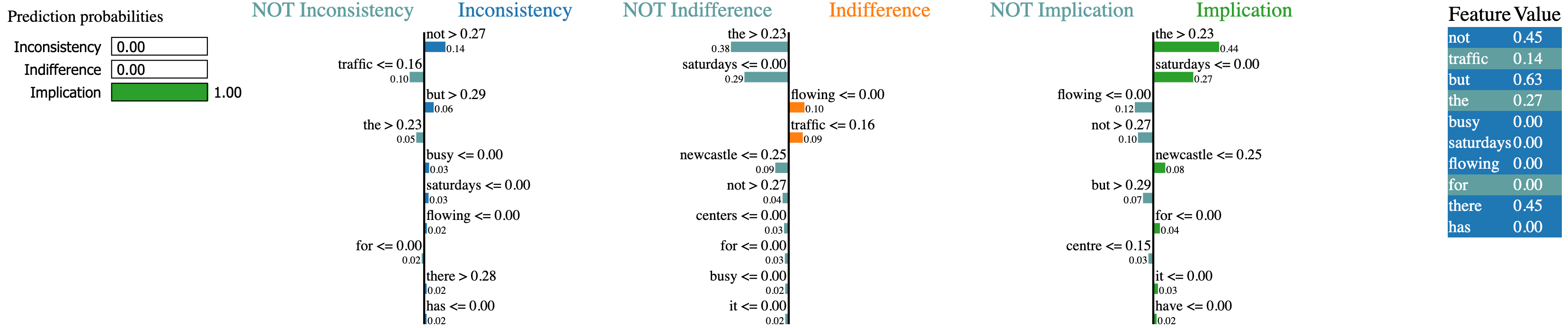}\quad 
		\label{fig:lime_tfidf_but_not}
	}
	\vspace{10pt}
	\subfloat[LIME explanation using DistilBERT+Train, highlighting the word ``\textit{not}'' leading to a \uline{NOT implication} and \uline{Inconsistency} classifications, ultimately resulting in an \uline{Inconsistency} classification.]{
		\includegraphics[width=1\textwidth]{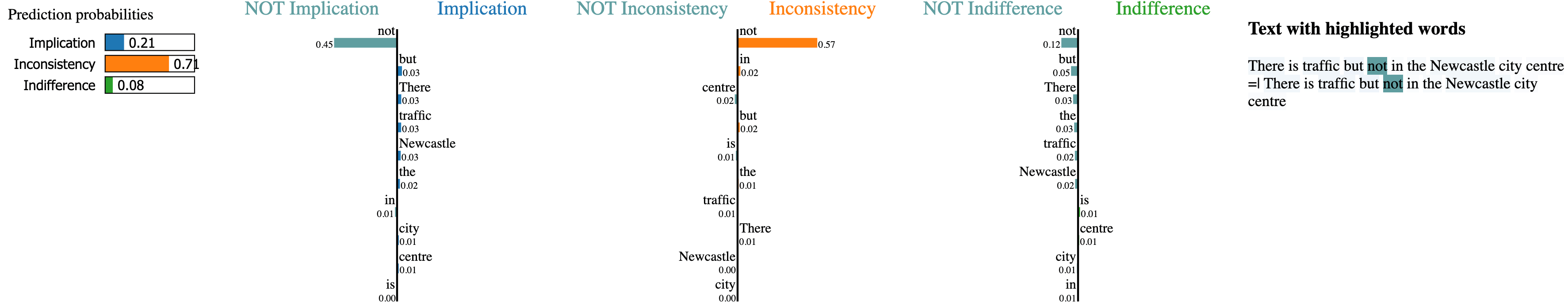}\quad 
		\label{fig:lime_distilbert_but_not}
	}
	\vspace{10pt}
	\subfloat[SHAP explanation using TF-IDFVec+DT, highlighting the model's confidence in the word ``\textit{the}'' leading to a \uline{Implication} classification (1 for the image above).]{
		\includegraphics[width=1\textwidth]{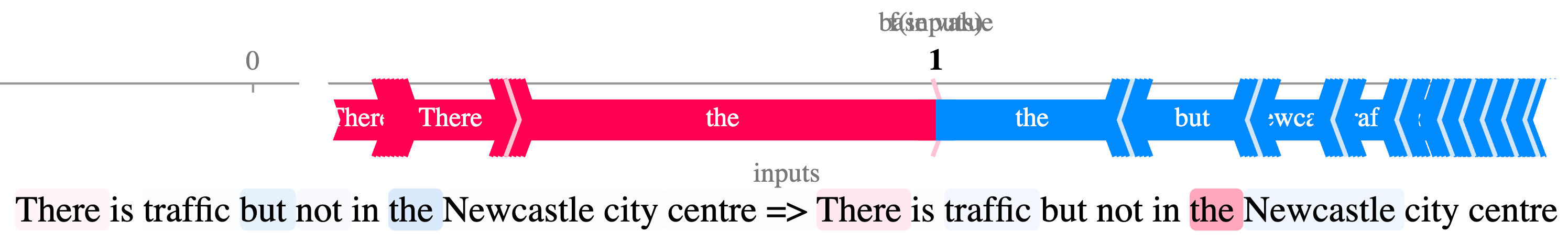}
		\label{fig:shap_tfidf_but_not}
	}
	\vspace{10pt}
	\subfloat[SHAP explanation using DistilBERT+Train, highlighting the model's confidence in the word ``\textit{not}'' leading to an \uline{Inconsistency} classification.]{
		\includegraphics[width=1\textwidth]{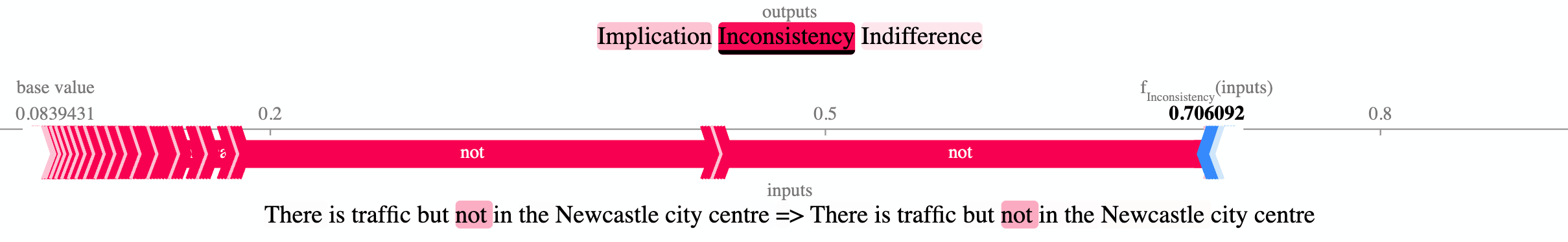}\quad 
		\label{fig:shap_distilbert_but_not}
	}
	\caption{LIME and SHAP explanations comparing ``\textit{There is traffic but not in the Newcastle city centre}'' against itself.}
	\label{fig:explanations_but_not}
\end{figure}

LIME plots display bar charts where each bar corresponds to a feature. The length of the bar illustrates the feature's importance for that particular prediction, and the colour indicates the direction of influence towards a class. By examining the features with the longest bars and their associated colours, we can understand which factors were most influential in the model's decision for that instance. Individual LIME plots are self-explanatory. However, when comparing across plots of different sentences and models (\figurename~\ref{fig:lime_tfidf_but_not} and \ref{fig:lime_distilbert_but_not}), legibility could have been improved if the order and the colour of classes could be fixed.

SHAP force plots (\figurename~\ref{fig:shap_tfidf_but_not} and \ref{fig:shap_distilbert_but_not}) show which features most influenced the model's prediction for a single observation: features coloured in red increase the confidence of the prediction, while ones in blue lower the estimation. These values are laid along a horizontal axis while meeting along the line reflecting the classification outcome.  This allows us to identify the most significant features and observe how their varying values correlate with the model's output. Unlike the LIME plots, the visualisation does not immediately display the force plot for all the classes, which must be manually selected from the graphical interface. Thus, we choose to present in this paper only the results for the predicted class; the full plots are available through OSF.io at the given URL.

\begin{figure}[!hp]
	\centering
	\subfloat[LaSSI explanation showing that sentence 2 is 50\% indifferent to sentence 11.]{
		\includegraphics[width=0.294\textwidth]{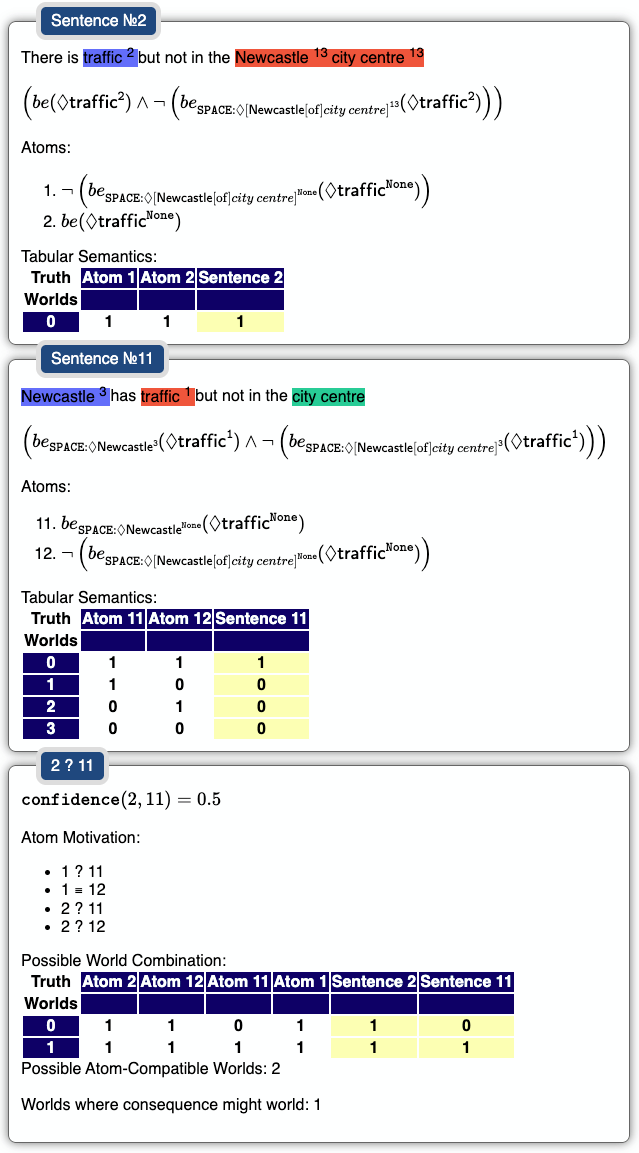}\quad 
		\label{fig:lassi_2?11}
	}
	\hfill
	\subfloat[LaSSI explanation shows 100\% implication between sentence 11 and sentence 2.]{
		\includegraphics[width=0.2975\textwidth]{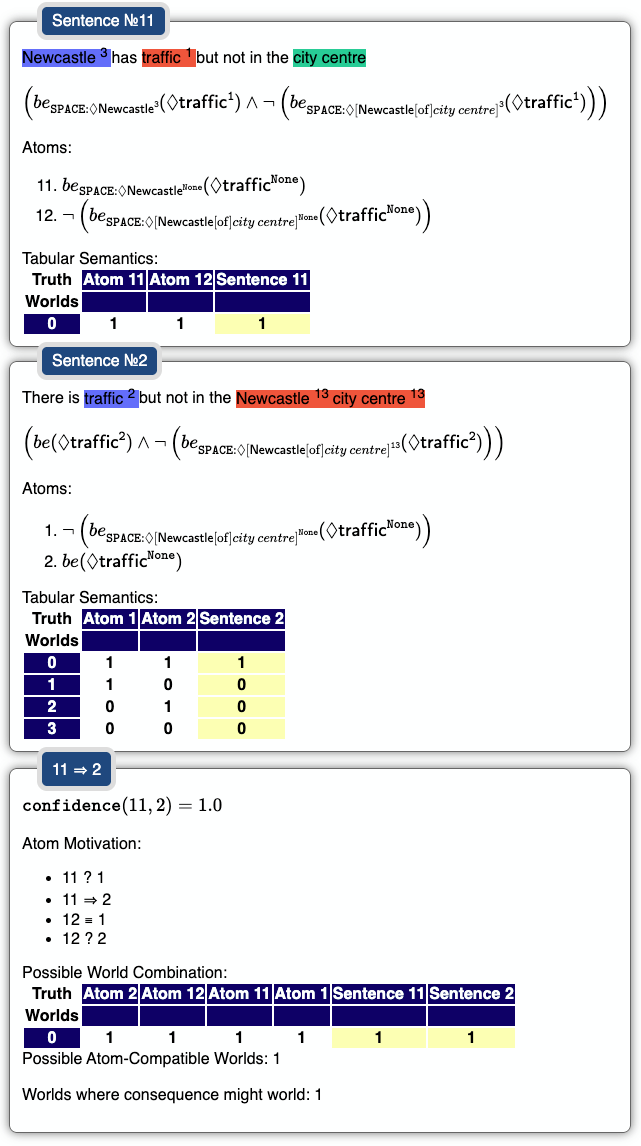}\quad 
		\label{fig:lassi_11=>2}
	}
	\hfill
	\subfloat[LaSSI explanation shows 100\% implication between sentence 11 and sentence 11.]{
		\includegraphics[width=0.3097\textwidth]{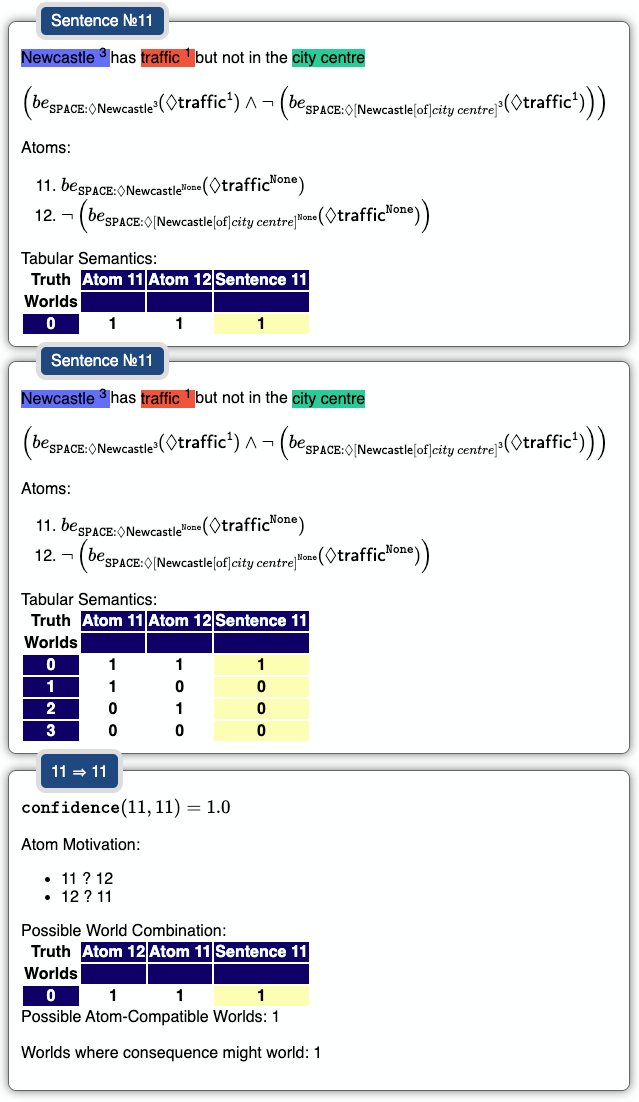}\quad 
		\label{fig:lassi_11=>11}
	}
	\caption{LaSSI explanations for indifference and implication between sentence 2 (``\textit{There is traffic but not in the Newcastle city centre}'') and sentence 11 (``\textit{Newcastle has traffic but not in the city centre}''). Columns in yellow provide the truth values for the sentences given the atoms' truth values on the left.}
	\label{fig:lassi_explanations}
\end{figure}

The LaSSI explanations in Figure \ref{fig:lassi_explanations} show how a sentence is changed into logic and how the subsequent confidence score is calculated without relying on an external tool for deriving the desired information. For the logical notation, we use the same syntax as per Equation \ref{folsyntax}, while equivalence ($\equiv$), implication ($\Rightarrow$), indifference (?) and inconsistency ($\neq$) show a different notation due to web browser limitations. The outputs for the logical rewriting are shown underneath each sentence. The process for extracting the atoms out of the logical representation is described in Example \ref{ex:alicebob}, while the description of how the ``Tabular Semantics'' are calculated from the atoms is found in the rest of Section \ref{sec:expostexmpl}. The coloured, highlighted words in the sentence correspond to all named entities in the logical representation. Differently from the other explainers, these do not directly remark on which words are relevant for the classification outcome but mainly reflect on the outcome of the \textit{a priori} phase. Given that computing the confidence score  requires  to assume the premise as true, we are interested in deriving the possible worlds where the premise's Sentence column is 1. Thus, given that all the sentences from our examples express a conjunction of their atoms, we are interested in the worlds where all atoms are true. \figurename~\ref{fig:lassi_explanations} reports, for all the first sentences being the rule premises, one row for their tabular semantics. For the sentence below representing the head of the implication rule, we list the whole combination of the possible worlds and the resulting truth values for the sentence.  ``Atom Motivation'' summarises the computation of the proposition equivalence as per Definition \ref{def:propequiv}, leading to the truth tables previously presented in Figure \ref{table:TT}. By naturally joining the tables derived for each pair of atoms with the two sentence tables above, we derive the  ``Possible World Combination'' as described in Example \ref{ex:alicebob}.

Without the former interpretation, it may not be intuitive to understand what these explanations show despite the improved logical soundness of the presented results. Still, they provide relevant insight into how the pipeline computes the confidence values provided in our former experiments.

\paragraph*{Explanation through Word Correlation}
Finally, we discuss \ref{conn}. This condition is trivially met for words occurring on the text and the expected class label, as the purpose of these explainers is to represent these correlations graphically. LaSSI achieves this through ``Atom Motivation'' and ``Possible World Combination''. Rather than relying on single words occurring within the text, we consider atoms to refer to text passages and not just single words alone, thus greatly improving the former. This also trivialises over \ref{simplereq}, as our model does not necessarily rely on one single word to draw its conclusions.

The rest of our discussion focuses on \ref{simplereq} for the competing solutions. 
TF-IDFVec+DT has substantial limitations in capturing semantics due to using a bag-of-words representation, ignoring grammar, word order, and semantic relationships like negation. A \gls{dt} operating on these non-semantic features cannot understand semantics but can partially reconstruct it by connecting occurring tokens while losing relevant semantic information. A \gls{dt} classifier learns rules by partitioning the feature space. If ``\textit{the}'' consistently correlates with the class \uline{Implication}, the DT can learn a rule like ``\texttt{IF TF-IDF\_score(`the') > threshold THEN class = `implication'}''. Nevertheless, this model can also explicitly correlate across different features when they are strong predictors for the class of interest. The TF-IDFVec+DT model classifies the example sentence as 100\% \uline{Implication}, attributing this to the presence of the word ``\textit{the}'':  \figurename~\ref{fig:lime_tfidf_but_not} shows \uline{Implication} has a bar for ``$\textit{the}>0.23$'' with $0.44$ confidence. We also see ``$\textit{the}>0.23$'' appearing in \uline{NOT Indifference} with $0.38$ confidence. Due to the explainer using all of the sentences related to Newcastle as training data, \figurename~\ref{fig:shap_tfidf_but_not} highlights that the lack of the word ``\textit{Saturdays}'' from each sentence also motivates the model's confidence in implication. The remaining words are either considered absent (values $\leq 0.0$) or have a considerably lower confidence score.

The DistilBERT+Train model classifies the same sentences as inconsistent, with 71\% confidence. This is primarily attributed to the word ``\textit{not}'' in \uline{NOT Implication} (and \uline{Inconsistency}) with $0.45$ (and $0.57$) confidence. 
Although it was designed to better capture word correlations by combining word tokenisation and an attention mechanism, its behaviour can be assimilated to the one given through TF-IDF Vectorisation, where single words are considered without learning to acknowledge the larger context. Unlike the former model, this shows that the two sentences are inconsistent by solely relying on the word ``\textit{not}''. As these two sentences are the same, the classification outcome of being \uline{Inconsistent} and its associated explanation is incorrect.




\startlandscape
\finishlandscape

Given the above, SHAP and LIME explanations are heavily hampered by the previously-trained model's ability to draw correlations between text and classification outcome beyond the single word of choice. As our explanation is automatically derived from the pipeline without the need to use any additional visualisation tool, it minimises the chances of providing a less meaningful representation of the semantic understanding of the sentence.

\subsubsection{Final Considerations}

\tablename~\ref{tab:finalsumm} summarises the findings from our previous evaluation: despite our proposed methodology not providing intuitive ways to visualise the derivations leading to the classification outcome, our approach meets all of the aforementioned correctness requirements. Future works will then postulate improved ways to better visualise the classification outcomes similarly to SHAP and LIME explainers.

\begin{table}[!h]
	\caption{Comparison table from the artefact evaluation. \CIRCLE = satisfies requirement; \LEFTcircle = partially satisfies requirement;  \Circle = does not satisfy requirement.}\label{tab:finalsumm}\centering
	\begin{adjustbox}{max width=1\textwidth}
		\begin{tabular}{ll|cccc}
			\toprule
			Explainer & Model & \ref{perfdegr} & \ref{intuitive} & \ref{conn} & \ref{simplereq} \\
			\midrule
			\multirow{2}{*}{SHAP} & TF-IDFVec+DT & \LEFTcircle & \CIRCLE & \LEFTcircle & \Circle \\ 
			& DistilBERT+Train & \Circle & \CIRCLE & \LEFTcircle & \Circle \\ 
			\multirow{2}{*}{LIME} & TF-IDFVec+DT & \LEFTcircle & \CIRCLE & \LEFTcircle & \Circle \\ 
			& DistilBERT+Train & \Circle & \CIRCLE & \LEFTcircle & \Circle \\ 
			\multicolumn{2}{c}{LaSSI} & \CIRCLE & \Circle &\CIRCLE &  \CIRCLE \\
			
			\bottomrule
		\end{tabular}
	\end{adjustbox}
	
\end{table}

Given that current state-of-the-art explainers still heavily rely on training another model to derive correlations between input and output data. Consequently, minor differences in the data sampling strategies might produce highly dissimilar explanations even for closely related data points \cite{10.5555/3666122.3669249}. Most importantly, it is widely known that both LIME and SHAP explainers can be easily manipulated to hide evident biases from the data  \cite{10.1145/3375627.3375830}. As any learning process naturally introduces biases when learning, more rigid control through rule-based mechanisms, such as the ones presented in this paper, should be used to mitigate such effects.


\section{Conclusions and Future Works}\label{sec:conclusion}

\gls{lassi} 
offers hybrid explainability within an  \gls{nlp} pipeline, which generates logical representations for 
factoid sentences 
which are left unparsed while generating common sense knowledge-based networks.
This remarks the possibility of further enhancing current \gls{nlp} parsers to better structure their internal representation: additional experiments in \appendixname~\ref{sub:benchmark} showcase the possibility of parsing sentences that were originally treated as single nodes rather than being expressed as full subgraphs in ConceptNet.
Through exploiting a curated Upper Ontology including both common-sense knowledge (ABox) and rewriting rules (TBox), we can easily capture the essence of similar sentences while remarking on the limitations of the existing state-of-the-art approaches.  Preliminary experiments suggest the inability of transformers to solve problems as closed Boolean expressions expressed in natural language and to reason paraconsistently when using cosine similarity metrics \cite{10.1162/tacl_a_00663}. Compared with pre-trained language models, we showcased their inability to fully capture deep sentence understanding after training over large corpora, even for short factoid sentences. 

Our current experiments clearly remark that, notwithstanding the possibility of pre-trained language models potentially supporting arbitrary sentence structure, they cannot soundly reason over sentences representable in propositional calculus, while also being affected by minimal text perturbations (Supplement \ref{debertaprocessing}). On the other hand, our approach remarks the possibility of reasoning over these representations while overcoming limitations of such approaches. In fact, if we cannot reason on factoid sentences representable on propositional calculus, then  we would not be able to reason on anything more complex as \gls{fol}. As propositional calculus is a subset of \gls{fol} and given the limitations of pre-trained models over propositional calculus, we clearly show their impossibility of providing sound reasoning on \gls{fol}, over which we can represent arbitrary sentences \cite{Montague+1975+94+121}. 

As we intend to test this pipeline on a large scale using real-world crawled data, future works will aim to further improve the presented results regarding this pipeline by first analysing the remaining sentences from ConceptNet, and then attempting to analyse other, more realistic scenarios. Furthermore, we will extend to a more diverse range of datasets: given that we are challenging a three-way classification distinguishing between logical implication, indifference, and inconsistency for the first time. As they stand, currently available datasets are unsuitable to test our system under our premises. In particular, we need to completely re-annotate these datasets to encompass the three-fold classification. 
As Parmenides is an Upper Ontology, mainly defined for lexicographical and grammatical analysis, it can be easily be extended to support Wikidata and DBpedia concepts and relationships through recent advancements in ontology alignment \cite{10.1007/978-3-031-81221-7_2}, thus enriching the current definition with more complex representations of the world. 

 In future work, we will also consider the possibility of bridging abductive reasoning \cite{Tsamoura_Hospedales_Michael_2021}, allowing for rule generalisation, as well as adopting relational learning algorithms, enabling learning over dirty data without cleaning \cite{Picado2019,picado2}. To the best of our knowledge, neither  abductive reasoning nor relational learning mechanisms provide learning capabilities over logical formulae potentially containing negations: this might be risky, as it could lead to the explosion problem as well as the derivation of unlikely conclusions over probabilistic data. In future works, we will thus consider integrating such approaches with the paraconsistency capabilities exploited in this study, as well as considering whether a rule-based approach can be used to minimise the effects of semantic leakage problem \cite{gonen-etal-2025-liking}.

Multidimensional scaling is a well-known methodology for representing distances between pairs of objects embedded as  points into Euclidean space \cite{multidim1,multidim2}, allowing for the derivation of object embeddings given their pairwise distances. Despite the existence of non-metric solutions \cite{DBLP:journals/jmlr/AgarwalWCLKB07,DBLP:journals/jmlr/QuistY04}, none of these approaches consider arbitrary divergence functions supporting neither the triangular inequality nor the symmetry requirement, as is the case for the confidence metric used in this paper. The possibility of finding such a technique can be expected to streamline the retrieval of similar constituents within the expansion phase by exploiting kNN queries \cite{DBLP:journals/geoinformatica/CostaNS18}, which can be further streamlined using vector databases and vector-driven indexing techniques \cite{DBLP:journals/geoinformatica/BoteaMNS08}. This can be then used to either overcome the mathematical limitations of current pre-trained language models, or to streamline and index the intermediate representations generated by \gls{lassi}.

Finally, we can see that deep semantic rewriting cannot be encompassed by the latest \gls{gql} standards \cite{bonifati2024transformingpropertygraphs}, nor generalised graphs  \cite{math12172677}, as both are context-free languages \cite{DBLP:books/aw/HopcroftU79} that cannot take into account the contextual information that can be only derived through semantic information. As these rewriting phases are not driven by \gls{fol} inferences, the inference step boils down to querying a \gls{kb} to disambiguate the sentence context and semantics; however,  most human languages are context-sensitive. This leads to the postulation that further automating and streamlining the rewriting steps detailed in Section \ref{sec:recursive}, Section \ref{sec:lsa}, and Section \ref{sec:logrewr}  will require the definition of a context-sensitive query language, which is currently missing from the existing database literature. All of the remaining rewriting steps were encoded using context-free languages. In future works, we will investigate the appropriateness of this language and whether it can be characterised through a straightforward algorithmic extension of \gls{ggg}.

%

\vspace{6pt} 





\authorcontributions{Conceptualisation, G.B.; methodology, O.R.F. and G.B.; software, O.R.F. and G.B; validation, O.R.F.; formal analysis, O.R.F., G.B.; investigation, O.R.F. and G.B.; resources, G.B.; data curation, O.R.F.; writing---original draft preparation, O.R.F.; writing---review and editing, O.R.F. and G.B.; visualisation, O.R.F.; supervision, G.B. and G.M.; project administration, G.B.; funding acquisition, G.B. All authors have read and agreed to the published version of the manuscript.}

\funding{O.R.F.'s research is sponsored by a UK EPSRC PhD Studentship.}

\institutionalreview{Not applicable.}

\informedconsent{Not applicable.}

\dataavailability{The dataset is publicly available at \url{https://osf.io/g5k9q/} (Accessed on 1 April 2025). The repository is available through GitHub (\url{https://github.com/LogDS/LaSSI}, Accessed on 28 April 2025).} 

\acknowledgments{The authors thank Tom McCone (MDPI) for English editing.}

%
%


\conflictsofinterest{The authors declare no conflicts of interest.} 



\abbreviations{Abbreviations}{
The following abbreviations are used in this manuscript:
\printglossaries
%
}

\appendixtitles{yes} 
\appendixstart
\appendix

\section{Classical Semantics, Bag Semantics, and Relational Algebra}

\subsection{Tabular Semantics per Sentence $\tau(\alpha)=A$}\label{app:tabular-semantics}
Given the impossibility of potentially enumerating all the possible infinite conditions where a specific situation might hold, we focus our interest on the worlds encompassed by the two formulae considered within the application of the confidence function. After extracting all the unary $\mathcal{U}(A)$ or binary $\mathcal{B}(B)$ atoms occurring within each formula and logical representation $A$, we can only consider the set of possible worlds covered by the truth or falsehood of each of these atoms. Thus, the set of all possible worlds where $A$ holds is the set of the worlds where each of the atoms within the formula holds while interpreting each formula using Boolean-valued classical semantics \cite{asperti}. Such semantics are showcased in Example \ref{ex:alicebob}.

As a next step, we want to derive whether each atom $p_i\in \mathcal{B}(A)\cup\mathcal{U}(A)$ occurring in each formula $A$ entails (\textbf{implication}), is equivalent to (bijective \textbf{implication}), is indifferent to (\textbf{indifference}, or is inconsistent (\textbf{inconsistency}) to another atom $p_j\in \mathcal{B}(B)\cup\mathcal{U}(B)$ in the other formula of interest. As the atoms in the original sentence can be further decomposed into other atoms that are either equivalent or logically entailing the former, to control that the machine produces sensible and correct rules from the data given, we exploit machine teaching techniques \cite{DBLP:journals/corr/SimardACPGMRSVW17,Ramos01112020} to ensure the machine derives correct propositions by exploiting a human-in-the-loop approach \cite{Mosqueira-Rey2023}. 
 To achieve this, we opt for rule-based semantics \cite{DBLP:conf/tac/ZhangSSHLPLZWWJ18,bergami2019frameworksupportingimprecisequeries}, expanding each distinct atom $p_i$ and $p_j$ mined from the full text. We exploit this as a common approach when designing Upper Ontologies \cite{DBLP:conf/fois/NilesP01}, where TBox reasoning rules are hardcoded to ensure the correctness of the inference approach.

Given the inability of description logic to represent matching and rewriting statements \cite{DBLP:phd/it/Bergami2018a} and given the lack of support flexible properties associated with predicates of current knowledge expansion algorithms that require a fixed schema \cite{DBLP:conf/tac/ZhangSSHLPLZWWJ18,bergami2019frameworksupportingimprecisequeries}, we perform this inference by exploiting an explicit pattern matching and rewriting mechanism. To achieve this, we exploit the Python library  FunctionalMatch (\url{https://github.com/LogDS/FunctionalMatch/releases/tag/v1.0}, Accessed on 30 March 2025) as a derivation mechanism $\mathcal{E}_{\Gamma,\mathcal{K}}(p_i)=\{p_j|\mathcal{K}\vDash p_j\}$ generating propositions $p_j$ out of the expansion rules of interest $\mathcal{K}$ representing common-sense information and relationships between the entities. We then apply expansion rules $\Gamma^\Rightarrow$ and $\Gamma^\equiv$, where the first derives a set of logically entailing propositions $E(p_i)=\mathcal{E}_{\Gamma^\Rightarrow,\mathcal{K}}(p_i)$, while the latter derives a set of equivalent propositions $T(p_i)=\mathcal{E}_{\Gamma^\equiv,\mathcal{K}}(p_i)$.

At this stage, we define the semantic equivalence $p_1\equiv^\textsf{prop} p_2$ between the expanded propositions using the Parmenides \gls{kb}, thus deriving a merely semantic correspondence between such propositions. For conciseness, this is detailed in \appendixname~\ref{propSem}. This provides a discrete categorisation of the general relationships that might occur between two propositions while remarking whether one entails the other ($\mstar$), if they are equivalent ($\eq$), if they are inconsistent ($\nq$), or if they are indifferent ($\omega$). Unlike in our previous paper, we then categorise the previous cases in order of priority as follows:
\begin{description}
\item[Equivalence:] if $p_1$ is structurally equivalent to $p_2$.
\item[Inconsistency:] if either $p_1$ or $p_2$ is the explicit negation of the other, or whether their negation appears within the expansion of the other ($T(p_1)$ and $E(p_2)$, respectively).
\item[Implication:] if $p_1$ occurs in one of the expansions $E(p_2)$.
\end{description}

If none of the above conditions holds, we compare the $T(p_1)$ and the $E(p_2)$ expanded propositions. Given $\varsigma$, the function prioritising the comparison outcomes over a set of compared values (Eq. \ref{varsigmadef} in the Appendix), we obtain the comparison outcome as: $\varsigma(\{p\equiv^\textsf{prop}q|p\in T(p_1),q\in E(p_2)\})$. After this, we associate each pair of propositions with a relational table from \figurename~\ref{table:TT}, from which we select only the possible worlds of interest where it is plausible to find the worlds occurring, where \textbf{indifference} is derived if none of the above conditions holds.

	\begin{figure}[!h]
		\centering
		\resizebox{.12\linewidth}{!}{
			\begin{tabular}{|>{\columncolor{lightGray}}c|c|}
				\hline \multicolumn{2}{|c|}{$R_{p_1\mstar p_2}$} \\
				\hline
				\rowcolor{lightGray}
				\cellcolor{darkGray}    $p_1$ & $p_2$\\ 
				\hline 
				0 & 0\\
				0 & 1\\
				1 & 1\\
				\hline
			\end{tabular}}\quad 		\resizebox{.095\linewidth}{!}{
			\begin{tabular}{|>{\columncolor{lightGray}}c|c|}
				\hline \multicolumn{2}{|c|}{$R_{p_1\omega p_2}$} \\
				\hline
				\rowcolor{lightGray}
				\cellcolor{darkGray}    $a_i$ & $b_j$\\
				\hline 
				0 & 0\\
				0 & 1\\
				1 & 0\\
				1 & 1\\
				\hline
			\end{tabular}}\quad 		\resizebox{.15\linewidth}{!}{
			\begin{tabular}{|>{\columncolor{lightGray}}c|c|}
				\hline \multicolumn{2}{|c|}{$R_{(a_i\nq b_j)}$} \\
				\hline
				\rowcolor{lightGray}
				\cellcolor{darkGray}    $a_i$ & $b_j$\\ 
				\hline 
				0 & 1\\
				1 & 0\\
				\hline
			\end{tabular}}\quad 		\resizebox{.15\linewidth}{!}{
			\begin{tabular}{|>{\columncolor{lightGray}}c|c|}
				\hline \multicolumn{2}{|c|}{$R_{(a_i\eq b_j)}$} \\
				\hline
				\rowcolor{lightGray}
				\cellcolor{darkGray}    $a_i$ & $b_j$\\ 
				\hline 
				1 & 1\\
				0 & 0\\
				\hline
			\end{tabular}}
		\caption{\label{table:TT}Truth tables over admissible and compatible worlds for implying, indifferent, inconsistent, and equivalent atoms.}
	\end{figure}

\subsection{Enumerating the Set of Possible Worlds Holding for a Formula}\label{app:formulartabular}

For relational algebra, we denote $\times$ as the cross product,  $\bowtie$ as the natural equi-join,  $\sigma_P$ as the select/filter operator over a binary predicate $P$, and $\pi_L$ as the projection operator selecting only the attributes appearing in $L$. We define $\textsf{Calc}_{f\textup{\,as\,}\texttt{A}}(T)$ as the non-classical relational algebra operator extending the relational table $T$ with a new attribute \texttt{A} while extending each tuple $(v_1,\dots,v_n)\in T$ via $f$ as $(v_1,\dots,v_n,f(v_1,\dots,v_n))$.  Given a logical formula $\varphi(s)$ and some truth assignments to logical predicates through a function $\Gamma$, we use $\sem{\varphi(s)}(\Gamma)$ to denote the \textit{valuation function} computed over $\varphi(s)$ via the truth assignments in $\Gamma$.

 	After determining all the binary or unary predicates $a_1,\dots,a_n$ associated to a \gls{lr} $\varphi(s)$ of a factoid sentence $s$, we derive a truth table $T_s(a_1,\dots,a_n,s)$ for $\varphi(s)$, represented as a relational table  $T_s=\textsf{Calc}_{\sem{\varphi(s)}\textup{\,as\,}\texttt{s}}(\times_{a_i\in \{a_1,\dots,a_n\}} \{0,1\})$, by assuming each proposition to be completely independent of the others without any further background knowledge. 

\subsection{Knowledge-Based Driven Propositional Semantics}\label{propSem}
Following in the footsteps of our previous research \cite{bergami2019frameworksupportingimprecisequeries}, we derive the following conditions through which we discretise the notion of logical implication between propositions and their properties: 

\begin{definition}[Multi-Valued Semantics]
Given a \gls{kb} $\mathcal{K}$, we say that two propositions or properties are equivalent ($\eq$) if either they satisfy the structural equivalence $\equiv$ formally defined as Leibniz equivalence \cite{Asperti_Ricciotti_Sacerdoti_Coen_2014}, or if they appear within a transitive closure of equivalence relationships in $\mathcal{K}$. We also say that these are inconsistent ($\nq$) if either one of the two is a negation and its argument is Leibniz equivalent to the other or, after a transitive closure of equivalence or logical implications, we derive that one argument is inconsistent in $\mathcal{K}$. Then, we distinguish multiple types of logical implications by considering the structure of the proposition extended with data properties:
\begin{description}
\item[$\mig$:] If we lose specificity due to some missing copula information.
\item [$\miss$:] By interpreting a missing value from one of the property arguments as missing information entailing any possible value for this field, whether the right element is a non-missing value.
\item[$\minst$:] If we interpret the second object as a specific instance of the first one.
\item[$\impl$:] A general implication state that cannot be easily categorised into any of the former cases while including any of them.
\end{description}
If none of the above applies, then we state that the two propositions or properties are indifferent, and we denote them with $\ndiff$. We denote any generic implication relationship as $\mstar$.
\end{definition} 

\subsubsection{Multi-Valued Term Equivalence}\label{mvte}
Please observe that, in contrast to the standard logic literature, we also consider the negation of terms---implicitly entailing any term being different from the one specified. This shorthand is required, as human language not only expresses negations over entire propositions or formulae, but also over single terms.

When the assessment of multi-valued semantics between two items $a$ and $b$ is determined through a \gls{kb} $\mathcal{K}$, we denote the outcome of such comparison as $a\equiv^\mathcal{K} b$. In the forthcoming steps, we provide the definitions for the notion of equivalence derivable from the analysis of the structural properties of the propositions and their terms through the \gls{kb} values given above. The \textit{transform when negated} function $\eta(v)$  implements the intended semantics of the negation of a multi-valued semantics comparison value: $\eq$ becomes $\nq$ and vice versa, while all other remaining cases are mapped to $\ndiff$ (as the negation of an implication does not necessarily entail non-implication, and no further information can be derived). The following definition offers an initial definition of term equivalence encompassing the name associated with it, potential specifications associated with them, as well as adjectives generally referred to as copulae:

\begin{definition}[Term Equivalence]
Given a \gls{kb} $\mathcal{K}$ of interest, we denote the discrete notion of term equivalence as follows:
\begin{equation}
a\equiv^\textsf{ter}b=\begin{cases}
\eq & a\equiv b \vee c=n=s\\
\nq & a\equiv \neg b \vee b\equiv \neg a\\
\eta(a'\equiv^\textsf{ter}b) & a\equiv\neg a'\\
\eta(a\equiv^\textsf{ter}b') & b\equiv\neg b'\\
\miss & a\equiv\textbf{None}\\
\ndiff & b\equiv\textbf{None}\vee \neg\texttt{isTerm}(a)\vee \neg\texttt{isTerm}(b)\\
\minst & n=\ndiff\wedge b.\textbf{all} \wedge a.\textbf{all} \wedge \tilde{nc}=\eq\wedge \\
&\qquad a.\textsf{name}\neq\textbf{None} \wedge b.\textsf{specification}\neq\textbf{None}\\
\minst & n=\ndiff\wedge b.\textbf{all} \wedge \neg 
a.\textbf{all} \wedge \tilde{nc}=\eq\wedge \\
&\qquad a.\textsf{name}\neq\textbf{None} \wedge b.\textsf{specification}\neq\textbf{None}\\
\tilde{n} & n=\ndiff\wedge b.\textbf{all} \wedge \neg 
a.\textbf{all} \wedge \tilde{n}=\mstar\\
\minst & n=\ndiff\wedge \neg b.\textbf{all} \wedge \neg 
a.\textbf{all} \wedge \tilde{n}=\eq\\
s & n=\eq \wedge s=c \wedge (b.\textbf{all}=a.\textbf{all}\vee b.\textbf{all})\\
\mig & n=s=\eq\wedge c=\miss\\
c & n=\eq\\ 
n & n=\mstar \wedge s=c=\eq \wedge \neg b.\textbf{all} \wedge a.\textbf{all}\\
\nq & n=\mstar \wedge s = \eq \wedge c=\nq\\
\minst & n=\mstar \wedge \tilde{nc} = \eq \wedge b.\textsf{specification}=\textbf{None}\wedge b.\textbf{all}\\
\nq & n=\nq \wedge f=c=\eq\\
\ndiff & \textrm{oth.}
\end{cases}
\end{equation}
where we use $c$ as a shorthand for the comparison between copulae occurring within flipped terms $b.\textsf{cop}\equiv^\textsf{var} a.\textsf{cop}$, $n$ as a shorthand for the name comparison within terms $a.\textsf{name}\equiv^\mathcal{K}b.\textsf{name}$, $s$ as a shorthand for the  specification comparison within terms $b.\textsf{specification}\equiv^\mathcal{K}a.\textsf{specification}$, and ${nc}$ as a shorthand of the comparison between name and specification $a.\textsf{name}\equiv^\textsf{var} b.\textsf{specification}$. Given any of the former symbols $x$, we denote by $\tilde{x}$ the flipped variant of the former where the first and second argument are swapped, while always referring to the same arguments (e.g., $\tilde{nc}=b.\textsf{name}\equiv^\textsf{var} a.\textsf{specification}$).

\end{definition}

\subsubsection{Multi-Valued Proposition Equivalence}
First, we can consider the equivalence of propositions by types, while ignoring their property arguments $p$ and $p'$: while this boils down to the sole argument's term equivalence for unary propositions (Eq. \ref{eq:un}), for binary predicates, by implicitly considering them as a functional dependency of the first argument over the second, we consider them according to this priority order (Eq. \ref{eq:bin}). In both cases, if the propositions differ in terms of the relationship name, we ignore the comparison with $\ndiff$.

\begin{equation}\label{eq:un}
r_p(s)\equiv^\textsf{un}r'_{p'}(s')=\begin{cases}
\ndiff & r\neq r'\\
s\equiv^\textsf{ter}s' & \textup{oth.}\\
\end{cases}
\end{equation}

\begin{equation}\label{eq:bin}
r_p(s,d)\equiv^\textsf{bin}r'_{p'}(s',d')=\begin{cases}
\ndiff & (r\neq r') \vee (s\equiv^\textsf{ter}s'=\ndiff) \vee (t\equiv^\textsf{ter}t'=\ndiff)\\
\ndiff & (s\equiv^\textsf{ter}s'=\nq) \vee (t\equiv^\textsf{ter}t'=\nq)\\
\nq & (s\equiv^\textsf{ter}s'=\nq) \wedge (t\equiv^\textsf{ter}t'\neq\ndiff)\\
\nq & (t\equiv^\textsf{ter}t'=\nq) \wedge (s\equiv^\textsf{ter}s'\neq\ndiff)\\
s\equiv^\textsf{ter}s' & t\equiv^\textsf{ter}t'=\eq \\
t\equiv^\textsf{ter}t' & s\equiv^\textsf{ter}s'=\eq \\
\varsigma(\{s\equiv^\textsf{ter}s',t\equiv^\textsf{ter}t'\}) & \textup{oth.}\\
\end{cases}
\end{equation}

Given a set of comparison outcomes $S$ referring to comparison outcomes of terms referring to the same key within a property, we define $\sigma$ as the function simplifying the comparison outcomes with the most specific multi-valued equivalence output summarising the evidence collected so far:
\begin{equation}\label{varsigmadef}
\varsigma(S)=\begin{cases}
\ndiff & S=\emptyset\\
\nq & \nq \in S\\
\eq & \eq \in S\\
\mig & \mig\in S \wedge  \miss\notin S \wedge \minst\notin S\wedge \impl\notin S\\
\miss & \mig\notin S \wedge  \miss\in S \wedge \minst\notin S\wedge \impl\notin S\\
\minst & \mig\notin S \wedge  \miss\notin S \wedge \minst\in S\wedge \impl\notin S\\
\impl & \mig\in S \vee  \miss\in S \vee \minst\in S\vee \impl\in S\\
\end{cases}
\end{equation}
We also define another function, $\varsigma'$, for further summarisation of these considerations across all key values within the same property; the purpose of this function is to provide the general notion of proposition similarity at the level of the properties, where the main differences rely on the order of application of the rules. If both properties have no properties, they are deemed equivalent on those premises:
\begin{equation}
\varsigma'(S)=\begin{cases}
\eq & S=\emptyset\\
\ndiff & \ndiff \in S\\
\nq & \nq \in S\\
\varsigma(S) & \eq\notin S \wedge (\mig\in S \vee  \miss\in S \vee \minst\in S\vee \impl\in S)\\
\eq & \eq\in S\\
\ndiff & \textup{oth.}
\end{cases}
\end{equation}

After this, we can define two functions, summarising the implication relationships within properties $p$ and $p'$ for the first and second propositions, respectively: $\kappa^r$ considers the verse of the implication from the first term towards the second (Eq. \ref{kr}), while $\kappa^i$ considers the inverse order (Eq. \ref{ki}):

\begin{equation}\label{kr}
\begin{split}
\kappa^r(p,p')=&[k\mapsto \varsigma(\{a\equiv^\textsf{ter}b|a\in p(k),b\in p'(b)\}) ]_{k\in\dom(p)\cap\dom(p')}\\
&\circ [k\mapsto\ndiff]_{k\in\dom(p)\wedge k\notin\dom(p')} \circ [k\mapsto\impl]_{k\in\dom(p')\wedge k\notin\dom(p)}
\end{split}\end{equation}\begin{equation}\label{ki}
\begin{split}
\kappa^i(p,p')=&[k\mapsto \varsigma(\{a\equiv^\textsf{ter}b|a\in p(k),b\in p'(b)\}) ]_{k\in\dom(p)\cap\dom(p')}\\
&\circ [k\mapsto\impl]_{k\in\dom(p)\wedge k\notin\dom(p')} \circ [k\mapsto\ndiff]_{k\in\dom(p')\wedge k\notin\dom(p)}
\end{split}
\end{equation}

\[,\kappa'\gets\varsigma'(\cod(\kappa^i(p,p')))\]

Given these ancillary definitions, we can now define the concept of discrete equivalence between propositions occurring within the formula:

\begin{definition}\label{def:propequiv}
Given a \gls{kb} $\mathcal{K}$ and two propositions $a$ and $b$, where the first is associated with a property $p$ and the second with a property $p'$, we can define the following discrete equivalence function:

\begin{equation}
a\equiv^\textsf{prop}b=\begin{cases}
\eq & a\equiv b\\
\nq & a\equiv \neg b \vee b\equiv \neg a\\
\eta(a'\equiv^\textsf{prop}b) & a\equiv \neg a'\\
\eta(a\equiv^\textsf{prop}b') & b\equiv \neg b'\\
\ndiff & a\equiv\textbf{None}\\
\mstar & b\equiv\textbf{None}\\
a'\equiv^\textsf{prop}b' & a\equiv \neg a'\wedge b\equiv \neg b'\wedge (a'\equiv^\textsf{prop}b')\neq \mstar\\
\ndiff & a\equiv \neg a'\wedge b\equiv \neg b'\wedge (a'\equiv^\textsf{prop}b')= \mstar\\
\ndiff & \texttt{isBinary}(a)\neq\texttt{isBinary}(b)\\
\ndiff & a=r(t_1)\wedge b=r'(t_2) \wedge t_1\neq t_2\\
\ndiff & \gamma=\ndiff\\
\kappa & \gamma=\eq \wedge f = \eq\wedge \kappa=\mstar\wedge c\neq \eq\\
\ndiff & \gamma=\eq \wedge f = \eq\wedge \kappa=\mstar\wedge c=\eq\wedge \ndiff\in\cod(\kappa^r(p,p'))\\
\ndiff & \gamma=\eq \wedge f = \eq\wedge \kappa=\mstar\wedge c=\eq\wedge \mig\in\cod(\kappa^r(p,p'))\\
\kappa & \gamma=\eq \wedge f = \eq\wedge (\kappa=\minst\vee \minst\in\cod(\kappa^r(p,p')))\\
\ndiff & \gamma=\eq \wedge f = \eq\wedge \kappa=\mstar\wedge \textup{oth.}\\
\minst & \gamma=\eq \wedge f = \eq\wedge \kappa'=\mig\\
\kappa & \gamma=\eq \wedge f = \eq\wedge \kappa'\neq\mig\wedge \kappa\neq\mstar\\
\ndiff & \gamma=\mstar \wedge \kappa\neq \eq \wedge \ndiff\in \kappa^r(p,p')\\
\minst & \gamma=\mstar \wedge \kappa\neq \eq \wedge \kappa' = \mig \\
\kappa & \gamma=\mstar \wedge \kappa\neq \eq \wedge \kappa' \neq \mig \wedge\ndiff\notin \kappa^r(p,p') \\
\ndiff & \gamma=\neq \wedge (\kappa=\ndiff\vee \kappa=\neq)\\
\gamma & \textup{oth.}
\end{cases}
\end{equation}
where $\gamma$ is shorthand for the comparison outcome between either binary ($a\equiv^\textsf{bin}b$) or unary ($a\equiv^\textsf{un}b$) propositions, $f$ is shorthand for the comparison of the first argument considered as a term ($s\equiv^\textsf{ter}s'$),  $c$ is a comparison of the first arguments' copulae if occurring ($s.\textsf{cop}\equiv^\textsf{ter}s.\textsf{cop}$), and $\kappa$ summarises the comparison between the first proposition's properties with the one of the second ($\varsigma'(\cod(\kappa^r(p,p')))$), while $\kappa'$ flips this comparison ($\kappa\gets\varsigma'(\cod(\kappa^r(p',p)))$).
\end{definition}

\section{Proofs}\label{proofs}

\begin{proof}[Proof (for Lemma \ref{def:cosine})]
	Given a binary relationship $x\Re y \Leftrightarrow (x,y)\in \Re$, 
	\textbf{symmetry} is a property of $\Re$; that is, if $x$ and $y$ are related in either direction, then $x$ and $y$ are equivalent. Formally,
	due to the commutative property of the dot product, $\mathcal{S}_c(A,B) = \mathcal{S}_c(B,A)$ (\url{https://mathinsight.org/dot_product}, Accessed on 07 July 2025).
\end{proof}

\begin{proof}[Proof (for Lemma \ref{lemma:symm})]
	If $B$ implies $A$ under a symmetric metric $\mathcal{S}$, then $A$ also implies $B$. This is because the similarity value remains the same regardless of the order of arguments.	Formally,
	\begin{align*}
		\varphi_\tau(\beta,\alpha) &\iff \mathcal{S}(\tau(\beta),\tau(\alpha))>\theta  \quad \text{(by Definition \ref{phitau}, $\overline{\theta}$)}\\
		&\iff \mathcal{S}(\tau(\alpha),\tau(\beta))>\theta  \quad \text{(by Lemma \ref{lemma:symm}, $symm$})\\
		&\iff \varphi_\tau(\alpha,\beta)  \quad \text{(by Definition \ref{phitau})}
	\end{align*}
\end{proof}

\begin{proof}[Proof (for Lemma \ref{lem:eqsameconf})]
	\begin{itemize}
		\item [$\Rightarrow$]
		Left to right: If two equations are the same, they will have the same set of the possible worlds. Given this, their intersection will be always equivalent to one of the two sets, from which it can be derived that the support of either of the two formulas is always true. 
		
		\[W(A)=W(B)\therefore W(A)\cap W(B)=W(A)\Rightarrow \frac{|W(A) \cap W(B)|}{|W(A)|}=1\]
		\item [$\Leftarrow$]
		Right to left: When the confidence is 1, then by definition, both the numerator and denominator are of the same size. Therefore, $|W(A)\cap W(B)|=|W(A)|$ and $|W(B)\cap W(A)|=|W(B)|$. By the commutativity of the intersection, we then derive that $|A|\ |B|$ are of the same size and represent the same set. Thus, it holds that $A\equiv B$.
	\end{itemize}
\end{proof}

\begin{proof}[Proof (for Corollary \ref{coroll:one})]
	The following notation implies that, if there exists $x$ that belongs to $W(A)$ and not to $W(B)$, then their intersection size will be less than the size of $A$ and, if $W(A)\cap W(B)$ has less items then $W(A)$, then it follows that $B$ contains elements not occurring in $A$:
	\[\exists x. x \in W(A) \cap x \notin W(B) \iff |W(A) \cap W(B)|<|W(A)|\iff \texttt{confidence}(A,B)<1\]
	
	Then, we want to characterise the logical formula supporting a set of elements belonging to $A$ but not to $B$. We can derive this from the bag semantics of our logical operators, inferring that the below condition holds when $A$ does not necessarily imply $B$:
	\begin{align*}
		W(\neg(A\Rightarrow B)) &= S\\
		&=W(A\cap \neg B)\\
		&=W(A)\cap \complement W(B)
	\end{align*}
	$\therefore$ when $A$ holds $B$ does not necessarily hold.
\end{proof}

\section{Sentence Scalability Rewriting}\label{sub:benchmark}
\begin{figure}[H]
\centering
    \includegraphics[width=.7\linewidth]{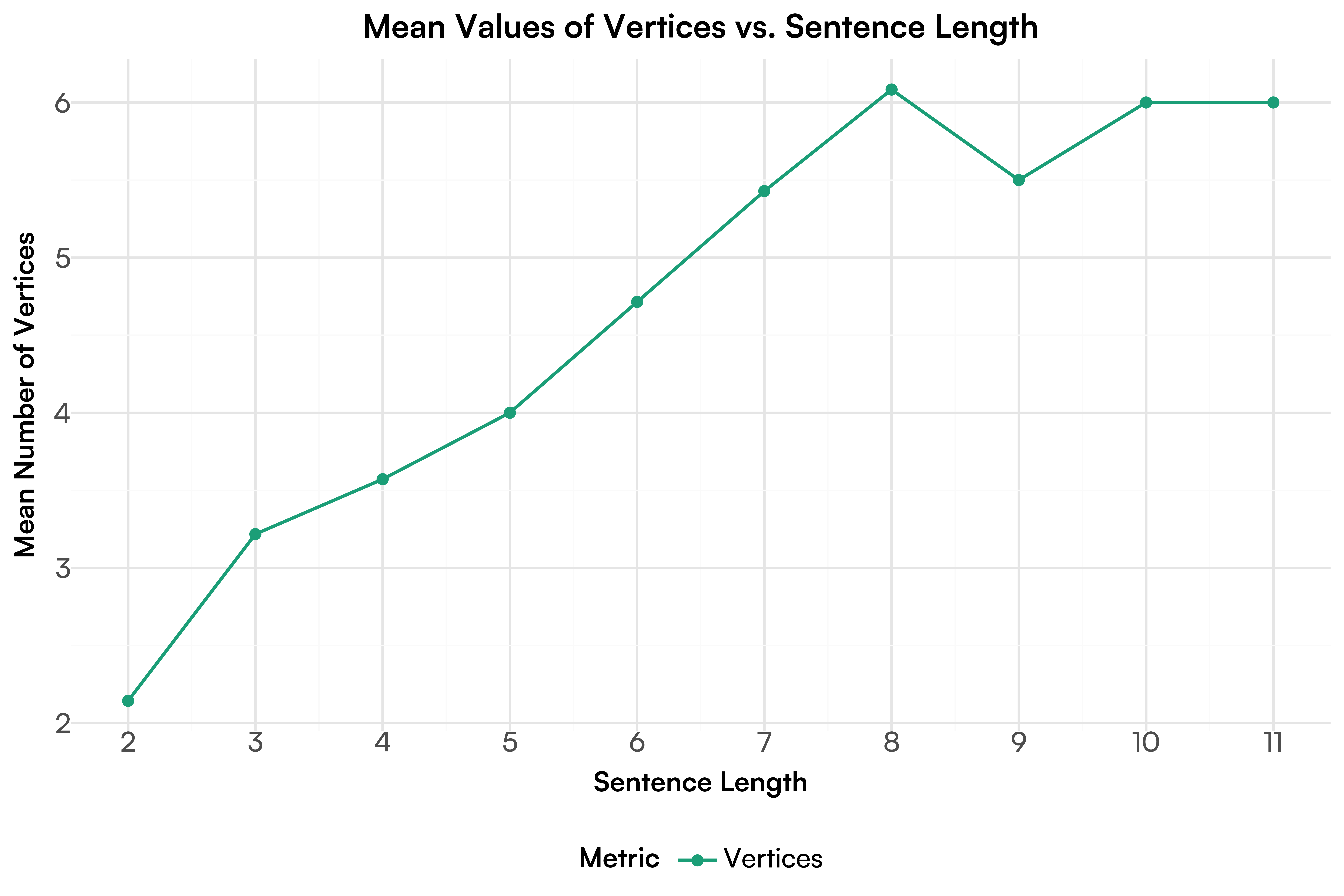}
    \caption{{The average (mean) number of vertices for varying sentence lengths (based on the number of nodes in each graph).}}
    \label{fig:vertices}
\end{figure}

We recorded the individual running times of each phase in the \gls{lassi} pipeline for each sentence within the 200 sentences taken from ConceptNet \cite{DBLP:conf/aaai/SpeerCH17} containing either dependant clauses or noisy representations: this is either due to typos occurring in the sources, or by faults of the \gls{kb} extracting algorithm \cite{bergami2019frameworksupportingimprecisequeries}. We also recorded the sentence length, as characterised by the number of nodes in each graph representation of the sentence and its character length. \figurename~\ref{fig:vertices} plots the average number of vertices as sentence length increases, which almost presents a linear correlation. As the \textit{ad hoc} phase of our pipeline operates on the graph representation of the sentence, rather than on its character length, while the GSM generation through StanfordNLP dependency parsing operates on the character length, we made attempts to normalise the forthcoming results.

\begin{figure}[H]
    \includegraphics[width=1\linewidth]{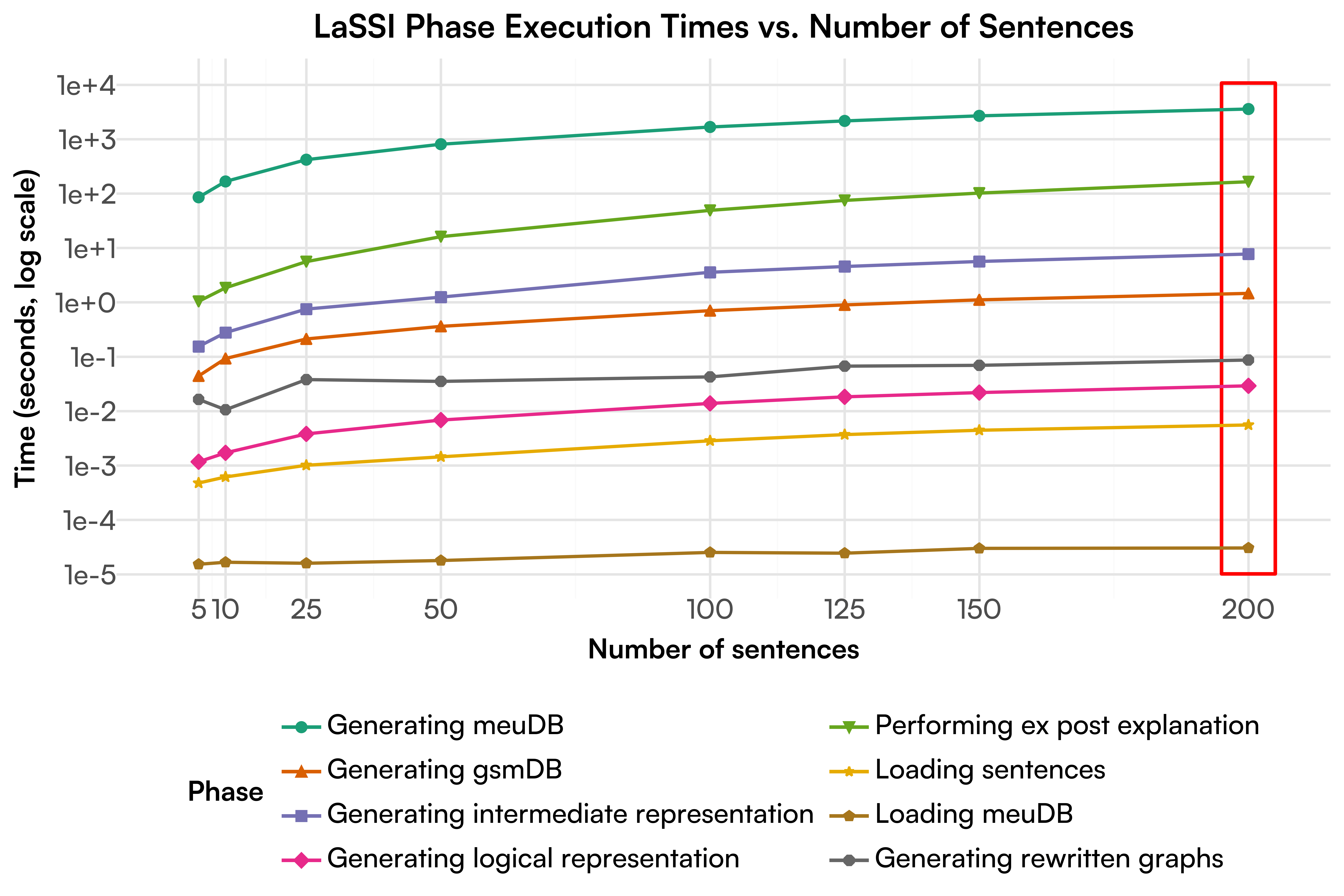}
    \caption{Plot of scalability tests for the logical representation in log-norm scale.}
    \label{fig:metrics}
\end{figure}

\figurename~\ref{fig:metrics} provides the scalability results over the number of the sentences. We now provide a discussion for each of the pipeline's phases.

\textit{Loading sentences} refers to injecting a .yaml file containing all sentences to be transformed into Python strings.

\textit{Generating \gls{meudb}} (Section \ref{sec:san}) takes the longest time to run, as it relies on Stanza to gather POS tagging information for every word in the full text. We also perform fuzzy string matching on the words to obtain every possible multi-word interpretation of each entity, the complexity of which increases dramatically as the number of sentences increases. However, once generated, it is saved to a JSON file, which is then loaded into memory for subsequent executions, represented by the \textit{loading \gls{meudb}} phase.

\textit{Generating the gsmDB} (Section \ref{gsmrewr}) uses the learning-based pipeline from StanfordNLP, which still achieves a rapid execution time. When generating our rewritten graphs, it performs even better, as this process is not dependent on any learning-based pipeline.

Generating the \textit{intermediate representation} (Section \ref{sec:adhoc} except Section \ref{sec:logrewr}) ends up being slower than generating the \gls{gsm} graphs, which is most likely because the pipeline is throttled when lemmatising verbs at multiple stages in the pipeline. This is due to the pipeline attempting to lemmatise each single word alone occurring within the text, while the GSM generation via StanfordNLP considers entire sentences as a whole. We boosted this process using the LRU caching mechanism \cite{DBLP:conf/sigmod/ONeilOW93,DBLP:conf/vldb/JohnsonS94} from \texttt{functools}, allowing words that have already been lemmatised to be reused from memory. However, as the number of sentences increases, so does the number of new words. As the other rewriting steps are streamlined through efficient algorithms, thus providing efficient implementations for rule matching and rewriting mechanisms, future works will address better strategies for enhancing the current rewriting step.

Given the plots, all processing steps discussed were found to exhibit a linear running time complexity {trend} modulo time fluctuations over the sentence length. The \textit{latency} for LaSSI refers to the time it takes to produce an output from providing an input, this latency is not currently known for large scale data, which is still a problem within the current implementation, but reasoning and knowledge expansion can be expressed on relational databases, which should help with scalability \cite{10.14778/3007263.3007284}. Future work will address this by migrating the expansion over \gls{rdbms}, while this phase considered the possibility of doing so. Previous research postulates how paraconsistent reasoning can be carried out through Relational Database queries \cite{bergami2019frameworksupportingimprecisequeries}.

\begin{figure}[H]
	\centering
    \includegraphics[width=.7\linewidth]{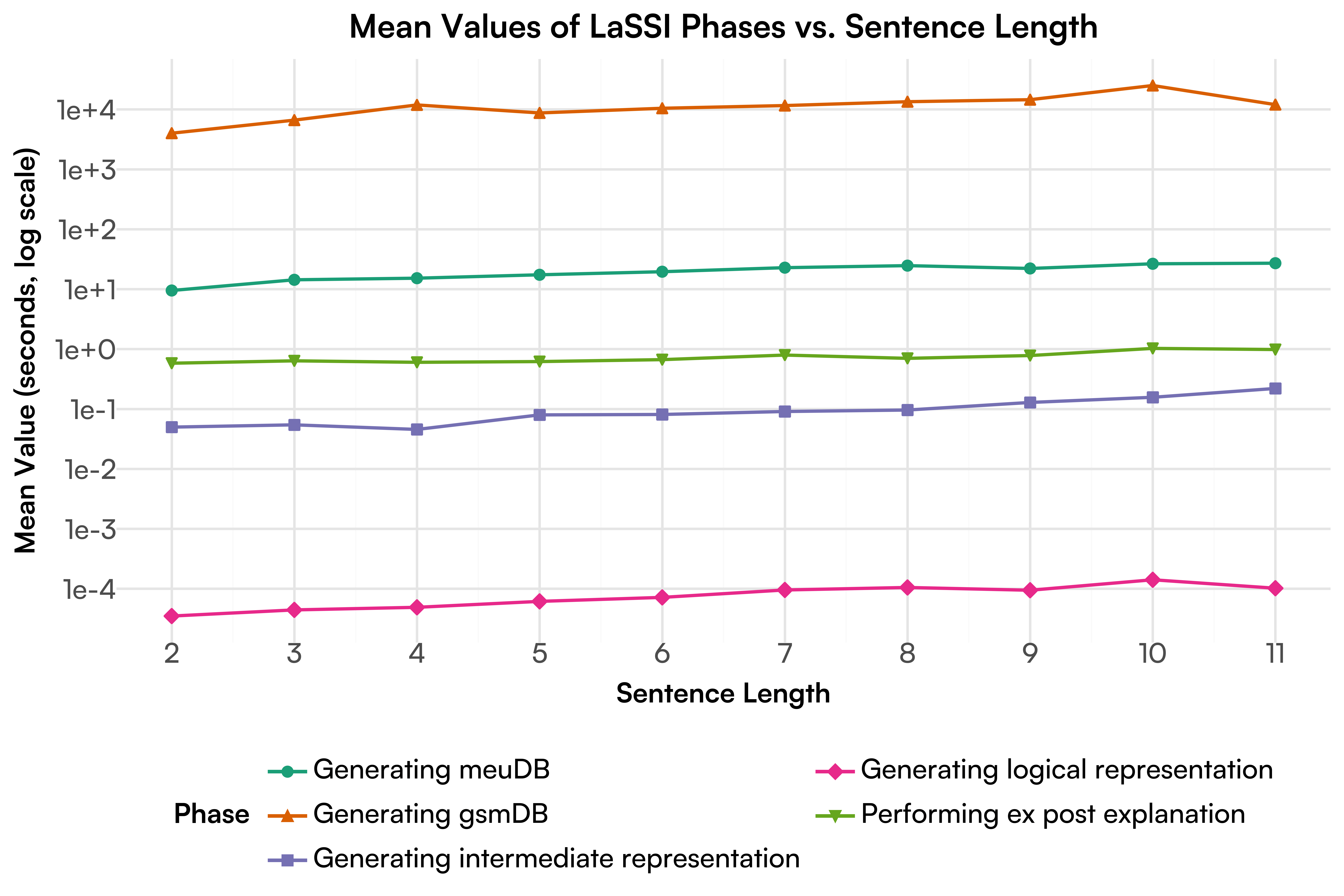}
    \caption{{Performance of each stage of the \gls{lassi} pipeline for the logical representation (in log-norm scale) for sentences of different lengths (based on the number of nodes in each graph).}}
    \label{fig:sentence_benchmarks}
\end{figure}

We now intuitively discuss the optimal expected time complexity, given the nature of the problem at stake. As all the algorithms detailed in this paper mainly entail the rewriting of original sentences by reading them from the beginning to the end independent of their representation, the least time cost is comparable to a linear scan of the data representation for each sentence.   
Using this sentence length, we grouped the phases in order to present a trend of how the running time changes as the sentence length increases, as presented in \figurename~\ref{fig:sentence_benchmarks}. Each phase shows a linear trend, except for the generation of the intermediate representation and \textit{ex post}  phases, which present exponential trends. This demonstrates the need to further improve the pipeline if we want to deal with longer sentences, especially regarding the generation of the intermediate representation. With respect to the exponential time of the \textit{ex post}  explanation, this is in line with the expected exponential time for propositional logic; in fact, at this stage, existential variables are mainly treated as variables to be matched, rather than being resolved by querying the \gls{kb}. Consequently, the only way to further improve upon these results is to implement novel heuristic approaches using the semantic-based representation of the data.

As highlighted in the experiments, the \gls{meudb} generation step was detrimental to the running time of our pipeline, although once it is generated, the running times significantly decrease. Therefore, further investigations will be performed to reduce the time complexity through a better algorithm, thus ensuring that the \gls{lassi} pipeline is as efficient as possible in the future.

\section{Clustering Results}\label{app:clustering}

Through clustering, we want to test ifpre-trained language models are well suited for expressing the notion of logical equivalence. Given that the training-based approach cannot ensure a perfect notion of equivalence (sentence transformers are specifically designed to represent this), returning a perfect score of $1$ for extremely similar sentences, we can relax the argument by requiring that there exists any suitable clustering algorithm allowing us to group certain sentences based on distance values, providing the closest match to the group of expected equivalent sentences selected. According to Definition \ref{phitau}, we can then derive a minimum threshold value $\theta$, with the minimum similarity score between the elements belonging to the same cluster derived from the chosen methodology. Given our purposes, we require to analyse $k=3$ clusters.

We test our hypothesis over two clustering algorithms supporting distance metrics as input rather than a collection of points in the $d$-dimensional Euclidean space, \gls{ahc}, and $k$-Medoids. The graphs visualised in Figures \ref{fig:clustering_dend_ab}, \ref{fig:clustering_dend_cm}, and \ref{fig:clustering_dend_newc} in Supplement \ref{sup:support_dendrograms} are \textit{dendrograms}, which allow us to explore relationships between clusters at different levels of granularity as returned by \gls{ahc}. For \gls{ahc}, we considered the complete link criterion for which the distance between two clusters is defined as the distance between two points, each for one single cluster, maximising the overall distance \cite{Zaki_Meira_Jr_2020}. 
We consider the $k$-Medoids algorithm with an initialisation procedure that is a variant of that of k-means++ \cite{DBLP:conf/soda/ArthurV07}. It considers the initial $k$-Medoids clusters to avoid poor initialisations resulting in the selection of weak initial random points by sampling these initial points over ``\textit{an empirical probability distribution of the points' contribution to the overall inertia}'' (\url{https://scikit-learn.org/stable/modules/generated/sklearn.cluster.KMeans.html}, Accessed on 29 March 2025). 
No further hyperparameter tuning is required under these assumptions, as these annihilate the need for carrying out multiple tests with multiple different random seeds.  \supplementaryname~\ref{supp:algosclust} provides additional details for such clustering algorithms.

{To ensure the correctness of the alignment score used to match the nodes in the graph semantically, we exploited this as a clustering measure and compared the score to others available in the current literature. In this case, rather than considering the alignment between a set of edges within the graph, we considered the alignment between sets of clusters. } The correctness of the returned clustering match was computed using a well-defined set-based alignment metric from \cite{logicalLearning}, which is a refinement of Eq. \ref{commonEq} using $\epsilon(s,t)=\mathbf{1}_{s=t}$, where $\mathbf{1}_P$ is the indicator function returning $1$ if and only if condition $P$ holds and $0$ otherwise \cite{Kleene1952-KLEITM}. We pair this metric with the classical Silhouette \cite{ROUSSEEUW198753}, Purity \cite{Manning_Raghavan}, and \gls{ari} \cite{DBLP:journals/jmlr/NguyenEB10}, as discussed in \supplementaryname~\ref{supp:clusteringmtric}.

\subsection{\ref{rqn2a} (Propositional Calculus)}
As for this dataset, we did not expect any clusters to form; the clustering experiments mainly aimed to determine whether any sensible clusters might be incorrectly derived.
When analysing \tablename~\ref{table:clustering_result_ab}, almost all approaches perfectly align with the expected clusters when using \gls{ahc}. However, this is determined by the nature of the algorithm as it starts with this exact configuration, where each element is its own cluster. It inherently matches the ground truth at the beginning; therefore, the algorithm achieves perfect alignment before any merging occurs. The silhouette score can only be calculated if the number of labels (sentences) is $2 \leq \text{labels} \leq \text{clusters} - 1$ (\url{https://scikit-learn.org/stable/modules/generated/sklearn.metrics.silhouette_score.html}, Accessed on 25 April 2025). In our case, the length of the clusters (8) is equal to the number of distinct labels (8); therefore, we cannot return the silhouette score. While using k-Medoids clustering, \gls{sg} and DeBERTaV2+AMR-LDA  (T5) yielded incorrect clusters, resulting in the grouping of sentences 2 and 3 and 1 and 4, respectively. This leads to the incorrect derivation of silhouette scores. For purity and \gls{ari}, we see perfect scores again when using \gls{ahc}; only when using $k$-Medoids do we see a lower value for \glspl{sg} and T5 (DeBERTaV2+AMR-LDA). As our \glspl{sg} do not encompass logical connectives entirely, the clustering scores are expected to be lower.

\begin{table}[!h]
\caption{Clustering scores for \ref{rqn2a} sentences{, with the best value for each row highlighted in bold blue text and the worst values highlighted in red}.}
\label{table:clustering_result_ab}\centering
\begin{tblr}{
  width = 0.5\textwidth,
  column{3-11} = {colsep=2pt,font=\footnotesize},
  row{1} = {font=\normalsize},
  cells = {c},
  row{1} = {Mercury},
  cell{2}{1} = {r=2}{Mercury},
  cell{2}{2} = {Gallery},
  cell{3}{2} = {Gallery},
  cell{4}{1} = {r=2}{Mercury},
  cell{4}{2} = {Gallery},
  cell{5}{2} = {Gallery},
  cell{6}{1} = {r=2}{Mercury},
  cell{6}{2} = {Gallery},
  cell{7}{2} = {Gallery},
  cell{8}{1} = {r=2}{Mercury},
  cell{8}{2} = {Gallery},
  cell{9}{2} = {Gallery},
  vlines,
  hline{1-2,4,6,8,10} = {-}{},
  hline{3,5,7,9} = {2-11}{},
}
\textbf{Metric} & \textbf{Clustering} & \textbf{\glspl{sg}} & \textbf{\glspl{lg}} & \textbf{Logical} & \textbf{T1}   & \textbf{T2}   & \textbf{T3}   & \textbf{T4}   & \textbf{T5}   & \textbf{T6}   \\
Alignment       & \textbf{\gls{ahc}}  & \textbf{\textcolor{blue}{1.00}}       & \textbf{\textcolor{blue}{1.00}}       & \textbf{\textcolor{blue}{1.00}}    & \textbf{\textcolor{blue}{1.00}} & \textbf{\textcolor{blue}{1.00}} & \textbf{\textcolor{blue}{1.00}} & \textbf{\textcolor{blue}{1.00}} & \textbf{\textcolor{blue}{1.00}} & \textbf{\textcolor{blue}{1.00}} \\
                & \textbf{$k$-Medoids}  & \textcolor{red}{0.40}                & \textbf{\textcolor{blue}{1.00}}       & \textbf{\textcolor{blue}{1.00}}    & \textbf{\textcolor{blue}{1.00}} & \textbf{\textcolor{blue}{1.00}} & \textbf{\textcolor{blue}{1.00}} & \textbf{\textcolor{blue}{1.00}} & 0.81          & \textbf{\textcolor{blue}{1.00}} \\
Silhouette      & \textbf{\gls{ahc}}  & -                   & -                   & -                & -             & -             & -             & -             & -             & -             \\
                & \textbf{$k$-Medoids}  & 0.26                & -                   & -                & -             & -             & -             & -             & \textcolor{red}{0.21}          & -             \\
Purity          & \textbf{\gls{ahc}}  & \textbf{\textcolor{blue}{1.00}}       & \textbf{\textcolor{blue}{1.00}}       & \textbf{\textcolor{blue}{1.00}}    & \textbf{\textcolor{blue}{1.00}} & \textbf{\textcolor{blue}{1.00}} & \textbf{\textcolor{blue}{1.00}} & \textbf{\textcolor{blue}{1.00}} & \textbf{\textcolor{blue}{1.00}} & \textbf{\textcolor{blue}{1.00}} \\
                & \textbf{$k$-Medoids}  & \textcolor{red}{0.40}                & \textbf{\textcolor{blue}{1.00}}       & \textbf{\textcolor{blue}{1.00}}    & \textbf{\textcolor{blue}{1.00}} & \textbf{\textcolor{blue}{1.00}} & \textbf{\textcolor{blue}{1.00}} & \textbf{\textcolor{blue}{1.00}} & 0.81          & \textbf{\textcolor{blue}{1.00}} \\
\gls{ari}       & \textbf{\gls{ahc}}  & \textbf{\textcolor{blue}{1.00}}       & \textbf{\textcolor{blue}{1.00}}       & \textbf{\textcolor{blue}{1.00}}    & \textbf{\textcolor{blue}{1.00}} & \textbf{\textcolor{blue}{1.00}} & \textbf{\textcolor{blue}{1.00}} & \textbf{\textcolor{blue}{1.00}} & \textbf{\textcolor{blue}{1.00}} & \textbf{\textcolor{blue}{1.00}}          \\
                & \textbf{$k$-Medoids}  & \textcolor{red}{0.40}                & \textbf{\textcolor{blue}{1.00}}       & \textbf{\textcolor{blue}{1.00}}    & \textbf{\textcolor{blue}{1.00}} & \textbf{\textcolor{blue}{1.00}} & \textbf{\textcolor{blue}{1.00}} & \textbf{\textcolor{blue}{1.00}} & 0.81          & \textbf{\textcolor{blue}{1.00}} 

\end{tblr}

\end{table}

\subsection{\ref{rqn2b} (Active vs. Passive)}
Analysing \tablename~\ref{table:clustering_results_cm}, we can see that all stages of the LaSSI pipeline achieved perfect alignment with \gls{ahc} and $k$-Medoids, effectively capturing the logical equivalence between sentences 0 and 1, and 2 and 3. This was mainly ascribable to the \gls{ggg} rewriting phase, which captures the notion of active vs. passive sentence, while rewriting the sentences in a uniform graph representation. Low silhouette scores for \glspl{sg} and \glspl{lg} indicate high intra-cluster heterogeneity with respect to similarity values. Furthermore, high levels of alignment, purity, and ARI demonstrate their ability to match the expected clusters. In agreement with the visual analysis of the dendrograms above, all pre-trained language models showed lower alignment with \gls{ahc} and $k$-Medoids, thus including DeBERTaV2+AMR-LDA (T5). While they captured some semantic similarity, they struggled to fully grasp the logical equivalence between the sentences in the same way that LaSSI does, even with some simplistic rewriting using our pipeline. None of these transformers produced zones with 0 similarity, as they determined all the sentences to be related to each other, also misinterpreting that ``\textit{The cat eats the mouse}'' and ``\textit{The mouse eats the cat}'' as similar. The pipelines for these given transformers ignore stop words, which may also impact the resulting scores. We also recognise that the similarity is heavily dominated by the entities occurring, while the sentence structure is almost not reflected: transformers  only consider one sentence to be a collection of entities without taking its structure into account, whereby changing the order will yield similar results and, therefore,  sentence structure cannot be derived. Interpreting similarity with compatibility, graph-based measures entirely exclude the possibility of this happening, while logic-based approaches are agnostic.

\begin{table}[h]
        \caption{Clustering scores for \ref{rqn2b} sentences, with the best value for each row highlighted in bold, blue text and the worst values highlighted in red. }
    \label{table:clustering_results_cm}\centering
    \begin{tblr}{
      width = 0.5\textwidth,
      column{3-11} = {colsep=2pt,font=\footnotesize},
      row{1} = {font=\normalsize},
      cells = {c},
      row{1} = {Mercury},
	  cell{2}{1} = {r=2}{Mercury},
	  cell{2}{2} = {Gallery},
	  cell{3}{2} = {Gallery},
	  cell{4}{1} = {r=2}{Mercury},
	  cell{4}{2} = {Gallery},
	  cell{5}{2} = {Gallery},
	  cell{6}{1} = {r=2}{Mercury},
	  cell{6}{2} = {Gallery},
	  cell{7}{2} = {Gallery},
	  cell{8}{1} = {r=2}{Mercury},
	  cell{8}{2} = {Gallery},
	  cell{9}{2} = {Gallery},
      vlines,
      hline{1-2,4,6,8,10} = {-}{},
      hline{3,5,7,9} = {2-11}{},
    }
    \textbf{Metric} & \textbf{Clustering} & \textbf{\glspl{sg}} & \textbf{\glspl{lg}} & \textbf{Logical} & \textbf{T1} & \textbf{T2} & \textbf{T3} & \textbf{T4} & \textbf{T5} & \textbf{T6} \\
    Alignment       & \textbf{\gls{ahc}}  & \textbf{\textcolor{blue}{1.00}}       & \textbf{\textcolor{blue}{1.00}}       & \textbf{\textcolor{blue}{1.00}}    & \textcolor{red}{0.38}        & \textcolor{red}{0.38}        & \textcolor{red}{0.38}        & \textcolor{red}{0.38}        & 0.50        & 0.67        \\
                    & \textbf{$k$-Medoids}  & \textbf{\textcolor{blue}{1.00}}       & \textbf{\textcolor{blue}{1.00}}       & \textbf{\textcolor{blue}{1.00}}    & 0.38        & 0.38        & 0.38        & 0.38        & \textcolor{red}{0.29}        & 0.67        \\
    Silhouette      & \textbf{\gls{ahc}}  & 0.17                & \textcolor{red}{0.15}                & \textcolor{blue}{\textbf{0.67}}    & 0.55        & 0.51        & 0.63        & 0.62        & 0.53        & 0.23        \\
                    & \textbf{$k$-Medoids}  & 0.17                & \textcolor{red}{0.15}                & 0.67    & 0.55        & 0.51        & 0.63        & 0.62        & \textbf{\textcolor{blue}{1.00}}        & 0.23        \\
    Purity          & \textbf{\gls{ahc}}  & \textbf{\textcolor{blue}{1.00}}       & \textbf{\textcolor{blue}{1.00}}       & \textbf{\textcolor{blue}{1.00}}    & \textcolor{red}{0.67}        & \textcolor{red}{0.67}        & 0.67        & 0.67        & 0.67        & 0.67        \\
                    & \textbf{$k$-Medoids}  & \textbf{\textcolor{blue}{1.00}}       & \textbf{\textcolor{blue}{1.00}}       & \textbf{\textcolor{blue}{1.00}}    & 0.38        & 0.38        & 0.38        & 0.38        & \textcolor{red}{0.29}        & 0.67        \\
    \gls{ari}       & \textbf{\gls{ahc}}  & \textbf{\textcolor{blue}{1.00}}       & \textbf{\textcolor{blue}{1.00}}       & \textbf{\textcolor{blue}{1.00}}    & \textcolor{red}{-0.15}       & \textcolor{red}{-0.15}       & \textcolor{red}{-0.15}       & \textcolor{red}{-0.15}       & \textcolor{red}{-0.15}       & \textcolor{red}{-0.15}       \\
                    & \textbf{$k$-Medoids}  & \textbf{\textcolor{blue}{1.00}}       & \textbf{\textcolor{blue}{1.00}}       & \textbf{\textcolor{blue}{1.00}}    & \textcolor{red}{0.38}        & \textcolor{red}{0.38}        & \textcolor{red}{0.38}        & \textcolor{red}{0.38}        & 0.67        & 0.67        
    \end{tblr}

\end{table}

\subsection{\ref{rqn2c} (Spatiotemporal)}
From \tablename~\ref{table:clustering_results_newc},  our \glspl{lg} and logical approaches outperformed the transformers, with our final logical approach achieving 100\% alignment against the expected clusters. Differently from the previous experimental set-up (\tablename~\ref{table:clustering_results_cm}), the sentence transformer approaches exhibited generally good behaviour in terms of their purity and ARI score, while DeBERTaV2+AMR-LDA (T5) provided a consistently worse behaviour than the former. This is likely ascribable to the simplistic logic being used to train the system, making it unable to fully capture the semantic complexity of our proposed logic for describing the semantic behaviour through a \gls{kb} back-up (Appendix \ref{mvte}), as well as not adequately rewriting the sentences to capture the equivalent sentences with different phrasing.


\begin{table}[!h]
\caption{Clustering scores for \ref{rqn2c} sentences{, with the best value for each row highlighted in bold, blue text and the worst values highlighted in red}.}
\label{table:clustering_results_newc}\centering
\begin{tblr}{
  width = 0.5\textwidth,
  column{3-11} = {colsep=2pt,font=\footnotesize},
  row{1} = {font=\normalsize},
  cells = {c},
  row{1} = {Mercury},
  cell{2}{1} = {r=2}{Mercury},
  cell{2}{2} = {Gallery},
  cell{3}{2} = {Gallery},
  cell{4}{1} = {r=2}{Mercury},
  cell{4}{2} = {Gallery},
  cell{5}{2} = {Gallery},
  cell{6}{1} = {r=2}{Mercury},
  cell{6}{2} = {Gallery},
  cell{7}{2} = {Gallery},
  cell{8}{1} = {r=2}{Mercury},
  cell{8}{2} = {Gallery},
  cell{9}{2} = {Gallery},
  vlines,
  hline{1-2,4,6,8,10} = {-}{},
  hline{3,5,7,9} = {2-11}{},
}
\textbf{Metric} & \textbf{Clustering} & \textbf{\glspl{sg}} & \textbf{\glspl{lg}} & \textbf{Logical} & \textbf{T1} & \textbf{T2} & \textbf{T3} & \textbf{T4} & \textbf{T5} & \textbf{T6} \\
Alignment       & \textbf{\gls{ahc}}  & 0.74                & 0.91                & \textbf{\textcolor{blue}{1.00}}    & 0.80        & 0.80        & 0.80        & 0.80        & \textcolor{red}{0.61}        & 0.80        \\
                & \textbf{$k$-Medoids}  & 0.78                & 0.74                & \textbf{\textcolor{blue}{1.00}}    & 0.80        & 0.80        & 0.63        & 0.62        & \textcolor{red}{0.57}        & 0.80        \\
Silhouette      & \textbf{\gls{ahc}}  & \textcolor{red}{0.21}                & 0.18                & \textbf{\textcolor{blue}{0.46}}             & 0.24        & 0.21        & 0.21        & 0.24        & 0.22        & 0.32        \\
                & \textbf{$k$-Medoids}  & 0.16                & 0.24                & \textbf{\textcolor{blue}{0.46}}    & 0.21        & 0.21        & 0.18        & 0.16        & \textcolor{red}{0.09}        & 0.32        \\
Purity          & \textbf{\gls{ahc}}  & 0.85                & 0.92                & \textbf{\textcolor{blue}{1.00}}    & 0.92        & 0.92        & 0.92        & 0.92        & \textcolor{red}{0.69}        & 0.92        \\
                & \textbf{$k$-Medoids}  & 0.78                & 0.74                & \textbf{\textcolor{blue}{1.00}}    & 0.80        & 0.80        & 0.63        & 0.62        & \textcolor{red}{0.57}        & 0.80        \\
\gls{ari}       & \textbf{\gls{ahc}}  & 0.32                & 0.71                & \textbf{\textcolor{blue}{1.00}}    & 0.71        & 0.71        & 0.71        & 0.71        & \textcolor{red}{-0.08}       & 0.71        \\
                & \textbf{$k$-Medoids}  & 0.78                & 0.74                & \textbf{\textcolor{blue}{1.00}}    & 0.80        & 0.80        & 0.63        & 0.62        & \textcolor{red}{0.57}        & 0.80        
\end{tblr}

\end{table}

%
%


\reftitle{References}


\bibliography{mdpi25}

\PublishersNote{}

\pagebreak\pagebreak
\supplement\label{supplstarts}

\doparttoc 
\faketableofcontents 
\part{} 
\parttoc 

	\section{Natural Language Processing (NLP)}\label{nlpsec}

\textit{\gls*{pos} tagging} algorithms work as follows: each word in a text (corpus) is marked up as belonging
to a specific part of speech based on the term's definition and context \cite{Sun14}. Words can be then categorised as nouns, verbs, adjectives, or adverbs. In \textit{Italian linguistics} (Supplement \ref{sec:itling}), this phase is referred to as the \textit{grammatical analysis} of a sentence structure and is one the most fine-grained analyses. As an example for \gls*{pos} tags, we can retrieve these initial annotations for the sentence ``\textit{Alice plays football}'' from Stanford CoreNLP \cite{manning-etal-2014-stanford}, identifying ``\textit{Alice}'' as a proper noun (NNP), ``\textit{plays}'' as a verb (VBZ---present tense, third-person singular), and ``\textit{football}'' as a noun (NN) and thus determining the subject--verb--object relationship between these words. 

\textit{Dependency parsing} \cite{jm3} refers to the extraction of language-independent grammatical functions expressed through a minimal set of binary relationships connecting \gls*{pos} components within the text, thus allowing a semistructured, graph-based representation of the text. These dependencies are beneficial in inferring how one word might relate to another. We can also extract these \glspl*{ud} through Stanford CoreNLP, whereby we obtain annotations for each word in the sentence, giving us relationships and types. For example, a \texttt{conj} \cite{conj_ud} dependency represents a \textit{conjunct}, which is a relation between two elements connected by a \texttt{cc} \cite{cc_ud} (a \textit{coordination} determining what type of group these elements should be in). 

As shown in \figurename~\ref{fig:classic_sentence} and \ref{fig:negation_sentences}, relationships are labelled on the edges, and types are labelled underneath each word. Looking at \figurename~\ref{fig:classic_sentence}, \texttt{Newcastle} are all children of \texttt{have}, through \texttt{nsubj} and \texttt{dobj} relationships, respectively. Types are determined from \gls*{pos} tags \cite{jm3}, so we can identify that \texttt{have} is a verb as it is annotated with \texttt{VBP} (a verb of the present tense and not third-person singular). The \texttt{nsubj} relation stands for the \textit{nominal subject}, and \texttt{dobj} is the \textit{direct object}, meaning that \texttt{Newcastle} acts upon \texttt{traffic} by \textit{having} traffic.  \texttt{Brighton} is also a child of \texttt{Newcastle} through a \texttt{conj} relation, and \texttt{Newcastle} has a \texttt{cc} relation with \texttt{and}, implying that these two subjects are related. Consequently, if we know Newcastle has traffic, then it holds that Brighton does as well. These \gls*{pos} tags also indicate \texttt{Newcastle} and \texttt{Brighton} are proper nouns as they both have \texttt{NNP} types.

\textit{\acrfull*{amr}} graphs proposed by Goodman et al. \cite{goodman-etal-2016-noise} provide a straightforward sentence representation as graphs, which mainly connects the sentence verb to the arguments belonging to the sentence. Although this representation can be enhanced to support full Multi-Word Entity Resolution using background knowledge (\texttt{wiki} relationship in \figurename~\ref{fig:AMRbn}, missing from \figurename~\ref{fig:classic_sentence}),  and despite its recent application in \glspl*{llm} for achieving logical reasoning abilities \cite{bao-etal-2024-abstract}, this representation discards relevant semantic relationships between words in the text, which might be relevant to faithfully capturing the distinction between subjects, (direct) objects, and other adverbial phrases occurring within the sentence. In comparing \figurename~\ref{fig:amrNCC} with \figurename~\ref{fig:negation_sentences}, it is clear that specific propositions such as ``\textit{in}'', useful for extracting information concerning a space-related adverbial phrase, are discarded from the \gls*{amr} graph but retained in the \gls*{ud} graph. 
Furthermore, both the subject of the main sentence (``traffic'') and the space-related adverbial phrase are marked with the same relationship label \texttt{ARG1}, while \gls*{ud} graphs distinguish these two logical functions with two distinct relationships, \texttt{nsubj} and \texttt{nmod}. For this reason, our approach uses \glspl*{ud} rather than \glspl*{amr} for retaining complex semantic representations of sentences. To overcome \gls*{ud}'s only shortcoming, we provide a preliminary \textit{a priori} explanation phase, enabling multi-word entity recognition using well-known NLP tools and vocabularies (Section \ref{sub:apriori}).

Capturing syntactical features through training is challenging. \gls*{nn}-based approaches are not proficient in precisely capturing relationships within text, as they fall down the same limitations as vector-based representations of sentences. \figurename~\ref{fig:negation_sentences} shows how \gls*{ai} struggles with understanding negation from full text: the sentence was fed into a natural language parser \cite{lexparser}, and the result shows no sign of a negated (\texttt{neg}) dependency, despite ``\textit{but not}'' being contained within the sentence. Still, we can easily identify and fix these issues before returning the \gls*{dg} to our LaSSI pipeline.

\section{Linguistics and Grammatical Structure} \label{sub:grammatical_structure}

The notion of the systematic and rule-based characterisation of human languages pre-dates modern studies on language and grammar: Aṣṭādhyāyī by Pāṇini utilised a derivational approach to explain the Sanskrit language, where speech is generated from theoretical, abstract statements created by adding affixes to bases under specific conditions \cite{panini}. This further supports the idea of representing most human languages in terms of grammatical derivational rules, from which we can identify the grammatical structure and functions of the terms occurring in a piece of text \cite{Christensen_2019}. This induces the recursive structure of most Indo-European languages, including English, which should be addressed to better analyse the overall sentence structure into its minimal components.

\begin{figure}[H]
	\centering
	\includegraphics[width=0.85\linewidth]{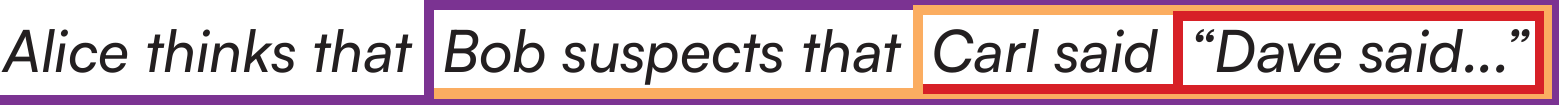}
	\caption{Example of a recursive sentence, highlighted with coloured bounding boxes \cite{DBLP:phd/it/Bergami2018a}.}
	\label{fig:recursive_sentence}
\end{figure}

Consider the example in \figurename~\ref{fig:recursive_sentence}; this could continue infinitely as a consequence of recursion in language due to the lack of an upper bound on grammatical sentence length \cite{facultyOfLanguage}. As the full text can be injected with semantic annotations, these can be further leveraged to derive a logical representation of the sentence \cite{10.1007/978-3-031-38499-8_29}.

Richard Montague developed a formal approach to natural language semantics, which later became known as \gls*{mg}, where natural and formal languages can be treated equally \cite{Montague+1975+94+121} to allow for the rewriting of a sentence in a logical format by assuming the possibility of \gls*{pos} tagging. \gls*{mg} assumes languages can be rewritten given their grammar \cite{Montague+1975+94+121}, preserving logical connectives and expressing verbs as logical predicates.  \gls*{mg} then provides a set of rules for translating this into a logical form; for instance, a sentence ($S$) can be formed by combining a noun phrase ($NP$) and a verb phrase ($VP$). We can also find the meaning of a sentence obtained by the rule $S \colon NP\ VP$, whereby the function for $NP$ is applied to the function $VP$. \gls*{mg} uses a system of types to define different kinds of expressions and their meanings. Some common types include $t$, denoting a term (a reference to an entity), and $f$, denoting a formula (a statement that can be true or false). The meaning of an expression is obtained as a function of its components, either by applying the function or by constructing a new function from the functions associated with the component. This compositionality makes it possible to assign meanings reliably to arbitrarily complex sentence structures, enabling us to then extract predicate logic from this, so the sentence ``\textit{Alice plays football}'' becomes: \texttt{play(Alice, football)}. 

However, \gls*{mg} only focuses on the \textit{logical form} of full text, overlooking the nuances of context and real-world knowledge. For example, does ``\textit{Newcastle}'' refer to ``\textit{Newcastle-upon-Tyne, United Kingdom}'', ``\textit{Newcastle-under-Lyme, United Kingdom}'', ``\textit{Newcastle, Australia}'', or ``\textit{Newcastle, Northern Ireland}''? Without an external \gls*{kb} or ontology, it is difficult to determine which of these it could be unless the full text provides relevant explicit information. Therefore, providing a dictionary of possible matches for all words in the sentence can significantly improve the \gls*{meu} recognition, meaning known places, people, and organisations can be matched to enhance the understanding of the syntactic structure of a given full text. At the time of this paper's writing, no \gls*{gql} can combine semantic utilities related to entity resolution alongside structural sentence rewriting. Therefore, this forces us to address minimal sentence rewriting through \glspl*{gql}, while considering the main semantic-driven rewritings in our given Python code base, where all of these are accounted for.

\subsection{Italian Linguistics}\label{sec:itling}
Not all grammatical information can be captured from \gls*{mg} alone: we can identify words that are \textit{verbs} and \textit{pronouns}, but these can both be broken down into several sub-categories that infer different rewriting that is not necessarily apparent from the initial structure of the sentence. For instance, a \textit{transitive verb} is a verb that can take a direct object, ``\textit{the cat \textbf{eats} the mouse}'', so when rewriting into the logical form, we know that a direct object must exist: \texttt{eat(cat, mouse)}, where \textbf{eat} is acting on the \textit{mouse}. However, if the verb is \textit{intransitive}, ``\textit{\textbf{going} across the street}'', then the logical form must not have a direct object and is thus removed, as the target does not reflect the direct object. Therefore, this sentence becomes: \texttt{go(?)[(SPACE:street[(det:the), (type:motion through place)])]}, as \textbf{go} does not produce an action on the street. The target is removed from the rewriting to reflect the nature of intransitive verbs. All these considerations are not accounted for in current \gls*{nlp} pipelines for \gls*{qa} \cite{10.1007/978-3-031-38499-8_29}, where merely simple binary relationships are accounted for, and the logical function of the \gls*{pos} components is not considered. 

In Italian linguistics, the examination of a proposition, commonly referred to as \textit{logical analysis}, is the recognition process for the components of a proposition and their logical function within the proposition itself \cite{analisilogica}. In this regard, this analysis recognises each clause/sentence as a predicate with its (potentially passive) subject, where we aim to determine the function of every single component: the verb, the direct object, and any other ``indirect complement'' that can refer to either an indirect object, adverbial phrase, or a locative \cite{Syntactic_stack}. This kind of analysis aims to determine the type of purpose the text is communicating and characterises each adverbial phase based on the information conveyed (e.g., limitation, manner, time, space) \cite{analisilogica}. This significantly differs from the \textit{POS tagging} of each word appearing in a sentence, through which each word is associated to a specific \gls*{pos} (adjective, noun, verb), as more than just one single word could participate in providing the same description. Concerning \figurename~\ref{fig:classic_sentence}, both \textit{Newcastle} and \textit{Brighton} are considered part of the same subject, \textit{Newcastle and Brighton}, while \textit{in Newcastle} is recognised as a \texttt{space} adverbial of time \textit{stay in place} given that the preposition \textit{it} and the verb \textit{is} are not indicating a motion rather than a state. Concerning \figurename~\ref{fig:negation_sentences}, this analysis considers ``\textit{but not in the city centre}'' (\figurename~\ref{fig:negation_sentences}) a separate coordinate clause, where ``\textit{There is (not) traffic}'' is subsumed from the main sentence. We argue that the possibility of further extracting more contextual knowledge from a sentence via \textit{logical analysis} tagging helps the machine to categorise the text's function better, thus providing both machine- and human-readable explanations. Although there is no support in the English language literature for these sentence-linguistic functions, since such functions are almost standard in all Indo-European languages, they can naturally be defined for the English language. In support of this,  \tablename~\ref{tab:logicalFunction} showcases the rendition of such functions and, given the lack of such characterisation in English, we freely exploit the characterisation found in Italian linguistics and contextualise this for the English language. 

\begin{table}[H]
\centering
	\caption{Some examples of logical functions considered in the current LaSSI pipeline beyond the simple subject--verb--direct object categorisation. These are then categorised and explained in our Parmenides ontology (Section \ref{sec:lsa}), where these are not only defined by the POS tags and the associated prepositions but also by the type of verb and entity associated with them.}\label{tab:logicalFunction}
	\begin{tabular}{ll|l}
		\toprule
		\textit{Logical Function} & \textit{(Sub)Type} & \textit{Example (underlined)}\\
		\midrule
		Space & Stay in place & ``\textit{I sit \underline{\textbf{on} a tree}.}''\\
		Space & Motion to place & ``\textit{I go \underline{\textbf{to} Bologna}.}''\\
		
		Space & Motion from place & ``\textit{I come \underline{\textbf{from} Italy}.}''\\
		Space & Motion through place & ``\textit{Going \underline{\textbf{across} the city center}.}''\\
		
		Cause & -- & ``\textit{Newcastle is closed \underline{\textbf{for} congestion}}''\\
		Time & Continuous & ``\textit{The traffic lasted \underline{\textbf{for} hours}}''\\
		Time & Defined & ``\textit{\underline{\textbf{On} Saturdays}, traffic is flowing}''\\
		
		\bottomrule
		
	\end{tabular}
	
\end{table}

To distinguish between Italian and English linguistic terminology, we refer to the characterisation of such sentence elements beyond the subject--verb--direct object characterisation as logical functions. Section \ref{sec:lsa} provides additional information on how such linguistic functions are recognised from a rewritten intermediate graph representation within LaSSI for the English language. We define such linguistic functions and how they can be matched through rewriting rules expressed within our ontology, Parmenides.

\section{Generation of \texttt{SetOfSingletons}}

\subsection{Multi-Word Entities}\label{mweandextras}
\begin{figure}[!h]
	\centering
	\includegraphics[width=0.6\linewidth]{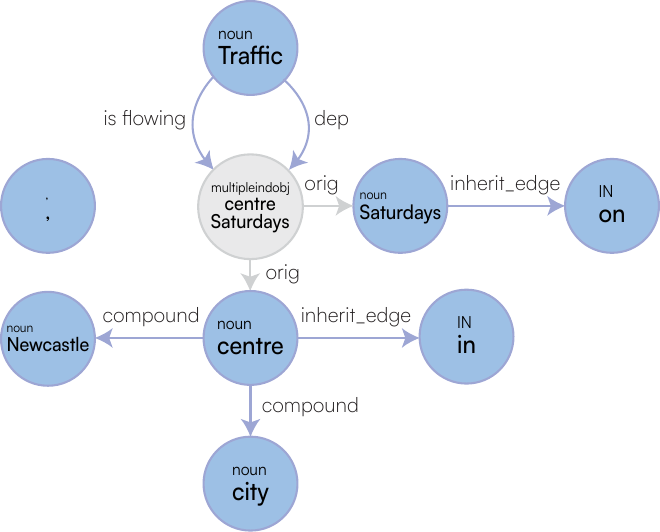}
	\caption{\gls*{ggg} output for ``\textit{Traffic is flowing in Newcastle city centre, on Saturdays}''.}
	\label{fig:multipleindobj}
\end{figure}

\texttt{Compound} types are labelled as a type \texttt{GROUPING}. There are two scenarios where \texttt{compound} edges present themselves: first, in a chain $(a)\xrightarrow{compound}(b)\xrightarrow{compound}(c)$, or second, a parent node with multiple children directly descending from it, $(b)\xleftarrow{compound}(a)\xrightarrow{compound}(c)$ (Newcastle city centre from Figure \ref{fig:multipleindobj}). To detect these structures, we use \gls*{dfs}, as entities that may have children that present extra information to their parents should be identified before the resolution of the parents. In the pipeline, these edges are removed, with the children either appended to the parent node's name or added as an \texttt{extra} property of the parent. These can be further refined to separate the main entity from any associated specification (Supplement~\ref{typehier}).

\begin{example}\label{ex:trarunning}
	Looking at the second case, we can focus on ``\textit{Newcastle city centre}'', which has edges \\$(\texttt{city})\xleftarrow[]{compound}(\texttt{centre})\xrightarrow[]{compound}(\texttt{Newcastle})$. Here, \texttt{Newcastle} crucially needs to be resolved, to identify it is a \texttt{\gls*{gpe}} for when we merge all the \texttt{Singletons} in Algorithm \ref{alg:merge_singletons}.
\end{example}

\subsection{Multiple Logical Functions}\label{app:mlf}
Concerning the detection of multiple logical functions, \figurename~\ref{fig:multipleindobj} has just one sentence with ``\textit{is flowing}'' as a verb, where multiple logical functions have the nodes  ``\textit{centre}'' and ``\textit{Saturdays}'' as entry points and ancestors. 
The node \texttt{centre Saturdays} would become a \texttt{SetOfSingletons} with \texttt{entities} \texttt{Newcastle} and \texttt{Saturdays}. 
As we have a node with \texttt{compound} edges in \figurename~\ref{fig:multipleindobj}, these are resolved into one \texttt{Singleton} per Example \ref{ex:trarunning}: \texttt{Newcastle[extra:city centre}]. Finally, we create our new logical relationships: \texttt{is flowing(traffic, Newcastle[\texttt{extra:city centre}])} and: \texttt{dep(traffic, Newcastle[\texttt{extra:city centre}])}.

\subsubsection{Handling Extras}\label{typehier}
When dealing with multi-word entities, we must identify whether a subset of these words acts as a specification (\texttt{extra}) to the primary entity of interest or whether it should be treated as a single entity. For example, ``\textit{Newcastle upon Tyne}'' would be represented as one \texttt{Singleton} with no \texttt{extra} property, whereas ``\textit{Newcastle city centre}'' has the {core} entity ``\textit{Newcastle}'' and an \texttt{extra} ``\textit{city centre}''.

To derive which part of the multi-named entity provides a specification for the main entity, we use Algorithm \ref{alg:merge_singletons}. The input takes a \texttt{node}, which is the \texttt{SetOfSingletons} to be resolved, and the \texttt{meu\_db} referring to the specific sentence of interest. The entities from the given \texttt{SetOfSingletons} are first sorted by the position of each node occurring within the full text, and a list of alternative representations is created from the power set of its associated entities from which \texttt{Singletons} and empty sets are removed. For example, given an array of three elements $[x, y, z]$, our \texttt{layered\_alternatives} from Line \ref{line:get_alternatives} are $[[x, y], [y, z], [x, y, z]]$. 	

\begin{example}[Example \ref{ex:trarunning} cont.]\label{ex:trarunning2}
	For ``\textit{Newcastle city centre}'', we  obtain \\\noindent $[[\texttt{Newcastle}, \texttt{city}],$ $[\texttt{city},\texttt{centre}], [\texttt{Newcastle}, \texttt{city}, \texttt{centre}]]$, representing all possible combinations of each given \texttt{Singleton} within the \texttt{SetOfSingletons} to extract the \texttt{extra} from the main entity.
\end{example}

We based our inference on a type hierarchy to differentiate between the components providing the specification (\texttt{extra}) to the main entity.	The hierarchy of types seen in use on Line \ref{line:most_specific_type} employs a \texttt{Most Specific Type (MST)} function to return a \texttt{Specific Type (ST)} from a curated entity-type hierarchy as follows: \texttt{VERB} $<:$ \texttt{GPE} $<:$ \texttt{LOC}(ation) $<:$ \texttt{ORG} $<:$ \texttt{NOUN} $<:$ \texttt{ENTITY} $<:$ \texttt{ADJECTIVE}. If none of these are met, then the type is set to \texttt{None}. This is updated from our previous pipeline so that \texttt{VERB} is now the \textit{most} specific type. Adjectives are also captured, as these were missing from the previous hierarchy.

Lastly, we look through the \gls*{meudb}, comparing the minimum and maximum values of the given alternative for a corresponding match that has the highest confidence value (Line \ref{line:get_score_from_meu}).  These confidence scores lead to how we calculate which alternative should be used for resolving later in our \texttt{Singleton} (Line \ref{line:should_add_alternative}). We check for whether the current candidate in the loop has a greater confidence score than the total confidence score, which is calculated from the product of all confidence scores within the entities.

\section{Recursive Sentence Rewriting}\label{app:recursiveSentenceRewr}
\textit{We now describe in greater detail any subroutine required by Algorithm \ref{alg:overall_kernel_construction} to derive further subsequent computational steps.}

{Algorithm \ref{alg:overall_kernel_construction} sketches the implementation of this phase, nesting the different relationships retrieved through a \texttt{Singleton}.} After identifying which edges are the candidates (containing verbs) to become relationships (\texttt{edges}, Line \ref{alg:get_kernel_edges}), which are the main entry points for each of these to extract a relevant subgraph (\texttt{top\_ids}, Line \ref{alg:get_top_ids}), we can now create all relationships representing each verb from the full text to then connect them into a single \texttt{Singleton}-based tree. After initialising the pre-conditions for the recursive analysis of the sentences (\texttt{settings} on Line \ref{line:loop_settings_init}) for each sentence entry point to be analysed ($n$, an ID), we collect all relevant nodes and edges associated with it: $d$, all descendant nodes of $n$ retrieved from a \gls*{bfs} from our \texttt{new\_edges} (see Algorithm \ref{alg:overall_kernel_construction}). We then filter the edges by ensuring the following: the source and target are contained within the descendant nodes ($d$) or the target node is in our preposition labels, and the source and target have not already been used in a previous iteration or they \textit{have} been used in a previous iteration but we have at least one preposition label within the text (Line \ref{line:filter_edges}). However, if our loop \texttt{settings} contain a relationship, {we use it as our \texttt{filtered\_edges}, as we need to create a new one}.

	\begin{algorithm}[!h]
	\caption{After our \textit{a priori} phase, we move to \textbf{creating our final kernel}. This is how our sentence is represented before transforming into our final logical representation.}
	\label{alg:overall_kernel_construction}
	\begin{algorithmic}[1]
		\State \texttt{new\_edges, true\_targets, preposition\_labels} $\gets$ \Call{GetKernelEdges}{edges, nodes}\label{line:get_kernel} \Comment{\textit{Algorithm \ref{alg:get_kernel_edges}}}
		\State \texttt{top\_ids} $\gets$ \Call{GetTopologicalRootNodeIDs}{\texttt{new\_edges}, \texttt{true\_targets}}  \Comment{\textit{Algorithm \ref{alg:get_top_ids}}}\label{line:twoalg2}
		\State \texttt{used\_edges} $\gets\ \emptyset$
		\State \texttt{acl\_map} $\gets\ \{\}$
		\State \texttt{position\_pairs} $\gets$ \Call{GetPositionPairs}{\texttt{new\_edges}} \label{line:get_position_pairs}
		
		\For{$n \in\ \texttt{top\_ids}$}
		\State $\texttt{settings} \gets\ \{\texttt{shouldLoop}=\textbf{true},\ \texttt{edgeForKernel}=\textbf{none}\}$  \label{line:loop_settings_init}
		\While{\texttt{settings.shouldLoop}}
		\State $d \gets \Call{BFS}{\texttt{new\_edges}, n}$ \label{line:bfs}
		\State \texttt{filtered\_edges} $\gets$ \Call{FilterEdges}{$d$, \texttt{new\_edges}, \texttt{preposition\_labels}} \label{line:filter_edges}
		
		\If{\texttt{settings.edgeForKernel} $\ne$ \textbf{none}}
		\texttt{filtered\_edges} $\gets\ [\texttt{settings.edgeForKernel}]$
		\EndIf
		
		\State \texttt{used\_edges} $\gets$ \texttt{filtered\_edges}
		\State \texttt{kernel, settings, acl\_map} $\gets$ \Call{CreateSentence}{\texttt{filtered\_edges}, $d$, $n$,

			\texttt{preposition\_labels}, \texttt{settings}, \texttt{acl\_map}} \Comment{\textit{Algorithm \ref{alg:create_sentence_func}}}\label{line:create_sentence}
		
		\State \texttt{kernel} $\gets$ \Call{PostProcessing}{\texttt{kernel}, \texttt{position\_pairs}} \label{line:post_processing}
		
		\If{$|\texttt{top\_ids}| > 1$}
		\State \texttt{kernel} $\gets$ \Call{CheckIfEmptyKernel}{\texttt{kernel}} \label{line:empty_kernel}
		\If{\texttt{kernel} $\ne$ \textbf{none}} \texttt{nodes[$n$]} $\gets$ \texttt{kernel} \EndIf
		\Else\
		\texttt{nodes[$n$]} $\gets$ \texttt{kernel}
		\EndIf
		
		\State \texttt{nodes} $\gets\ \Call{RemoveRootFromAllNodes}{d}$ \label{line:remove_roots}
		\If{\texttt{kernel} $\ne \textbf{none} \land\ |\texttt{top\_ids}| > 1 \land\ n = \textbf{last}(\texttt{top\_ids})$}
		\texttt{top\_ids.pop()}
		\EndIf
		\EndWhile
		\EndFor
		
		\State \texttt{final\_kernel} $\gets$ \texttt{nodes[}\textbf{last}\texttt{(top\_ids)]} \label{line:select_kernel}
		
		\State {$\rhd$ Post-Processing}
		\State \texttt{final\_kernel} $\gets$ \Call{AclReplacement}{\texttt{final\_kernel, acl\_map}} \label{line:acl_rep}
		\State \texttt{final\_kernel} $\gets$ \Call{CheckForActionNode}{\texttt{final\_kernel}} \label{line:check_action}
		\State \texttt{final\_kernel} $\gets$ \Call{RemoveDuplicateProperties}{\texttt{final\_kernel}} \label{line:duplicate_props}
		\State \texttt{final\_kernel} $\gets$ \Call{RewritePropertiesLogically}{\texttt{final\_kernel}} \Comment{\textit{Algorithm \ref{alg:rewrite_props}}} \label{line:rewrite_properties}
		\State \texttt{final\_kernel} $\gets$ \Call{CheckForAdv}{\texttt{final\_kernel}} \label{line:check_for_adv}
		\State \Return \texttt{final\_kernel}
		
		
	\end{algorithmic}
\end{algorithm}

\textsc{CreateSentence} only handles the rewriting of at most one kernel, whereas another may be contained within its properties; 
therefore, we handle this by returning a possible new kernel through ``\texttt{settings.edgeForKernel}'' and create a new sub-sentence to be rewritten with the same root ID, which is determined from our conditions set out in Supplement \ref{sub:binary_rels}. At this stage, we assign our \texttt{used\_edges} to our collected \texttt{filtered\_edges} for the next (possible) iteration. 

After considering only the edges relevant to generating logical information for the sentence and after electing those to become a relationship across \texttt{Singletons} as our \texttt{kernel} (Line \ref{line:create_sentence}), we further refine the content of the selected edge and carry out a post-processing phase also considering \textit{semi-modal verbs} \cite{semimodal} (Line \ref{line:post_processing}). 
We also check if we have multiple verbs leading to multiple relationships generating new \texttt{kernels}; if so, we check if this current relationship has no appropriate source or target  (Line \ref{line:empty_kernel}). We refer to these relationships as \textit{empty}. However, we check within the properties of this kernel to see if a kernel is present within these properties and whether this can be used as our new kernel instead. Following this, we remove all root properties from the nodes used in this iteration to avoid being considered in the next step 
and produce duplicate rewritings. Finally, if we are considering more than one kernel, and the last rewritten kernel is \textbf{none}, then we remove it to ensure that the last successfully rewritten kernel is used for our final kernel. The kernel is selected by taking the ID of the last occurring ID in \texttt{top\_ids}, 
which is the first relationship in topological order for a given full text (Line \ref{line:select_kernel}).
Finally, we check if the edge label is a verb; if not, it is replaced with \textbf{none}. Otherwise, we return the kernel.

The final stage of the kernel creation is additional \textit{post-processing} to further standardise the final sentence representation: after resolving the pronouns with the entities they are referring to (Line \ref{line:acl_rep}), we generate relationships as verbs as either occurring as edge graphs or node properties (Line \ref{line:check_action}). After cleaning redundant properties (Line \ref{line:duplicate_props}), we rewrite such properties as Logical Functions of the sentence (Line \ref{line:rewrite_properties}, Section \ref{sec:lsa}). Last, we associate propositions to the verbs when forming phrasal verbs (Line \ref{line:check_for_adv}).

	
	\subsection{Promoting Edges to Binary/Unary Relationships}\label{appsub:promote_edge}
	\textit{Algorithm \ref{alg:get_kernel_edges}} collects all the appropriate edges to be considered when creating the kernel, and ensures that all edge labels to be considered are verbs while labels that are not verbs are disregarded. The first step in creating these kernels is obtaining all the appropriate edges and determining which nodes within those edges are \textit{already} roots, or should be considered if they are not already. This is presented in Line \ref{func:get_k_edges}. Edges that are labelled with `dep' are skipped in Line \ref{line:ignore_dep} as, for this type of edge, it is `\textit{impossible to determine a more precise relation}' \cite{dep_ud}, and they can therefore be ignored. Next, we check every edge label for \textit{prototypical prepositions}, retrieved from our ontology in Line \ref{line:proto_ontol}, which determines target nodes to be roots which are not already handled by dependency parsing. These prototypical prepositions are a single word (a preposition) that precedes a noun phrase complement, which expresses spatial relations \cite{d462da49-0226-3286-9df6-376b8db5d528}; for example: ``\textit{the mouse is eaten \textbf{by} the cat}'' or ``\textit{characters \textbf{in} movies}''. We check whether a edge contains a \textit{prototypical preposition} in Line \ref{line:check_proto}: we want to determine whether the preposition is contained within a wider edge label, but is not exactly equal to it; for example, if we had the label ``\textit{to steal}'', this \textbf{contains} ``\textit{to}'', so it would return \textbf{true}. However, if we had ``\textit{like}'', this is exactly equal to a preposition
	and, therefore, \textbf{false} would be returned. We also check for when the target does \textbf{not} contain a \texttt{case} property as, if we imply that a target \textit{with} a \texttt{case} property \textit{is} a sub-sentence, then we may lose additional information necessary for the logical rewriting phase in Section \ref{sec:lsa}. We also perform an additional check for whether the edge label ends with ``\textit{ing}'' and does not contain an auxiliary as a condition, in order to accept a true state for a prototypical preposition. 
	We carry out this process such that sub-sentences can still be captured when a preposition is not contained within the full text, but still introduced by a verb in its gerund form (-ing). We ensure that this does not contain an auxiliary, such that we do not incorrectly consider that a new sentence is being introduced.
	
	\begin{algorithm}[h]
		\caption{Edges collected from our \textit{a priori} phase need to be analysed to ensure that they are all relevant and structured correctly, such that our kernel best represents the given full text. This function checks for prototypical prepositions within edge labels, in order to possibly rewrite targets of a number of edges.}
		\label{alg:get_kernel_edges}
		\begin{algorithmic}[1]
			\Function{GetKernelEdges}{edges, nodes}\Comment{1. Get Edges to Use in Kernel Creation} \label{func:get_k_edges}
			\If{$|\mathtt{edges}| > 0$}
			\State \texttt{prototypical\_prepositions} $\gets$ \Call{GetPrototypicalPrepositions}{ } \label{line:proto_ontol}
			\State $L'(s, t) := \{e\texttt{.edgeLabel.name} \mid e \in \texttt{edges},\ e\texttt{.source.id} = s,\ e\texttt{.target.id} = t \}$ 
			\State $L(s, t) := (\texttt{``dep''} \in L'(s, t)) \land (|L'(s, t)| > 1)$ \label{line:edge_labels}
			\For{$e \in$ \texttt{edges}}
			\If{$e\mathtt{.edgeLabel.name} = \texttt{``dep''} \land L(e\texttt{.source.id},\ e\texttt{.target.id})$}
			\textbf{continue} \label{line:ignore_dep}
			\EndIf
			\State \texttt{new\_edge} $\gets$ \textbf{none}
			
			\For{$\texttt{p} \in \texttt{prototypical\_prepositions}$}
			\If{\Call{CheckForPrototypicalPreposition}{\texttt{p}, $e$}} \label{line:check_proto}
			\State \texttt{found\_sub\_sentence} $\gets$ \textbf{true}
			\State $e$\texttt{.target.addRootProperty()}
			\State \texttt{preposition\_labels[$e$.target.id]} $\gets\ e\texttt{.edgeLabel.name}$
			\State \texttt{new\_edge} $\gets\ e\texttt{.updateTarget(nodes[}e\texttt{.target.id])}$
			\State \textbf{break}
			\EndIf
			\EndFor
			
			\If{\Call{ShouldRemoveTargetRoot}{}} \label{line:remove_target_root}
			\State \texttt{new\_edge} $\gets\ e\texttt{.updateTarget(}e\texttt{.target.removeRootProperty())}$
			\EndIf
			
			\If{\texttt{new\_edge} $=$ \textbf{none}}
			\State \texttt{new\_edges} $\gets$ \texttt{new\_edges} $\cup\ \{e\}$
			\If{\textbf{not }\texttt{kernel\_in\_props($e$.target)}}
			\texttt{true\_targets} $\gets$ $e$.target.id
			\EndIf
			\Else
			\State \texttt{new\_edges} $\gets$ \texttt{new\_edges} $\cup\ \{\texttt{new\_edge}\}$ 
			\EndIf
			\EndFor
			\EndIf
			
			\State \Return new\_edges, true\_targets, preposition\_labels
			\EndFunction
		\end{algorithmic}
	\end{algorithm}
	
	If we find a preposition, we add a root property to the target node of the given edge and map the target ID to the full edge label name in a \texttt{prepositions} list, which is to be used when determining which edge to use when creating the sentence in Algorithm \ref{alg:create_sentence_func}.
	
	There might be situations where roots are \textit{incorrectly} identified, which is reflected in Line \ref{line:remove_target_root}. We check the following: if we have \textbf{not} found a preposition in the current edge label, the edge label is a verb, the source is a root, and the target is not an existential variable. If these conditions are all true, then the target's root property (if present) is removed. We are removing something that is incorrectly recognised as a root; however, this does not affect verbs as in Algorithm \ref{alg:get_top_ids}, and Line \ref{line:check_root} will always return true if the given entity is a verb regardless of whether it contains a root property.
	
	\subsection{Identifying the Clause Node Entry Points}\label{sup:icnep}
	Now that we have a list of all edges that should be considered for our given full text, we need to find the IDs of the nodes that should be considered roots from which we generate all kernels. For this purpose, Algorithm \ref{alg:get_top_ids} iterates over each source and target node within every edge collected in the previous step (Algorithm \ref{alg:get_kernel_edges}). Then, we filter each node by checking if they have a root property and are not contained within the list of true targets, in order to mitigate the chance of duplicate root IDs being added to \texttt{filtered\_nodes} on Line \ref{line:check_root}. We also perform an additional check for whether the node is a \texttt{SetOfSingletons}, and remove any node from the list that may be a child of the \texttt{SetOfSingletons}, as it would be incorrect to consider them roots given that they are contained within the parent of the \texttt{SetOfSingletons}. 
	
	\begin{algorithm}[h]
		\caption{To properly encompass the recursive nature of the sentence, we find the root IDs within the given edges in topological order, ensuring that we maintain structural understanding when rewriting the sentence. (\textit{Algorithm \ref{alg:overall_kernel_construction} cont.})}
		\label{alg:get_top_ids}
		\begin{algorithmic}[1]
			\Function{GetTopologicalRootNodeIDs}{edges, true\_targets} \Comment{2. {Filter Root Node IDs}} \label{func:get_top_ids}
			\State $\texttt{filtered\_nodes} \gets \emptyset$
			\State $\texttt{top\_ids} \gets \emptyset$
			
			\For{$e \in \bigcup_{i \in \texttt{edges}} \{\texttt{$i$.source},\ \texttt{$i$.target}\}$}
			\If{$e \ne \textbf{none} \land e\texttt{.id} \notin \texttt{true\_targets} \land \Call{IsKernelInProps}{e}$} \label{line:check_root}
			\State \texttt{filtered\_nodes} $\gets$\ \texttt{filtered\_nodes} $\cup\ \{e\texttt{.id}\}$
			\If{$e \in$ \texttt{SetOfSingletons}}
			\For{$\texttt{entity} \in e\texttt{.entities}$}
			\State $\texttt{filtered\_nodes} \gets \texttt{filtered\_nodes} \setminus\ \{e\texttt{.id}\}$
			\EndFor
			\EndIf
			\EndIf
			
			\If{$e =$ \textbf{none} $\lor\ e\texttt{.id} \in$ \texttt{top\_ids}} \textbf{continue} \EndIf
			
			\State \texttt{top\_ids} $\gets$ \texttt{top\_ids} $\cup$ $\{e$\texttt{.id}$\}$
			\EndFor
			
			\State $\texttt{top\_ids} \gets [x\texttt{.id} \mid x \in \texttt{nodes},\ x\texttt{.id} \in \texttt{filtered\_nodes}]$ \label{line:sort_e_top_nodes}
			\If{$|\texttt{top\_ids}| = 0$} \label{line:valid_roots}
			\If{$|\texttt{nodes}| = 1$}
			\State $\texttt{top\_ids} \gets [\texttt{last}(\texttt{nodes})]$ 
			\Else
			\State $\texttt{top\_ids} \gets [\texttt{nodes[$x$].id} \mid x \in \texttt{nodes},\ \Call{IsKernelInProps}{\texttt{nodes[$x$]}}]$
			\EndIf
			\State \texttt{top\_ids} $\gets$ \Call{RemoveDuplicates}{\texttt{top\_ids}}
			\EndIf
			\State \Return \texttt{top\_ids}
			\EndFunction
		\end{algorithmic}
	\end{algorithm}
	
	With the \texttt{filtered\_nodes} collected, we need to sort them in topological order. In this way, when creating our logical representation, the structure is compliant with the original representation of the full text. As we have performed a topological sort on the list of nodes when first parsing the \gls*{gsm} {to generate a graph}, we create a list of \texttt{top\_ids} in Line \ref{line:sort_e_top_nodes}. If we have not found any root nodes based on the filtering performed, then, if only one node remains, this becomes our root node; otherwise, we collect all nodes that have a root property as a fallback.
	
	\subsection{Generating Binary Relationships (Kernels)}\label{sub:binary_rels}

	\begin{algorithm}[H]
		\caption{Construct final relationship (kernel) (\textit{Algorithm \ref{alg:overall_kernel_construction} cont.})} 
		\begin{algorithmic}[1]
			
			\Function{CreateSentence}{\texttt{edges}, \texttt{nodes}, \texttt{root\_id}, \texttt{preposition\_labels}, \texttt{prev\_settings}, \texttt{acl\_map}} \Comment{3. Create All Kernel: Create Kernel Object} \label{func:create_k}
			\State $\texttt{settings} \gets\ \{\texttt{shouldLoop}=\textbf{false}\}$
			
			\State \texttt{root\_node} $\gets$ \texttt{nodes[root\_id]}
			\If{\texttt{root\_node.type} $=$ `verb' $\land\ |\texttt{edges}| = 0$}
			\State \Return \Call{CreateEdgeKernel}{\texttt{root\_node}}, \texttt{settings}, \texttt{acl\_map} \label{line:edge_kernel}
			\ElsIf{$|\texttt{edges}|=0$}
			\State \Return \texttt{root\_node}, \texttt{settings}, \texttt{acl\_map}
			\EndIf
			
			\State \texttt{kernel} $\gets$ \Call{AssignKernel}{\texttt{edges}, \texttt{kernel}, \texttt{nodes}, \texttt{root\_id}, \texttt{preposition\_labels}} \Comment{\textit{Algorithm \ref{alg:assign_kernel}}}
			\State \texttt{kernel, kernel\_nodes} $\gets$ \Call{AnalyseNodes}{\texttt{kernel}} \label{line:analyse_k}
			\State \texttt{properties} $\gets\ \{\}$
			\For{$e \in$ \texttt{edges}}
			\State \texttt{kernel}, \texttt{properties}, \texttt{kernel\_nodes} $\gets$ \Call{AddToProps}{\texttt{kernel}, $e$} \label{line:add_to_props}
			
			\If{$e$\texttt{.edgeLabel.name} $\in\ \{$`acl\_relcl', `nmod', `nmod\_poss'$\}$}
			\If{$e$\texttt{.edgeLabel.name} $=$ `acl\_relcl'}
			\texttt{acl\_map} $\gets \texttt{acl\_map} \cup \{e\texttt{.source}\}$ \label{line:append_acl}
			\EndIf

			\State \texttt{edge\_kernel} $\gets$ \Call{ConvertToKernel}{$e$\texttt{.edgeLabel}} \label{line:convert_to_k}
			\State \texttt{kernel}, \texttt{properties}, \texttt{kernel\_nodes} $\gets$ \Call{AddToProps}{\texttt{kernel}, \texttt{edge\_kernel}}
			\EndIf
			
			\If{\Call{ShouldLoop}{$e$, \texttt{kernel}}} \label{line:check_for_loop}
			$\texttt{settings} \gets\ \{\texttt{shouldLoop}=\textbf{false},\ \texttt{edgeForKernel}=e\}$
			\EndIf
			\EndFor
			
			\State \texttt{prev\_kernel} $\gets$ \texttt{prev\_settings.previousKernel} \label{line:check_for_prev_kernel}
			\If{\texttt{prev\_kernel} $\ne$ \textbf{none}}
			\If{\textbf{position(\texttt{kernel}) $<$ \textbf{position(\texttt{prev\_kernel})}}} \label{line:top_position_check}
			\State \texttt{kernel}, \texttt{properties}, \texttt{kernel\_nodes} $\gets$ \Call{AddToProps}{\texttt{prev\_kernel}, \texttt{kernel}}
			\Else
			\State \texttt{kernel}, \texttt{properties}, \texttt{kernel\_nodes} $\gets$ \Call{AddToProps}{\texttt{kernel}, \texttt{prev\_kernel}}
			\EndIf
			\EndIf
			
			\State \texttt{new\_kernel, properties} $\gets$ \Call{CheckForNewEdgeLabel}{\texttt{properties}}
			
			\State \texttt{final\_kernel} $\gets$ \Call{ConstructSingleton}{\texttt{root\_id}, \texttt{kernel}, \texttt{properties}}
			
			\If{\texttt{new\_kernel} $\ne$ \textbf{none}}
			\texttt{final\_kernel} $\gets$ \Call{UpdateKernel}{\texttt{final\_kernel}, \texttt{new\_kernel}}
			\EndIf
			
			\If{\texttt{settings.shouldLoop}}
			\texttt{settings.previousKernel} $\gets$ \texttt{final\_kernel} \label{line:prev_kernel_set}
			\EndIf
			
			\State \Return \texttt{final\_kernel}, \texttt{settings}, \texttt{acl\_map}
			\EndFunction
		\end{algorithmic}
		\label{alg:create_sentence_func}
	\end{algorithm}
	
		To create our intermediate binary relationships, Algorithm \ref{alg:create_sentence_func} retrieves the node from the passed \texttt{root\_id} (our $n$ in Algorithm \ref{alg:overall_kernel_construction}), and check whether we have any edges to use in rewriting. If we do not have any edges and the node is a verb, we create an `edge kernel' on Line \ref{line:edge_kernel}; as an example, if we had the node `work' with no other edges, then we return the kernel \texttt{work(?, None)}, as we have no further information at this stage. If it is \textbf{not} a verb and there are no edges, then we simply return the node as we cannot perform any rewriting. 
	
	If these conditions are not met, then we \textbf{assign our kernel}, whereby we find the source, target, and edge label to be used (as detailed in Algorithm \ref{alg:assign_kernel}). From this returned kernel, we create a list of \texttt{kernel\_nodes}, which are the nodes to be considered before adding to the properties of the kernel. Here, if a property to be added is in the \texttt{kernel\_nodes}, it should be ignored as the information is already contained in the entire kernel. Therefore, we call a function to `analyse' the source and target in Line \ref{line:analyse_k}, which adds the source and target to \texttt{kernel\_nodes}. 
	
	
	\begin{algorithm}[H]
		\caption{Given a list of edges, we find the most relevant edge (using a set of rules narrated in the text) that should be used as our kernel. (\textit{Algorithm \ref{alg:create_sentence_func} cont.})}
		\label{alg:assign_kernel}
		\begin{algorithmic}[1]
			\Function{AssignKernel}{\texttt{edges}, \texttt{kernel}, \texttt{nodes}, \texttt{root\_id}, \texttt{preposition\_labels}} 
			\Comment{4. Assign Kernel} \label{func:assign_k}
			\State \texttt{chosen\_edge} $\gets$ \textbf{none}
			\For{$e \in \texttt{edges}$} \label{line:get_chosen_edge}
			\If{$e$\texttt{.edgeLabel.type} $=$ `verb' $\land (\texttt{root\_id}\in\texttt{preposition\_labels} \lor e\texttt{.edgeLabel.name} \notin \texttt{preposition\_labels}) \land e\texttt{.source.id} = \texttt{root\_id}$}
			\texttt{chosen\_edge} $\gets e$
			\EndIf
			\EndFor
			
			\For{$e \in \texttt{edges}$}
			\If{$((e\texttt{.edgeLabel.type}=\text{`verb'} \lor e\texttt{.source.type}=\text{`verb'})\land\texttt{chosen\_edge}=\textbf{none}) \land (\texttt{root\_id} \in \texttt{preposition\_labels} \lor e\texttt{.edgeLabel.name} \notin \texttt{preposition\_labels})$} \label{line:existential_source}
			\If{$e$\texttt{.edgeLabel.type} $=$ `verb'}
			\texttt{edge\_label} $\gets\ e$\texttt{.edgeLabel}
			\Else\
			\texttt{edge\_label} $\gets\ e$\texttt{.source}
			\EndIf
			
			\If{\textbf{not} \Call{SemiModal}{$e$\texttt{.source.name}} $\land\notin\texttt{nodes}\lor\ e\texttt{.source}=\texttt{edge\_label}$}
			\State \texttt{edge\_source} $\gets$ \Call{NewExistentialVariable}{ } \label{line:create_exist}
			\Else
			\State \texttt{edge\_source} $\gets\ e\texttt{.source}$
			\EndIf
			
			
			
			\State \texttt{kernel} $\gets$ \Call{CreateRelationship}{\texttt{edge\_source}, $e$\texttt{.target}, \texttt{edge\_label}, $e$\texttt{.isNegated}}
			\If{\textbf{not} \Call{IsTransitive}{\texttt{edge\_label.name}}}
			\texttt{kernel.target} $\gets$ \textbf{none}
			\EndIf
			\EndIf
			\EndFor
			
			\If{\texttt{kernel} $\ne$ \textbf{none}}
			\texttt{kernel} $\gets$ \Call{FindExistential}{\texttt{edges}} \label{line:find_exist}
			\EndIf
			
			\If{\texttt{kernel} $\ne$ \textbf{none}}
			\State \texttt{kernel} $\gets$ \Call{TryCreateExistential}{\texttt{edges}} \label{line:try_create_exist}
			\If{\texttt{kernel} $=$ \textbf{none}}
			\texttt{kernel} $\gets$ \textbf{last}(\texttt{edge})
			\EndIf
			\EndIf
			
			\If{\Call{CaseInProperties}{\texttt{kernel.source}}}
			\State \texttt{kernel.source} $\gets$ \Call{NewExistentialVariable}{ }
			\ElsIf{\Call{CaseInProperties}{\texttt{kernel.target}}}
			\If{\texttt{kernel.target} $\notin$ \texttt{SetOfSingletons}} \texttt{kernel.target} $\gets$ \textbf{none}
			\Else\
			\texttt{kernel.target} $\gets$ \Call{FindValidTarget}{\texttt{kernel.target}} \label{line:find_valid_targ}
			\EndIf
			\EndIf
			
			\If{\texttt{kernel.edgeLabel.type} $\ne$ `verb'} \texttt{kernel.edgeLabel} $\gets$ \textbf{none} \EndIf
			
			\State \Return \texttt{kernel}
			\EndFunction
		\end{algorithmic}
	\end{algorithm}
	
	Next, we iterate over each edge $e$, and add the source and target nodes of each $e$ to \texttt{properties} (Line \ref{line:add_to_props}). If a given $e$ has an edge label equal to `acl\_relcl', `nmod', `nmod\_poss', then we rewrite this edge as a kernel within a \texttt{Singleton} (Line \ref{line:convert_to_k}) and add the entire edge to the properties, to later be rewritten in Algorithm \ref{alg:logical_function}. If the edge label is `acl\_relcl', then we append this to the \texttt{acl\_map}, used in Algorithm \ref{alg:overall_kernel_construction}, in Line \ref{line:acl_rep}.
	
	While iterating, we check in Line \ref{line:check_for_loop} for cases where $e$'s \texttt{properties} necessitate the creation another kernel from the edges.
	If $e$ is a verb and not already in the \texttt{kernel\_nodes}, we use this edge in the next iteration within Algorithm \ref{alg:create_sentence_func} using the same root node; in particular, \texttt{previousKernel} is assigned to the current \texttt{kernel} at Line \ref{line:prev_kernel_set}. We then check for this in Line \ref{line:check_for_prev_kernel}, where we compare the current and previous kernels in order to determine whether the previous kernel should become the root or be added as a property; this is determined by the positions of each (Line \ref{line:top_position_check}).
	
	\subsubsection{Kernel Assignment}\label{app:kerass}
	
	To assign our kernel, Algorithm \ref{alg:assign_kernel} first needs to determine which edge applies to the current kernel in Line \ref{line:get_chosen_edge}, iterating over all the edges to find our \texttt{chosen\_edge}. We then iterate again over the edges and, once we reach either our chosen edge or a verb when no chosen edge is found, we determine the attributes of our kernel. As long as our edge label is a verb, it remains the same; otherwise, we use the source of the current $e$. According to the condition on Line \ref{line:existential_source}, our source becomes an existential (Line \ref{line:create_exist}); otherwise, it remains as the source of $e$. We then construct our kernel and check whether the edge label is a \textit{transitive} verb or not (Section \ref{sub:grammatical_structure}) and, if it is \textbf{not}, then we remove the target as it reflects the direct object.
	
	If no kernel can be constructed, we look for an existential in Line \ref{line:find_exist} by checking the source and target for existential properties. If we \textit{still} cannot construct a kernel (i.e., no existential is found); then, we forcefully create the existential by looking for a verb within the list of \texttt{nodes} in Line \ref{line:try_create_exist}.
	
	At this stage, we check whether a \texttt{case} property is contained within the source or target and, if so, we remove the node (which is later appended as a property of the kernel). If the source is removed, it is replaced with an existential; if the target is removed, it is replaced with \textbf{none}. However, if our target is a \texttt{SetOfSingletons}, then we look for a valid target within the entities target, where we use the first occurring element by its position in the entities and append the rest of the entities as properties (Line \ref{line:find_valid_targ}).

	\section{Rewriting Semantics for Logical Function Rules in Parmenides}\label{sec:fullSemantics}

	We now discuss the entailed semantics for the application of rules capturing logical functions as per Algorithm~\ref{alg:rewrite_props}: for each relationship $k$ generated in the previous phase, we select all the \texttt{Singletons} (Line \ref{line:loop_over_key}) and \texttt{SetOfSingletons} (Line \ref{line:sos}) within its properties. For each of the former, we consider them in declaration order (\texttt{rule order}). Once we find a rule matching some preconditions (\texttt{premise}), we apply the rewriting associated with it and skip the testing for the other rules. When such a condition is met, we establish an association between the logical function determined by the rule and the matched \texttt{Singleton} or \texttt{SetOfSingletons} within the relationship properties. If this is differently stated at the level of the rule, we then move such a property to the level of the properties of another \texttt{Singleton} within the relationship of interest (\figurename~\ref{code:logical_function_space}).  We perform these steps recursively for any further nested relationship as part of the properties (Line \ref{line:ker1}).

	\begin{algorithm}[p]
		\caption{Properties contained within the kernel at this stage are not entirely covered logically. Therefore, this function determines, under a set of rules within the text, how they should be rewritten and appended to the properties of the kernel in order to be properly represented.}
		\label{alg:rewrite_props}
		\begin{algorithmic}[1]
			\Function{RewritePropertiesLogically}{$k$} \Comment{where $k$ is our Relationship}
			\State \texttt{properties} $\gets \{\}$
			\For{\texttt{key} $\in\ k$\texttt{.properties}} \label{line:for_key}
			\If{\texttt{key} $=$ `extra'} \label{line:check_extra}
			\State \texttt{properties} $\gets k\texttt{.properties[key]}$
			\State \textbf{continue}
			\EndIf	
			
			\For{$n \in k$\texttt{.properties[key]}} \label{line:loop_over_key}
			\If{$n \in \texttt{Singleton}$}\label{line:singleton}
			\If{$n$\texttt{.kernel} $\ne$ \textbf{none}}
			\If{$n$\texttt{.kernel.edgeLabel.name} $\in \{\text{`nmod'}, \text{`nmod\_poss'}\}$} \label{line:has_nmod}
			\State \texttt{properties} $\gets$ \Call{RewriteNodeLogically}{$k$, $n$, \texttt{properties}, \textbf{true}, \textbf{false}} \Comment{\textit{Algorithm \ref{alg:create_sentence_func}}} \label{line:rewrite_nmod}
			\State $k$, \texttt{properties} $\gets$ \Call{PropertyReplacement}{$k$, \texttt{properties}} \label{line:property_replace}
			\ElsIf{$n$\texttt{.type} $=$ `SENTENCE'}
			\State $n \gets$ \Call{RewritePropertiesLogically}{$n$}
			\State \texttt{properties[key]} $\gets$ \texttt{properties[key]} $\cup\ \{n\}$
			\EndIf
			\Else
			\State \texttt{properties} $\gets$ \Call{RewriteNodeLogically}{$k$, $n$, \texttt{properties}, \textbf{false}, \textbf{false}}\label{line:ker1}
			\EndIf
			\ElsIf{$n \in \texttt{SetOfSingletons}$}\label{line:sos}
			\State \texttt{rewritten\_entities} $\gets$ \texttt{[]} \label{line:start:rewrite_set}
			\For{$e \in$ \texttt{prop\_node.entities}}
			\State $e$, \texttt{key\_to\_use} $\gets$ \Call{RewriteNodeLogically}{$k$, $e$, $\{\}$, \textbf{false}, \textbf{true}} \label{line:rewrite_node}
			\State \texttt{rewritten\_entities} $\gets$ \texttt{rewritten\_entities} $\cup\ [e]$
			\EndFor
			\State \texttt{prop\_node.entities} $\gets$ \texttt{prop\_node.entities} $\cup\ \{\texttt{rewritten\_entities}\}$ \label{line:end:rewrite_set}
			\If{\texttt{key\_to\_use} $\ne$ \textbf{none}}
			\State \texttt{properties[key\_to\_use]} $\gets$ \texttt{properties[key\_to\_use]} $\cup\ \newline\{\Call{CreateSetOfSingletons}{\texttt{prop\_node}, \texttt{key}}\}$ \label{line:new_set_of}
			\Else\
			\texttt{properties[key]} $\gets$ \texttt{properties[key]} $\cup\ \{\texttt{prop\_node}\}$
			\EndIf
			\EndIf
			\EndFor
			\EndFor
			
			\State $k$\texttt{.kernel.source} $\gets$ \Call{RewritePropertiesLogically}{$k$\texttt{.kernel.source}}
			\State $k$\texttt{.kernel.target} $\gets$ \Call{RewritePropertiesLogically}{$k$\texttt{.kernel.target}}
			\State $k$\texttt{.kernel.properties} $\gets$ \texttt{properties}
			
			\State \Return $k$
			\EndFunction
		\end{algorithmic}
	\end{algorithm}

	We now provide a more in-depth discussion of this algorithm. It shows how all the properties for a given kernel are rewritten and ensures recursiveness for any given kernel by ensuring that properties of properties are accounted for. Each key (typically) contains an array of entities, where $n$ is each entity in the key iterated over in Line \ref{line:loop_over_key}. If $n$ is a \texttt{Singleton}, we check that this contains a kernel and rewrite it accordingly; if the node is an \texttt{nmod} or \texttt{nmod\_poss}, then we perform an additional check to determine whether the rewritten property can be replaced within the source or target of the current kernel ($k$).
	In Line \ref{line:check_extra}, we perform a check to determine whether the key is \texttt{extra}, which signifies that the current properties have already been logically rewritten and, therefore, can be skipped. 
	In Line \ref{line:new_set_of}, we create a new \texttt{SetOfSingletons} so long as the function in Line \ref{line:rewrite_node} returns a \texttt{key\_to\_use}. This will have an ID matching \texttt{prop\_node}, and contain one element (which is \texttt{prop\_node}) and a type (which is the \texttt{key} from Line \ref{line:for_key}). For example, the sentence: ``\textit{There is traffic but not in the Newcastle city centre}'' is initially:
	
	\begin{center}
		\texttt{be(traffic, ?)[(AND:NOT(Newcastle[(extra:city centre), (6:in), (det:the)]))]}
	\end{center}
	
	Therefore, through Lines 19--23, as \texttt{(6:in)} is contained within \texttt{Newcastle}, this is rewritten as: 
	\begin{center}
		\texttt{be(traffic, ?)[(SPACE:AND(NOT(Newcastle[(extra:city centre), (type:stay in place), (det:the)])))]}
	\end{center}
	
	\begin{algorithm}[h]
		\caption{Given a node, taken from our kernel, we try to match a rule from our Parmenides ontology. From this rule we get the type which determines the rewriting function, that should be applied to the given node. This function also determines whether the rewriting should be added to the properties of the given \texttt{Singleton}, or to the entire kernel.}
		\label{alg:logical_function}
		\begin{algorithmic}[1]
			\Function{RewriteNodeLogically}{\texttt{kernel}, \texttt{initial\_node}, \texttt{properties}, \texttt{has\_nmod}, \texttt{return\_key}}
			\State \texttt{rule} $\gets$ \texttt{getLogicalRule(kernel, initial\_node, has\_nmod)}
			\State \texttt{type} $\gets$ \texttt{rule.logicalConstructName} \Comment{e.g. `space', `time'}
			\State $\texttt{property} \gets \texttt{rule.logicalConstructProperty}$ \Comment{e.g. `motion to place', `defined'}
			\State \texttt{prop\_node} $\gets$ \texttt{kernel.target} \textbf{if} \texttt{has\_nmod} \textbf{else} \texttt{initial\_node }
			\State \texttt{function} $\gets$ \texttt{getLogicalFunction(type, property)}
			
			\If{\texttt{function.attachTo} = "\texttt{Singleton}"} \Comment{where the property should be added}
			\If{\texttt{property} is not None}
			\If{\texttt{type} = "Specification"}
			\If{\texttt{property} = "inverse"}
			\State \texttt{prop\_node.properties[}"extra"\texttt{]} $\gets$ \texttt{[kernel.source]}
			\Else
			\State \texttt{prop\_node.properties[}"extra"\texttt{]} $\gets$ \texttt{[kernel.target]}
			\EndIf
			\Else
			\State \texttt{prop\_node.properties[}"type"\texttt{]} $\gets$ \texttt{type}
			\EndIf
			\EndIf
			\State \texttt{properties[type]} $\gets$ \texttt{prop\_node}
			\ElsIf{\texttt{function.attachTo} = "Kernel" \textbf{and} \texttt{property} is not None}
			\State \texttt{prop\_node.properties[}"type"\texttt{]} $\gets$ \texttt{property}
			\State \texttt{properties[type]} $\gets$ \texttt{prop\_node}
			\EndIf
			
			\If{\texttt{return\_key}}
			\State \Return \texttt{prop\_node}, \texttt{type}
			\Else
			\State \Return \texttt{properties}
			\EndIf
			\EndFunction
		\end{algorithmic}
	\end{algorithm}
	
	Algorithm \ref{alg:logical_function} then rewrites a matched kernel according to the ontology information, as per the example listed in Figure \ref{code:logical_function_space}.
	The algorithm returns properties with the (potentially) new `type key'---referring to the logical rewriting---which is then added to the kernel. The parameters passed include the entire \texttt{kernel}, the \texttt{initial\_node} (being the node to be rewritten), \texttt{properties} (which is the set of properties to be added to the entire kernel), \texttt{has\_nmod} (a Boolean from Line \ref{line:has_nmod} in Algorithm \ref{alg:rewrite_props}), and \texttt{return\_key} (used for rewriting \texttt{SetOfSingletons}). Additional rules are also included to deal with swapping the target with the properties, as discussed in more detail in Section \ref{sub:incorrect_sv_rel}.	
	
	\section{Ad Hoc Pipeline Sub-Routines}
	This section provides more in-depth details over specific sub-routines for the \textit{ad hoc}  phase.
	
	\subsection{Initial Graph Construction}\label{sec:igc}
	This step builds an initial graph representation of the sentence based on the syntactic relationships extracted with Stanford CoreNLP through dependency parsing (Supplement \ref{nlpsec}).
	Each word in the sentence is represented as a node in the graph, and the syntactic relationships between words (e.g., \glspl*{ud}) are represented as edges connecting these nodes. This graph provides a structured representation of the sentence's syntactic structure, which is then rewritten as described in the following subsection. 
	Crucially, this process identifies \glspl*{ud}, which are essential for our final logical sentence analysis (Section \ref{sec:lsa}), as they determine the kernel of our sentence and collect additional information associated with it.
	
	\subsection{Details for Algorithm \ref{alg:overall_kernel_construction}}\label{algo2det}
	
	If the relationship enclosed in this contains a \textit{semi-modal verb} \cite{semimodal} in its label, then we check the following: 
	\begin{itemize}
		\item If it does have a kernel as a property, and this kernel's position is contained within our position pairs (collected on Line \ref{line:get_position_pairs}), then we update the target of our kernel to this property to make an entire kernel describing the action associated with the subject, introduced by the semi-modal verb.
		\item 	Otherwise, we check if the edge label is \textbf{none}, and whether the source is an adjective and the target is an entity, or vice versa. If so, we update the entity with the properties of the adjective. For example, the text ``\textit{clear your vision}'' is initially rewritten as \texttt{None(clear, vision)} and is transformed into \texttt{be(vision[JJ: clear], ?)} based on this condition.
	\end{itemize}
	
	We now discuss the last post-processing operations (Lines \ref{line:acl_rep}-\ref{line:check_for_adv}):
	
	\begin{description}
		\item[Line \ref{line:acl_rep}:] It replaces any occurrence of an \texttt{acl\_relcl} edge within the properties, where the source ID is contained within the \texttt{acl\_map}, appended to on Line \ref{line:append_acl} in Algorithm \ref{alg:create_sentence_func}, and thus replaces the node associated in the map. 
		This enables the replacement of any pronoun with the exact entity it is referring to: as our pipeline retains provenance information, this does not come at the cost of losing any information under the circumstance that there are multiple instances of the same entity. We discuss how this has changed from our previous pipeline in Section \ref{sub:improve_prev}.
		\item[Line \ref{line:check_action}:] It checks if we do not have a relationship (where \texttt{kernel} is \textbf{none}) and rewrites this into an \textit{edge kernel} (Algorithm \ref{alg:create_sentence_func}). An edge kernel is where a node of type verb is rewritten to an edge label with no source or target. For example, if we had the node `work' with no other edges, then we return the following kernel: \texttt{work(?, None)}. Otherwise, we deal with verbs that were not rewritten as an edge in our \gls*{ggg} phase and, due to the grammatical structure, were represented as an entity property: \texttt{action} (or \texttt{actioned}) indicates that the entity performs (or receives) the action indicated in \texttt{action} (or \texttt{actioned}).  If we have a node with \texttt{action} (or \texttt{actioned}), it becomes the source (or target) of the relationship.
		\item[Line \ref{line:duplicate_props}:] It removes duplicated properties occurring in both relationship arguments and their properties: if an entity is contained within the relationship's additional properties but is present as either the source or the target of the relationship, then it is removed from the properties. This is performed recursively for all kernels. 
		\item[Line \ref{line:rewrite_properties}:] It rewrites the properties within each kernel as their logical functions. Section \ref{sec:lsa} discusses this important phase in more detail.
		\item[Line \ref{line:check_for_adv}:] Lastly, we deal with phrasal verbs with their adverb separated from the edge relationship name and occurring within the relationship property.  If one is found, it is appended to the end of the edge label, which works for some cases, such as `\textit{come back}', which is initially \texttt{come(?[adv:back], None)}, and therefore becomes \texttt{come back(?, None)}. However, it can produce some grammatically incorrect edge label names. For `\textit{how to use it}', we obtain `\texttt{to use(?[adv:how], it)}', which should have ``\textit{how}'' appended to the beginning of the edge label, but currently, we obtain `\texttt{to use how(?, it)}'. This is easily fixed by considering whether the resulting edge belongs to a phrasal verb, and only under such circumstances can we retain such a change. 
	\end{description}
	
	\section{Evaluation Details}
	This section provides additional details referring to the metrics used for either clustering or classification purposes.
	
	\subsection{Clustering Algorithms of Choice}\label{supp:algosclust}
	\gls*{ahc} \cite{10.1093/comjnl/20.4.364} is a hierarchical clustering method that uses a `bottom-up' approach, starting with creating the clusters from the data points with the shortest distance. In considering each data point as a separate cluster, the data are iteratively merged into the shortest-distance pair of clusters according to a given linkage criterion until we reach the number of the desired clusters, at which point the clustering algorithm will stop \cite{Nielsen2016}. 
	
	$k$-Medoids \cite{kmedoids} is a variation of $k$-Means Clustering where, as a first step, the algorithm starts by picking actual data points as cluster centres, suitable for supporting distance metrics by design. 
	In setting these initial medoids as the centres of their clusters, each point is assigned to the cluster identified by the nearest medoid. At each step, we might consider swapping a medoid with another node if this minimises the clustering assignment costs. The iteration converges after a maximum number of steps or when we reach stability, i.e., no further clusters are updated. As this approach also minimises a sum of pairwise dissimilarities rather than a sum of squared Euclidean distances like $k$-Means, $k$-Medoids  is more robust to noise and outliers than $k$-Means.

	\gls*{ahc} is implemented in \texttt{scikit-learn} v1.6.0 (\url{https://scikit-learn.org/stable/whats\_new/v1.6.html\#version-1-6-0}, Accessed on 29 March 2025); however, given that this library does not implement $k$-Medoids, we used \\\texttt{scikit-learn-extra} v0.3.0 (\url{https://scikit-learn-extra.readthedocs.io/en/stable/changelog.html\#unreleased}, Accessed on 29 March 2025). 
	
	\subsection{Clustering Metrics}\label{supp:clusteringmtric}
	{\textbf{Silhouette} score \cite{ROUSSEEUW198753} measures intra-cluster variance  by taking into account both cohesion (how close the object is to others in the same cluster) and separation (how far it is from the nearest neighbouring cluster). Given the average distance between a point and all other points in the same cluster, $a$, and the average distance between that point and all points in the nearest neighbouring cluster, $b$, we define the score as}
	\[\text{Silhouette Score}=\frac{b-a}{\max{(a, b)}}\]
	
	As this score does not take into account the alignment between the mined clusters and the expected ones, usually referred to as classes, this score is referred to as an \textit{internal} measure. The following metrics are therefore \textbf{external}, thus providing a numerical score for the aforementioned comparison.
	
	{\textbf{Purity} \cite{Manning_Raghavan} measures the extent to which clusters contain only items belonging to one single expected cluster. This is computed as follows: for each cluster, collect the number of data points from the most common expected cluster, sum over all clusters, and divide by the total number of data points (\url{https://medium.com/@vincydesy96/evaluation-of-supervised-clustering-purity-from-scratch-3ce42e1491b1}, Accessed on 22 April 2025). Given the mined clusters $M$ and the expected ones $D$, both partitioning $N$ data points, purity is defined as}
	\[\frac{1}{N} \sum_{m \in M} \max_{d \in D} |m \cap d|\]
	
	Thus, while $1$ indicates that all clusters contain only objects from one class, $0$ indicates the opposite.

	{Similarly to the alignment score, the \textbf{\gls*{ari}} \cite{DBLP:journals/jmlr/NguyenEB10} quantifies the similarity between true and predicted clusters.  To calculate the \gls*{ari}, we must first calculate the \gls*{ri}: Given that $N$ is the number of samples, let $C_1$ and $C_2$ represent two clustering assignments. This is formulated as the ratio $RI$ between the total number of agreements between $C_1$ and $C_2$ over the number of all possible pairs $\binom{N}{2}$. Then, the expected value, $E$, for the \gls*{ri} is computed as $E=\frac{\sum { \binom{n_i}{2}} \times \sum {\binom{m_j}{2}}}{\binom{N}{2}}$, where $n_i$ and $m_j$ are the sizes of clusters $C1$ and $C2$, respectively.} {Thus, the \gls*{ari} is computed as:}
	\[ARI=\frac{RI-E}{1-E}\]
	
	{Here, $1$ indicates perfect agreement, $0$ suggests a random one, and negative values suggest less-than-random performance, meaning it is sub-optimal. 
		We used the implementation of the \gls*{ari} and Silhouette score in \texttt{scikit-learn} v1.6.0 (\url{https://scikit-learn.org/stable/whats\_new/v1.6.html\#version-1-6-0}, Accessed on 22 April 2025)}. 

	\subsection{Classification Metrics}\label{supp:classmetric}
	
	\textit{In this section, we first introduce all the metrics used for the classification task in terms of the binary classification problem and then generalise this to an n-ary problem, per the scope of this paper.} 
	
	\textbf{Accuracy} is the proportion of all classifications that are correct, whether positive or negative, and is defined as:
	\[\text{Accuracy} =
	\frac{\text{correct classifications}}{\text{total classifications}}
	= \frac{TP+TN}{TP+TN+FP+FN}\] 
	{where TP (or TN) stands for the True Positives (or Negatives), and FP (FN) stands for the False Positives (or Negatives).}
	
	\textbf{Precision}, for a binary classification problem, is the proportion of all positive classifications that are actually positive, defined as:
	\[\text{Precision} =
	\frac{\text{correctly classified actual positives}}
	{\text{everything classified as positive}}
	= \frac{TP}{TP+FP}\] 
	
	\textbf{Recall} for a binary classification problem is the proportion of all actual positives classified correctly as positives, defined as:
	\[\text{Recall} =
	\frac{\text{correctly classified actual positives}}{\text{all actual positives}}
	= \frac{TP}{TP+FN}\] 
	
	The \textbf{F1} score for binary classification is the harmonic mean of precision and recall, balancing the two metrics in a single number, defined as:
	\[\text{F1 Score} = 2\times\frac{\text{Precision}\times\text{Recall}}{\text{Precision}+\text{Recall}}\]
	
	{In our experiments, for a multiclass classification problem, we employ two different averaging techniques to calculate the precision, recall, and F1 score: macro and weighted. The macro average is calculated using the unweighted mean, potentially impacting the model if the performance in minority classes is subpar. The weighted average weights each measure by the frequency of the occurrence of the class within the testing dataset. As this takes into account the number of true instances in each class to cope with class imbalance, it usually favours the majority class \cite{averages}. }

	{Our problem is a multi-classification problem; therefore, we average across the precision for each class (\url{https://www.evidentlyai.com/classification-metrics/multi-class-metrics}, Accessed on 22 April 2025):}
	\[\text{Precision}_{Macro~Average} =
	\frac{\text{Precision}_{Class~A}+\text{Precision}_{Class~B}+\ldots+\text{Precision}_{Class~N}}
	{N}\]

	
	{Similarly to precision, we must average across all $N$ classes to obtain the recall for a multi-classification problem:
		\[\text{Recall}_{Macro/Weighted~Average} =
		\frac{\text{Recall}_{Class~A}+\text{Recall}_{Class~B}+\ldots+\text{Recall}_{Class~N}}{N}\]}
	
	{To calculate the F1 score for a multi-classification problem, we take the precision and recall for each class  and average all F1 scores over $N$ classes (\url{https://www.baeldung.com/cs/multi-class-f1-score}, Accessed on 22 April 2025):
		\[\text{F1 Score}_{Macro/Weighted~Average} =
		\frac{\text{F1 Score}_{Class~A}+\text{F1 Score}_{Class~B}+\ldots+\text{F1 Score}_{Class~N}}{N}\]}
	
	We implemented all of the above metrics using \texttt{scikit-learn} v1.6.0 (\url{https://scikit-learn.org/stable/whats\_new/v1.6.html\#version-1-6-0}, Accessed on 22 April 2025). 
	
	\section{Additional Ex Post Similarity Metrics}\label{sup:add-ex-post}
	\subsection{Sentence Transformers}\label{sec:fulltexcos}
	Vector-based similarity systems most commonly use \textbf{cosine similarity}  for expressing similarities for vectors expressing semantic notions \cite{DBLP:conf/iccsci/JatnikaBS19,DBLP:journals/corr/abs-1301-3781}, as two almost-parallel normalised vectors will lead to a near-one value, while extremely dissimilar values lead to negative values \cite{ROSENBERGER2025127043}. This induces the possibility of seeing zero as a threshold boundary for separating similar from dissimilar data. This notion is also applied when vectors represent hierarchical information \cite{DBLP:journals/cai/LiuBX11} with some notable exceptions \cite{DBLP:conf/nips/NickelK17}. Given $A$ and $B$ are vector representations (i.e., embeddings) from a transformer $\tau$ for sentences $\alpha$ and $\beta$, this is $\mathcal{S}_c(A,B)=\frac{A\cdot{B}}{\left\|{A}\right\|\left\|{B}\right\|}$.
	Still, a proper similarity metric should return non-negative values \cite{logicalLearning}. Given the former considerations, we can consider only values above zero as relevant and return zero otherwise, thus obtaining:

	\begin{equation}\label{eq:normcos}
		\mathcal{S}_c^+(A,B)=ReLU(\mathcal{S}_c(A,B))=\max\{\mathcal{S}_c(A,B),0\}
	\end{equation}
	
	Different transformers generate different vectors, automatically leading to different similarity scores for the same pair of sentences.

	\subsection{Neural \gls*{ir}}\label{neural-ir}
	When considering neural \gls*{ir} approaches, we require an extra loading phase, where all the sentences within the datasets are treated as the corpora of documents $\mathcal{D}$ to be considered. Then, the documents are indexed through their associated vectors. In this scenario, we also consider the same sentences as the queries of interest. As ColBERTv2 yields $A=\tau(\alpha)$, which is a set of vectors for a given sentence $\alpha$, the authors defined the ranking score as $\mathcal{S}_{\textup{nir}}(A,B)=\sum_{\vec{u}\in A}\max_{\vec{v}\in B}\mathcal{S}_c(\vec{u},\vec{v})$, which is not necessarily normalised between $0$ and $1$. 
	We now consider the following normalisation:
	
	\[\mathcal{S}_{\textup{nir}}^+(A,B)=\frac{\mathcal{S}_{\textup{nir}}(A,B)-m}{M-m}\]
	where $m$ and $M$, respectively, denote the minimum  ($\min_{\alpha,\beta\in\mathcal{D}}\mathcal{S}_\textup{nir}(\tau(\alpha),\tau(\beta))$) and the maximum \\($\max_{\alpha,\beta\in\mathcal{D}}\mathcal{S}_\textup{nir}(\tau(\alpha),\tau(\beta))$) query-document alignment score.
	
	\subsection{Generative \acrfull*{llm}}\label{debertaprocessing}
	When considering classifiers such as DeBERTaV2+AMR-LDA based on a generative \gls*{llm}, we express the classification for the sentence pair $\alpha$ and $\beta$ as ``$\alpha\textrm{. }\beta\textrm{.}$'', which is then used in the classification task. Any other unexpected representation of sentences may lead to misleading classification results; for example, changing the prompts to ``if $\alpha$ then $\beta$'' leads to completely wrong results. This returns a confidence score per predicted class $k$: $\mathbb{P}(k|\alpha\textrm{. }\beta\textrm{.})$. Thus, the class predicted is the one associated with the highest score; that is, $\argmax_k\mathbb{P}(k|\alpha\textrm{. }\beta\textrm{.})$. As the representation of interest only classifies logical entailment\footnote{See Appendix A.2 for further details on this notation.} ($\mstar$) or indifference ($\omega$), and given that all former approaches work under the assumption that the same given score alone can be used to determine the similarity of two sentences, we map the predicted score for the logical entailment class between 0.5 and 1. In contrast, we map the indifference score between 0 and 0.5. 
	
	\[\mathcal{S}_{\textrm{glm}}(\alpha,\beta)=\begin{cases}
		\frac{\mathbb{P}(\mstar|\alpha\textrm{. }\beta\textrm{.})}{2}+0.5 & \mstar=\argmax_{k}\mathbb{P}(k|\alpha\textrm{. }\beta\textrm{.})\\
		\frac{\mathbb{P}(\omega|\alpha\textrm{. }\beta\textrm{.})}{2} & \textup{oth.}
	\end{cases}\]
	
	\subsection{Simple Graphs (SGs) vs. Logical Graphs (LGs)}\label{sec:galignment}
	Given that our graphs of interest can be expressed as a collection of labelled edges, we reduce our argument to edge matching \cite{DBLP:journals/dpd/VirgilioMT15}. Given an edge distance function $\epsilon$, an edge $e$, and a set of edges $A$ obtained from the pipeline as a transformation of the sentence, the best match for $e$ is an edge $e'\in A$ minimising the distance $\epsilon$, i.e., $m_\epsilon(e,A)=\argmin_{e'\in A}\epsilon(e,e')$. We can then express the best matches of edges in $A$ over another set $A'$ as a set of matched edge pairs $M_\epsilon(A,A')=\{(e,m(e,A'))|e\in A\}$. Then, we denote $D_\epsilon(A,A')$ as the set of edges not participating in any match. The matching distance between two edge sets shall then consider both the sum of the distances of the matching edges as well as the number of the unmatched edges \cite{logicalLearning}. Given an edge-based representation $A$ and $B$ for two sentences $\alpha$ and $\beta$ generated like in Section \ref{gsmrewr}, we derive the following edge similarity metric as the basis of any subsequent graph-based matching metric:
	
	\begin{equation}\label{commonEq}
		\mathcal{S}_g^\epsilon(A,B)=\left(1-\frac{\epsilon(M_\epsilon(A,B))}{|A|}\right)\cdot D_\epsilon(A,B)\vert^s_N
	\end{equation}
	
	Given a node representing a \texttt{(SetOf)Singleton(s)} $\nu$, an edge label $\varepsilon$, and normalised similarity metric ignoring the negation information, we refine  $\epsilon$ from Eq.~\ref{commonEq} by conjoining the similarity among the edges' sources and targets, while considering  edge label information. We annihilate such  similarity if the negations associated to the edges do not match by multiplying such similarity by 0; then, we negate the result for transforming this similarity into a distance:
	
	\begin{equation}\label{usingNu}
		\epsilon_{\nu,\varepsilon}((s,t),(s',t'))=\begin{cases}
			\nu(s,s')\nu(t,t')\varepsilon(\lambda(s,t),\lambda(s,'t')) & \textsf{neg}(\lambda(s,t))= \textsf{neg}(\lambda(s',t'))\\
			0 & \textup{oth.}\\
		\end{cases}
	\end{equation}
	where $(s,t)$ represents an edge, and $\lambda(s,t) $, its associated label. This metric can be instantiated in different ways for simplistic graphs 
	and \glspl*{lg} using a suitable definition for $\nu$ and $\varepsilon$.
	
	\paragraph*{Simple Graphs (SGs)} For graphs, all \texttt{SetOfSingletons} are flattened to \texttt{Singletons}, including the nodes containing information related to logical operators. In these cases, we use $\nu$ and $\epsilon$ as $\mathcal{S}_c^+$ from Eq. \ref{eq:normcos}. At this stage, we still have a symmetric measure.
	
	\paragraph*{Logical Graphs (LGs)} 
	{We introduce notation from \cite{logicalLearning} to guarantee the soundness of the normalisation of distance metrics: we denote $d\vert_N=\frac{d}{d+1}$ the normalisation of a distance value between 0 and 1, and $d\vert_N^s=1-d\vert_N$ its straightforward conversion to a similarity score.}
	
	Now, we extend the definition of $\nu$ from Eq.~\ref{usingNu} as a similarity $\nu'(u,v):=\delta_\nu(u,v)|^s_N$ where $\delta_\nu$ is the associated distance function defined in Eq.~\ref{refEq}, where we leverage the logical structure of \texttt{SetOfSingletons}. We approximate the confidence metric via an asymmetric node-based distance derived using fuzzy logic metrics with matching metrics for score maximisation. We return the maximum distance 1 for all the cases when one logical operator cannot necessarily entail the other.
	\begin{equation}\label{refEq}
		\delta_\nu(u,v)=\begin{cases}
			1-\nu(u,v) & \textsf{singleton}(u),\textsf{singleton}(v)\\
			\delta_\nu(u,m_{\delta_\nu}(u,v)) & u\equiv\wedge_i x_i,\textsf{singleton}(v)\\
			1-\delta_\nu(x,v) & u\equiv\neg x,\textsf{singleton}(v)\\
			1-\delta_\nu(u,y) & \textsf{singleton}(u),v\equiv\neg y\\
			\delta_\nu(u,m_{\delta_\nu}(u,v)) & \textsf{singleton}(u),v\equiv\vee_i y_i\\
			\delta_\nu(x,y) & u=\neg x,v=\neg y\\
			\textup{avg}\, \delta_\nu(M_{\delta_\nu}(u,v)) & u=\wedge_i x_i,v=\wedge_i y_i,|M_{\delta_\nu}(u,v)|=|u|\\
			\textup{avg}\, \delta_\nu(M_{\delta_\nu}(u,v)) & u=\wedge_i x_i,v=\vee_i y_i\\
			1-E_{\delta_\nu}(u,v) & u=\vee_i x_i,v=\vee_i y_i\\
			1 & \textup{oth.}
		\end{cases}
	\end{equation}

	\section{Pipeline Considerations}\label{supp:consider}
	
	We now discuss some preliminary results that we obtained by analysing the 200 real-world sentences benchmarked in the last experiment (Appendix C), while sentences explicitly mentioning Newcastle and traffic scenarios were derived from our Spatiotemporal reasoning dataset (\figurename~\ref{subf:sentences_newc}). For the former dataset, we were able to accurately describe about 80\% of the sentences in logical format, thus improving over our previous implementation, and the associated differences are detailed below.
	
	\subsection{Handling Incorrect Verbs} 
	We cannot rely on StanfordNLP's dependency parsing, as it may fail to distinguish the proper interpretation of a sentence \cite{DBLP:conf/emnlp/ChenM14}, severely impacting our representation when verbs are incorrectly identified. 
	
	\begin{figure}[p]
		\centering
		\subfloat[In the sentence ``\textit{emilia's short tailed opossum}''{, we show how ``\textit{tailed}'' (underlined) is initially identified as a verb from our StanfordNLP input graph and, thus,  is transformed into an edge (underlined) in our result graph through \gls*{ggg} rewriting.}]{
			\label{fig:opossum}
			\begin{minipage}{0.42\textwidth}
				\centering
				\includegraphics[width=\linewidth,height=8cm,keepaspectratio]{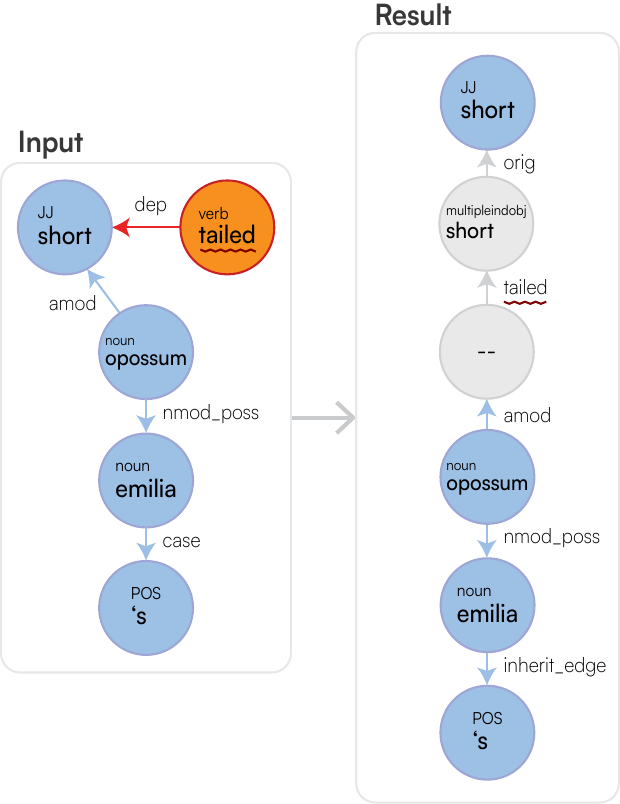}
			\end{minipage}
		}
		\hfill
		\subfloat[{In the sentence }``\textit{you are visiting Griffith Park Observatory}''{, ``\textit{Park}'' is correctly identified as a noun from StanfordNLP; however, this was identified as a verb in our \gls*{meudb}. Therefore, we ensure through specified rules when to use StanfordNLP's type over our \gls*{meudb}.}]{
			\label{fig:park}
			\begin{minipage}{0.6\textwidth}
				\centering
				\includegraphics[width=\linewidth,height=5cm,keepaspectratio]{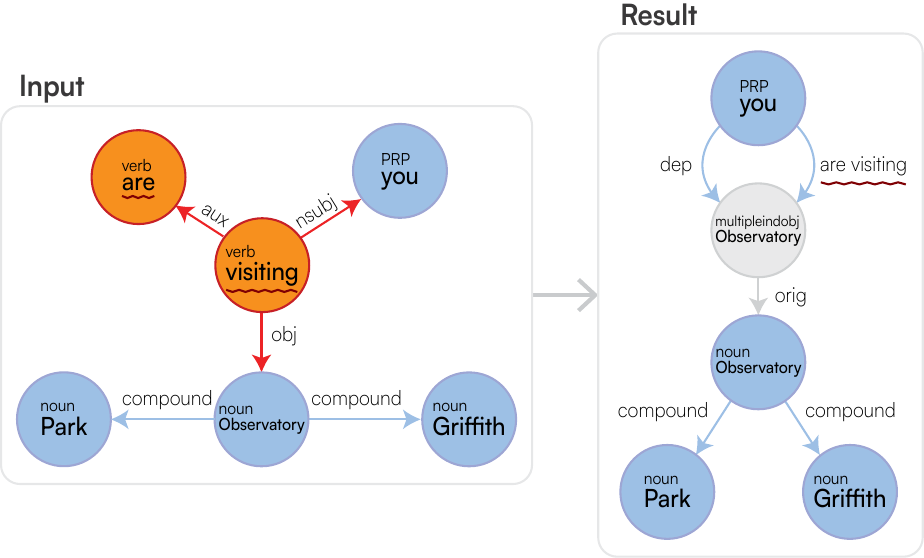}
			\end{minipage}
		}
		\vspace{10pt}
		
		\subfloat[{In the sentence }``\textit{attend an interview}''{, ``\textit{interview}'' is correctly identified as a noun from StanfordNLP; however, this was identified as a verb in our \gls*{meudb}. Therefore, we ensure through specified rules when to use StanfordNLP's type over our \gls*{meudb}.}]{
			\label{fig:interview}
			\begin{minipage}{\textwidth}
				\centering
				\includegraphics[width=\linewidth,height=6cm,keepaspectratio]{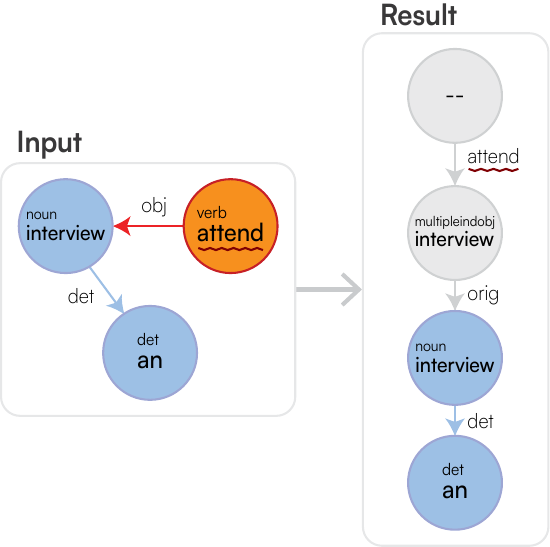}
			\end{minipage}
		}
		
		\vspace{20pt}
		
		\begin{minipage}{0.65\textwidth}
			\centering
			\subfloat[Legend of symbols used for the images given above.]{
				\label{fig:node_key}
				\centering
				\includegraphics[width=\linewidth,height=8cm,keepaspectratio]{images/discussion/key.pdf}
			}
		\end{minipage}
		\caption{Graph inputs (StanfordNLP) and outputs ({after} \gls*{ggg} rewriting) for three {example} sentences{, which are discussed within the captions and in more detail within the text.}}
		\label{fig:verb_graphs}
	\end{figure}
	
	In \figurename~\ref{fig:opossum}, ``\textit{short}'' and ``\textit{tailed}'' should both be identified as \texttt{amod}'s of \texttt{opossum}; however, we instead get ``\textit{tailed}'' as a verb, rewritten as an edge towards the adjective short. Therefore, given this interpretation, LaSSI rewrites the sentence to: 
	
	\begin{center}
		\texttt{be(opossum[(extra:emilia[(2:'s)])], ?2)[(SENTENCE:tail(?1[(cop:short)], None))]}    
	\end{center}
	
	We can see that, due to ``\textit{tailed}'' being identified as a verb, a kernel that is incorrect  is created. However, given the interpretation from the parser, this is the best rewriting we can produce, as we currently do not have a way to differentiate whether ``\textit{tailed}'' should be a verb or not. Thanks to our explainability part, we are able to backtrack through the stages of the pipeline to identify that the dependency parsing phase is causing this problem, thus allowing a relevant analysis to be carried out. Given this, we return this partially correct representation of the sentence in logical form:
	\[\left(\textit{tail}(\lozenge \left[\textsf{opossum} [\textup{of}] \textit{emilia}\right]) \wedge \textit{be}(\lozenge \left[\textsf{opossum} [\textup{of}] \textit{emilia}\right])\right)\]

	There are also occasions where our \gls*{meudb} might produce incorrect identifications of verbs; for example, in \figurename~\ref{fig:park}, we can see that ``\textit{Park}'' is \textit{correctly} identified as a noun from StanfordNLP while, in our Parmenides ontology, it is labelled as a VERB with the highest confidence of \texttt{1.0}. Due to our type hierarchy (see \appendixname~\ref{typehier}), this is chosen as being the correct type. Therefore, we have derived rules from a syntactic point of view in order to ensure that, when a verb type is selected from the \gls*{meudb}, it is certainly a verb. These rules are as follows:
	\begin{enumerate}[]
		\item A given node does not contain a \texttt{det} property \textbf{AND};
		\item A given node does not contain `on' within a found \texttt{case} \cite{case_ud} property \textbf{AND};
		\item One of the following conditions is met:
		\begin{enumerate}
			\item A given node has at least one incoming edge \textbf{AND} has a \texttt{case} property;
			\item A given node is the last occurring in the graph \textbf{AND} has a \texttt{root} property;
			\item A given node is the last occurring in the graph \textbf{AND} has a parent with a \texttt{root} property which is connected by a compound edge.
		\end{enumerate}
	\end{enumerate}
	Therefore, for the stated example in \figurename~\ref{fig:park}, ``\textit{Park}'' cannot be a verb as the parent does \textbf{not} have a root property attached to it. In another case, \figurename~\ref{fig:interview} shows the correct identification of ``\textit{interview}'' but, again, the \gls*{meudb} identified it as a verb; however, as the node has a \texttt{det} property attached to it, we ignore the resolution and keep it as a noun.
	
	\subsection{Handling Incorrect Subject--Verb Relationships}\label{sub:incorrect_sv_rel} While creating our kernel (detailed in Algorithm \ref{alg:create_sentence_func}), we apply \textbf{post-processing} on Line \ref{line:post_processing}. In this way, we check whether the created kernel's target is an adjective, as well as whether a pronoun or entity is contained within the properties. If so, we swap these nodes around to better reflect the sentence representation. This happens as the mined \glspl*{ud} might wrongly refer a subject to an adjective, rather than to the main sentence verb. 
	
	\begin{figure}[H]
		\centering
		\includegraphics[width=0.5\linewidth]{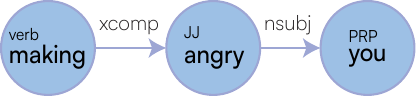}
		\caption{Input graph from StanfordNLP for the sentence: ``\textit{making you angry}''.}
		\label{fig:make_angry}
	\end{figure}
	
	For example, the text ``\textit{making you angry}'' is initially rewritten as: \texttt{make(?, angry)}\\\texttt{[PRONOUN: you]} at this stage. In \figurename~\ref{fig:make_angry} we can see that the verb ``\textit{making}'' is referring to ``\textit{angry}'', and ``\textit{angry}'' has an \texttt{nsubj} relation to ``\textit{you}'', which is incorrect. Therefore, our post-processing fixes this such that ``\textit{angry}'' and ``\textit{you}'' are swapped to become: \texttt{make(?, you)[JJ: angry]}. This is then represented in logical form as follows:
	
	\[\exists ?1.\,\textit{make}_{\texttt{JJ}: \lozenge {\textsf{angry}}}(\lozenge {\textsf{?1}},\lozenge {\textsf{you}})\]
	
	\subsection{Handling Incorrect Spelling (Typos)} \label{supp:handleincspell} 
	Some of the words within the ConceptNet dataset are misspelt; for example: ``\textit{be on the interne}''. We can assume that the last word in this sentence should be ``\textit{internet}'', which is found within the \gls*{meudb}. However, it has a confidence value that matches other suggestions for replacing this word. In another example, ``\textit{a capatin}'', we could assume that this should be corrected to ``\textit{captain}''; however, this term is not included in the \gls*{meudb}. Without further information, we could not be sure whether any other matches are correct for replacement of the misspelt term within the sentence. At present, we do not have a method to identify words to be replaced with a possible match. However, we plan to develop such an approach by demultiplexing each sentence representation for any ambiguous text interpretation. Future works will also attempt to carry out probabilistic reasoning to determine the most suitable interpretation given the context of the sentence; this would require us to retain more complex information and semantics from the sentence, which would then be used to resolve the best matching entity depending on the context.

	\section{Improvements From Our Previous Solution}\label{sub:improve_prev} 
	This section outlines the major improvements from our previous solution \cite{ideas2024b}.
	
	\subsection{Extensions to \acrfull*{ggg} Language and Graph Grammar Rewriting Rules for \acrfullpl*{ud}}\label{sub:ex_ggg}

	To improve the final logical rewriting, the \gls*{ggg} language demonstrated in \cite{math12172677} was extended:  nodes can now copy properties from others and support better node variable matching and updates. Some \glspl*{ud} \cite{ud} that were not accounted for previously are now considered in our graph grammar rewriting rules: \texttt{parataxis} and \texttt{compound\_prt}. Furthermore, there was an issue that if an entity contained multiple \texttt{case} properties, they would not all appear in the node's properties, due to the properties being key--value associations, where the key can only take one value. To overcome this, the \texttt{case} property keys are labelled as their float positions occurring within the sentence. For example, for the sentence ``\textit{It is busy in Newcastle}'', our \texttt{case} property ``\textit{in}'' is at the 4th position, so ``\textit{Newcastle}'' would have a property \texttt{\{4:in\}}. This also means that more information is retained within the rewriting, as we can identify whether this property precedes or proceeds the given entity, which is not pertinent at the moment but how it will improve the pipeline in the future is discussed in Section \ref{sec:conclusion}.
	
	\subsection{Improving Multi-Word Entity Recognition with Specifications}\label{sec:griff}
	In our previous solution \cite{ideas2024b}, the most recently resolved entity was chosen, meaning that it could not account for scenarios where multiple resolutions with equal confidence occur.  Algorithm \ref{alg:merge_singletons} rectifies this: on each possible powerset of the \texttt{SetOfSingletons} \texttt{entities} (Line \ref{line:loop_alternatives}), we check if the current item score is greater than anything previously addressed (Line \ref{line:should_add_alternative}) and, if so, we use this resolution in place of the former. Furthermore, if we have multiple resolutions with the same score, we choose the one with the greatest number of entities (Line \ref{line:get_most_entities}). For example, the sentence ``\textit{you are visiting Griffith Park Observatory}'' contains a \texttt{SetOfSingletons} with entities \texttt{Griffith}, \texttt{Park}, and \texttt{Observatory}. Two resolutions with equal confidence scores are resolved, the first being \texttt{Observatory}, with an \texttt{extra} \texttt{Griffith Park}, and the second: \texttt{Griffith Park Observatory}. Thus, based on Line \ref{line:get_most_entities}, we use the latter as our resolved entity. This is a preliminary measure, which may change in the future with further testing; however, our current experiments demonstrate this to be a valid solution.

	\subsection{Characterising Entities by Their Logical Function}
	In our previous implementation, we did not categorise entities by their logical function but, instead, only by their type. This can be seen in the sentence above, with ``\textit{Saturdays}'' being defined as a \texttt{DATE}, rather than \texttt{TIME} in our new pipeline. However, not only have we made improvements to the rewriting process concerning spatio-temporal information, we  now also handle specifications such as causation and aims (to mention a few). For example, the sentence ``\textit{The Newcastle city centre is closed for traffic}'' was previously rewritten to:

	\begin{center}
		\texttt{close(traffic[(case:for)], Newcastle city centre[(amod:busy), (det:The)])}
	\end{center}

	In contrast, we now include a further specification, implementing \texttt{AIM\_OBJECTIVE} to demonstrate specifically \textit{what} is closing the Newcastle city centre: 
	\[\exists ?1.\,\textit{close}_{\texttt{AIM\_OBJECTIVE}: \lozenge {\textsf{traffic}}}(\lozenge {\textsf{?1}},\lozenge \left[\textsf{Newcastle} [\textup{of}] \textit{city centre}\right]^{\texttt{None}})\]
	
	This also relates to a better recognition of whether entities occurring within relationships' properties actually had to be considered the subject or direct object of a sentence. 
	For the sentence ``\textit{you are visiting Griffith Park Observatory}'', this would have been \texttt{visit(you, None)[(GPE:Observatory[(extra:Griffith Park)])]}, but has been improved to become \texttt{visit(you, Griffith Park Observatory)}, which is then rendered in logical form as follows:
	
	\[\textit{visit}(\lozenge {\textsf{you}},\lozenge {\textsf{Griffith Park Observatory}})\]

	As a side note, the correct identification of multi-word entities with their specification counterpart (\texttt{Griffith Park Observatory} rather than \texttt{Observatory}, with \texttt{Griffith Park} as an extra) is a direct consequence of the discussion in Section \ref{sec:griff}.

	This enhances the contextual information of the sentence, thanks to the use of Algorithm \ref{alg:rewrite_props}.
	
	\subsection{Dealing with Multiple Distinct Logical Functions}
	To fully capture the semantic and structural information from the sentences, our pipeline needed to be extended to include rules for scenarios that did not arise in our initial investigation. In Section \ref{subsub:setofsingletons}, we discussed the implementation of \texttt{multipleindobj}, which was not present beforehand.
	
	\begin{figure}[H]
		\centering
		\includegraphics[width=0.6\linewidth]{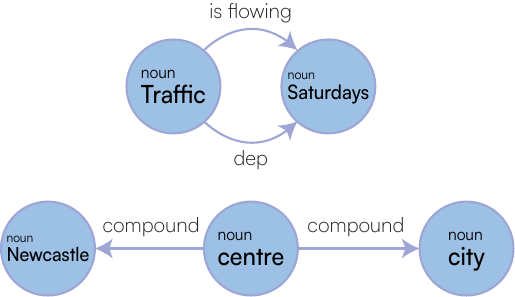}
		\caption{\gls*{ggg} output from our old pipeline \cite{ideas2024b} for ``\textit{Traffic is flowing in Newcastle city centre, on Saturdays}'', where \texttt{multipleindobj} is missing from the graph. This has been added into our new pipeline to reify n-ary relationships into a \texttt{SetOfSingletons} (\figurename~\ref{fig:multipleindobj}) and potentially merge different patterns being matched within the graph.}
		\label{fig:old_multi}
	\end{figure}
	
	Examining \figurename~\ref{fig:old_multi}, we see no \texttt{multipleindobj} type present in the graph, and the structure is fundamentally flawed compared to our amended output in \figurename~\ref{fig:multipleindobj}, as it does not capture the n-ary relationships that we need for rewriting.  Our old pipeline rewrote this to \texttt{flow in(Traffic, None)[DATE:Saturdays]}, from which  we can see that ``\textit{Newcastle city centre}'' is missing, due to the two separated connected components. On the other hand, we now return:
	
	\begin{center}
		\texttt{flow(Traffic, None)[(TIME:Saturdays[(type:defined)]), (SPACE:Newcastle[(type:stay in place), (extra:city centre)])]}
	\end{center}

	This is a sufficiently accurate representation of the given full text, leading to the following logical representation:
	\[\textit{flow}_{\texttt{SPACE}: \lozenge \left[\textsf{Newcastle} [\textup{of}] \textit{city centre}\right],\; \texttt{TIME}: \lozenge {\textsf{Saturdays}}}(\lozenge {\textsf{traffic}})\]
	
	\subsection{Pronoun Resolution}\label{sup:pronres}
	Pronouns were also not handled previously, meaning that sentences were not properly resolved, thus leading to an output that does not accurately represent the semantics of the full text. For example, ``\textit{Music that is not classical}'' was originally represented as \texttt{nsubj(classical, that)[(GPE:Music)]}, both not interpreting the pronoun \textit{and} mis-categorising ``\texttt{Music}'' as a \texttt{GPE}. This is ameliorated in the new pipeline, as discussed with respect to the post-processing part of Algorithm \ref{alg:overall_kernel_construction}, with the above example now represented as: \texttt{be(Music$^0$, ?1)[(SENTENCE:be(Music$^0$, NOT(classical$^6$)))]}. For the sentence ``\textit{Music that is not classical}'', the relationship passed into Line \ref{line:acl_rep} is: 
	
	\begin{center}
		\footnotesize
		\texttt{be(Music$^0$, ?)[(SENTENCE:be(that$^5$, NOT(classical$^6$)$^8$)), (acl\_relcl(Music$^0$, be(that$^5$, NOT(classical$^6$)$^8$)))]}
	\end{center}

	To identify whether a replacement should be made, we access the \texttt{acl\_map} using our kernel's source and target IDs as keys and, if the ID is not present, we return the given node. With reference to our previous example, our map contains one entry: \{\texttt{5: Music$^0$}\}. Here, we do not match a value in the source or target of the kernel, but loop through the properties containing a \texttt{SENTENCE}, with a source of ID: 5. So, we replace the source with the maps' value of \texttt{Music$^0$} and remove the \texttt{acl\_relcl} property, resulting in the new kernel:
	
	\begin{center}
		\texttt{be(Music$^0$, ?)[(SENTENCE:be(Music$^0$, NOT(classical$^6$)$^8$))]}
	\end{center}
	
	While \texttt{Music} is repeated,  we retain the ID of the nodes within this final representation, such that we can identify both inclusions of \texttt{Music} as the same entity. Then, this is rewritten in logical form as follows:
	\[\left( \neg \left(\textit{be}(\lozenge {\textsf{music}},\lozenge {\textsf{classical}})\right) \wedge \textit{be}(\lozenge {\textsf{music}})\right)\]

	\subsection{Dealing with Subordinate Clauses}
	If a given full text contained one or more dependent sentences (i.e., where two verbs are present), the pipeline would have failed to recognise this, and could only return at most one sentence. This is not a pitfall of the graph structure from GSM, but a needed improvement to the LaSSI pipeline, considering the recursive nature of language (as discussed in Algorithm \ref{alg:overall_kernel_construction}). We can show this through two examples, the first being ``\textit{attempt to steal someone's husband}''. This was rewritten to: \texttt{steal(attempt, None)[(ENTITY:someone), (GPE:husband)]}, where ``\textit{attempt}'' should be recognised as a verb and, again, another misclassification of ``\textit{husband}'' as a \texttt{GPE}. Our new pipeline recognises the presence of two verbs and the dependency between them, and correctly rewrites in its intermediate form as: \texttt{attempt(?2, None)[(SENTENCE:to steal(?1, husband[(extra:someone[(5:'s)])]))]}, which is then represented in FOL as:
	\[\exists ?1.\,\left(\textit{to steal}(\lozenge {\textsf{?1}},\lozenge \left[\textsf{husband} [\textup{of}] \textit{someone}\right]) \wedge \textit{attempt}(\lozenge {\textsf{?1}})\right)\]
	Second, the full text ``\textit{become able to answer more questions}'' also contains \textit{two} verbs; however, the old pipeline again did not account for this and thus rewrote it to: \texttt{become(?, None)}. Subsequently, it is now represented in an intermediate representation as: 
	
	\begin{center}
		\texttt{become(?2[(cop:able)], None)[(SENTENCE:to answer(?1, questions[(amod:more)]))]}
	\end{center}
	
	This then becomes the following in logical form:
	\[\exists ?1.\,\left(\textit{to answer}(\lozenge {\textsf{?1}}^{\texttt{None}}_{\texttt{JJ}: \lozenge {\textsf{able}}^{\texttt{None}}},\lozenge {\textsf{questions}}_{\texttt{JJ}: \lozenge {\textsf{more}}}) \wedge \textit{become}(\lozenge {\textsf{?1}}_{\texttt{JJ}: \lozenge {\textsf{able}}})\right)\]

	\subsection{Improvement Over the Logical Representation}
	With reference to the \ref{rqn2c} dataset, sentence 2 states that ``\textit{There is traffic but not in the Newcastle city centre}'', meaning there is traffic \textit{somewhere}, but specifies that it is not in the Newcastle city centre; meanwhile, sentence 11 states ``\textit{Newcastle has traffic but not in the city centre}'', comprising a subtle difference which was not captured by any other approach presented. Our previous implementation showed complete similarity for 2 $\implies$ 11, but a 50\% for 11 $\implies$ 2, thus wrongly entailing that knowing the general possibility of having traffic should entail at least the possibility of having traffic in Newcastle. Our current implementation, through better structuring of the direction of the implication through novel sentence semantics, flips the implication, as having traffic in Newcastle entails the general possibility of having traffic elsewhere, but not vice versa.

	\section{Dendrograms}\label{sup:dendrograms}
	\subsection{Supporting Dendrograms}\label{sup:support_dendrograms}
	\textit{The following sections present and explain dendrograms for the results displayed in Section \ref{sec:results}.}
	
	\subsubsection{\ref{rqn2a} (Logical Connectives)}
	\begin{figure}[!p] 
		\centering
		
\hspace*{-1cm}
		\subfloat[\glspl*{sg}]{
			\label{subfig:cluster_dend_ab_simple}
			\begin{minipage}{0.42\textwidth}
				\centering
				\includegraphics[width=\linewidth,height=5cm,keepaspectratio]{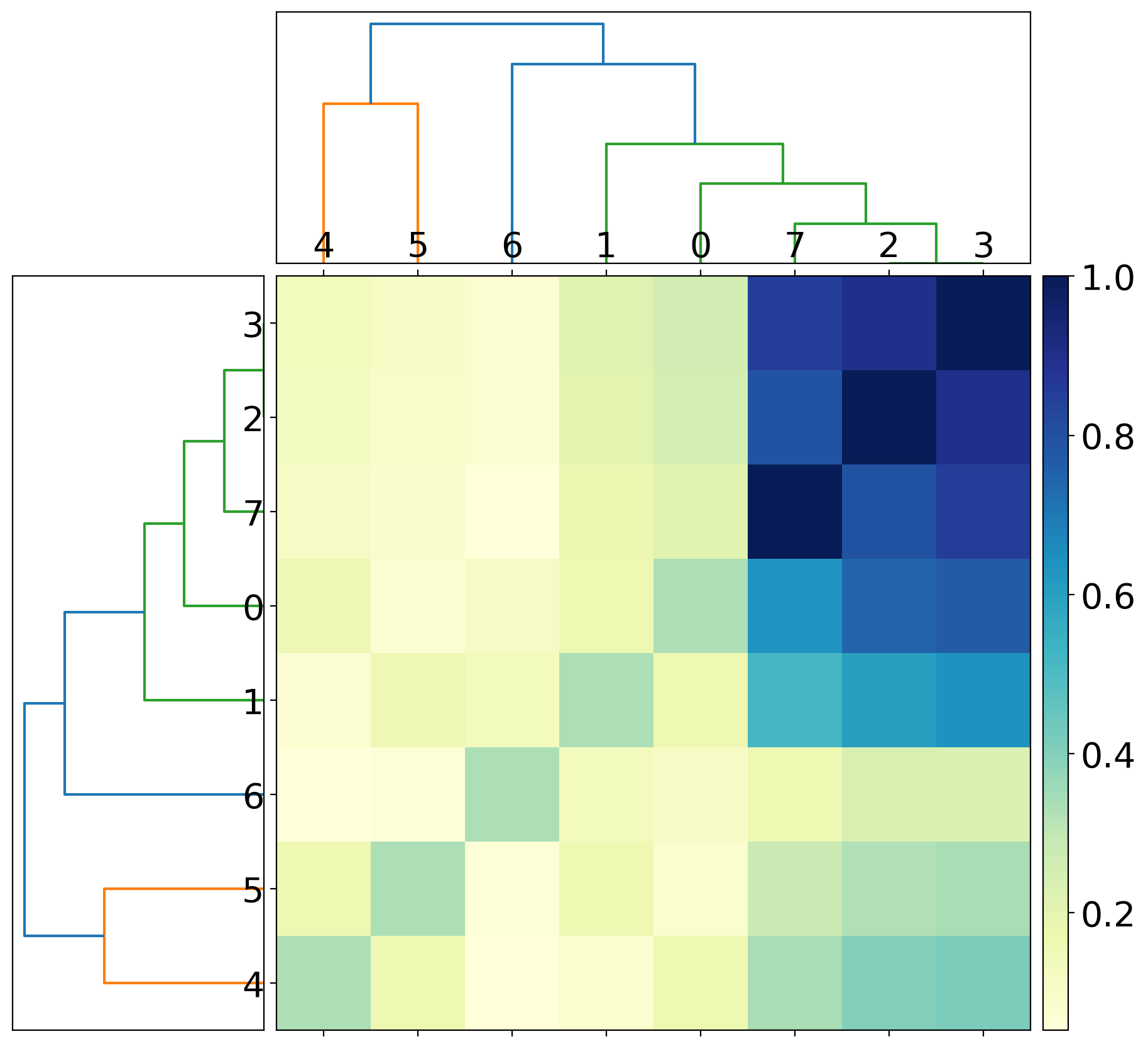}
			\end{minipage}
		}\hspace*{-1.5cm}
		\subfloat[\glspl*{lg}]{
			\label{subfig:cluster_dend_ab_logicalgraphs}
			\begin{minipage}{0.42\textwidth}
				\centering
				\includegraphics[width=\linewidth,height=5cm,keepaspectratio]{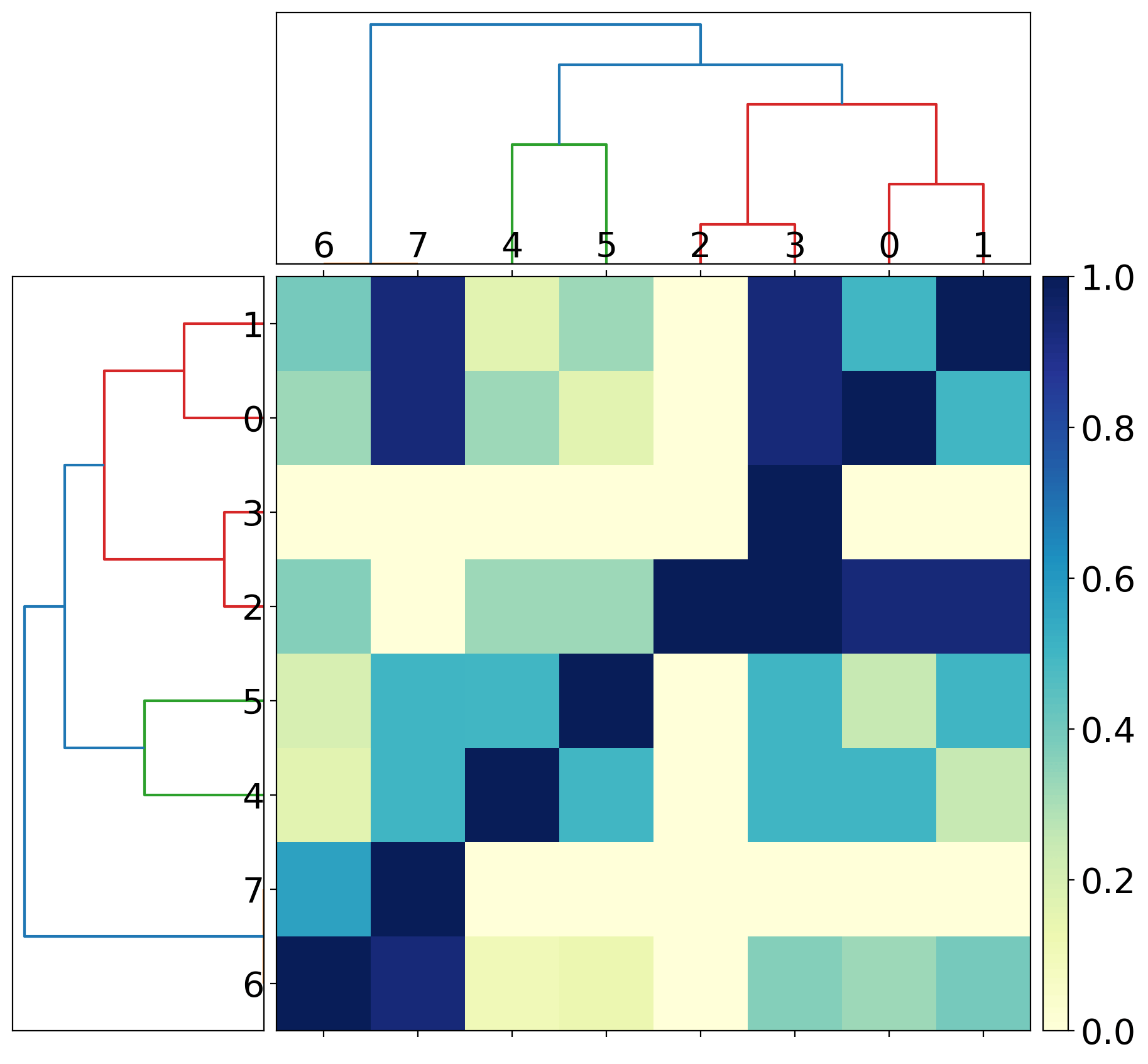}
			\end{minipage}
		}\hspace*{-1.5cm}
		\subfloat[Logical]{
			\label{subfig:cluster_dend_ab_logical}
			\begin{minipage}{0.42\textwidth}
				\centering
				\includegraphics[width=\linewidth,height=5cm,keepaspectratio]{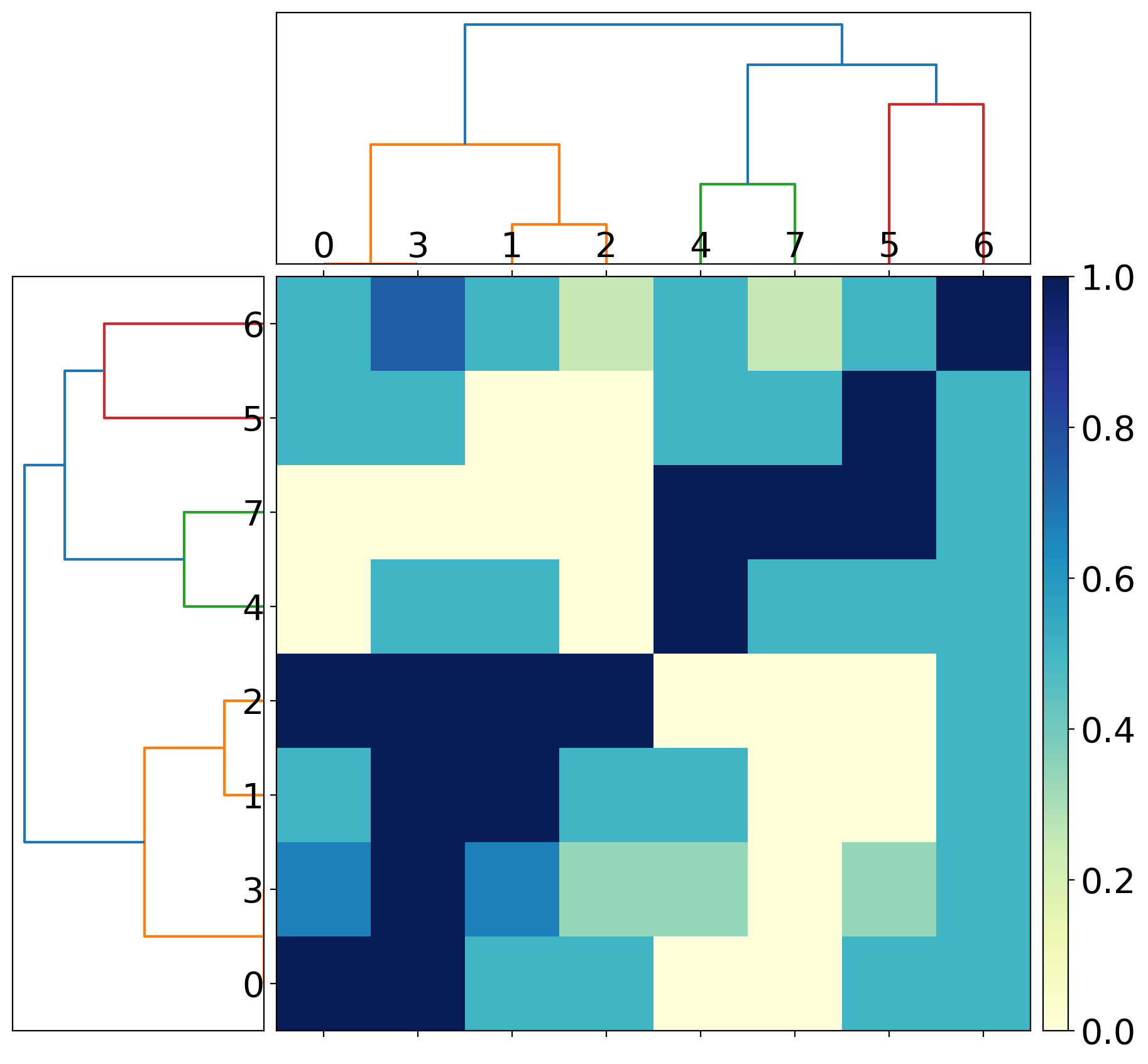}
			\end{minipage}
		}
		
		\vspace{12pt}
		
\hspace*{-1cm}
		\subfloat[\textbf{T1:} all-MiniLM-L6-v2]{
			\label{subfig:cluster_dend_ab_l6v2}
			\begin{minipage}{0.42\textwidth}
				\centering
				\includegraphics[width=\linewidth,height=5cm,keepaspectratio]{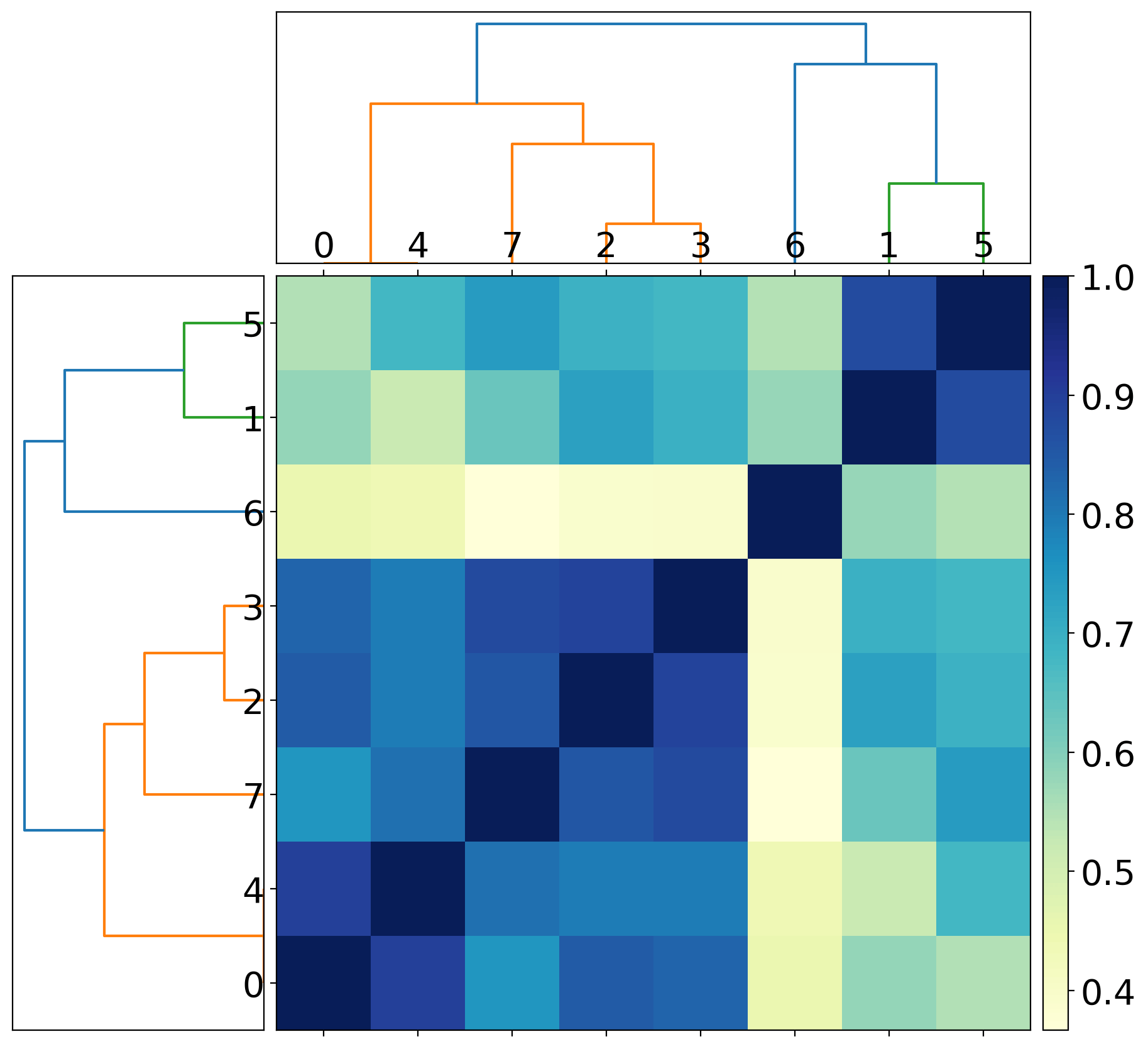}
			\end{minipage}
		}
		\hspace*{-1.5cm}
		\subfloat[\textbf{T2:} all-MiniLM-L12-v2]{
			\label{subfig:cluster_dend_ab_l12v2}
			\begin{minipage}{0.42\textwidth}
				\centering
				\includegraphics[width=\linewidth,height=5cm,keepaspectratio]{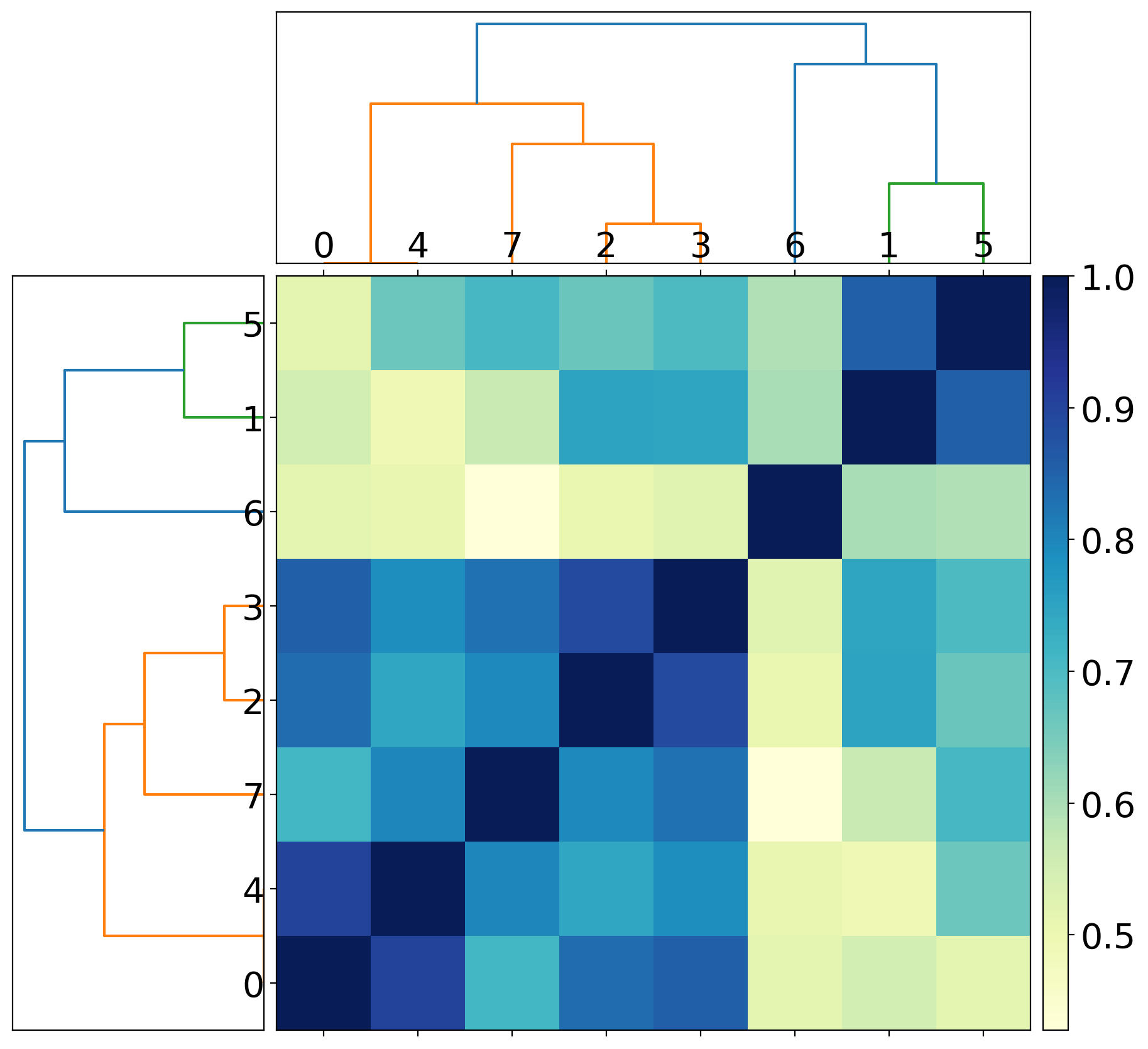}
			\end{minipage}
		}
		\hspace*{-1.5cm}
		\subfloat[\textbf{T3:} all-mpnet-base-v2]{
			\label{subfig:cluster_dend_ab_mpnet}
			\begin{minipage}{0.42\textwidth}
				\centering
				\includegraphics[width=\linewidth,height=5cm,keepaspectratio]{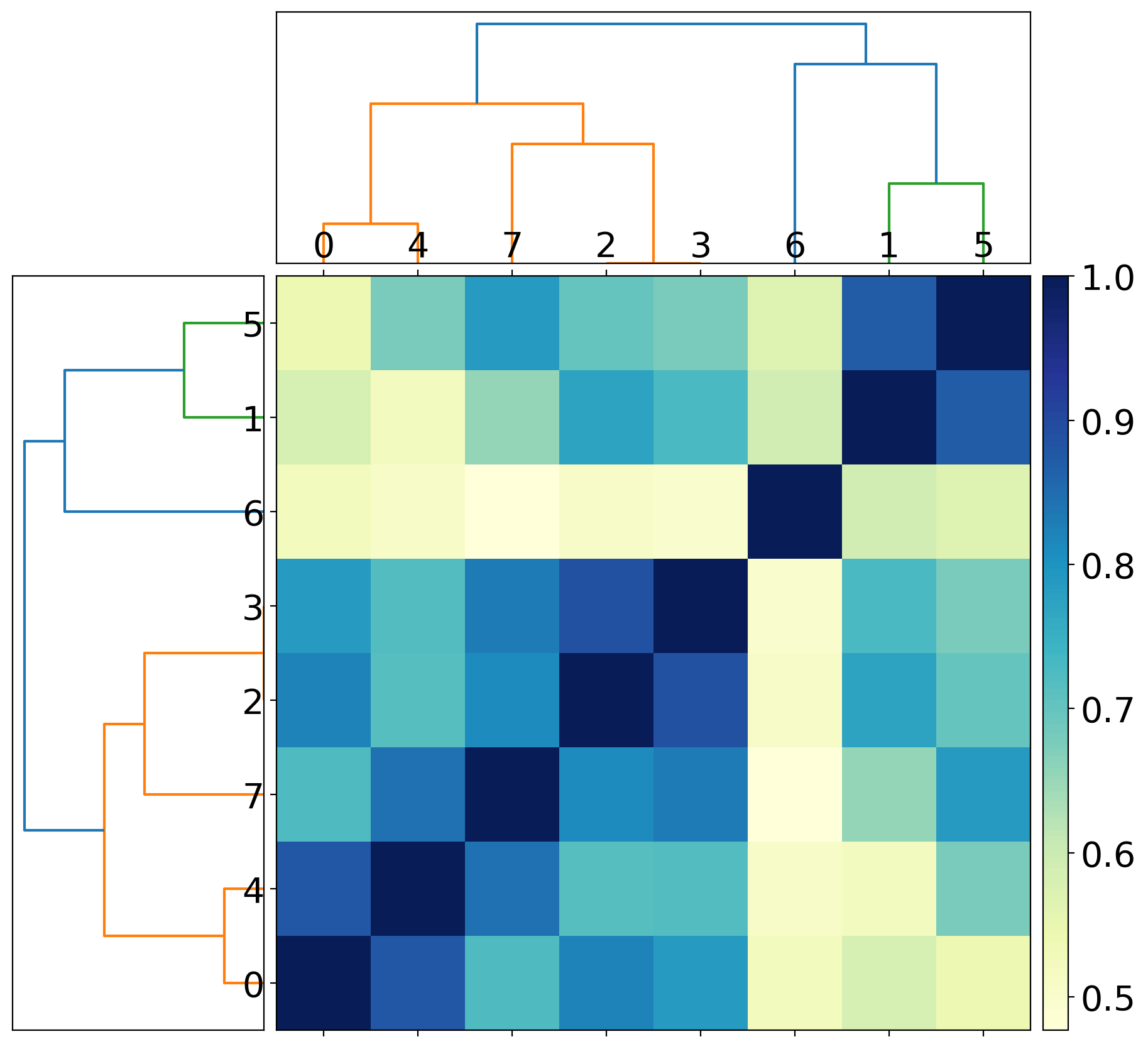}
			\end{minipage}
		}
		
		\vspace{12pt}
		
\hspace*{-1cm}
		\subfloat[\textbf{T4:} all-roberta-large-v1]{
			\label{subfig:cluster_dend_ab_roberta}
			\begin{minipage}{0.42\textwidth}
				\centering
				\includegraphics[width=\linewidth,height=5cm,keepaspectratio]{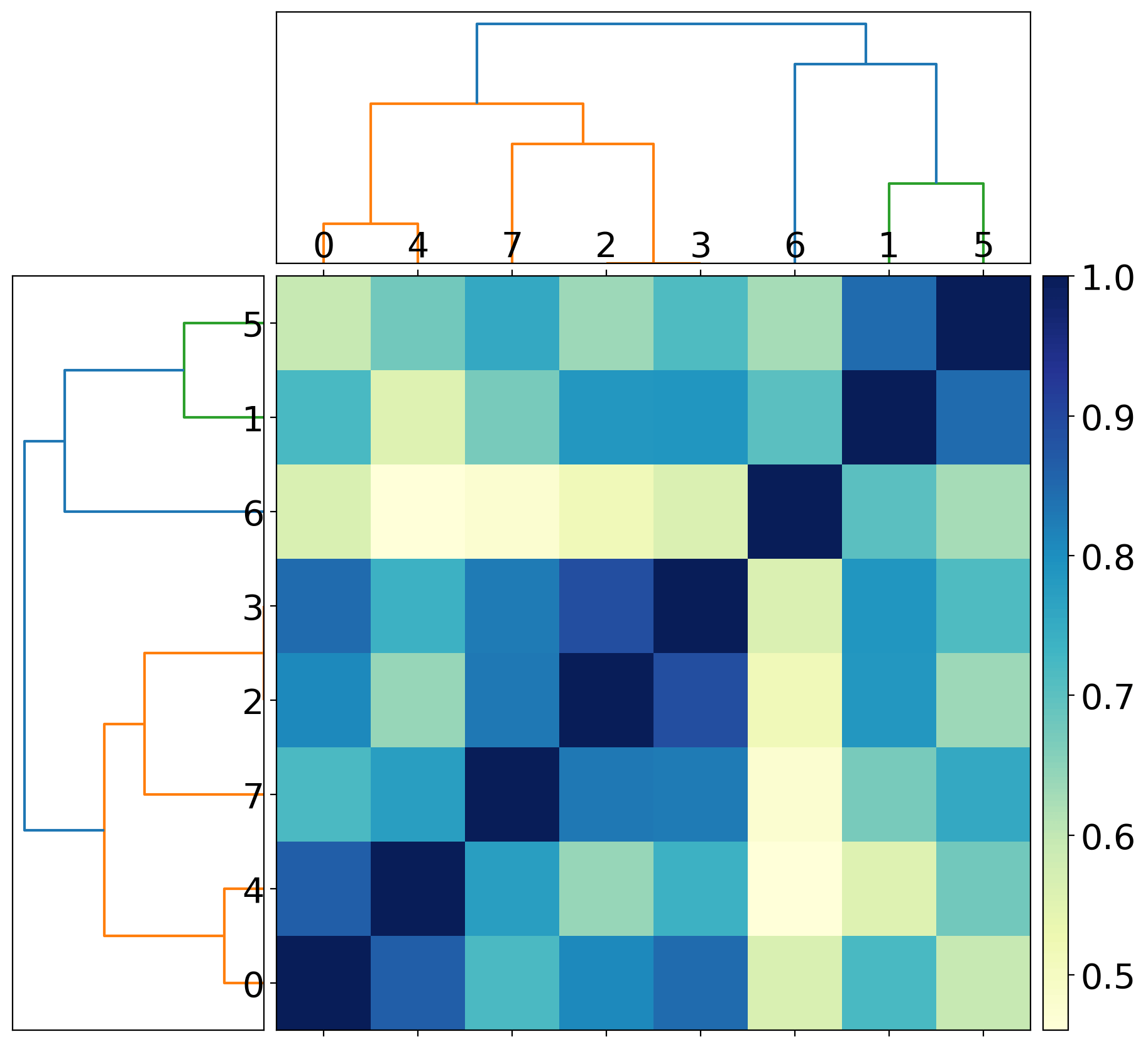}
			\end{minipage}
		}\hspace*{-1.5cm}
		\subfloat[\textbf{T5:} DeBERTaV2+AMR-LDA]{
			\label{subfig:cluster_dend_ab_DeBERTaV2}
			\begin{minipage}{0.42\textwidth}
				\centering
				\includegraphics[width=\linewidth,height=5cm,keepaspectratio]{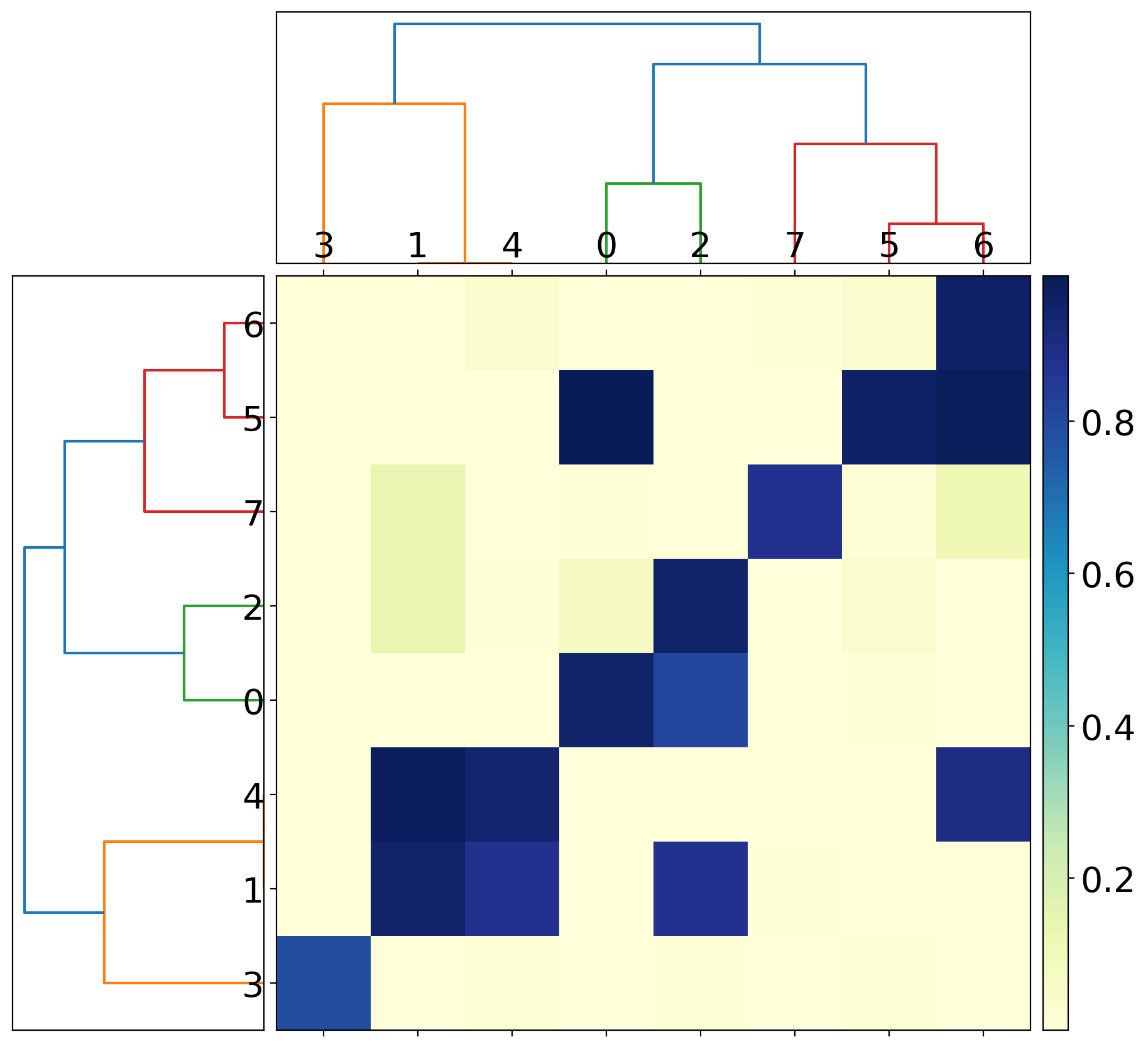}
			\end{minipage}
		}
		\hspace*{-1.5cm}
		\subfloat[\textbf{T6:} ColBERTv2+RAGatouille]{
			\label{subfig:cluster_dend_ab_ColBERTv2}
			\begin{minipage}{0.42\textwidth}
				\centering
				\includegraphics[width=\linewidth,height=5cm,keepaspectratio]{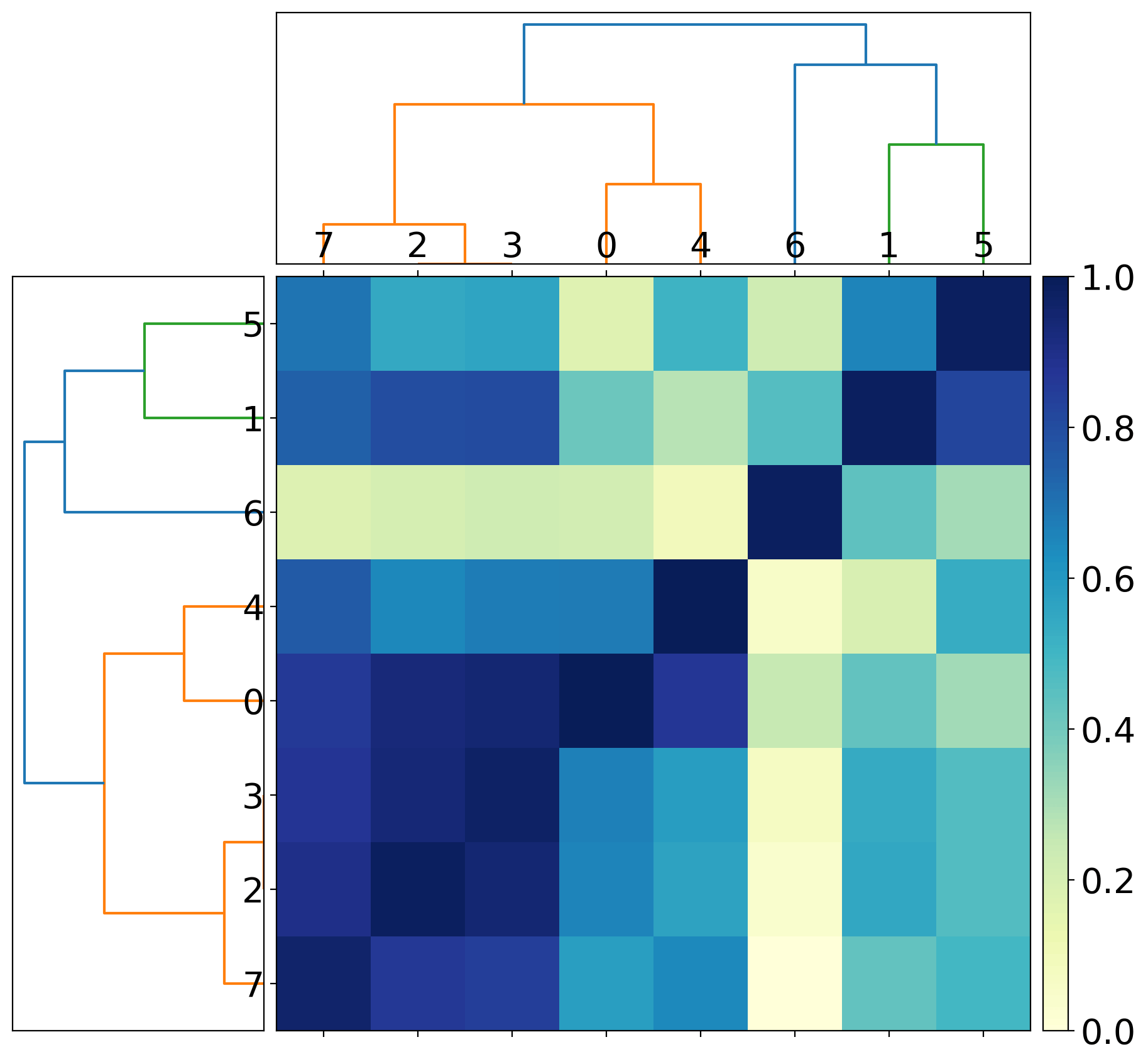}
			\end{minipage}
		}
		\caption{Dendrogram clustering for \ref{rqn2a} sentences{, where no clusters are highlighted, as there are none to present}.}
		\label{fig:clustering_dend_ab}
	\end{figure}
	
	In \figurename~\ref{fig:clustering_dend_ab}, we can see that for our \glspl*{sg} in \figurename~\ref{subfig:cluster_dend_ab_simple}, this approach fails to recognise directionality and does not produce appropriate clusters, even failing to recognise the similarity a sentence has with itself. The \glspl*{lg} shown in \figurename~\ref{subfig:cluster_dend_ab_logicalgraphs} improve, recognising similarity for 2 $\implies$ 0, for example; however, they indicate a similarity of 0 for 0 $\implies$ 2. Finally, our logical representation in \figurename~\ref{subfig:cluster_dend_ab_logical} successfully identifies directionality and appropriately groups sentences together. For example, 2 $\implies$ 0 is entirely similar; however, 0 $\implies$ 2 is only slightly similar, which entails that ``\textit{Alice and Bob play football}'' implies that ``\textit{Alice plays football}'', but it does not necessarily hold that \textbf{both} Alice and Bob play football if we only know that Alice plays football.
	
	The sentence transformer approaches displayed in Figures \ref{subfig:cluster_dend_ab_l6v2}, \ref{subfig:cluster_dend_ab_l12v2}, \ref{subfig:cluster_dend_ab_mpnet}, and \ref{subfig:cluster_dend_ab_roberta} all produce similar clustering results, with high similarities between sentences in the dataset that are incorrect; for example, the clusters of Alice \textbf{does} and \textbf{does not} play football are closely related according to the transformers. As mentioned above, this is most likely due to the transformers' use of tokenisation, which removes stop words and, therefore, does not capture a proper understanding of the sentences. In contrast, our approach does this by showing that they are dissimilar. DeBERTaV2+AMR-LDA (\figurename~\ref{subfig:cluster_dend_ab_DeBERTaV2}) has a dramatically different dendrogram from the others, showing very little notion of directionality and classifying the majority of sentences as completely dissimilar. This was unexpected for this approach, as it was explicitly trained to recognise logical entailment from simple factoid sentences. Moreover, ColBERTv2+RAGatouille (\figurename~\ref{subfig:cluster_dend_ab_ColBERTv2}) shows a similar dendrogram to those of the sentence transformer approaches, with a wider range of confidence determining similarity, with $0$ at its lower bound compared to $\approx${0.4} for the transformer approaches.
	
	\subsubsection{\ref{rqn2b} (Active vs. Passive)}
\begin{figure}[!p] 
		\centering
		
\hspace*{-1cm}
		\subfloat[\glspl*{sg}]{
			\label{subfig:cluster_dend_cm_simple}
			\begin{minipage}{0.42\textwidth}
				\centering
				\includegraphics[width=\linewidth,height=5cm,keepaspectratio]{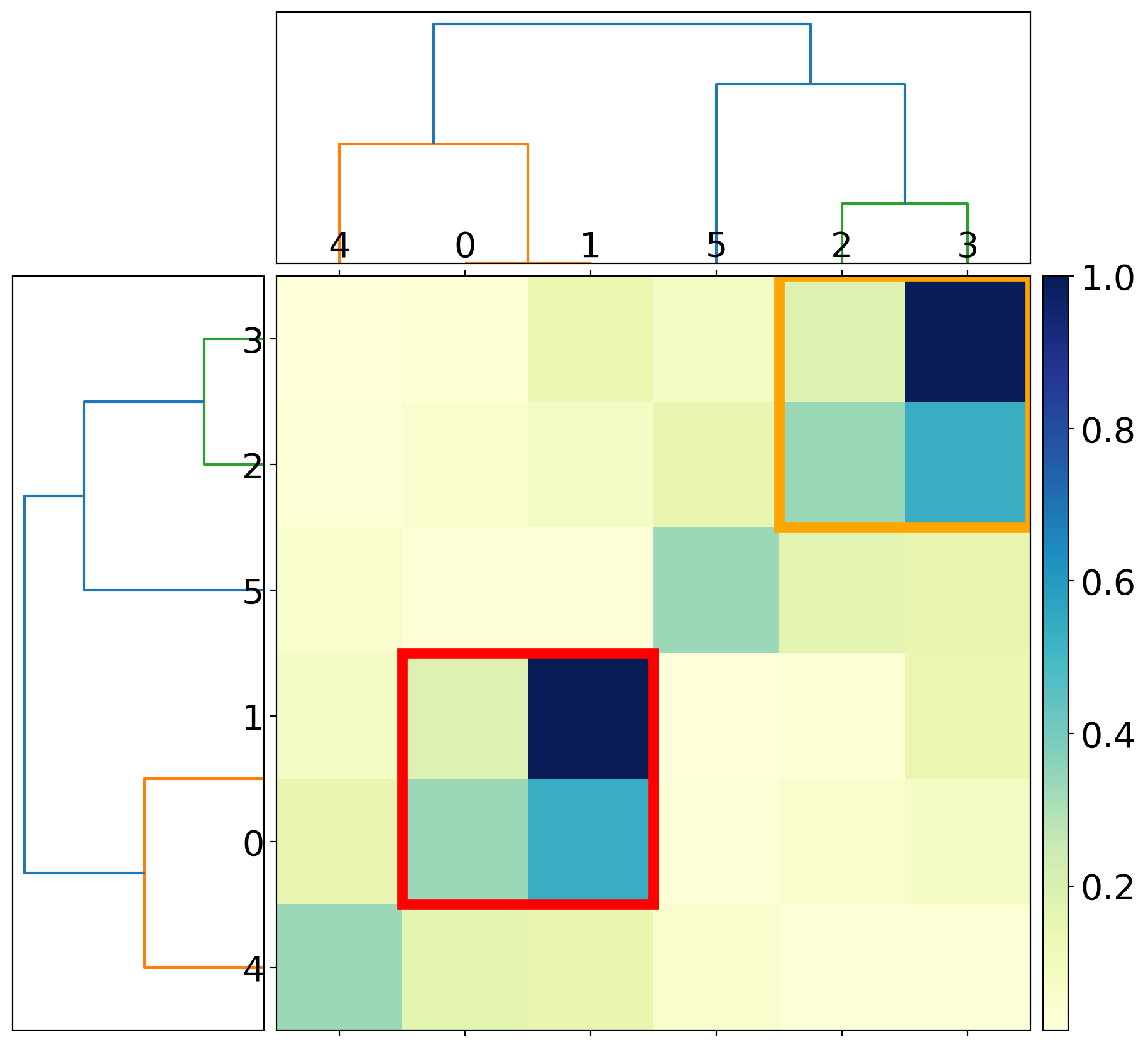}
			\end{minipage}
		}
		\hspace*{-1.5cm}
		\subfloat[\glspl*{lg}]{
			\label{subfig:cluster_dend_cm_logicalgraphs}
			\begin{minipage}{0.42\textwidth}
				\centering
				\includegraphics[width=\linewidth,height=5cm,keepaspectratio]{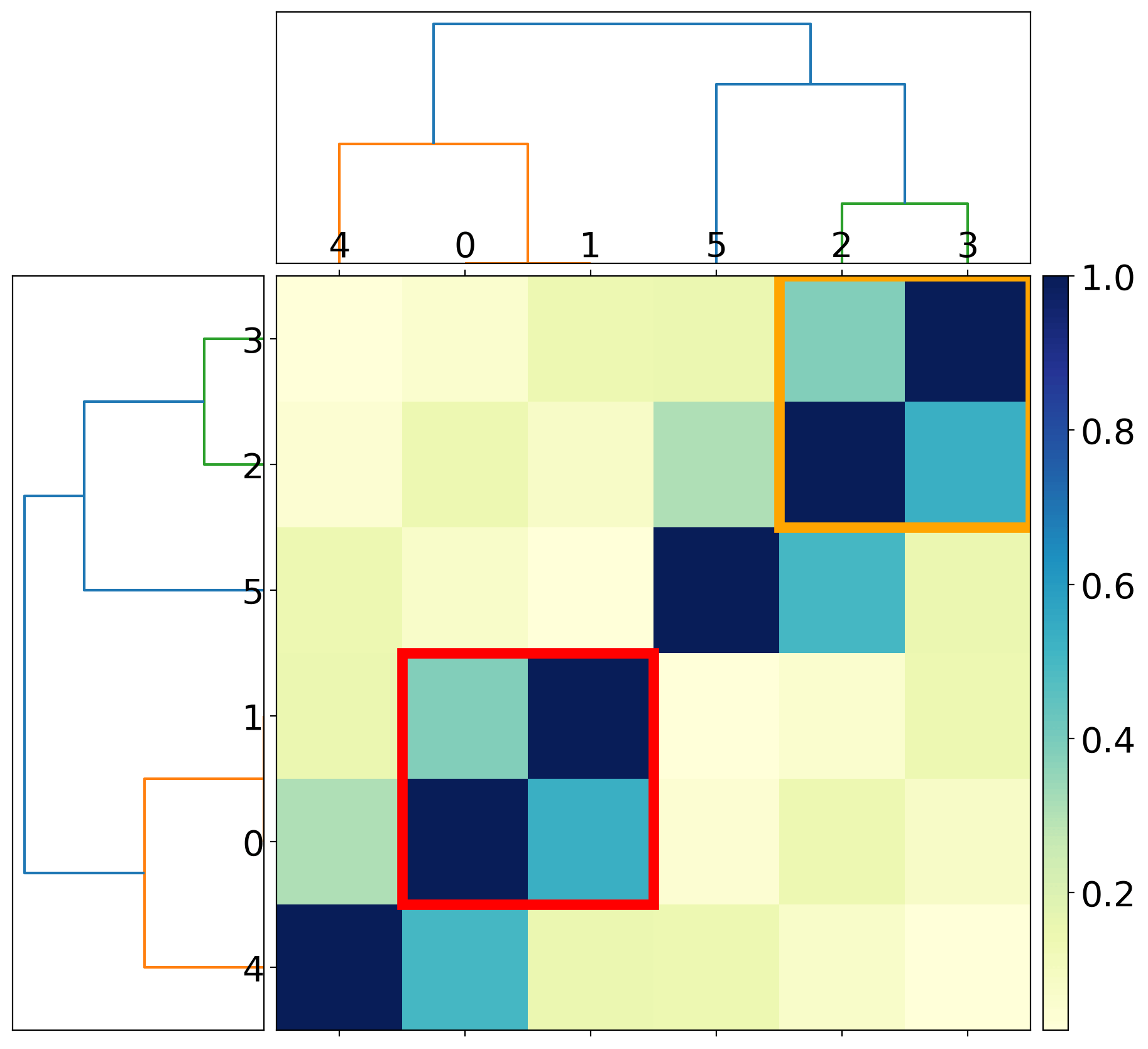}
			\end{minipage}
		}
		\hspace*{-1.5cm}
		\subfloat[Logical]{
			\label{subfig:cluster_dend_cm_logical}
			\begin{minipage}{0.42\textwidth}
				\centering
				\includegraphics[width=\linewidth,height=5cm,keepaspectratio]{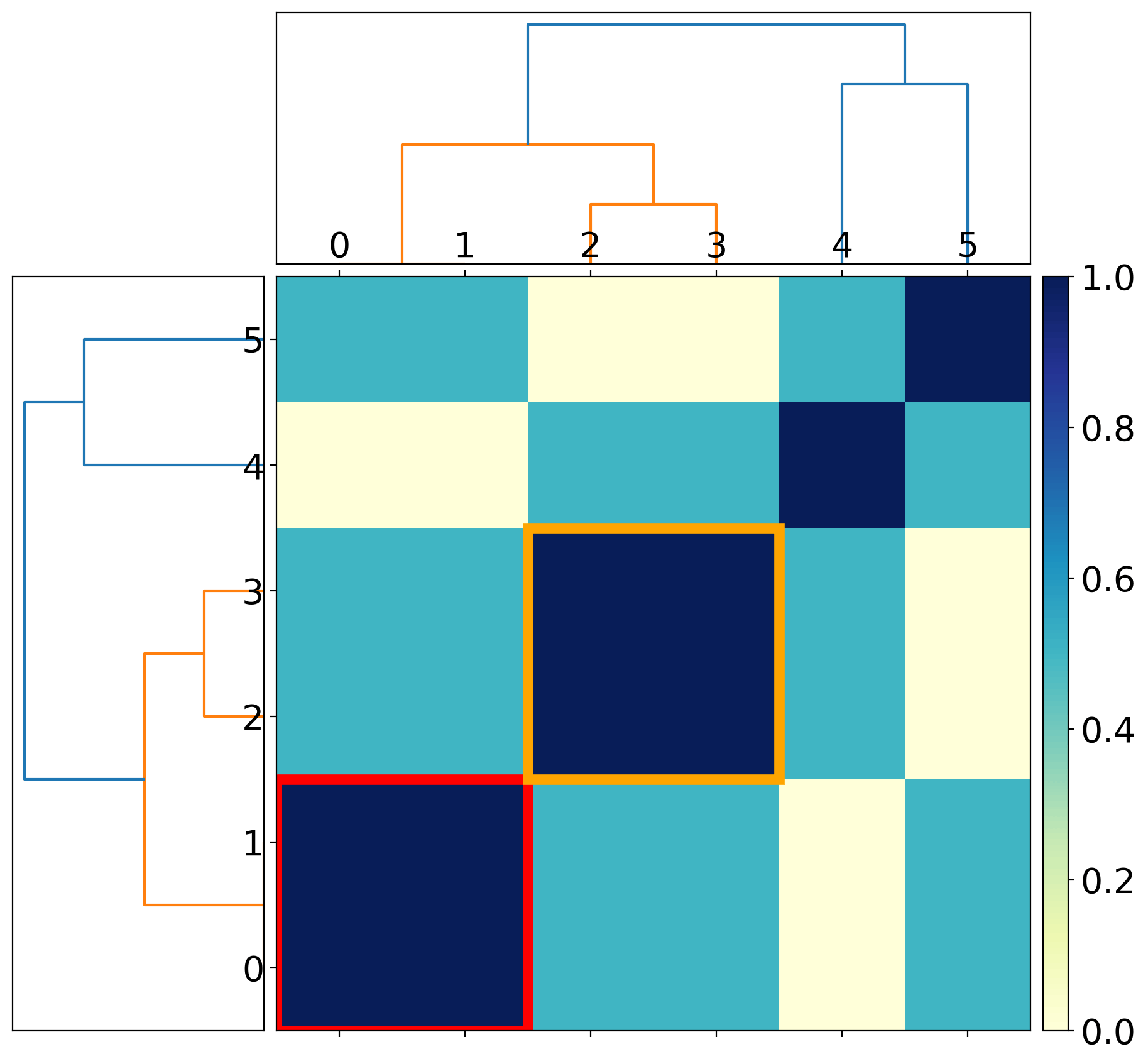}
			\end{minipage}
		}
		
		\vspace{12pt}
		
\hspace*{-1cm}
		\subfloat[\textbf{T1:} all-MiniLM-L6-v2]{
			\label{subfig:cluster_dend_cm_l6v2}
			\begin{minipage}{0.42\textwidth}
				\centering
				\includegraphics[width=\linewidth,height=5cm,keepaspectratio]{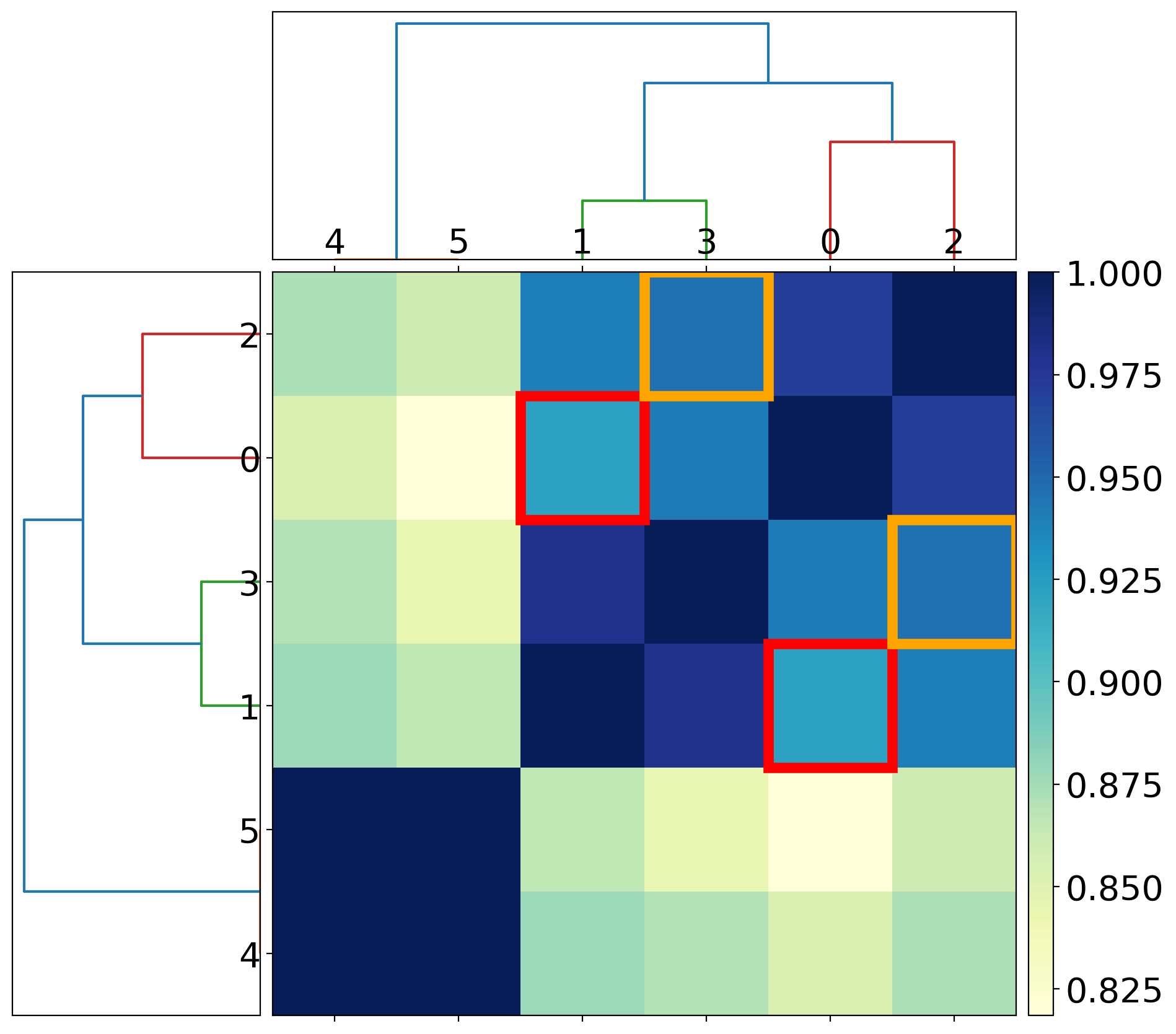}
			\end{minipage}
		}
		\hspace*{-1.5cm}
		\subfloat[\textbf{T2:} all-MiniLM-L12-v2]{
			\label{subfig:cluster_dend_cm_l12v2}
			\begin{minipage}{0.42\textwidth}
				\centering
				\includegraphics[width=\linewidth,height=5cm,keepaspectratio]{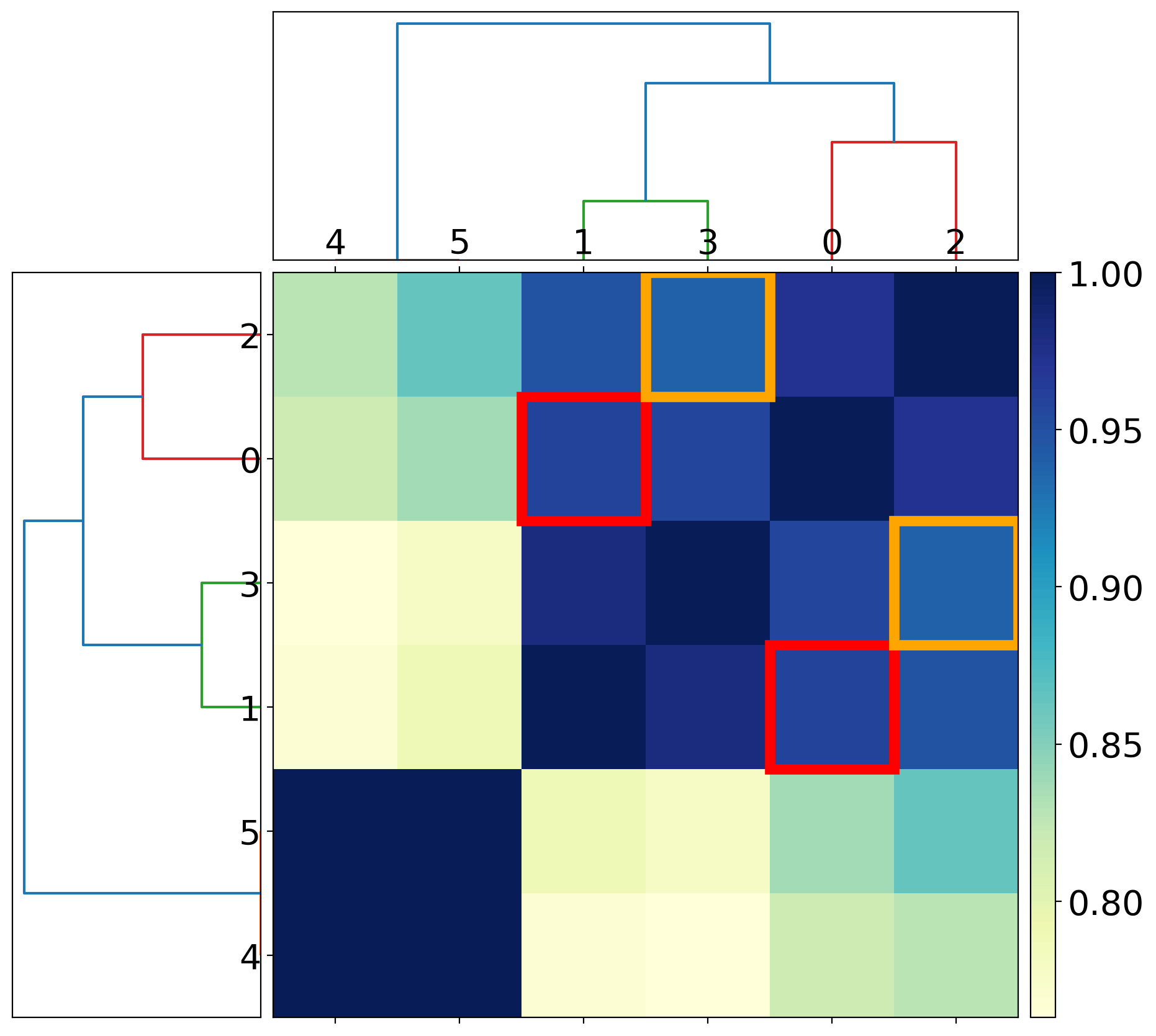}
			\end{minipage}
		}
		\hspace*{-1.5cm}
		\subfloat[\textbf{T3:} all-mpnet-base-v2]{
			\label{subfig:cluster_dend_cm_mpnet}
			\begin{minipage}{0.42\textwidth}
				\centering
				\includegraphics[width=\linewidth,height=5cm,keepaspectratio]{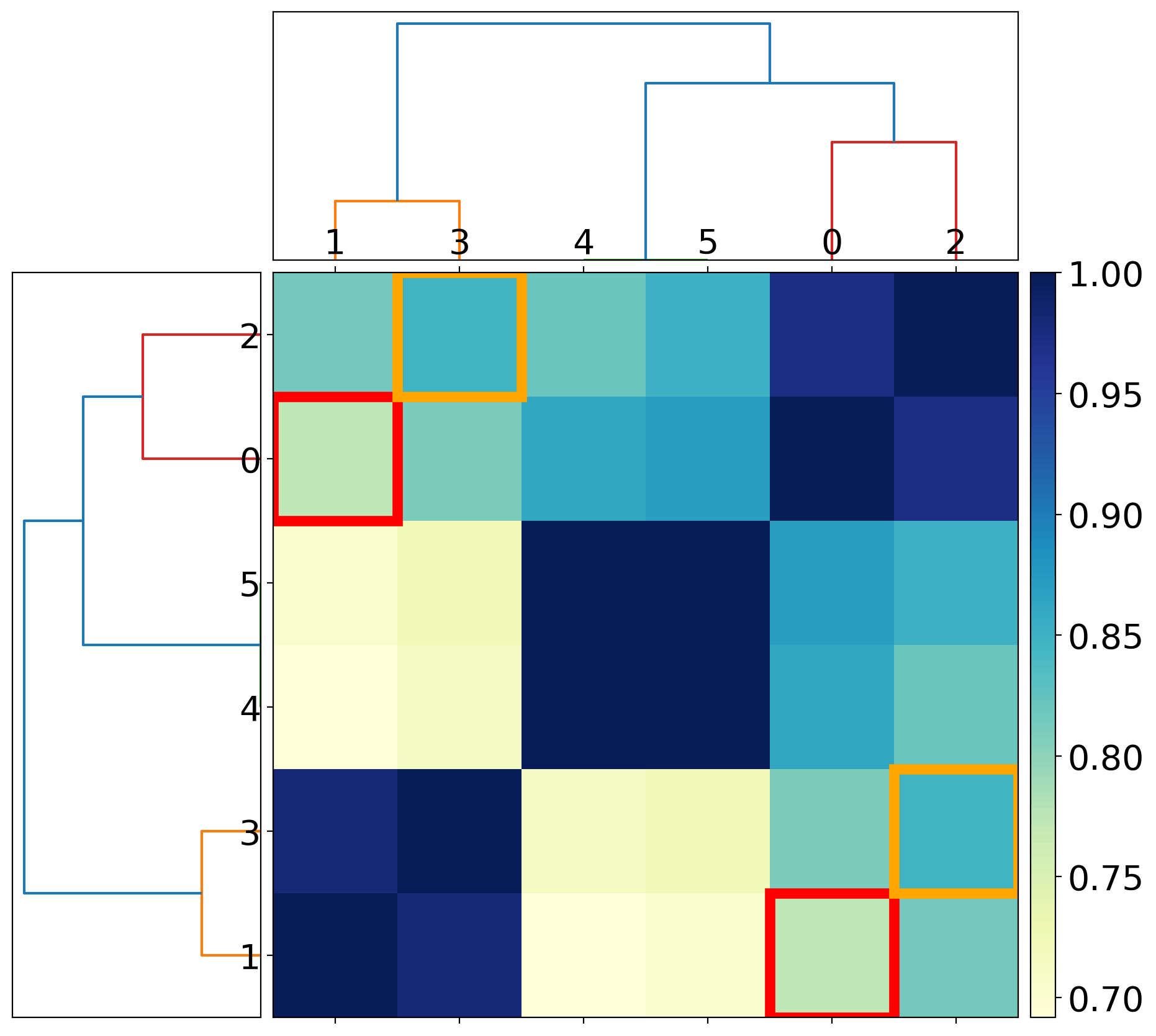}
			\end{minipage}
		}
		
		\vspace{12pt}
		
\hspace*{-1cm}
		\subfloat[\textbf{T4:} all-roberta-large-v1]{
			\label{subfig:cluster_dend_cm_roberta}
			\begin{minipage}{0.42\textwidth}
				\centering
				\includegraphics[width=\linewidth,height=5cm,keepaspectratio]{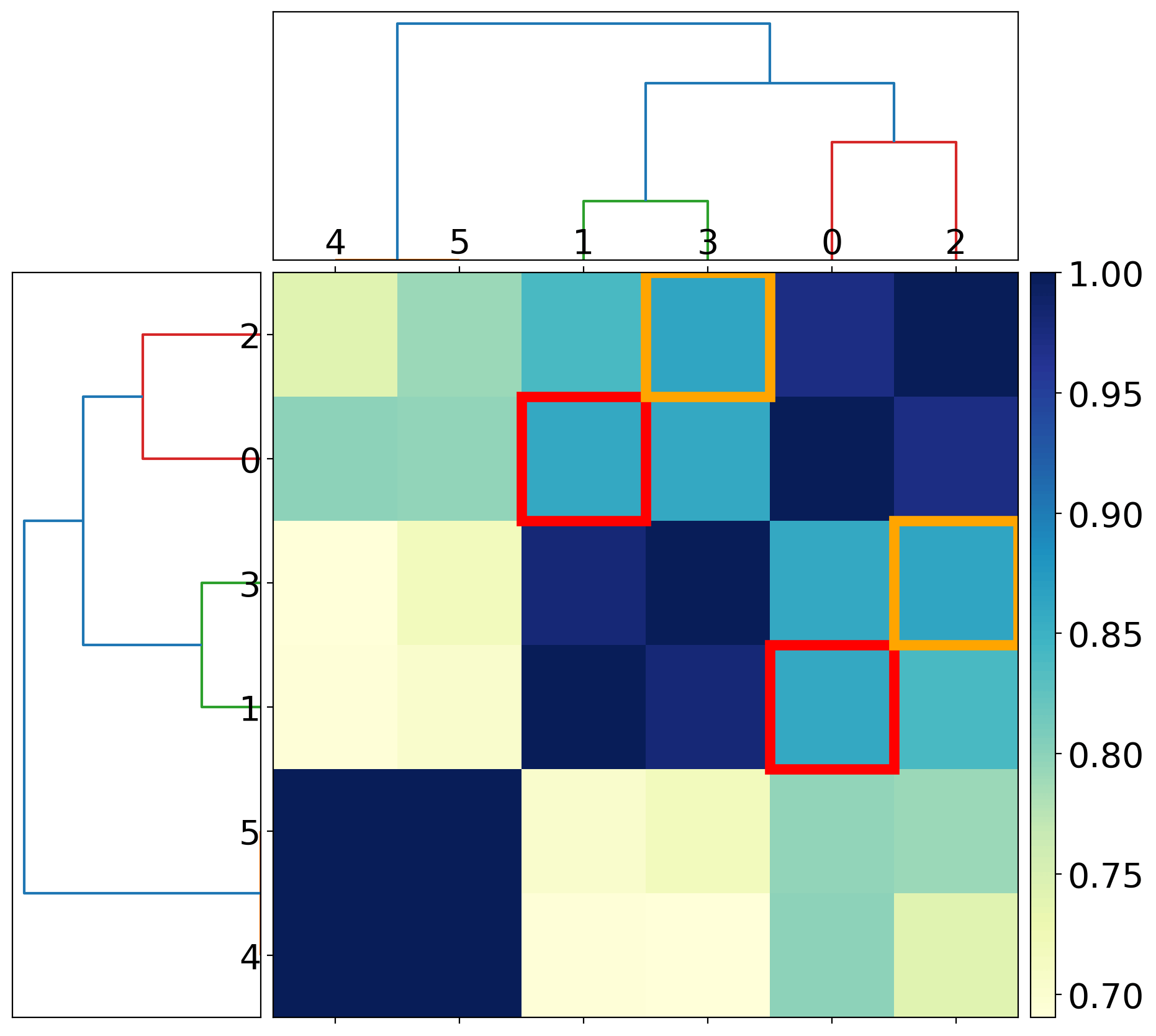}
			\end{minipage}
		}
		\hspace*{-1.5cm}
		\subfloat[\textbf{T5:} DeBERTaV2+AMR-LDA]{
			\label{subfig:cluster_dend_cm_DeBERTaV2}
			\begin{minipage}{0.42\textwidth}
				\centering
				\includegraphics[width=\linewidth,height=5cm,keepaspectratio]{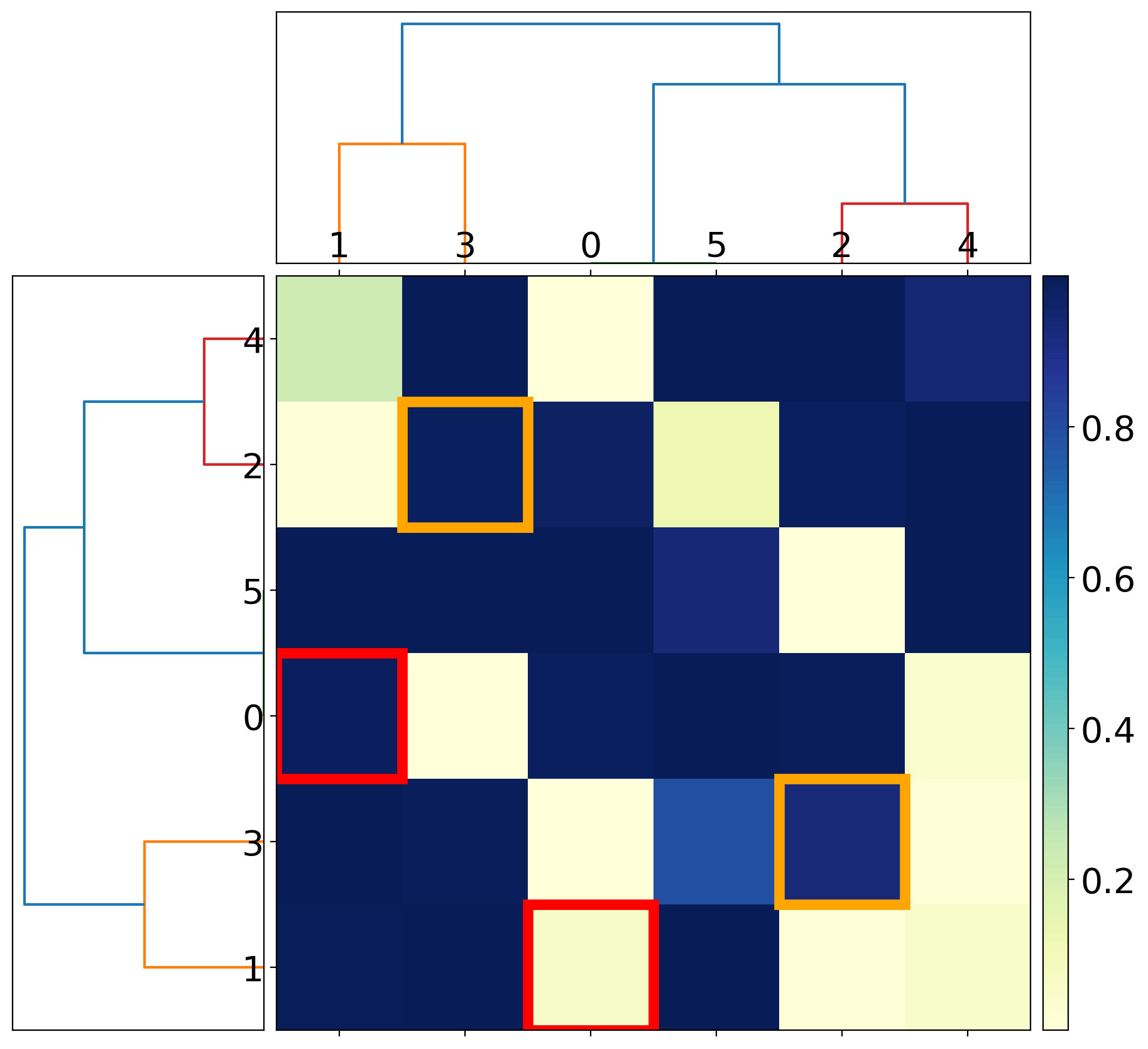}
			\end{minipage}
		}
		\hspace*{-1.5cm}
		\subfloat[\textbf{T6:} ColBERTv2+RAGatouille]{
			\label{subfig:cluster_dend_cm_ColBERTv2}
			\begin{minipage}{0.42\textwidth}
				\centering
				\includegraphics[width=\linewidth,height=5cm,keepaspectratio]{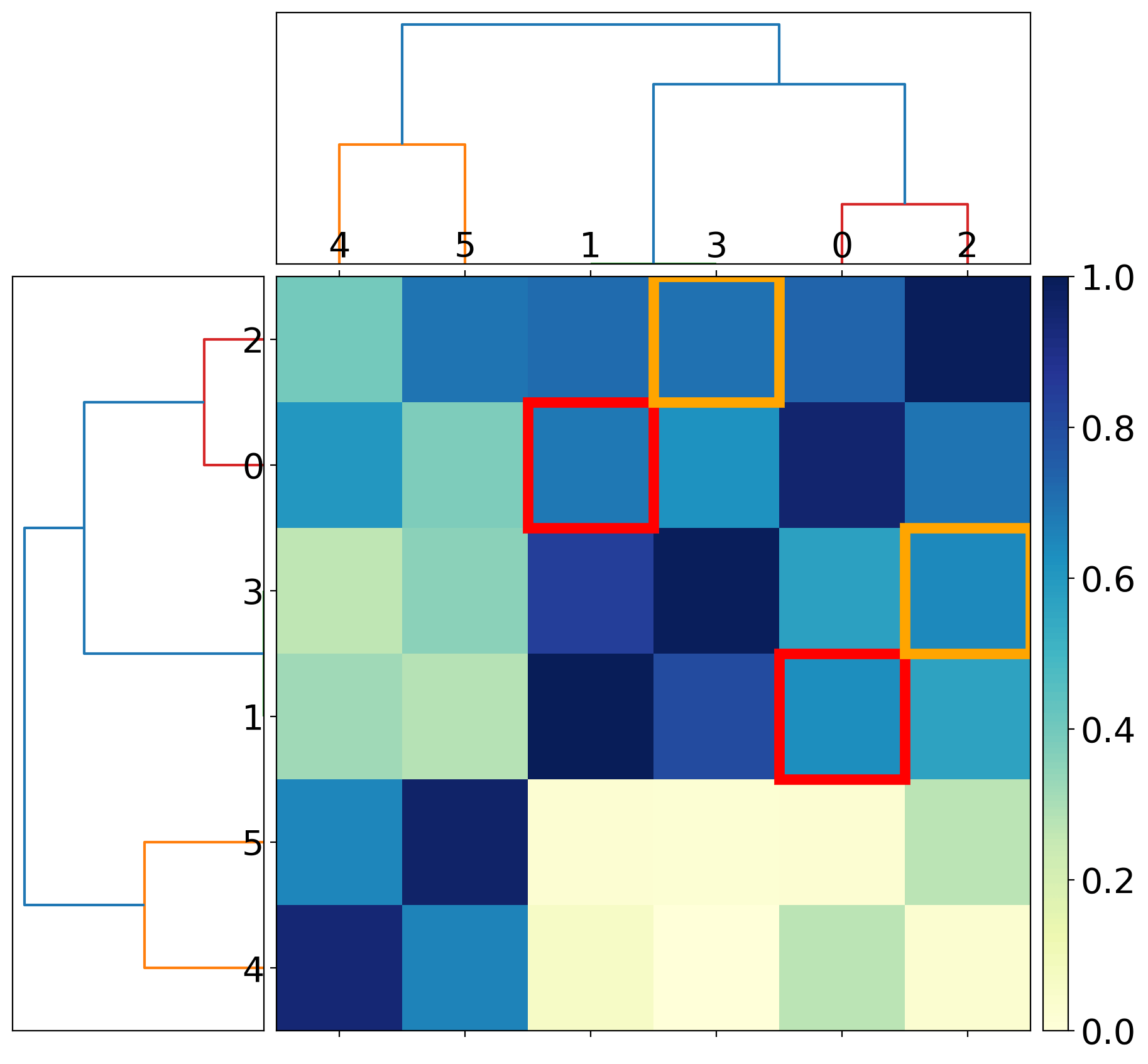}
			\end{minipage}
		}
		\caption{Dendrogram clustering for \ref{rqn2b} sentences{, where red and orange boxes represent the clusters for \{0,1\} and \{2,3\} respectively.}}
		\label{fig:clustering_dend_cm}
	\end{figure}

	We can see a gradual increase in the number of matched clusters in Figures \ref{subfig:cluster_dend_cm_simple}, \ref{subfig:cluster_dend_cm_logicalgraphs}, and \ref{subfig:cluster_dend_cm_logical}, as the contextual information within the final representation increases with each different sentence representation. Each representation shows a further refinement within the pipeline. The clustering shows values that are near to each other, due to their structural semantics, but does not determine that they are the same. The first two approaches (\glspl*{sg} and \glspl*{lg}) roughly capture the expected clusters that we are looking for, but do not encapsulate the entire picture regarding the similarity of the sentences in the whole dataset. Meanwhile, our final logical representation produces the expected perfect matches with \{0, 1\}, and \{2, 3\}. This suggests that our proposed logical representation can identify the same relationship between the ``\textit{cat}'' and ``\textit{mouse}'', despite different syntactic structures via active and passive voice. On the other hand, all pre-trained language models except for DeBERTaV2+AMR-LDA (T5) exhibited the following behaviour: clusters \{0,2\} and \{1,3\} always formed, thus suggesting that the attention mechanism favours the clustering of sentences around the verb, despite grouping together unrelated sentences. DeBERTaV2+AMR-LDA also made a similar choice by grouping together unrelated sentences: \{0,5\} and \{2,4\}.

	\begin{figure}[!p] 
		\centering
		
\hspace*{-1cm}
		\subfloat[\glspl*{sg}]{
			\label{subfig:cluster_dend_newc_simple}
			\begin{minipage}{0.42\textwidth}
				\centering
				\includegraphics[width=\linewidth,height=5cm,keepaspectratio]{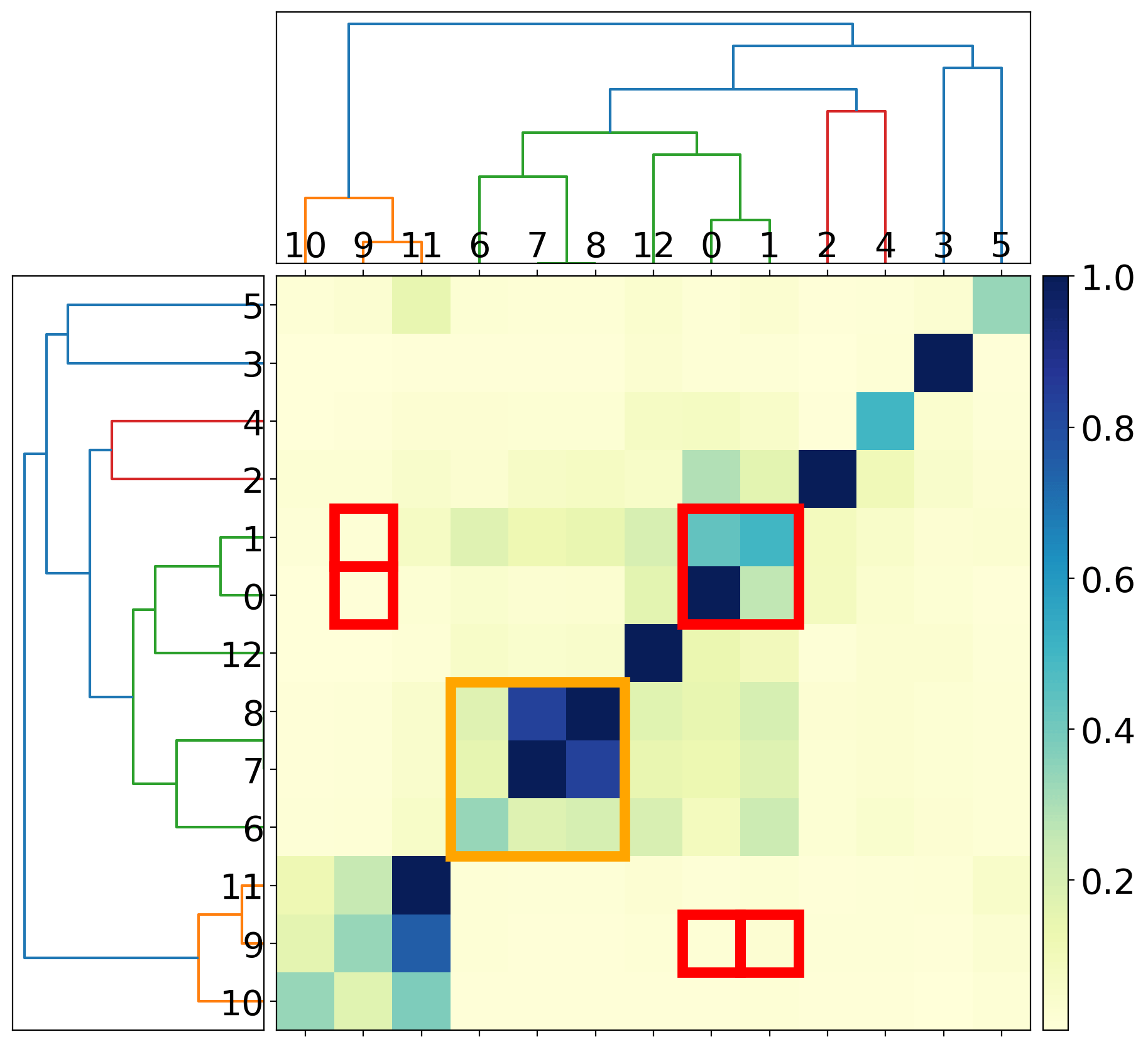}
			\end{minipage}
		}
		\hspace*{-1.5cm}
		\subfloat[\glspl*{lg}]{
			\label{subfig:cluster_dend_newc_logicalgraphs}
			\begin{minipage}{0.42\textwidth}
				\centering
				\includegraphics[width=\linewidth,height=5cm,keepaspectratio]{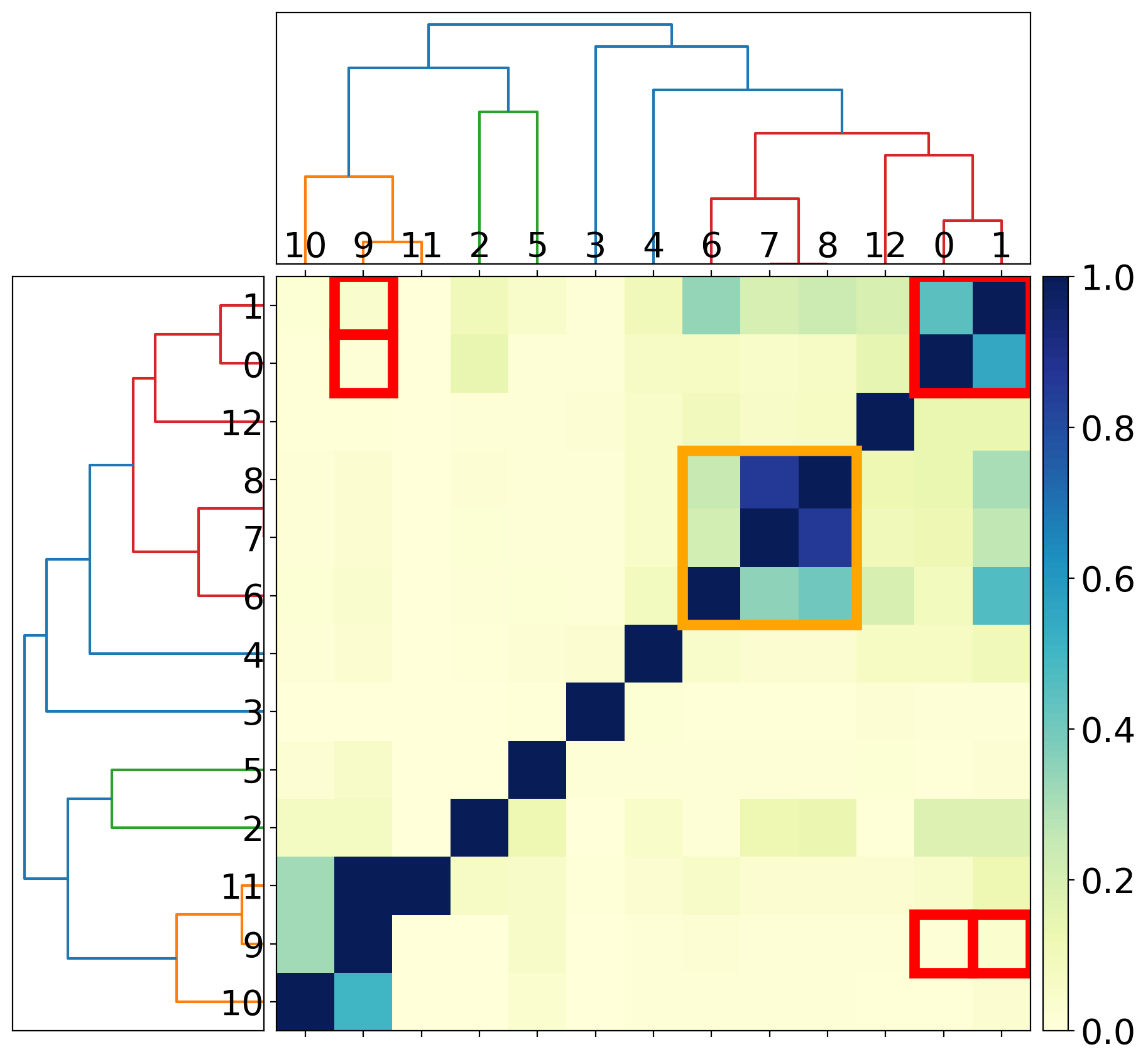}
			\end{minipage}
		}
		\hspace*{-1.5cm}
		\subfloat[Logical]{
			\label{subfig:cluster_dend_newc_logical}
			\begin{minipage}{0.42\textwidth}
				\centering
				\includegraphics[width=\linewidth,height=5cm,keepaspectratio]{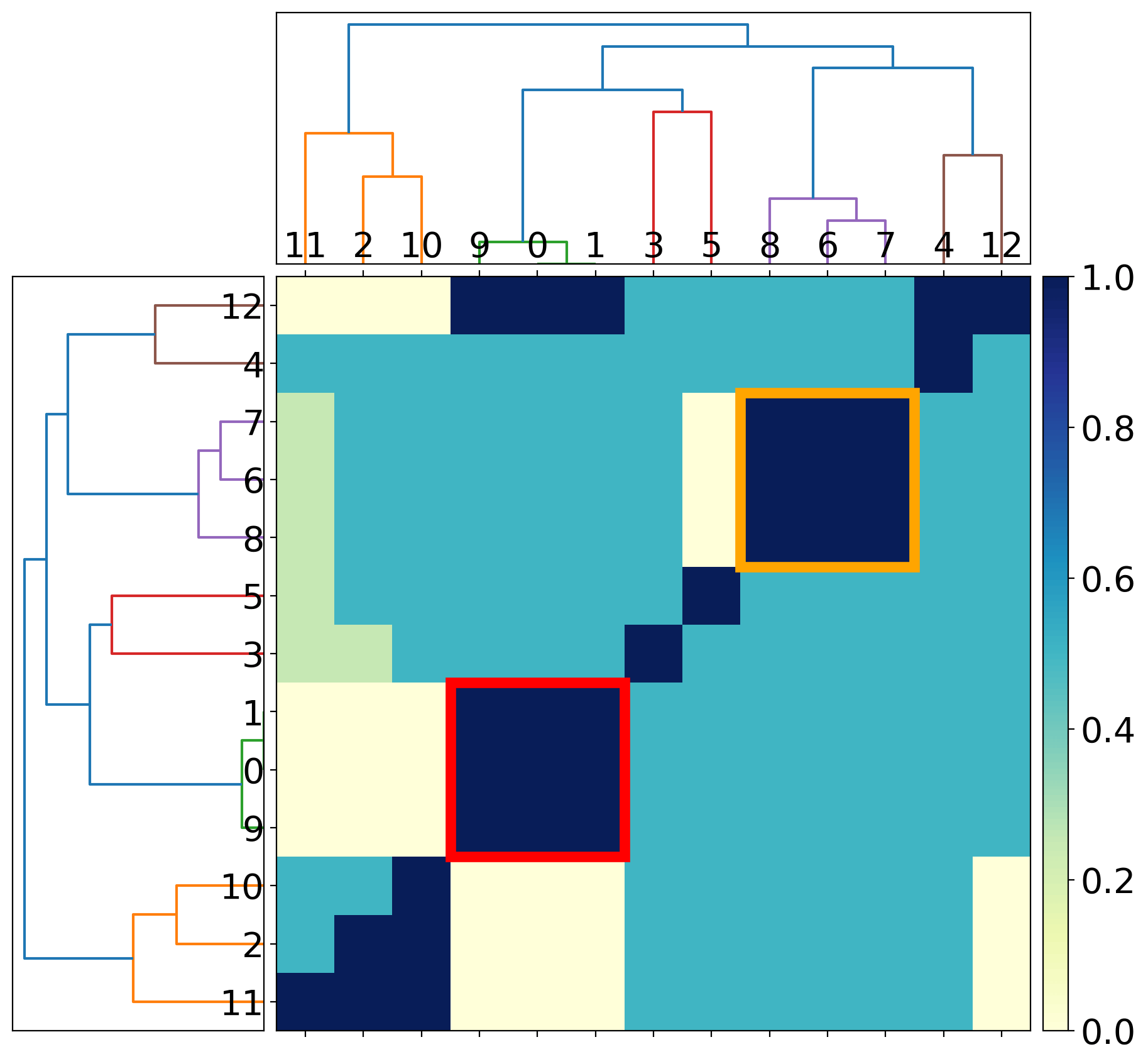}
			\end{minipage}
		}
		
		\vspace{12pt}
		
		\hspace*{-1cm}
		\subfloat[\textbf{T1:} all-MiniLM-L6-v2]{
			\label{subfig:cluster_dend_newc_l6v2}
			\begin{minipage}{0.42\textwidth}
				\centering
				\includegraphics[width=\linewidth,height=5cm,keepaspectratio]{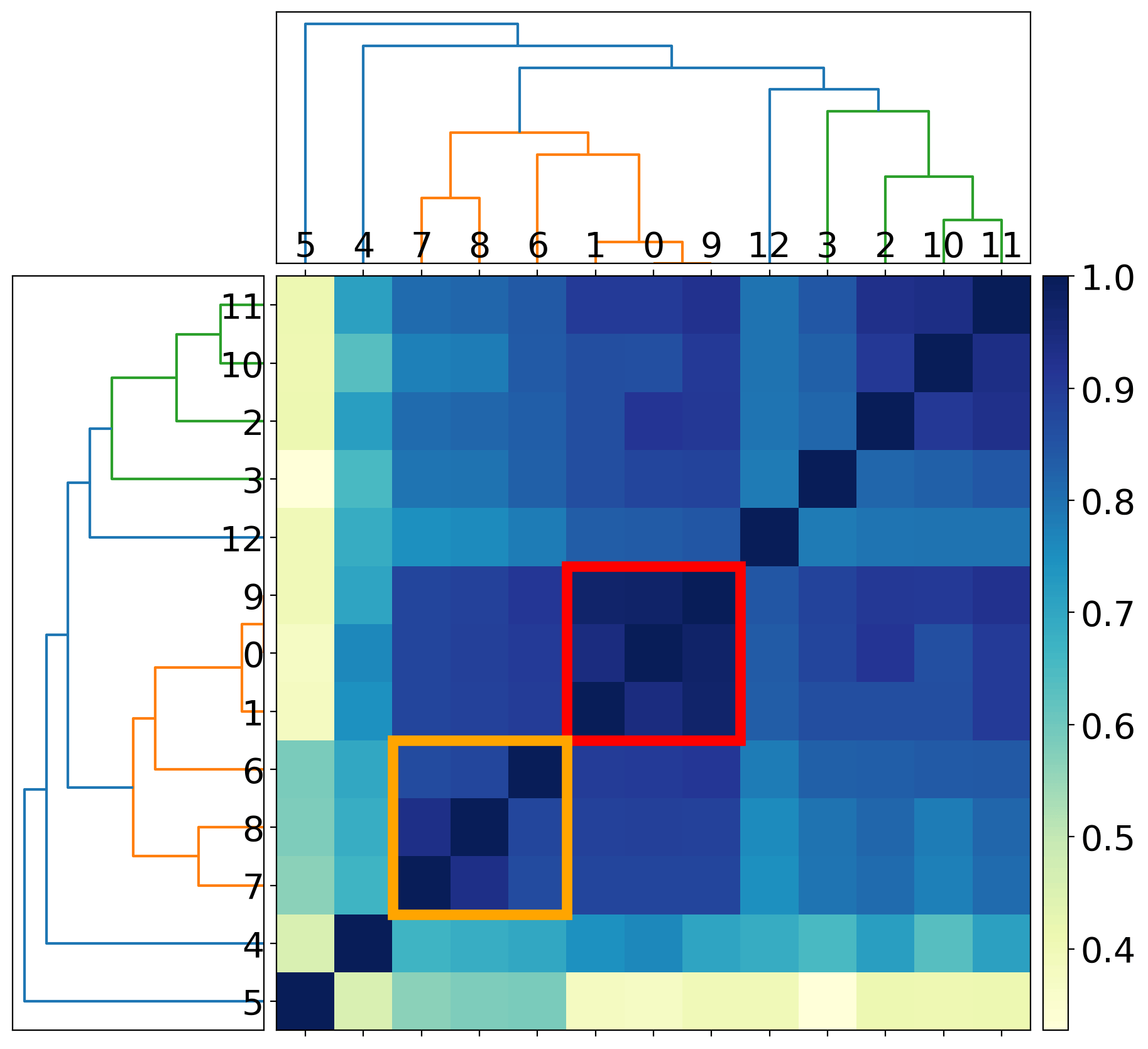}
			\end{minipage}
		}
		\hspace*{-1.5cm}
		\subfloat[\textbf{T2:} all-MiniLM-L12-v2]{
			\label{subfig:cluster_dend_newc_l12v2}
			\begin{minipage}{0.42\textwidth}
				\centering
				\includegraphics[width=\linewidth,height=5cm,keepaspectratio]{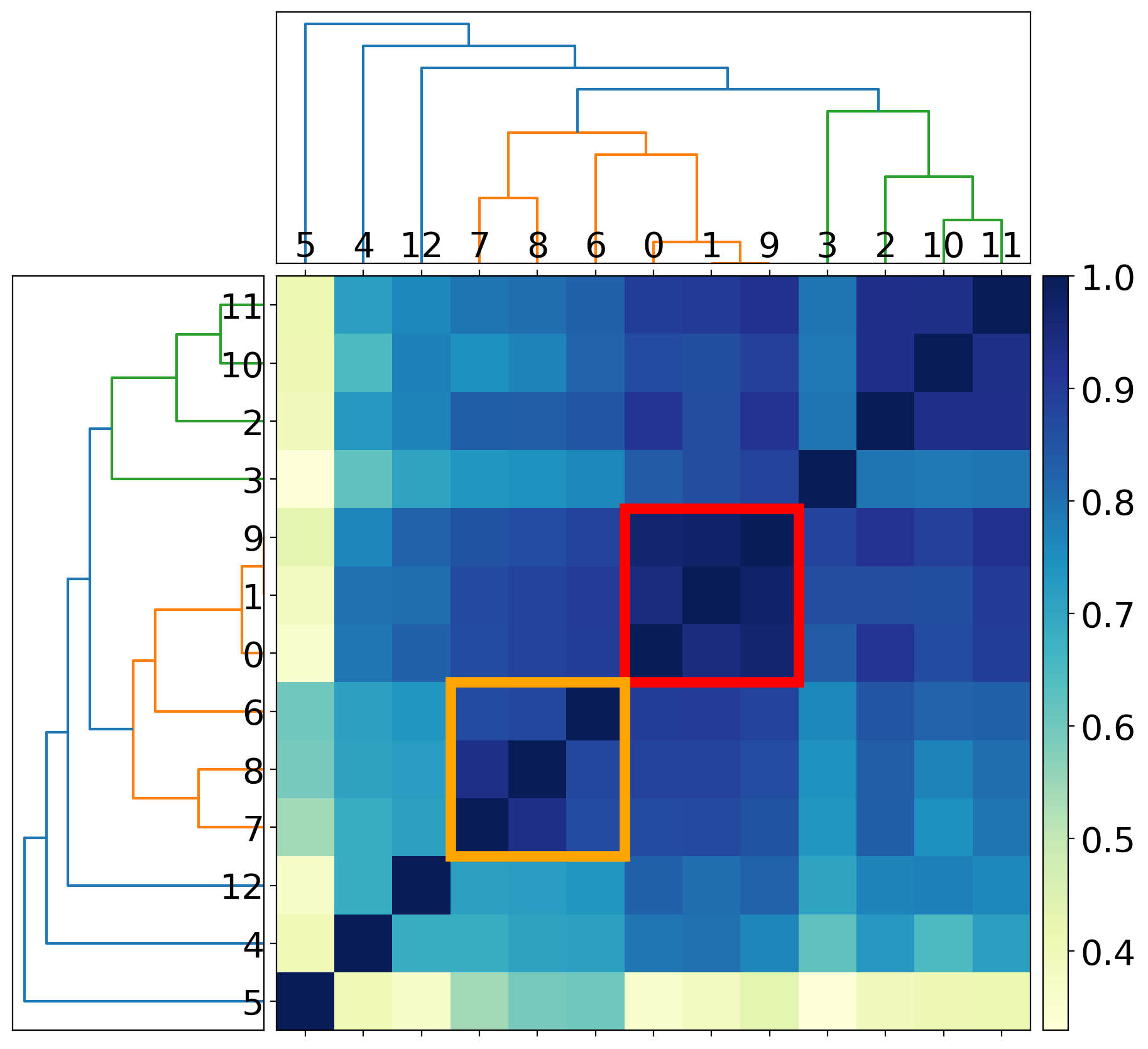}
			\end{minipage}
		}
		\hspace*{-1.5cm}
		\subfloat[\textbf{T3:} all-mpnet-base-v2]{
			\label{subfig:cluster_dend_newc_mpnet}
			\begin{minipage}{0.42\textwidth}
				\centering
				\includegraphics[width=\linewidth,height=5cm,keepaspectratio]{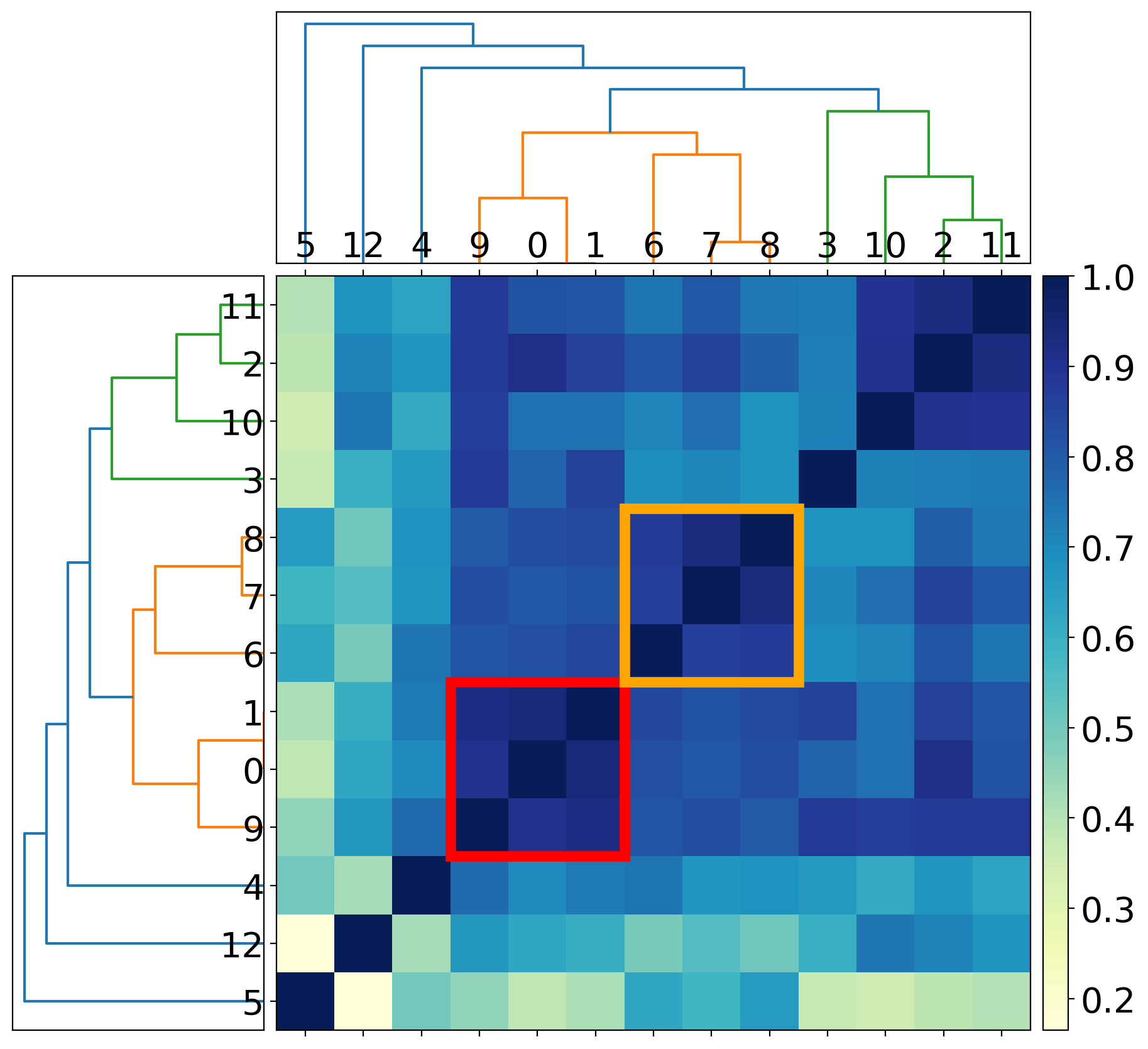}
			\end{minipage}
		}
		
		\vspace{12pt}
		
\hspace*{-1cm}
		\subfloat[\textbf{T4:} all-roberta-large-v1]{
			\label{subfig:cluster_dend_newc_roberta}
			\begin{minipage}{0.42\textwidth}
				\centering
				\includegraphics[width=\linewidth,height=5cm,keepaspectratio]{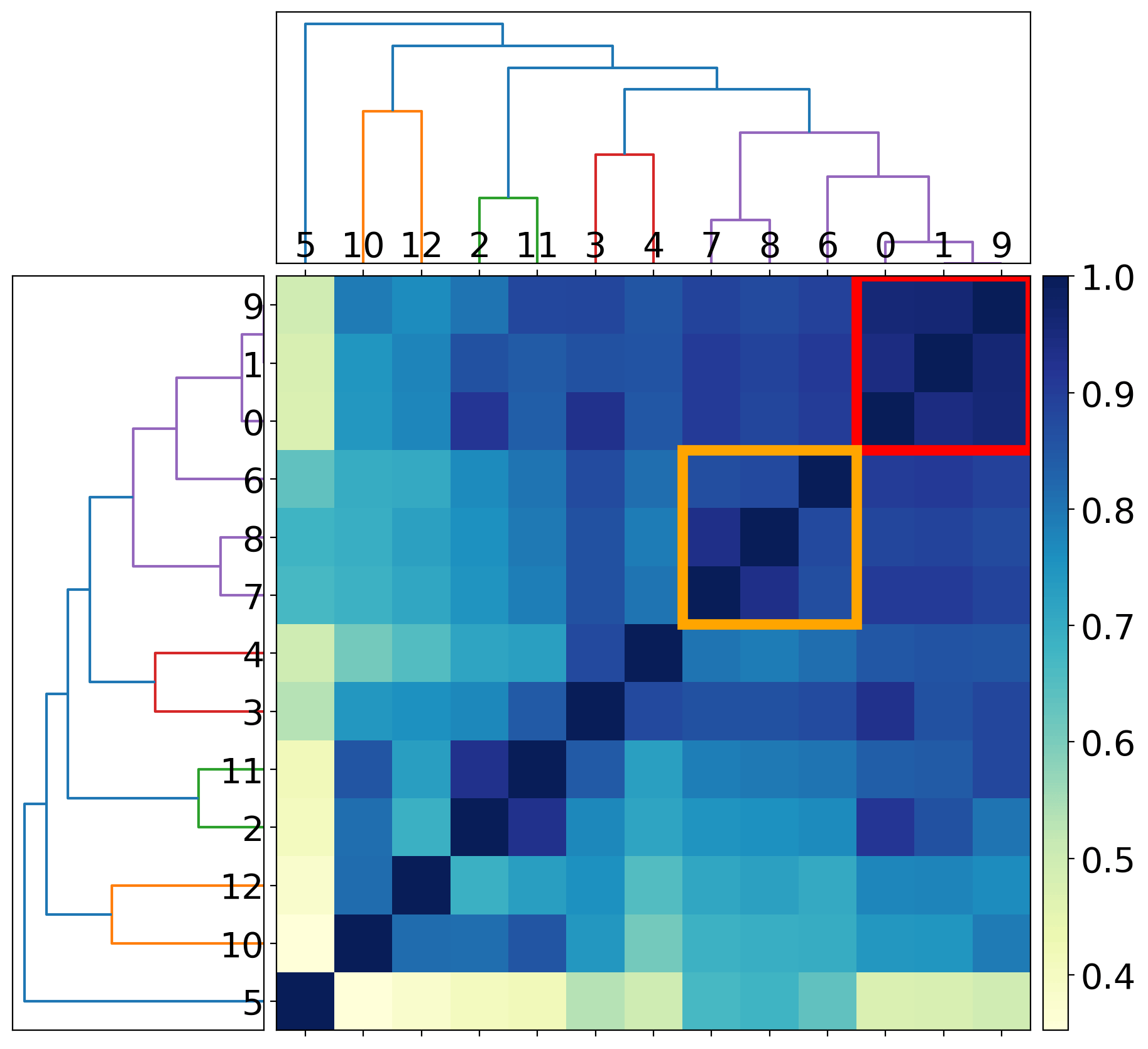}
			\end{minipage}
		}
		\hspace*{-1.5cm}
		\subfloat[\textbf{T5:} DeBERTaV2+AMR-LDA]{
			\label{subfig:cluster_dend_newc_DeBERTaV2}
			\begin{minipage}{0.42\textwidth}
				\centering
				\includegraphics[width=\linewidth,height=5cm,keepaspectratio]{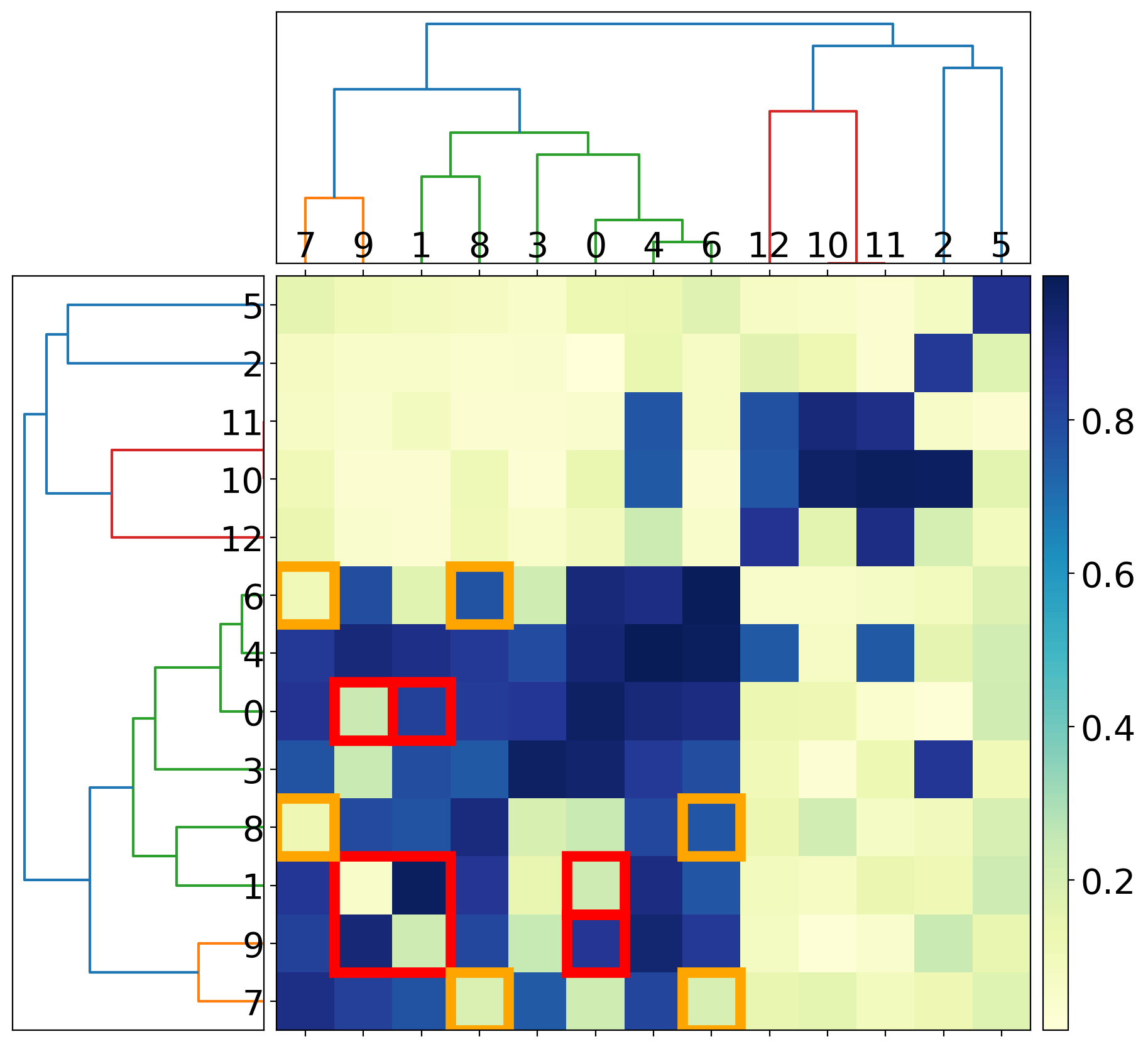}
			\end{minipage}
		}
		\hspace*{-1.5cm}
		\subfloat[\textbf{T6:} ColBERTv2+RAGatouille]{
			\label{subfig:cluster_dend_newc_ColBERTv2}
			\begin{minipage}{0.42\textwidth}
				\centering
				\includegraphics[width=\linewidth,height=5cm,keepaspectratio]{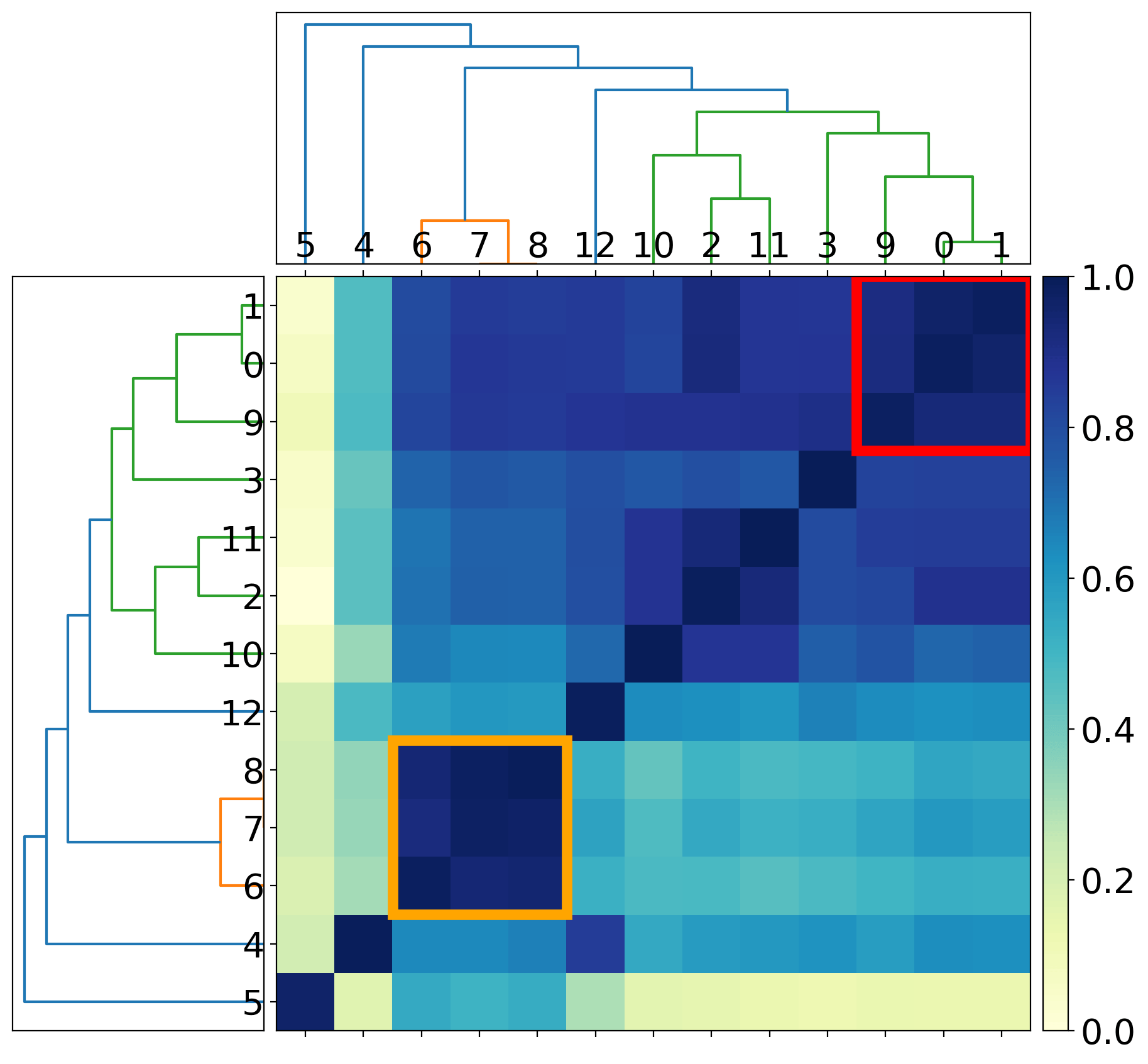}
			\end{minipage}
		}
		\caption{Dendrogram clustering for \ref{rqn2c} sentences{, where red and orange boxes represent the clusters for \{0,1,9\} and \{6,7,8\} respectively}.}
		\label{fig:clustering_dend_newc}
	\end{figure}
	
	\subsubsection{\ref{rqn2c} (Spatiotemporal)}
	
	For the \glspl*{sg} in \figurename~\ref{subfig:cluster_dend_newc_simple}, we can see that some similarity was detected between sentences 7 (``\textit{Traffic is flowing in Newcastle city centre, on Saturdays}'') and 8 (``\textit{On Saturdays, traffic is flowing in Newcastle city centre}''), which is correct; however it should be treated as having 100\% similarity, which it does not here. Furthermore, the majority of the dendrogram shows little similarity across the dataset. We see an increase in overall clustering similarity for both \gls*{ahc} and $k$-Medoids in the \glspl*{lg} approach, as shown in \figurename~\ref{subfig:cluster_dend_newc_logicalgraphs}, now also capturing similarity between a few more sentences, but overall still not performing as expected. Finally, our logical approach presents an ideal result, with 100\% clustering alignment and a dendrogram that presents clusters that match our expected outcome. \{0,1,9\} and \{6,7,8\} are clustered together, as we would expect, and we also see some further implications being presented. 
	
	The sentence transformer approaches produced a very high similarity for nearly all sentences in the dataset. There were discrepancies in the returned clusters, compared to what we would expect, for several reasons. The word embeddings might be capturing different semantic relationships. For example, ``\textit{busy}'' in sentence 4 (``\textit{It is busy in Newcastle}'') might be considered similar to the ``\textit{busy city centers}'' in sentence 5 (``\textit{Saturdays have usually busy city centers}''), leading to their clustering, even though the context of Newcastle being busy is more general than the usual Saturday busyness. We can see that sentences related to Saturdays (5, 6, 7, 8) form a relatively cohesive cluster in the dendrogram, which aligns with the desired grouping for 6, 7, and 8. However, sentence 5's early inclusion with the general ``\textit{busy}'' statement (4) deviates from the intended separation. Furthermore, the embeddings might be heavily influenced by specific keywords; for example, the presence of ``\textit{Newcastle city centre}'' in multiple sentences might lead to their clustering, even if the context regarding traffic flow or presence differs. As discussed previously, these transformers could not differentiate between the presence and absence of traffic as intended in the desired clusters. For example, sentence 10 (``\textit{Newcastle city center does not have traffic}''), was clustered with sentence 2 (``\textit{There is traffic but not in the Newcastle city centre}''), which is incorrect. 
	

	\subsection{Additional Dendrograms}\label{app:addendo}
\begin{figure}[h!]
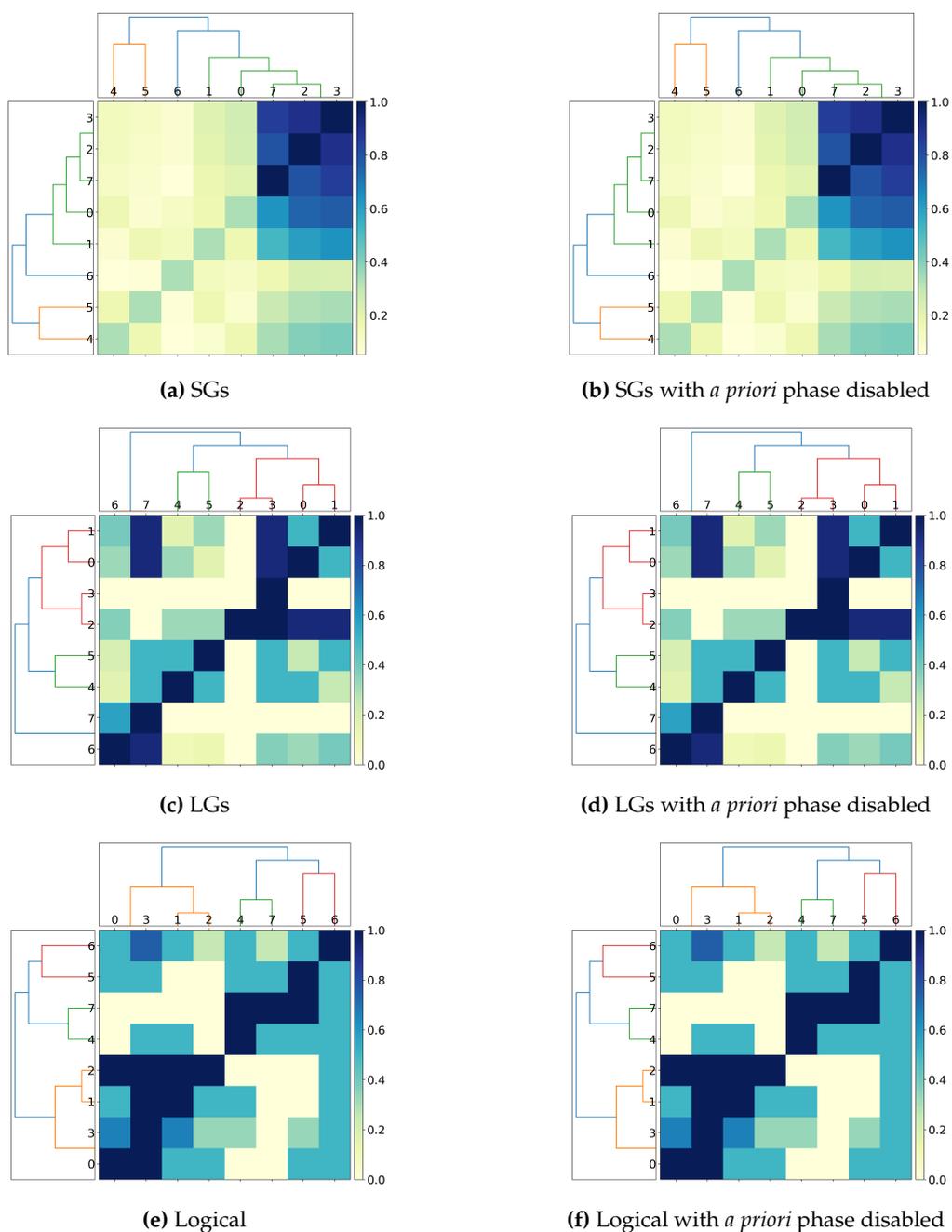
 
		\centering
		\subfloat[\glspl*{sg}]{\label{subfig:cluster_dend_ab_simplegraphs_2}
			\begin{minipage}{0.48\textwidth}
				\centering
				\includegraphics[width=\linewidth,height=5cm,keepaspectratio]{images/clustering/ab/SimpleGraph_dend.png}
			\end{minipage}
		}
		\hfill
		\subfloat[\glspl*{sg} with \textit{a priori} phase disabled]{\label{subfig:cluster_dend_ab_simplegraphs_disabled}
			\begin{minipage}{0.48\textwidth}
				\centering
				\includegraphics[width=\linewidth,height=5cm,keepaspectratio]{images/clustering/ab/SimpleGraphDisabledAdHoc_dend.png}
			\end{minipage}
		}
		\vspace{12pt}
		
		\subfloat[\glspl*{lg}]{\label{subfig:cluster_dend_ab_logicalgraphs_2}
			\begin{minipage}{0.48\textwidth}
				\centering
				\includegraphics[width=\linewidth,height=5cm,keepaspectratio]{images/clustering/ab/LogicalGraph_dend.png}
			\end{minipage}
		}
		\hfill
		\subfloat[\glspl*{lg} with \textit{a priori} phase disabled]{\label{subfig:cluster_dend_ab_logicalgraphs_disabled}
			\begin{minipage}{0.48\textwidth}
				\centering
				\includegraphics[width=\linewidth,height=5cm,keepaspectratio]{images/clustering/ab/LogicalGraphDisabledAdHoc_dend.png}
			\end{minipage}
		}
		\vspace{12pt}
		
		\subfloat[Logical]{\label{subfig:cluster_dend_ab_logical_2}
			\begin{minipage}{0.48\textwidth}
				\centering
				\includegraphics[width=\linewidth,height=5cm,keepaspectratio]{images/clustering/ab/Logical_dend.png}
			\end{minipage}
		}
		\hfill
		\subfloat[Logical with \textit{a priori} phase disabled]{\label{subfig:cluster_dend_ab_logical_disabled}
			\begin{minipage}{0.48\textwidth}
				\centering
				\includegraphics[width=\linewidth,height=5cm,keepaspectratio]{images/clustering/ab/LogicalDisabledAdHoc_dend.png}
			\end{minipage}
		}
		
		\caption{Dendrogram clustering graphs for \ref{rqn2a} sentences, with comparisons when the \textit{a priori}  phase is disabled. No clusters are highlighted, as there are none to present.}
		\label{fig:clustering_disabled_dend_alice_bob}
	\end{figure}

\begin{figure}[h!] 
		\centering
		\subfloat[\glspl*{sg}]{\label{subfig:cluster_dend_cm_simplegraphs_2}
			\begin{minipage}{0.48\textwidth}
				\centering
				\includegraphics[width=\linewidth,height=5cm,keepaspectratio]{images/clustering/cm/SimpleGraph_dend.png}
			\end{minipage}
		}
		\hfill
		\subfloat[\glspl*{sg} with \textit{a priori} phase disabled]{\label{subfig:cluster_dend_cm_simplegraphs_disabled}
			\begin{minipage}{0.48\textwidth}
				\centering
				\includegraphics[width=\linewidth,height=5cm,keepaspectratio]{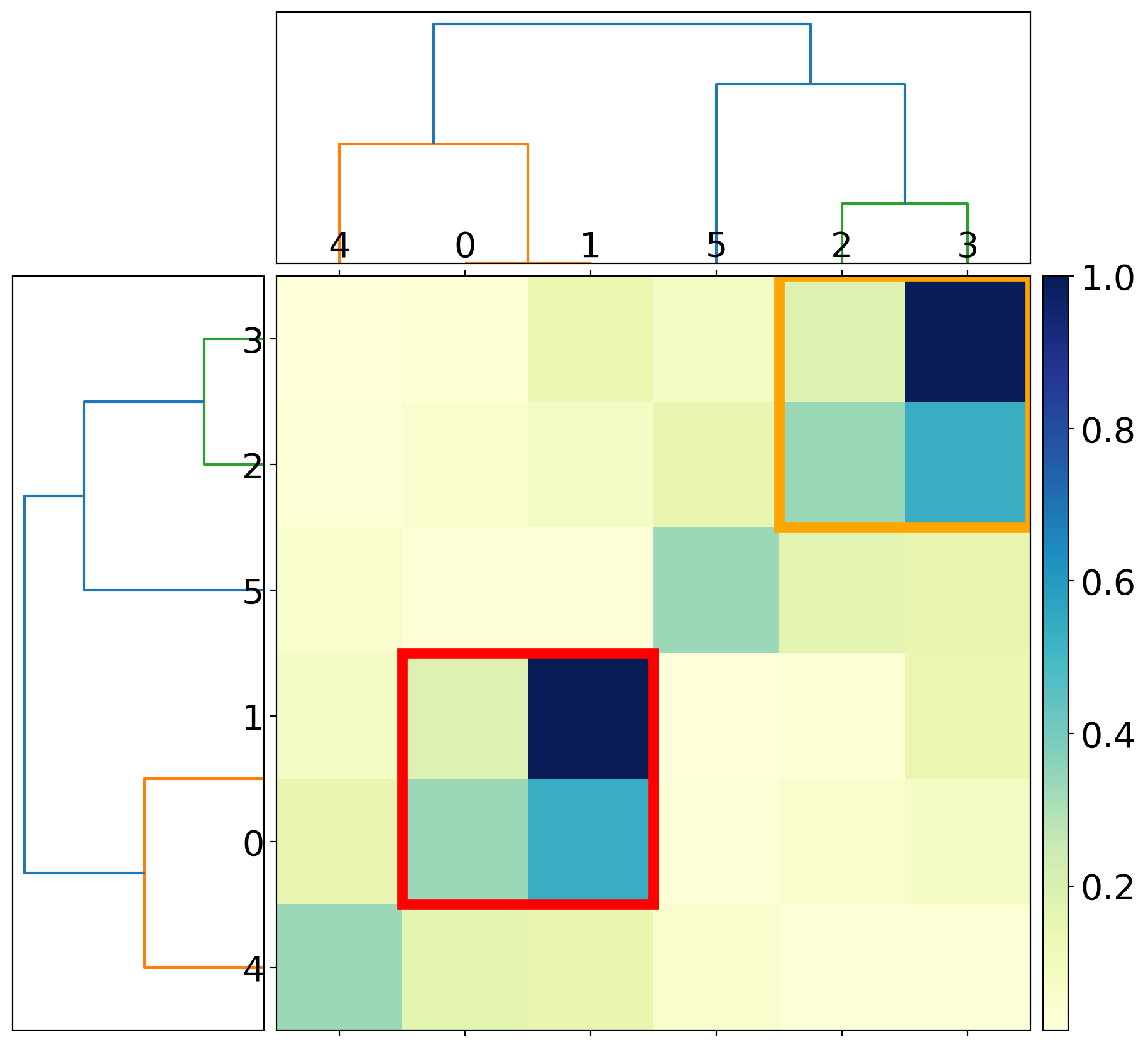}
			\end{minipage}
		}
		\vspace{12pt}
		
		\subfloat[\glspl*{lg}]{\label{subfig:cluster_dend_cm_logicalgraphs_2}
			\begin{minipage}{0.48\textwidth}
				\centering
				\includegraphics[width=\linewidth,height=5cm,keepaspectratio]{images/clustering/cm/LogicalGraph_dend.png}
			\end{minipage}
		}
		\hfill
		\subfloat[\glspl*{lg} with \textit{a priori} phase disabled]{\label{subfig:cluster_dend_cm_logicalgraphs_disabled}
			\begin{minipage}{0.48\textwidth}
				\centering
				\includegraphics[width=\linewidth,height=5cm,keepaspectratio]{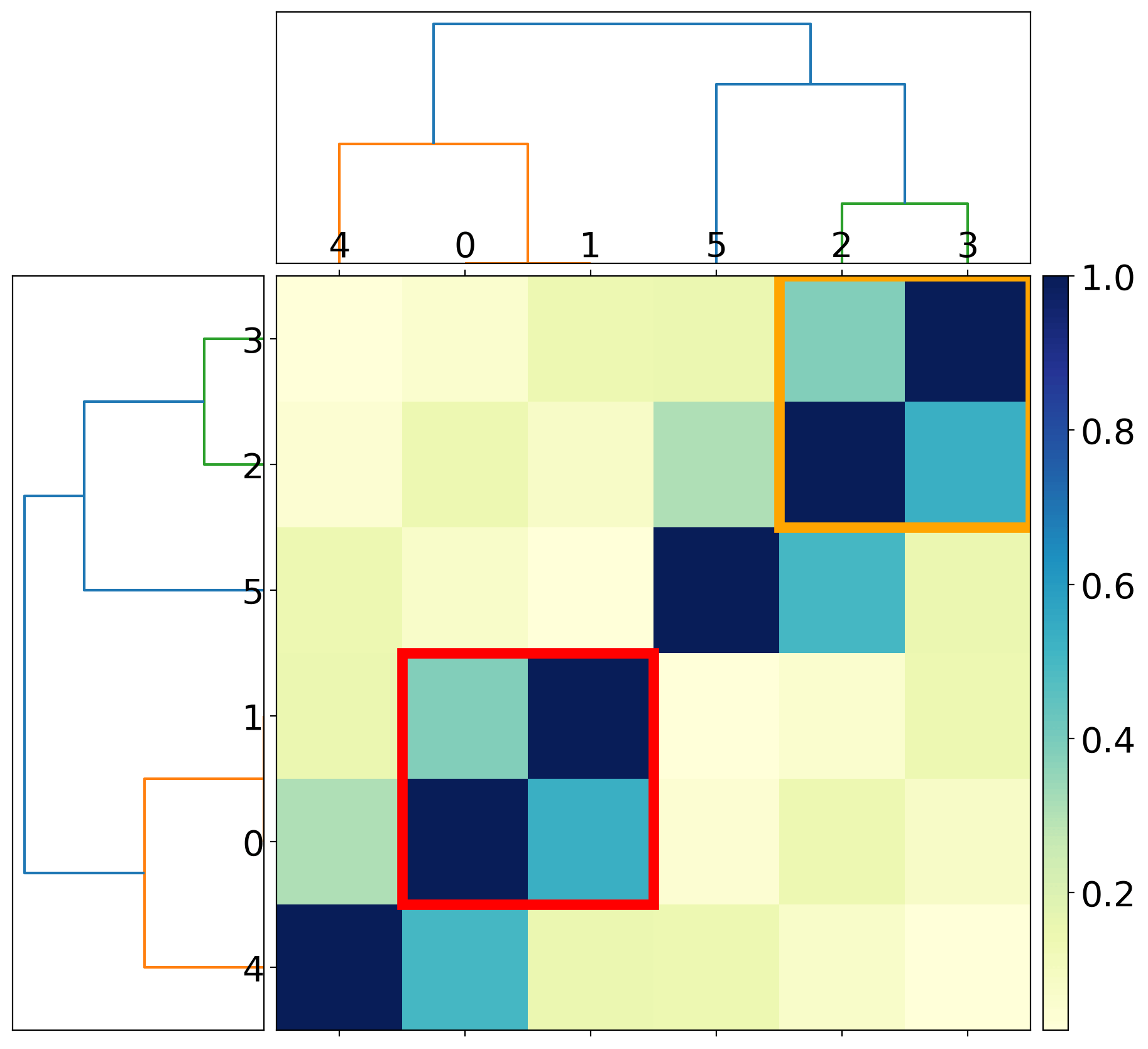}
			\end{minipage}
		}
		\vspace{12pt}
		
		\subfloat[Logical]{\label{subfig:cluster_dend_cm_logical_2}
			\begin{minipage}{0.48\textwidth}
				\centering
				\includegraphics[width=\linewidth,height=5cm,keepaspectratio]{images/clustering/cm/Logical_dend.png}
			\end{minipage}
		}
		\hfill
		\subfloat[Logical with \textit{a priori} phase disabled]{\label{subfig:cluster_dend_cm_logical_disabled}
			\begin{minipage}{0.48\textwidth}
				\centering
				\includegraphics[width=\linewidth,height=5cm,keepaspectratio]{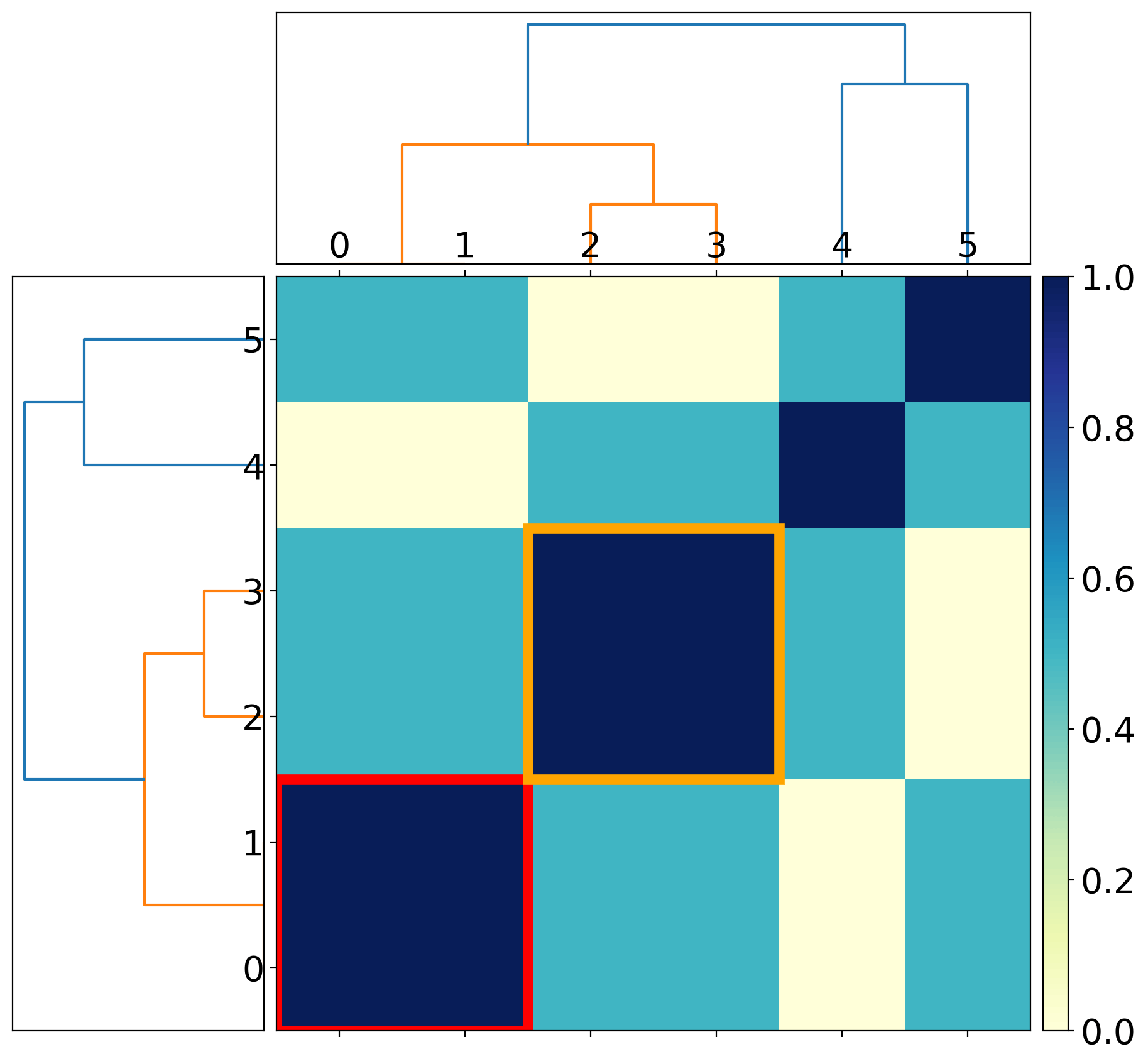}
			\end{minipage}
		}
		
		\caption{{Dendrogram clustering graphs for \ref{rqn2b} sentences, with comparisons when the \textit{a priori}  phase is disabled. Red and orange boxes represent the clusters for \{0,1\} and \{2,3\} respectively.}}
		\label{fig:clustering_disabled_dend_cat_mouse}
	\end{figure}

\begin{figure}[h!] 
		\centering
		\subfloat[\glspl*{sg}]{\label{subfig:cluster_dend_newc_simplegraphs_2}
			\begin{minipage}{0.48\textwidth}
				\centering
				\includegraphics[width=\linewidth,height=5cm,keepaspectratio]{images/clustering/newc/SimpleGraph_dend.png}
			\end{minipage}
		}
		\hfill
		\subfloat[\glspl*{sg} with \textit{a priori} phase disabled]{\label{subfig:cluster_dend_newc_simplegraphs_disabled}
			\begin{minipage}{0.48\textwidth}
				\centering
				\includegraphics[width=\linewidth,height=5cm,keepaspectratio]{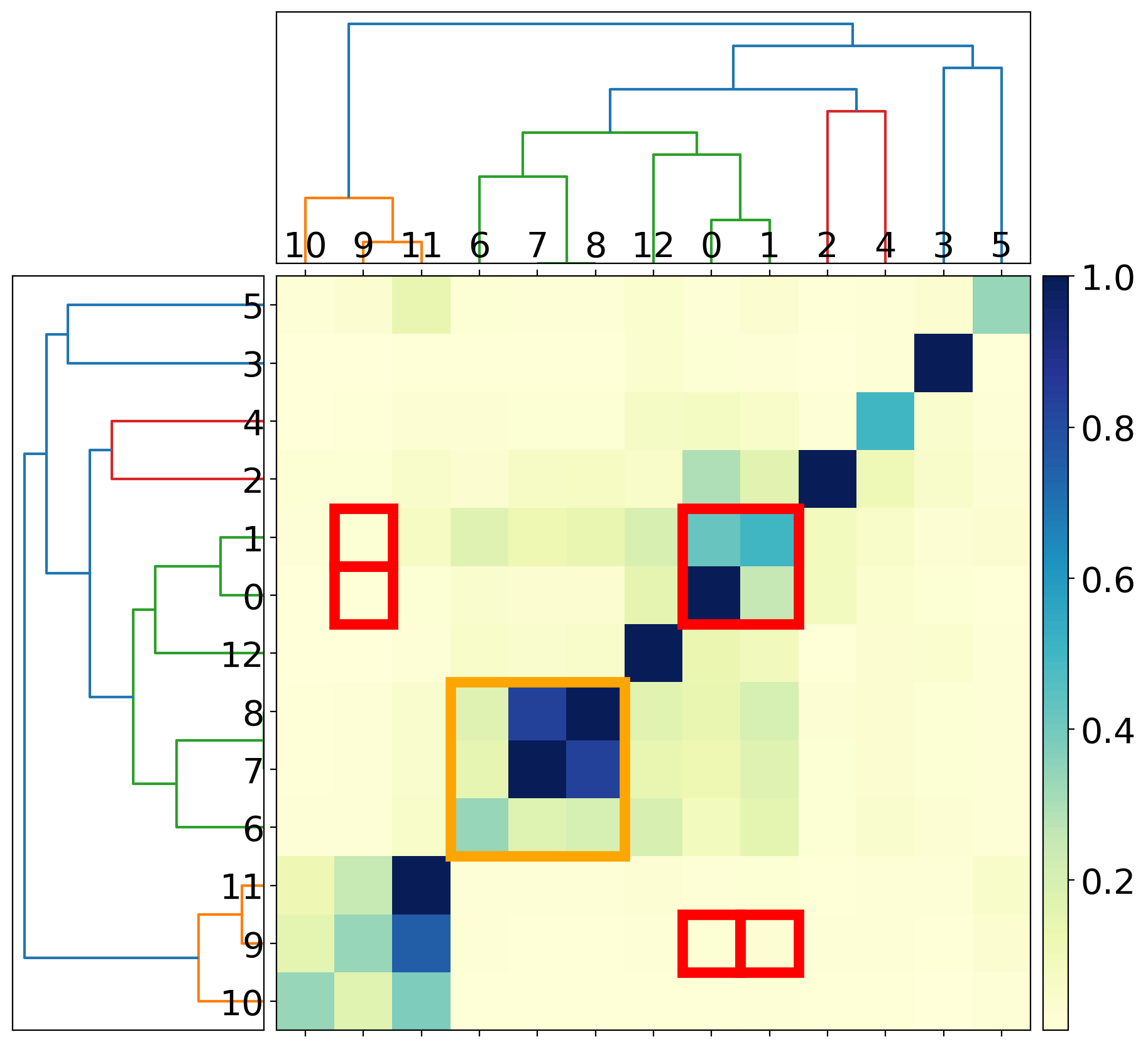}
			\end{minipage}
		}
		\vspace{12pt}
		
		\subfloat[\glspl*{lg}]{\label{subfig:cluster_dend_newc_logicalgraphs_2}
			\begin{minipage}{0.48\textwidth}
				\centering
				\includegraphics[width=\linewidth,height=5cm,keepaspectratio]{images/clustering/newc/LogicalGraph_dend.png}
			\end{minipage}
		}
		\hfill
		\subfloat[\glspl*{lg} with \textit{a priori} phase disabled]{\label{subfig:cluster_dend_newc_logicalgraphs_disabled}
			\begin{minipage}{0.48\textwidth}
				\centering
				\includegraphics[width=\linewidth,height=5cm,keepaspectratio]{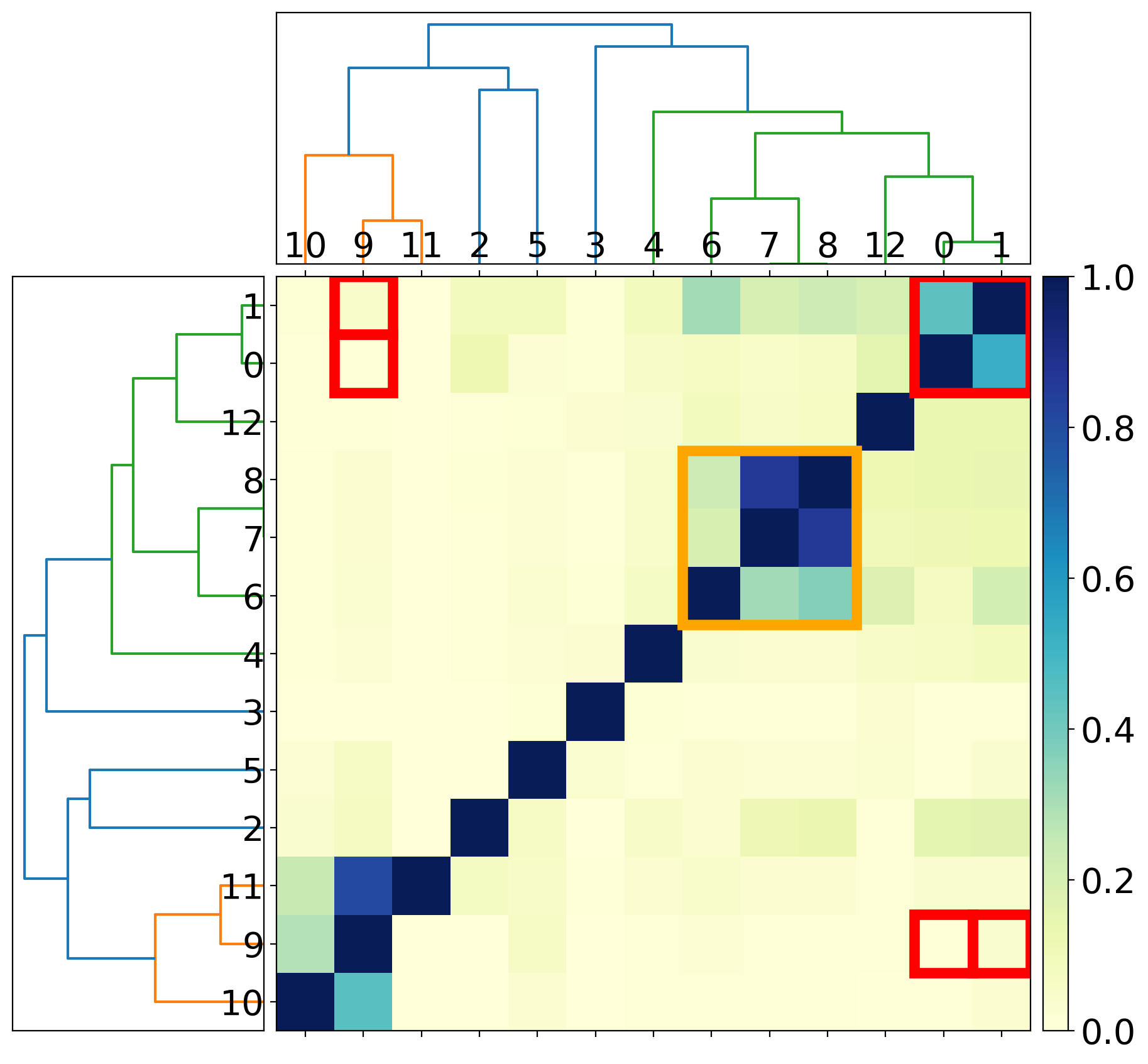}
			\end{minipage}
		}
		\vspace{12pt}
		
		\subfloat[Logical]{\label{subfig:cluster_dend_newc_logical_2}
			\begin{minipage}{0.48\textwidth}
				\centering
				\includegraphics[width=\linewidth,height=5cm,keepaspectratio]{images/clustering/newc/Logical_dend.png}
			\end{minipage}
		}
		\hfill
		\subfloat[Logical with \textit{a priori} phase disabled]{\label{subfig:cluster_dend_newc_logical_disabled}
			\begin{minipage}{0.48\textwidth}
				\centering
				\includegraphics[width=\linewidth,height=5cm,keepaspectratio]{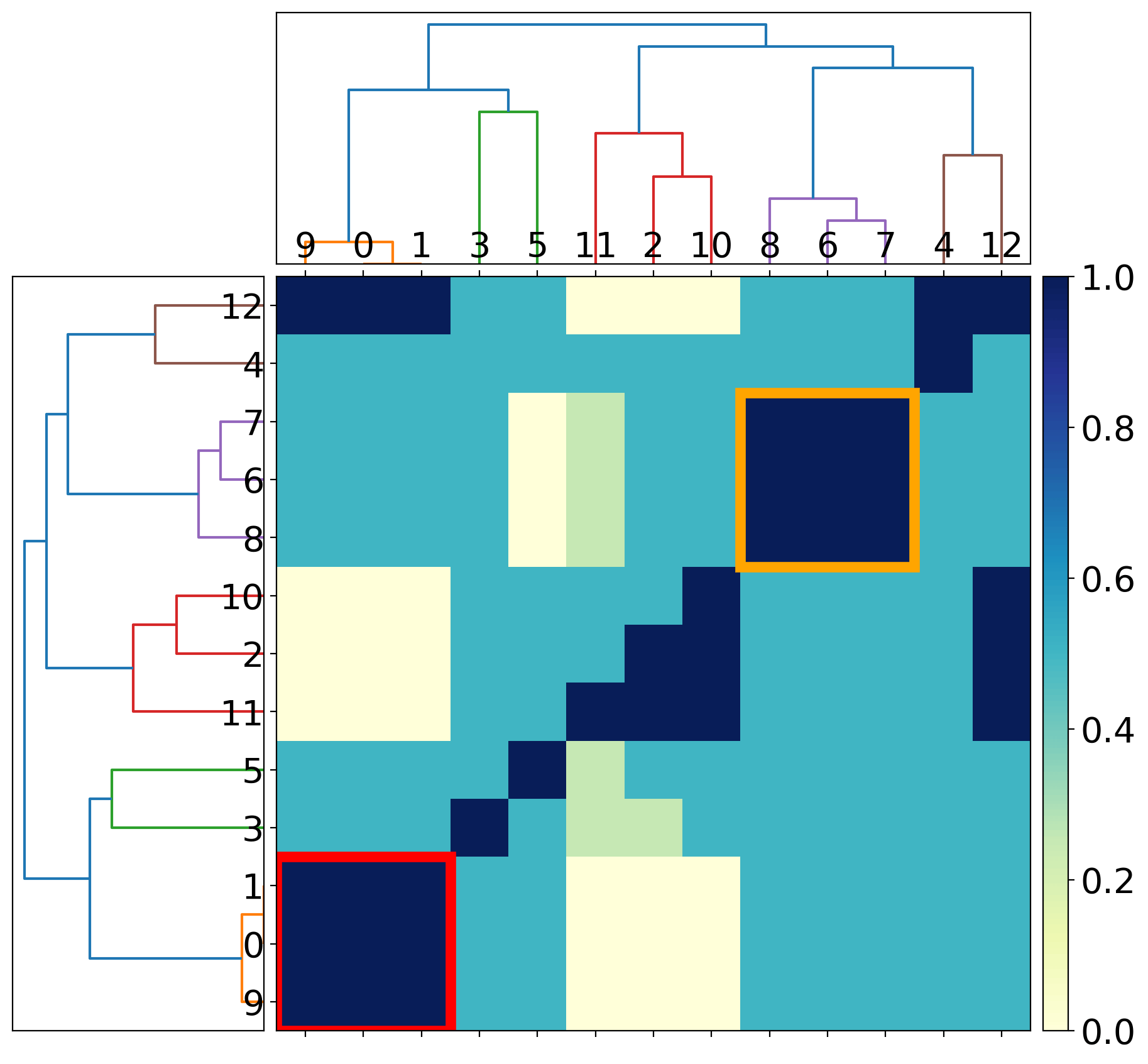}
			\end{minipage}
		}
		
		\caption{Dendrogram clustering graphs for \ref{rqn2c} sentences, with comparisons when the \textit{a priori}  phase is disabled. Red and orange boxes represent the clusters for \{0,1,9\} and \{6,7,8\} respectively.}
		\label{fig:clustering_disabled_dend_newcastle}
	\end{figure}

	\textit{Dendrograms in Figures \ref{fig:clustering_disabled_dend_alice_bob}, \ref{fig:clustering_disabled_dend_cat_mouse}, and \ref{fig:clustering_disabled_dend_newcastle} provide additional plots that were omitted in the main manuscript, as they showed similar results.}
	
	\clearpage
	\section{Additional Algorithms}
	\textit{Algorithms \ref{alg:merge_singletons} provide additional algorithms that were omitted in the main manuscript, they provide additional detail through pseudocode for different parts of the \gls*{lassi} pipeline.}
	
	\begin{algorithm}[!h]
		\caption{Given a \texttt{SetOfSingletons} node, this pseudocode shows how it is \textit{merged}, while also determining whether an `extra' should be added to the resulting merged \texttt{Singleton} node.}
		\label{alg:merge_singletons}
		\begin{algorithmic}[1]
			\Function{MultiWordEntityRecognition}{\texttt{node:SetOfSingletons}}
			\State \textbf{global} \texttt{meu\_db}
			\State \texttt{sorted\_entities} $\gets$ \textbf{sort} \texttt{(node.entities)} \textbf{by} \texttt{pos}
			\State \texttt{sorted\_entity\_names} $\gets$ \texttt{sorted\_entities.name}
			
			
			\For{$\texttt{layer} \in \textbf{sort} \{[s_i,\dots,s_{i+n-1}]\subseteq\texttt{sorted\_entities}|\,n>1\} \textbf{by}\, \textrm{length}$}  \label{line:loop_alternatives}\label{line:get_alternatives}
			\State \texttt{max\_confidence} $\gets$ -1
			
			\For{$\texttt{name} \in \texttt{layer}$}
			\State \texttt{all\_types} $\gets$ \texttt{sorted\_entities.type}
			\State $\texttt{ST} \leftarrow \texttt{MST}(\texttt{all\_types})$ \label{line:most_specific_type}
			
			\State \texttt{confidence, type} $\gets$\\ \qquad\qquad\qquad\texttt{meu\_resolution(name.min, name.max, ST, meu\_db)} \label{line:get_score_from_meu}
			
			\State $\texttt{all\_scores} \leftarrow \prod_{z \in \texttt{name}} \texttt{sorted\_entities}[z].\texttt{confidence}$
			
			\If{\Call{shouldAddAlternative}{}} \label{line:should_add_alternative}
			\State $\texttt{alternatives} \gets [(\texttt{name}, \texttt{type}, \texttt{total\_confidence})]$ \label{line:chosen_alternative}
			\State $\texttt{max\_confidence} \leftarrow \texttt{confidence}$
			\EndIf
			\EndFor
			
			\If{$|\texttt{alternatives}| > 0$}
			\State \texttt{alternatives} $\gets$ \textbf{sort} \texttt{alternatives} \textbf{by} \texttt{total\_confidence}
			\State $G(d) \gets \Set{(idx, x)|\texttt{sorted\_entity\_names}[idx] = x}$
			\State $\texttt{candidate\_delete} \leftarrow \emptyset$
			\State \texttt{x} $\gets$ \texttt{alternatives}[-1] \Comment{alternative with lowest confidence}
			\For{$k, v \in \texttt{d}.\texttt{items}()$}
			\If{$\texttt{isinstance}(k, \texttt{int})$}
			\If{$k \in \texttt{x}$}
			\State $\texttt{candidate\_delete} \leftarrow \texttt{candidate\_delete} \cup \{k\}$
			\ElsIf{$\texttt{isinstance}(k, \texttt{tuple})$}
			\If{$|\texttt{set}(\texttt{x.name}) \cap \texttt{set}(k)| > 0$}
			\State $\texttt{candidate\_delete} \leftarrow \texttt{candidate\_delete} \cup \{k\}$
			\EndIf
			\EndIf
			\EndIf
			\EndFor
			\For{$z \in \texttt{candidate\_delete}$}
			\State \textbf{remove} $z$ \textbf{from} \texttt{d}
			\EndFor
			\State \texttt{resolved\_d} $\gets$ \texttt{resolved\_d} $\cup\ \{(\texttt{d}, \texttt{total\_confidence}, \texttt{type})\}$
			\EndIf
			\EndFor
			
			\If{$|\texttt{resolved\_d}| > 1$} \label{line:multiple_resolutions}
			\State \texttt{d} $\gets$ \Call{getAlternativeWithMostEntities}{} \label{line:get_most_entities}
			\ElsIf{$|\texttt{resolved\_d}| = 1$}
			\State \texttt{d} $\gets$ \texttt{resolved\_d}
			\EndIf
			
			\For{\texttt{entity} $\in$ \texttt{sorted\_entities}}
			\If{\texttt{entity.name} $=$ \texttt{d[0].name}}
			\State \texttt{chosen\_entity} $\gets$ \texttt{entity} \label{line:chosen_entity}
			\Else	
			\State \texttt{extra\_name} $\gets$ \texttt{extra\_name} $+$ " " $+$ \texttt{entity.name}
			\EndIf
			\EndFor
			\State \
			\EndFunction
		\end{algorithmic}
	\end{algorithm}

\end{document}